%% file: lic_mk_digital.tex
\pdfoutput=1

\documentclass{LTHthesis}
\usepackage[T1]{fontenc}
\usepackage[utf8]{inputenc}  
\usepackage{color}
\usepackage{mathptmx, helvet}
\usepackage{standalone}
\usepackage{pgfplots}

\newsubfloat{figure}

\usepackage{amsmath,amsfonts,amssymb}
\pgfplotsset{compat=1.14}

\usepackage[capitalise]{cleveref} 

\crefname{section}{Sec.}{Secs.}
\Crefname{section}{Section}{Sections}

\crefname{figure}{Fig.}{Figs.}
\Crefname{figure}{Figure}{Figures}

\crefname{equation}{}{}
\Crefname{equation}{Equation}{Equations}

\usepackage{url}

\usepackage{rotating}

\usepackage[detect-all]{siunitx}
\usepackage{tikz}
\usepgfplotslibrary{groupplots}
\usetikzlibrary{plotmarks}
\usetikzlibrary{positioning}
\usetikzlibrary{shapes,arrows, fit, positioning}
\usetikzlibrary{narrow}
\usepgfplotslibrary{external}
\usepackage{tikz-3dplot}
\usepackage{tikzscale}
\usepackage{centeredtikzarrowbulletheads}
\tikzset{>=latex}
\tikzset{
 font={\fontsize{10pt}{12}\selectfont}}
\DeclareMathOperator{\dis}{d}

\newlength\figureheight 
\newlength\figurewidth 
\pgfplotsset{every tick label/.append style={font=\small}}
\pgfplotsset{legend style={font=\small}}
\pgfplotsset{every axis label/.append style={font=\small}}

\begin{document}
\include{chapters/frontmatter/frontmatter}
\include{chapters/intro/intro}

\include{chapters/publications/publications}

\include{chapters/discussion/discussion}

\include{chapters/conclusion/conclusion}
\begin{papers}
\include{chapters/paper1/paper1}
\include{chapters/paper2/paper2}

\include{chapters/paper3/paper3}
\end{papers}

\nocite{karlsson2017autonomous}
\nocite{karlsson2017dmp}
\nocite{karlsson2017detection}
\nocite{wadenback2017visual}
\nocite{karlsson2016fsw}
\nocite{bagge2016particle}
\nocite{karlsson2016cooperative}
\nocite{haage2016cognitive}
\nocite{karlsson2015sensor}
\printbibliography
\end{document}

%% file: chapters/frontmatter/frontmatter.tex
\begin{titlepages}
\author{Martin Karlsson}
\title{On Motion Control and Machine Learning for Robotic Assembly}
\year{2017}
\month{June}
\TFRT{3274}  
\printer{Media-Tryck}  
\type{Lic. Tech. Thesis}
\end{titlepages}

\input{chapters/abstract/abstract.tex}
\input{chapters/acknow/acknow.tex}

\tableofcontents

%% file: chapters/abstract/abstract.tex
\chapter*{Abstract}
Industrial robots typically require very structured and predictable working environments, and explicit programming, in order to perform well. Therefore, expensive and time-consuming engineering work is a major obstruction when mediating tasks to robots. This thesis presents methods that decrease the amount of engineering work required for robot programming, and increase the ability of robots to handle unforeseen events. This has two main benefits: Firstly, the programming can be done faster, and secondly, it becomes accessible to users without engineering experience. Even though these methods could be used for various types of robot applications, this thesis is focused on robotic assembly tasks.

Two main topics are explored: In the first part, we consider adjustment of robot trajectories generated by dynamical movement primitives (DMPs). The framework of DMPs as robot trajectory generators has been widely used in robotics research, because of their convergence properties and emphasis on easy modification. For instance, time scale and goal state can be adjusted by one parameter each, commonly without further considerations. In this thesis, the DMP framework is extended with a method that allows a robot operator to adjust DMPs by demonstration, without any traditional computer programming or other engineering work required. Given a generated trajectory with a faulty last part, the operator can use lead-through programming to demonstrate a corrective trajectory. A modified DMP is formed, based on the first part of the faulty trajectory and the last part of the corrective one. Further, a method for handling perturbations during execution of DMPs on robots is considered. Two-degree-of-freedom control is used together with temporal coupling, to achieve practically realizable reference trajectory tracking and perturbation recovery. In the second part of the thesis, a method that enables robots to learn to recognize contact force/torque transients acting on the end-effector, without using a force/torque sensor, is presented. A recurrent neural network (RNN) is used for transient detection, with robot joint torques as input. A machine learning approach to determine the parameters of the RNN is presented.

Each of the methods presented in this thesis is implemented in a real-time application and verified experimentally on a robot.

%% file: chapters/acknow/acknow.tex
\chapter*{Acknowledgments}
I would like to thank my supervisor Prof. Rolf Johansson, for your invaluable advice and support throughout this work. Because of your balance between pointing out interesting research directions and giving me freedom to define my work, I look forward to many more days as a PhD student. Thank you for sharing so much of your knowledge and experience.

My co-supervisor Prof. Anders Robertsson, thank you for all your guidance. Three years ago, you patiently taught me how to communicate with the internal controller of an industrial robot. Since then, you have supported me not only with theoretical insights, but also in overcoming practical difficulties, easily encountered in a laboratory environment.


Fredrik Bagge Carlson, I am very lucky to have you as a close colleague. It has been both fruitful and fun to do research together with you. We have a common interest not only in our work, but also in solid-state physics in general and defects in semiconductors in particular, and this has resulted in many interesting discussions and good times. Thank you!

Dr. Björn Olofsson, you deserve deep gratitude for contributing with your energy, your friendliness, and your patience with practicalities in the lab. Thank you for the many times you came to the rescue, when I was lost in the world of Terminal commands.

I would like to acknowledge my colleagues in the RobotLab, Dr. Mahdi Ghazaei Ardakani, Dr. Maj Stenmark, Asst. Prof. Mathias Haage, Prof. Jacek Malec, Assoc.~Prof.~Elin Anna Topp, Assoc. Prof. Klas Nilsson, Dr. Anders Nilsson, Anders Blomdell and Pontus Andersson, as well as former colleagues, Dr. Magnus Linderoth, Dr.~Olof Sörnmo, Martin Holmstrand, Dr. Karl Berntorp and Dr. Andreas Stolt. Thank you for great cooperation.

Dr. Mårten Wadenbäck, thank you for being a fantastic colleague, co-author, and friend, and for introducing me to the world of visual odometry. It was worth all the difficulties with cables and camera settings, and even two minor electric shocks.

I have the honor and pleasure of sharing office with Olof Troeng, Irene Zorzan and Victor Millnert. Thank you for creating such a nice working environment! Thanks to you, I enjoy workdays just as much as weekends.

I am grateful to the administrative staff at the department, Ingrid Nilsson, Mika Nishimura, Monika Rasmusson, Cecilia Edelborg Christensen, and Eva Westin, for your help with various things related to my work.

Leif Andersson, without your deep knowledge in \LaTeX, this thesis, as well as my other publications, would have looked significantly worse. Every time I think that I need to add a new package, you convince me to instead remove several unnecessary ones, always with successful results. On a related note, I thank Dr. Björn Olofsson, Axel Karlsson, Fredrik Bagge Carlson, and Dr. Mahdi Ghazaei Ardakani, for proofreading of this thesis.

I also thank all colleagues at the department. Each one creates a positive, inspiring, and fun working environment, and contributes to interesting discussions about control theory and related topics.

Last but not least, Daniel, Ann-Eli, Axel, and Johan Karlsson, have been a fantastic family through my entire life. You constantly provide me with insights and ideas from the by no means negligible world outside my research field. Heartfelt thanks for all the love and support!

\vfill

\section*{Financial Support}
Financial support is gratefully acknowledged from the European Commission, under the Framework Programme Horizon 2020 -- within grant agreement No 644938 -- SARAFun, and under the 7th Framework Programme -- within grant agreement No 606156 -- FlexiFab, as well as from the Swedish Foundation for Strategic Research through the SSF project ENGROSS. The author is a member of the LCCC Linnaeus Center, supported by the Swedish Research Council, and the ELLIIT Excellence Center, supported by the Swedish Government.

%% file: chapters/intro/intro.tex
\chapter{Introduction}
Compared to humans, typical industrial robots are very good at performing sequences of pre-defined movements, with high speed and high accuracy in the position domain. This has promoted automation of repetitive tasks where position control suffices, such as spray painting and welding. However, standard industrial robots perform well only in carefully structured workcells, specifically designed to fit the robot and the given task. In general, it is required that the task is highly repetitive, and possible deviations from the original plan must have been foreseen by the robot programmer. See, \textit{e.g.}, \cite{spong2006robot,siciliano2010robotics} for an introduction to robot modeling and control in general. Even under favorable conditions, traditional robot programming is time consuming and requires expert knowledge. As a result, human labor is still more cost effective than automation for many tasks, such as most assembly tasks. Even though these tasks might appear monotonous and predictable, small tolerances and tiny variations between similar parts make it inadequate to just perform a series of accurate movements. Robotic assembly has been addressed in, \emph{e.g.}, \cite{bjorkelund2011integration}. Further, there is a trend toward manufacturing a given product in a smaller volume and for a shorter time, and then changing to a new one. These circumstances have motivated the following two research objectives:
\begin{enumerate}
\item Enable easier and faster robot programming;
\item Enable robots to take proper action with respect to their surroundings.
\end{enumerate}
In this thesis, research toward these objectives is presented. Paper~I mainly addresses Objective 1. More specifically, it is investigated how a human could correct a faulty movement performed by a robot, by demonstrating a desired behavior. Paper~II deals with online replanning of robot movements to handle unforeseen events, and hence Objective 2 is mainly addressed. In both Paper~I and Paper~II, robot motion is modeled by dynamical movement primitives (DMPs). The concept of DMPs is introduced in \cref{sec:dmps}.

It should be noted that the two objectives are partly overlapping. For instance, if a robot is able to replan with respect to its workspace, the robot programmer does not have to take all eventualities into account, which reduces the required programming work. Vice versa, intuitive means of robot programming could allow for a human to mediate suitable behavior, given certain states or events, to the robot. In Paper~III, it is investigated how robots could learn to detect force transients from previous experience, and use this as a decision basis. A closely related topic is force estimation and control, see, \emph{e.g.}, \cite{olsson2002force,gamez2006generalized,stolt2012force}. In Paper~III, explicit programming of the detection model is eschewed, to make the human--robot interaction as easy as possible for the human. Instead, a machine learning approach is used. How machine learning can promote robot programming is described in \cref{sec:machine_learning}.

\section{Thesis Outline}
This thesis consists mainly of three papers, and it is organized as follows. In \cref{ch:publications}, the publications authored or co-authored by the thesis author are listed. A discussion and ideas for continuation of the work are presented in \cref{ch:discussion}, and a conclusion is presented in \cref{ch:conclusion}.

The first part of the thesis consists of Paper~I and Paper~II, where the DMP concept is augmented to support corrective demonstrations and enhance replanning capabilities. In this part, demonstrations and robot motion control are mainly considered. The second part of the thesis consists of Paper~III, and in contrast to the first part, the main focus is not on motion control. Instead, the aim is to enable recognition of sensor data sequences. Despite this difference between the two parts, all three papers present research toward faster and more intuitive mediation of skills from humans to robots, and consider assembly scenarios in particular.

\section{Dynamical Movement Primitives}
\label{sec:dmps}
Representation and execution of movements is an important area within robotics. An industrial robot program commonly consists of a sequence of movement instructions, each containing some details that specify the movement, such as velocity and end point. Further, the ability to handle deviations from the planned movement is usually very low. Instead, some motion supervision algorithm would typically stop the robot if it would be too far from its position reference or experience too large joint torques, for instance due to some unexpected physical contact.

To enhance real-time motion modulation, DMPs have been proposed in \cite{ijspeert2002humanoid,schaal2003computational}. The concept has been inspired by the biological movement models presented in \cite{giszter1993convergent,mussa1999modular}. It was used for robotic learning from unstructured demonstrations in \cite{niekum2015learning}, and for object handover in \cite{prada2014handover}. Trajectory-based reinforcement learning has been applied to automatically tune DMP parameters in, \emph{e.g.}, \cite{pastor2013dynamic,kober2008learning,abu2015adaptation,kroemer2010combining}.
A DMP is a movement model, defined by a weakly nonlinear dynamical system with attractor behavior, so that the state converges to a desired end point. Although there are other alternatives, the most ubiquitous movement model in the DMP framework, which is the one that Paper~I and Paper~II proceed from, is based on the following damped-spring system. 

\begin{equation}
\tau^2 \ddot{y} = \alpha(\beta(g-y)-\tau \dot{y}) + f(x)
\end{equation}
Here, $y$ denotes robot position, $g$ is the goal position, and $\alpha$ and $\beta$ are positive constants chosen such that the system is critically damped for $f(x)=0$. Further, $f(x)$ is a learnable forcing term, with significant magnitude only in a finite time window, that allows for detours before reaching $g$. The evolution rate is scalable through the time parameter $\tau$, and explicit time dependence is avoided with the phase variable $x$. Once determined, a DMP can be used as a robot motion controller, by sending control signals so that the robot moves according to the evolution of $y$. As explained in \cite{ijspeert2013dynamical}, coupling terms are easily incorporated in the DMP framework while retaining its convergence properties, which facilitates online motion modulation with respect to the surroundings of the robot.

\subsection{Comparison with alternative movement representations}
Potential fields and splines are often brought up as two alternatives to DMPs for movement representation. Similar to DMPs, potential fields represent attractor landscapes, with convergence to a goal position and without explicit time dependence. Potential fields have been considered for robot control by many researchers, see, \emph{e.g.}, \cite{khatib1986real,koditschek1987exact,li1999passive}. Vector fields define the movement based on given positions, but determining the vector field given a desired behavior is not straight forward. Design of potential fields for some obstacle avoidance scenarios has been done in, \emph{e.g.}, \cite{koditschek1987exact}. Further, while DMPs allow for different control signals from the same position, this can not be achieved with conventional potential fields. 

For imitation learning, splines have been widely used. Splines are functions that are defined piecewise by polynomials, and retain smoothness where the polynomials connect. It has been shown in, \emph{e.g.}, \cite{miyamoto1996kendama,wada2004via} that demonstrated trajectories can be represented and successfully reproduced by means of splines. However, online replanning is not supported, and temporal and spatial scaling can be done only by recomputation of the spline polynomials.

\section{Supervised Machine Learning}
\label{sec:machine_learning}
Traditionally, computers and robots have been programmed by writing explicit code, specifying sets of rules and behaviors in detail. This works well in predictable scenarios, but most of the tasks that humans perform in their everyday life, are far too complex to mediate in such fashion. For instance, consider the task of distinguishing whether a certain image represents a car or a bicycle, which is in general easy for humans. Indeed, both categories could take different forms, and hand-crafting the rules for classification from raw image data would not be feasible.

It is better to address such problems with machine learning approaches. This field consists of two major parts; supervised and unsupervised learning, see \cite{bishop2007pattern,murphy2012machine}. In this thesis, supervised learning is considered. In general, supervised machine learning is used to approximate a given function, $y(x)$, with a parameterized function, $\hat{y}(x|\theta)$, which takes some input data $x$, and maps it to an output $\hat{y}$. Here, $\theta$ denotes the model parameters. In the example of image recognition, $x$ could be pixel values, $y$ would be the true image category, and $\hat{y}(x)$ could be interpreted as the probability distribution over the two categories, \emph{i.e.}, car and bicycle, given $x$.

In order to learn $\hat{y}(x)$, the model is exposed to a large data set of examples, called training data. In supervised learning, the training data consist of both input data and the corresponding known outputs, usually manually labeled. In the training phase, the elements of $\theta$ are adjusted to fit the training data by means of optimization. A loss function, $L$, in which some measurement of the error of $\hat{y}(x)$ compared to $y(x)$ is included, is minimized with respect to the model parameters.

Since the training data can only include a small subset of all possible data points, an important aspect of machine learning is generalization, \textit{i.e.}, to predict the output given input not used during training. In order to achieve this, the complexity of the model is typically restricted, by keeping the number of parameters low, or by penalizing the complexity by including it in $L$. Further, test data, not directly used to optimize the model parameters, are used to estimate how well models generalize. It is, however, common to determine some model hyperparameters based on the performance on test data. Therefore, it is good practice to use yet another data set to investigate the generalizability, without affecting the model in any way. Such data are called validation data.

This general approach is adopted in Paper~III, where the aim is to take a step toward more intuitive human--robot interaction. Ideally, a non-expert operator should be able to provide a robot with data, enabling it to learn from experience. Similar to the image classification example, a model is trained to determine the class of the data given $x$. In particular, $x$ consists of robot joint torques, and the task is to determine whether a certain force/torque transient, acting on the robot end-effector, is present or not. One important difference from the image recognition example is that $x$ consists of a time-series rather than a static representation, which should be taken into account when choosing the structure of the model. In Paper~III, a recurrent neural network (RNN) is used as classification model. Prior to the training phase, data are gathered by letting the robot experience the force/torque transient, while logging the joint torques. After that, the data sets are formed by labeling the data. Even though this implies some work by the operator, the required time and traditional programming skills can be reduced significantly with this method compared to explicitly programmed conditions for classification. Related approaches have been presented in \cite{rodriguez2010failure,rojas2012relative,stolt2015detection}, but then, the force/torque acting on the end-effector has been measured directly, which requires a force/torque sensor.


\section{Problem Formulation}
The first aim of this thesis is to answer the question of whether it is possible to automatically interpret a correction, made by an operator, of the last part of a robot trajectory generated by a DMP, while retaining the first part. The human--robot interaction must be intuitive, and the result of a correction predictable enough for its purpose. The result should be a new DMP, of which the first part behaves qualitatively as the first part of the original DMP, whereas the last part behaves according to the corrective demonstration. Discontinuities between the original and corrective trajectories must be mitigated.

Further, it should be investigated whether perturbations of trajectories generated by DMPs could be recovered from, while using control signals of moderate magnitudes only. In the absence of significant perturbations, the behavior should resemble that of the original DMP framework described in \cite{ijspeert2013dynamical}.

The above problems will be addressed by augmenting the original DMP framework. Meanwhile, the benefits of the DMP framework, \textit{i.e.}, scalability in time and space as well as guaranteed convergence to the goal $g$, should be preserved. 

It should also be investigated whether robot joint torques could be used to recognize contact force transients during robotic assembly, despite uncertainties introduced by, \emph{e.g.}, joint friction. Finally, it is desirable to explore how the performance of the detection algorithm is affected by the length of the joint torque sequences used as input.

\section{Thesis Contributions}
The main contributions of this thesis are:
\begin{itemize}
\item A framework for modification of DMPs by means of corrective demonstrations;
\item An augmentation of the DMP framework that enables recovery from perturbations during DMP execution;
\item A machine learning procedure for detecting force/torque transients acting on a robot end-effector, by measuring the robot joint torques.
\end{itemize}


%% file: chapters/publications/publications.tex
\chapter{Publications}
\label{ch:publications}
This licentiate thesis is based on the following three papers.

\section*{Paper I}
\begin{refsection}
    \nocite{karlsson2017autonomous}
    \printbibliography[heading=none]
\end{refsection}
In this publication, M.~Karlsson formulated the method for updating a partly faulty trajectory representation, based on a corrective demonstration. Further, M.~Karlsson implemented the method and verified it experimentally. A.~Robertsson and R.~Johansson contributed with comments on the research and the manuscript. 

\section*{Paper II}
\begin{refsection}
    \nocite{karlsson2017dmp}
    \printbibliography[heading=none]
\end{refsection}
M.~Karlsson and F.~Bagge Carlson identified the necessity of augmenting the existing DMP framework, to make related research approaches on DMP perturbation recovery practically realizable. M. Karlsson formulated the augmentation, and verified it in simulations and experimentally, while frequently discussing the work with F.~Bagge Carlson. Further, F.~Bagge Carlson implemented the method presented as an open-source Julia package, which can be found on \cite{dmp_julia}. Example code in Matlab, written by M.~Karlsson, can be found on \cite{dmp_matlab}. Throughout the work, A.~Robertsson and R.~Johansson supervised the research and assisted in structuring the manuscript.

\section*{Paper III}
\begin{refsection}
    \nocite{karlsson2017detection}
    \printbibliography[heading=none]
\end{refsection}
M. Karlsson formulated the transient detection method, implemented it, and performed the experimental work. A.~Robertsson and R.~Johansson contributed with insights regarding related work, provided comments on the research, and assisted in structuring the manuscript.\\ \\ 

The following publications, authored or co-authored by the author of this thesis, cover topics in robotics, nonlinear state estimation, and positioning. They are, however, not included in this thesis.

\begin{refsection}
    \nocite{karlsson2016fswjournal}
    \nocite{wadenback2017visual}
    \nocite{karlsson2016fsw}
    \nocite{bagge2016particle}
    \nocite{karlsson2016cooperative}
    \nocite{haage2016cognitive}
    \nocite{karlsson2015sensor}
    \printbibliography[heading=none]
\end{refsection}

%% file: chapters/discussion/discussion.tex
\chapter{Discussion and Future Work}
\label{ch:discussion}
Much more work is still required, both in terms of research and engineering, before robot programming could be considered non-problematic. In the future, one natural way to extend the DMP functionality presented here, would be to incorporate trajectory-based learning of suitable actions based on sensor data. Such learning could be warm-started by defining DMP parameters from one initial demonstration. Under the DMP execution, the robot could then deviate from the demonstrated trajectory, based on sensor feedback. Further, albeit the control algorithm in Paper~II worked satisfactorily in the simulations and experiments, and some stability properties were addressed, it remains to construct a formal proof of convergence to the goal state. 

It would be reasonable to expect that the method for force/torque transient recognition in Paper~III would work well for other tasks than that presented, since no assumptions were made in the method design regarding task, parts to be assembled, \emph{etc}., except that a transient would be generated. However, it remains to evaluate the approach experimentally on various tasks, \emph{e.g.}, assembly of other parts, to verify robustness and generalizability. Further, the data acquisition and labeling is still a bottleneck in this approach, requiring half a working day for the thesis author per task to learn. A natural continuation would therefore be to automatize the work flow as much as possible. For instance, the operator could be aided by a GUI, showing plots of the robot joint torques, and asking the operator to indicate where the transients occur. Perhaps, the operator could be provided with suggestions, already after labeling a few examples, by forming a preliminary, less complex, detection model. 

The movement before and after the transient, \emph{i.e.}, moving down in the scenario in Paper~III, does not necessarily take much time to implement, but requires traditional coding with the current setup. It would be more accessible, if it could be demonstrated by, \emph{e.g.}, lead-through programming. However, physical contact with the robot arm would affect the joint torques, and by that corrupting the training data and test data. A possible solution for this has been presented in \cite{mahdi2016trajectory}, where one of the two robot arms has been used for lead-through tele-operation of the other. That approach would allow one robot arm to perform the task, without physical contact with the operator, according to the operator's demonstration on the other arm. Ideally, from such demonstrations, the robot should not only learn recognition of transients, but also desired actions before and after these.  It would therefore be valuable to integrate the approach in \cite{mahdi2016trajectory} with that in Paper~III.

%% file: chapters/conclusion/conclusion.tex
\chapter{Conclusion}
\label{ch:conclusion}

The aim of the research presented in this thesis was to facilitate robot programming by enhancing the ability of robots to learn from demonstrations and from experience. More specifically, the DMP framework described in \cite{ijspeert2013dynamical} was augmented with two new functionalities. First, an algorithm was developed that allowed an operator to correct the last part of a faulty trajectory generated by a DMP, while retaining the first part. The correction could be done in an intuitive way by demonstrating a corrective trajectory. The first part of the resulting DMP behaved as the first part of the original DMP, and the last part behaved according to the correction. Discontinuities between the first and last parts were eschewed by formulating and solving a convex optimization problem. Secondly, a control algorithm that enabled trajectory tracking and perturbation recovery during execution of DMPs was presented. The algorithm was based on a combination of two-degree-of-freedom control and temporal coupling. Since the required control signals were of moderate magnitude only, the controller was practically realizable, which was the main benefit compared to state of the art. In the absence of perturbations, the controller behaved like the original DMP framework.

Further, a machine learning procedure was presented for model-based detection of force/torque transients acting on the robot end-effector without direct force/torque measurement. Instead, robot joint torque sequences were used as model inputs. Therefore, a force/torque sensor was not required, which was the main benefit as compared to previous research. A systematic approach for choosing a suitable length of the input sequences was presented. An RNN was used as classification model for the detection. The approach presented seems promising, since the resulting model showed high performance for the test data as well as during the experiments.

%% file: chapters/paper1/paper1.tex
    \paper[Autonomous Interpretation of Demonstrations...]{Autonomous Interpretation of Demonstrations for Modification of Dynamical Movement Primitives}
    \authors{Martin Karlsson \and Anders Robertsson \and Rolf Johansson}
\begin{abstract}
The concept of dynamical movement primitives (DMPs) has become popular for modeling of motion, commonly applied to robots. This paper presents a framework that allows a robot operator to adjust DMPs in an intuitive way. Given a generated trajectory with a faulty last part, the operator can use lead-through programming to demonstrate a corrective trajectory. A modified DMP is formed, based on the first part of the faulty trajectory and the last part of the corrective one. A real-time application is presented and verified experimentally.
\end{abstract}
    \vfill
    Originally published in the 2017 IEEE International Conference on Robotics and Automation (ICRA), May 29--June 3, Singapore. Reprinted with permission.
    \newpage

\section{Introduction}
High cost for time-consuming robot programming, performed by engineers, has become a key obstruction in industrial manufacturing. This has promoted the research toward faster and more intuitive means of robot programming, such as learning from demonstration, to which an introduction is presented in~\cite{argall2009survey}. It is in this context desirable to make robot teaching available to a broader group of practitioners by minimizing the engineering work required during teaching of tasks.

A costumary way to quickly mediate tasks to robots is to use lead-through programming, while saving trajectory data so that the robot can reproduce the motion. In this paper, the data are used to form dynamical movement primitives (DMPs). Early versions of these were presented in \cite{ijspeert2002humanoid}, \cite{schaal2003computational} and \cite{ijspeert2003learning}, and put into context in \cite{niekum2015learning}. Uncomplicated modification for varying tasks was emphasized in this literature. For example, the time scale was governed by one parameter, which could be adjusted to fit the purpose. Further, the desired final state could be adjusted, to represent a motion similar to the original one but to a different goal. DMPs applied on object handover with moving targets were addressed in \cite{prada2014handover}. The scalability in space was demonstrated in, \emph{e.g.}, \cite{ijspeert2013dynamical}.

\begin{figure}
	\centering
	\includegraphics[width=\columnwidth]{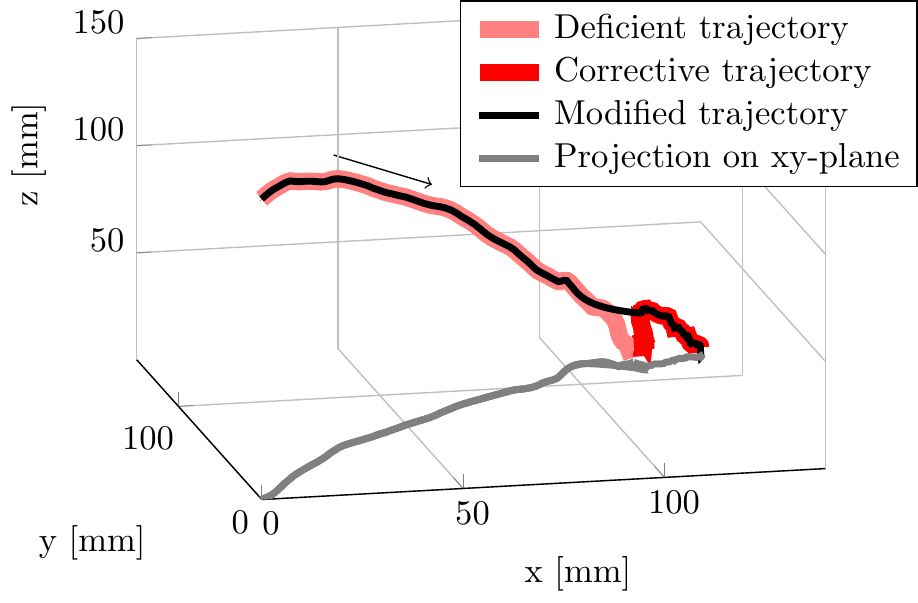}
	\caption{Trajectories of the robot's end-effector from one of the experiments. The arrow indicates the motion direction. The deficient trajectory was generated from the original DMP. After that, the operator demonstrated the corrective trajectory. Merging of these, resulted in the modified trajectory. The projection on the $xy$-plane is only to facilitate the visualization.} \label{fig:adjust_cart}
\end{figure}

The scenario considered in this paper is the unfavorable event that the last part of the motion generated by a certain DMP is unsatisfactory. There might be several reasons for this to occur. In the case where the starting points differ, the generated trajectory would still converge to the demonstrated end point, but take a modified path, where the modification would be larger for larger differences between the starting points. Further, the DMP might have been created in a slightly different setup, \emph{e.g.}, for a different robot or robot cell. There might also have been a mistake in the teaching that the operator would have to undo. If the complete last part of the trajectory is of interest, it is not enough to modify the goal state only. One way to solve the problem would be to record an entirely new trajectory, and then construct a corresponding DMP. However, this would be unnecessarily time consuming for the operator, as only the last part of the trajectory has to be modified. Instead, the method described here allows the operator to lead the manipulator backwards, approximately along the part of the trajectory that should be adjusted, followed by a desired trajectory, as visualized in \cref{fig:adjust_cart}.

Hitherto, DMPs have usually been formed by demonstrations to get close to the desired behavior, followed by trajectory-based reinforcement learning, as presented in, \emph{e.g.}, \cite{pastor2013dynamic,kober2008learning,abu2015adaptation,kroemer2010combining}. Compared to such refinements, the modification presented here is less time consuming and does not require engineering work. On the other hand, the previous work on reinforcement learning offers modulation based on sensor data, and finer  movement adjustment. Therefore, the framework presented in this paper forms an intermediate step, where, if necessary, a DMP is modified to prepare for reinforcement learning, see \cref{fig:workflow}. This modification can be used within a wide range of tasks. In this paper, we exemplify by focusing on peg-in-hole tasks.

In \cite{pastor2013dynamic}, online modulation, such as obstacle avoidance, was implemented for DMPs. This approach has been verified for several realistic scenarios, but requires an infrastructure for obstacle detection, as well as some coupling term parameters to be defined. It preserves convergence to the goal point, but since the path to get there is modified by the obstacle avoidance, it is not guaranteed to follow any specific trajectory to the goal. This is significant for, \emph{e.g.}, a peg-in-hole task.

The paper is outlined as follows. Two example scenarios in which the framework would be useful are presented in \cref{sec:motivexamples}, followed by a description of the method in \cref{sec:framework}. Experimental setup and results are described in \cref{sec:experiments,sec:results}, and finally a discussion and concluding remarks are presented in \cref{sec:discussion,sec:conclusions}, respectively.

\section{Problem Formulation}
\label{sec:problemformulation}
In this paper, we address the question whether it is possible to automatically interpret a correction, made by an operator, of the last part of a DMP trajectory, while still taking advantage of the first part. The human--robot interaction must be intuitive, and the result of a correction predictable enough for its purpose. The correction should result in a new DMP, of which the first part behaves qualitatively as the first part of the original DMP, whereas the last part resembles the last part of the corrective trajectory. Any discontinuity between the original and corrective trajectories must be mitigated.



\section{Motivating Examples}
\label{sec:motivexamples}
We here describe two scenarios where the framework proves useful. These are evaluated in \cref{sec:experiments,sec:results}, where more details also are given.

\subsection{Inadequate precision -- Scenario A}
\label{sec:inadequate}
\begin{figure}
\begin{minipage}{.48\columnwidth}
\label{fig:pm1}
\centering
\includegraphics[height=4cm]{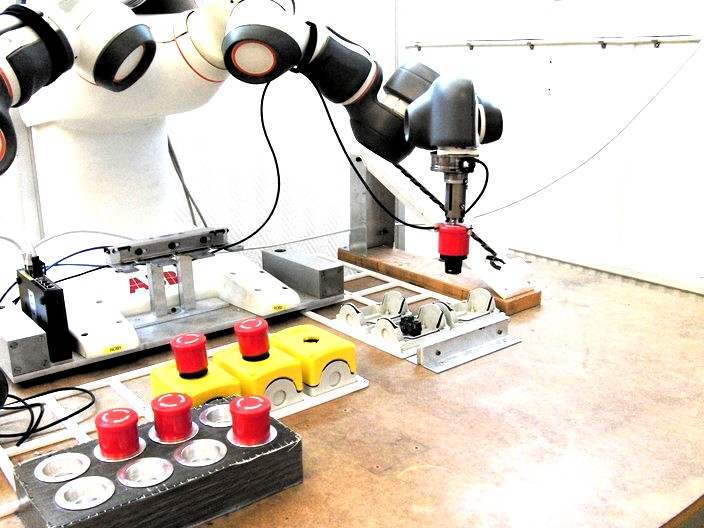}
\subcaption{}
\vspace{2mm}
\end{minipage}
\hfill
\begin{minipage}{.48\columnwidth}
\label{fig:pm2}
\centering
\includegraphics[height=4cm]{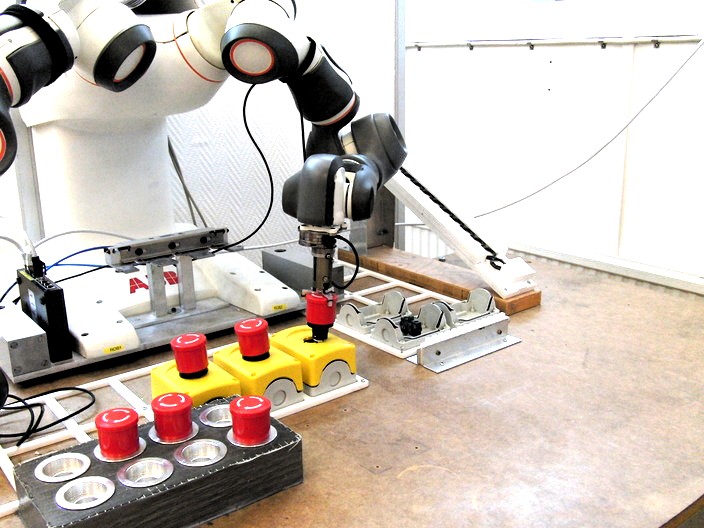}
\subcaption{}
\vspace{2mm}
\end{minipage}
\begin{minipage}{.48\columnwidth}
\label{fig:pm3}
\centering
\includegraphics[height=4cm]{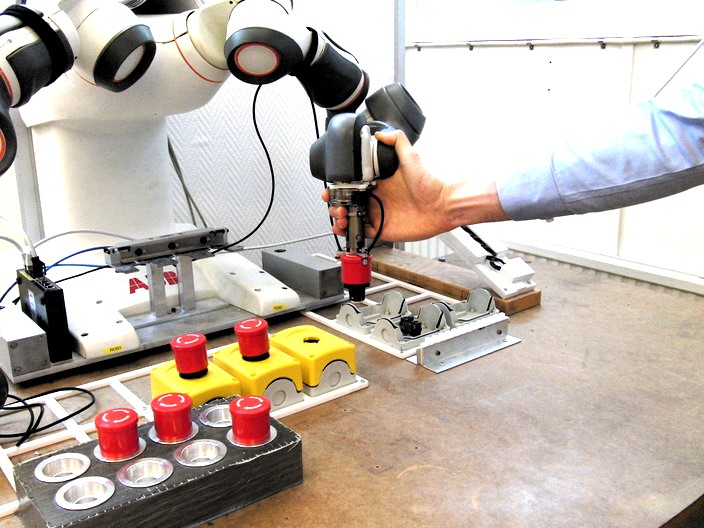}
\subcaption{}
\end{minipage}
\hfill
\begin{minipage}{.48\columnwidth}
\label{fig:pm4}
\centering
\includegraphics[height=4cm]{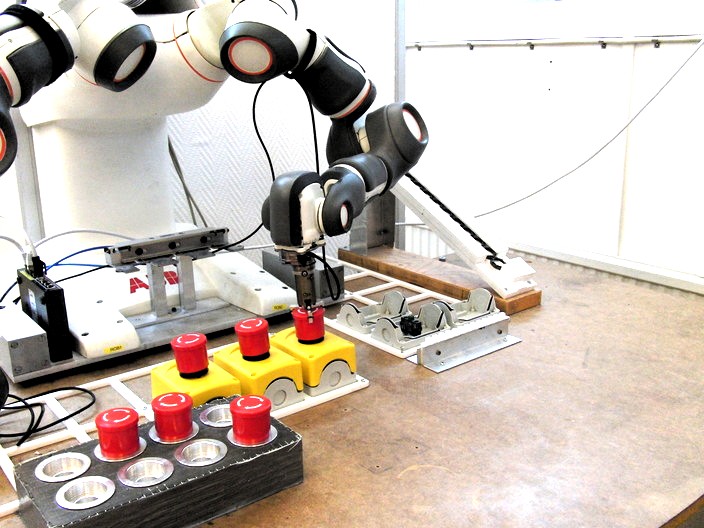}
\subcaption{}
\end{minipage}
\caption{Scenario A. The evaluation started in (a), and in (b) the robot failed to place the button in the hole due to inadequate accuracy. Between (a) and (b), the deficient trajectory was recorded. The operator led the robot arm backwards (c), approximately along a proportion of the deficient trajectory, and subsequently led it to place the button properly, while the corrective trajectory was recorded. The robot then made the entire motion, starting in a configuration similar to that in (a), and ending as displayed in (d).}
\label{fig:pm}
\end{figure}

Consider the setup shown in \cref{fig:pm}, where the button should be placed into the yellow case. A DMP was run for this purpose, but, due to any of the reasons described above, the movement was not precise enough, and the robot got stuck on its way to the target. Hitherto, such a severe shortcoming would have motivated the operator to teach a completely new DMP, and erase the old one. With the method proposed in this paper, the operator had the opportunity to approve the first part of the trajectory, and only had to modify the last part. This was done by leading the robot arm backwards, approximately along the faulty path, until it reached the acceptable part. Then, the operator continued to lead the arm along the desired path to the goal. When this was done, the acceptable part of the first trajectory was merged with the last part of the corrective trajectory. After that, a DMP was fitted to the resulting trajectory. Compared to just updating the target point, this approach also allowed the operator to determine the trajectory leading there. This scenario is referred to as Scenario~A.

\subsection{New obstacle -- Scenario B}
\label{sec:novelobstacle}
For the setup in \cref{fig:as}, there existed a DMP for moving the robot arm from the right, above the button that was already inserted, to a position just above the hole in the leftmost yellow case. However, under the evaluation the operator realized that there would have been a collision if a button were already placed in the case in the middle. A likely reason for this to happen would be that the DMP was created in a slightly different scene, where the potential obstacle was not taken into account. Further, the operator desired to extend the movement to complete the peg-in-hole task, rather than stopping above the hole. With the method described herein, the action of the operator would be similar to that described in~\cref{sec:inadequate}, again saving work compared to previous methods. This scenario is referred to as Scenario~B.

\section{Description of the Framework}
\label{sec:framework}

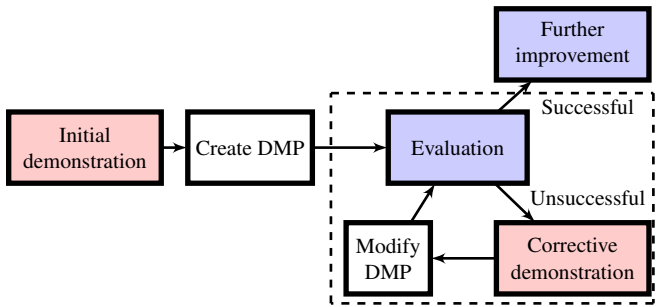
\begin{figure}
	\centering
    \input{chapters/paper1/figs/cascade_scheme.tex}
    \caption{Schematic visualization of the work flow, from an operator's perspective. A DMP was created based on a demonstration. Subsequently, the DMP was executed while evaluated by the operator. If unsuccessful, the operator demonstrated a correction, which yielded a modified DMP to be evaluated. Once successful, further improvement could be done by, \emph{e.g.}, trajectory-based reinforcement learning, though that was outside the scope of this work. Steps that required direct, continuous interaction by the operator are marked with light {\color{red} red} color. Steps that required some attention, such as supervision and initialization, are marked with light {\color{blue} blue}. The operations in the white boxes were done by the software in negligible computation time, and required no human involvement. The work in this paper focused on the steps within the dashed rectangle.}
\label{fig:workflow}
\end{figure}

In this section, the concept of DMPs is introduced. A method to determine what parts of the deficient and corrective trajectories to retain is presented, followed by a description of how these should be merged to avoid discontinuities. Finally, some implementation aspects are addressed. \Cref{fig:workflow} displays a schematic overview of the work flow of the application, from the user's perspective.

\subsection{Dynamical movement primitives}
A review of the DMP concept was presented in~\cite{ijspeert2013dynamical}, and here follows a short description of how it was applied in this paper. A certain trajectory, $y$, was modeled by the system
\begin{equation}
\tau \dot{y} = z
\label{eq:getting_ydot}
\end{equation}
where $z$ is determined by
\begin{equation}
\tau \dot{z} = \alpha_z(\beta_z(g-y)-z) + f(x)
\label{eq:getting_zdot}
\end{equation}
In turn, $f(x)$ is a function given by
\begin{equation}
f(x) = \frac{\sum_{i=1}^{N_b} w_i\Psi_i(x)}{\sum_{i=1}^{N_b} \Psi_i(x)} x \cdot (g-y_0)
\end{equation}
where the basis functions, $\Psi_i(x)$, take the form
\begin{align}
\Psi_i(x) &= \exp \left(-\frac{1}{2\sigma_i^2}(x-c_i)^2 \right) \\
\tau \dot{x} &= -\alpha_x x
\end{align}
Here, $\tau$ is a time constant, while $\alpha_z$, $\beta_z$, and $\alpha_x$ are positive constants. Further, $N_b$ is the number of basis functions, $w_i$ is the weight for basis function $i$, $y_0$ is the starting point of the trajectory $y$, and $g$ is the goal state; $\sigma_i$ and $c_i$ are the width and center of each basis function, respectively. Given a DMP, a robot trajectory can be generated from \cref{eq:getting_ydot,eq:getting_zdot}. Vice versa, given a demonstrated trajectory, $y_{\text{demo}}$, a corresponding DMP can be formed; $g$ is then given by the end position of $y_{\text{demo}}$, whereas $\tau$ can be set to get a desired time scale. Further, the solution of a weighted linear regression problem in the sampled domain yields the weights

\begin{equation}
w_i = \frac{\boldsymbol{s}^T \boldsymbol{\Gamma}_i \boldsymbol{f}_{\text{target}}}{\boldsymbol{s}^T \boldsymbol{\Gamma}_i \boldsymbol{s}}
\end{equation}
where
\begin{gather}
\boldsymbol{s} = 
\begin{pmatrix}
x^1 (g - y_{\text{demo}}^1) \\
x^2 (g - y_{\text{demo}}^1) \\
\vdots \\
x^{N} (g - y_{\text{demo}}^1)
\end{pmatrix}, \quad
\boldsymbol{\Gamma_i} = \text{diag}(\Psi_i^1, \Psi_i^2 \cdots \Psi_i^N) \\
\boldsymbol{f}_{\text{target}} = \tau^2 \ddot{y}_{\text{demo}} - a_z(b_z(g-y_{\text{demo}}) - \tau \dot{y}_{\text{demo}}) =
\begin{pmatrix}
f_{\text{target}}^1 \\
f_{\text{target}}^2 \\
\vdots \\
f_{\text{target}}^N
\end{pmatrix}
\end{gather}
Here, $N$ is the number of samples in the demonstrated trajectory.

\begin{figure}
\begin{minipage}{.48\columnwidth}
\label{fig:as1}
\centering
\includegraphics[height=37mm]{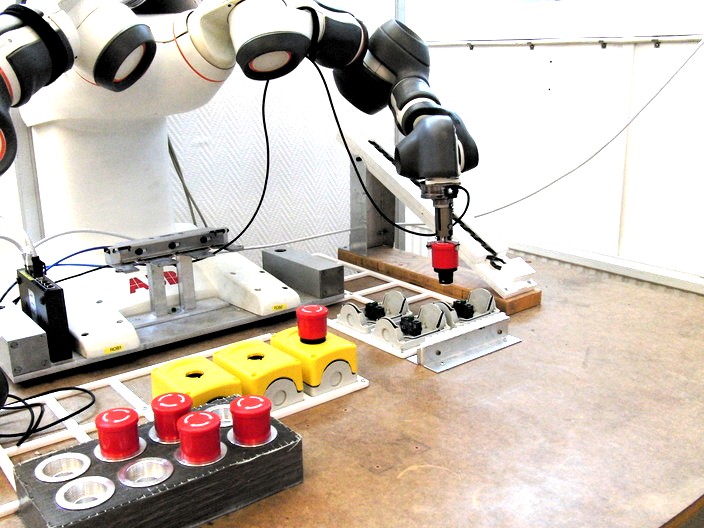}
\subcaption{}
\vspace{2mm}
\end{minipage}
\hfill
\begin{minipage}{.48\columnwidth}
\label{fig:as2}
\centering
\includegraphics[height=37mm]{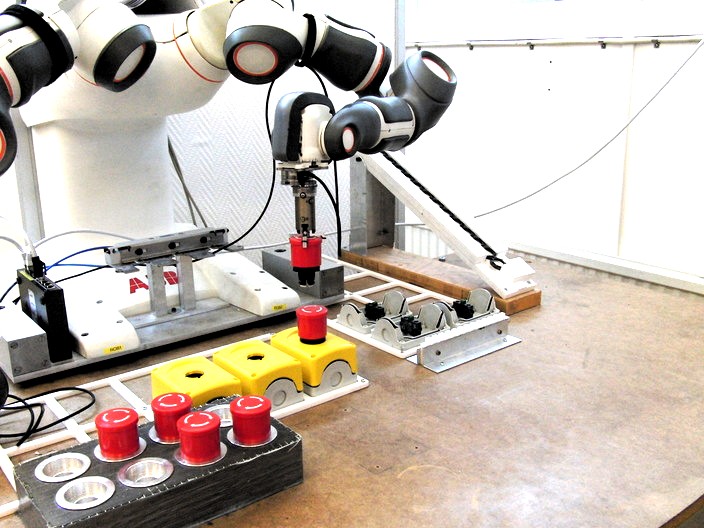}
\subcaption{}
\vspace{2mm}
\end{minipage}
\begin{minipage}{.48\columnwidth}
\label{fig:as3}
\centering
\includegraphics[height=37mm]{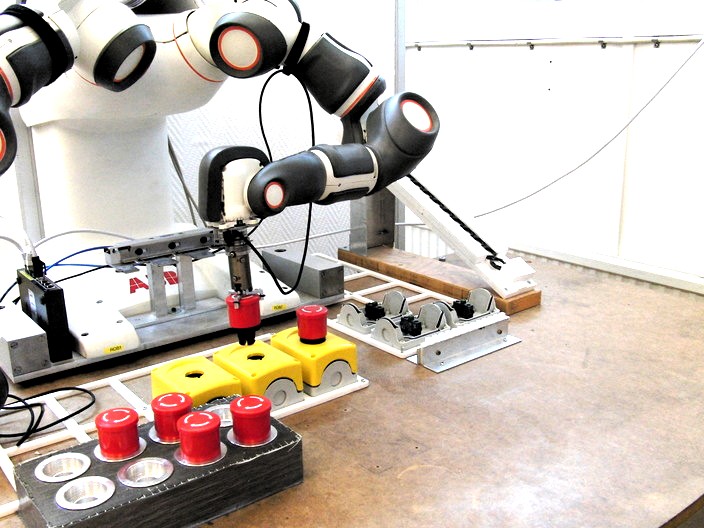}
\subcaption{}
\vspace{2mm}
\end{minipage}
\hfill
\begin{minipage}{.48\columnwidth}
\label{fig:as4}
\centering
\includegraphics[height=37mm]{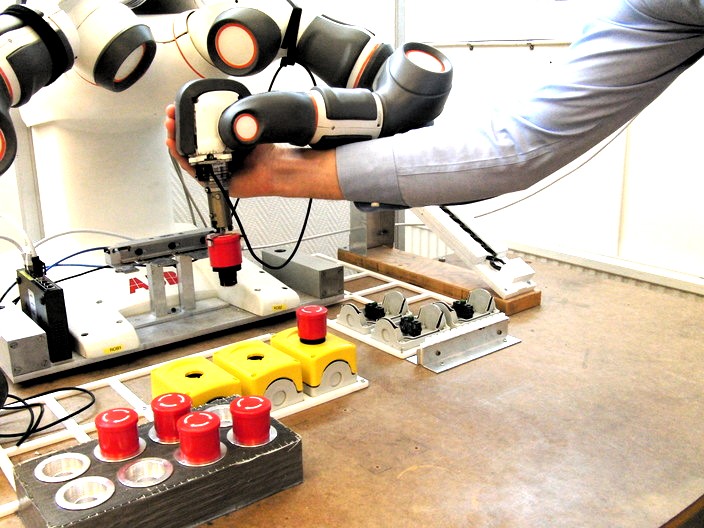}
\subcaption{}
\vspace{2mm}
\end{minipage}
\begin{minipage}{.48\columnwidth}
\label{fig:as5}
\centering
\includegraphics[height=37mm]{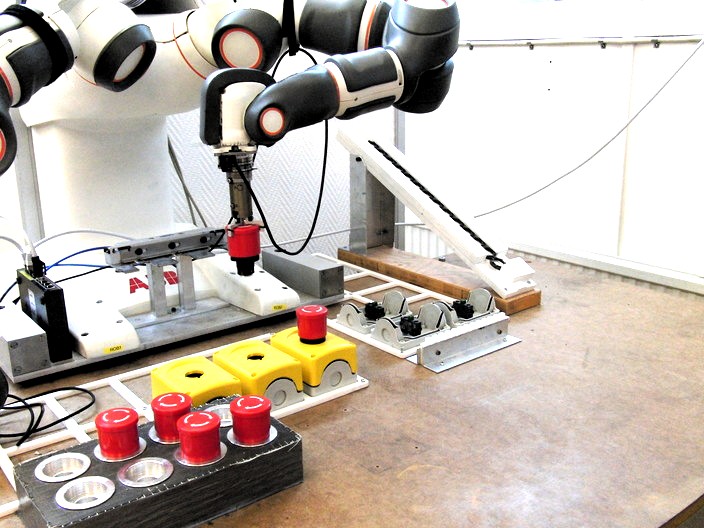}
\subcaption{}
\vspace{2mm}
\end{minipage}
\hfill
\begin{minipage}{.48\columnwidth}
\label{fig:as6}
\centering
\includegraphics[height=37mm]{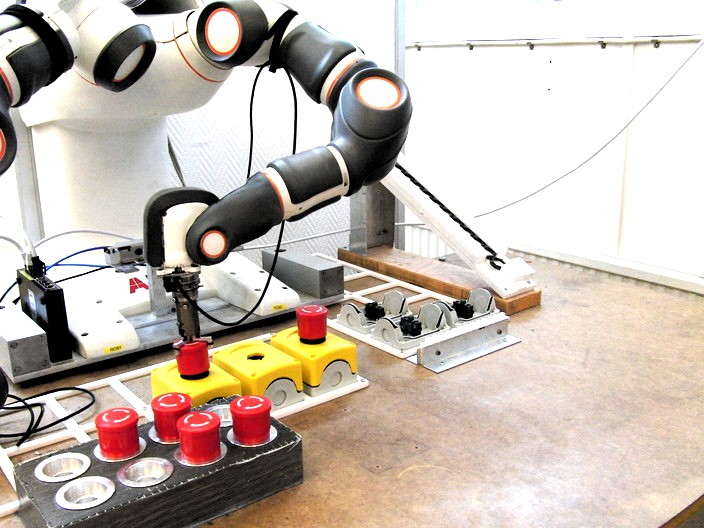}
\subcaption{}
\vspace{2mm}
\end{minipage}
	\caption{Scenario B. The initial goal was to move the button to the leftmost yellow case, above the hole, to prepare for placement. The evaluation started in (a), and in (b) the trajectory was satisfactory as the placed button was avoided. In (c), however, there would have been a collision if there was a button placed in the middle case. Further, it was desired to complete the peg-in-hole task, rather than stopping above the hole. Hence, the evaluated trajectory was considered deficient. In (d), the operator led the robot arm back, and then in a motion above the potential obstacle, and into the hole, forming the corrective trajectory. Based on the modified DMP, the robot started in a position similar to that in (a), avoided the potential obstacle in (e) and reached the new target in (f).} 
\label{fig:as}
\end{figure}

\subsection{Interpretation of corrective demonstration}
\label{sec:interpret}
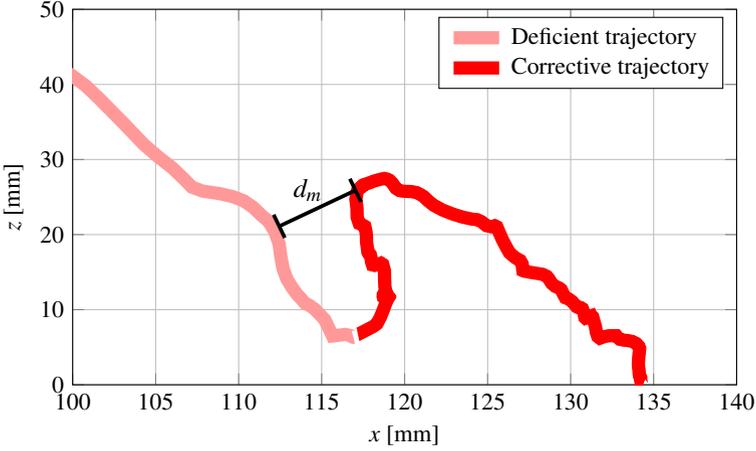
\begin{figure}
	\centering	
	\setlength{\figurewidth}{0.75\linewidth}
	\setlength{\figureheight}{5cm}
	\small
	\input{chapters/paper1/figs/adjust_notes.tex}
	\caption{Visualization of shortest distance, here denoted $d_m$, used to determine the left separation marker in Fig. \ref{fig:adjust_zoom}. The trajectories are the same as in Figs. \ref{fig:adjust_cart} and \ref{fig:adjust_zoom}, except that the modified trajectory is omitted. }
\label{fig:adjust_notes}
\end{figure}

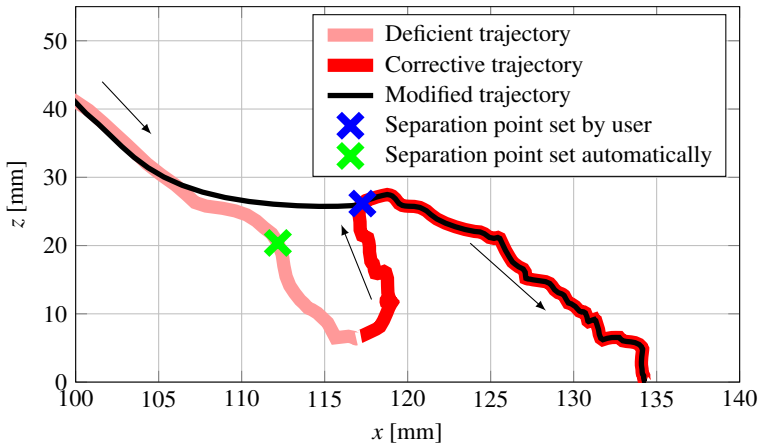
\begin{figure}
	\centering	
	\setlength{\figurewidth}{0.75\linewidth}
	\setlength{\figureheight}{5cm}
	\small
	\input{chapters/paper1/figs/adjust_zoom.tex}
	\caption{Same trajectories as in Fig. \ref{fig:adjust_cart}, but zoomed in on the corrective trajectory. Arrows indicate directions. The parts of the trajectories between the separation markers were not retained. The right, {\color{blue} blue}, separation point was determined explicitly by the operator during the corrective demonstration. The left, {\color{green} green}, separation point was determined according to \cref{eq:mindist}. Further, what was left of the deficient trajectory was modified for a smooth transition. However, the part of the corrective trajectory retained was not modified, since it was desired to closely follow this part of the demonstration. Note that the trajectories retained were not intended for direct play-back execution. Instead, they were used to form a modified DMP, which in turn generated a resulting trajectory, as shown in Figs.~\ref{fig:place1}, \ref{fig:place2} and \ref{fig:avoid1}.}  
\label{fig:adjust_zoom}
\end{figure}

If the evaluation of a trajectory was unsuccessful, a corrective demonstration and DMP modification should follow, as in \cref{fig:workflow}. Denote by $y_{d}$ the deficient trajectory, and by $y_{c}$ the corrective one, of which examples are shown in Figs.~\ref{fig:adjust_cart}, \ref{fig:adjust_notes}, and \ref{fig:adjust_zoom}. A trajectory formed by simply appending $y_{c}$ to $y_{d}$ was likely to take an unnecessary detour. Thus, only the first part of $y_{d}$ and the last part of $y_{c}$ were retained. This is illustrated in \cref{fig:adjust_zoom}. Denote by $y_{cr}$ the retained part of the corrective trajectory. The operator signaled where to separate the corrective trajectory, during the corrective demonstration. In the current implementation, this was done by pressing a button in a terminal user interface, when the robot configuration corresponded to the desired starting point of $y_{cr}$, denoted $y_{cr}^{1}$.

The next step was to determine which part of $y_{d}$ to retain. This was chosen as the part previous to the sample of $y_{d}$ that was closest to $y_{cr}^{1}$, \emph{i.e.},
\newcommand{\argmin}{\operatornamewithlimits{argmin}}
\begin{align}
y_{dr}^{m} &= y_{d}^{m}, \hspace{4mm} \forall m \in [1;M]
\intertext{where}
M &= \argmin\limits_{k = 1\dots K} \hspace{1mm} \dis(y_{d}^{k}, y_{cr}^1)
\label{eq:mindist}
\end{align}
Here, $\dis$ denotes distance, and $K$ is the number of samples in $y_d$, see \cref{fig:adjust_notes} for an illustration. The approach of using the shortest distance as a criterion, was motivated by the assumption that the operator led the robot arm back, approximately along the deficient trajectory, until the part that was satisfactory. At this point, the operator separated the corrective demonstration, thus defining $y_{cr}^{1}$ (see right marker in \cref{fig:adjust_zoom}). By removing parts of the demonstrated trajectories, a significant discontinuity between the remaining parts was introduced. In order to counteract this, $y_{dr}$ was modified into $y_{m}$, of which the following features were desired:
\begin{itemize}
\item $y_{m}$ should follow $y_{dr}$ approximately;
\item The curvature of $y_{m}$ should be moderate;
\item $y_{m}$ should end where $y_{cr}$ began, with the same movement direction in this point.
\end{itemize}
To find a suitable trade-off between these objectives, the following convex optimization problem was formulated and subsequently solved:
\begin{alignat}{2}
& \underset{y_{m}}{\text{minimize}}
&\quad &\lVert y_{dr} - y_{m} \rVert_2 + \lambda 
\lVert T_{(\Delta^2)} y_{m} \rVert_2 \label{eq:opt1}\\
& \text{subject to}
& & y_{m}^{M} = y_{cr}^1 \label{eq:opt2}\\
& & & y_{m}^{M} - y_{m}^{M-1} = y_{cr}^2 - y_{cr}^1 \label{eq:opt3}
\end{alignat}
Here, $\lambda$ denotes a constant scalar, and $T_{(\Delta^2)}$ is a second-order finite difference operator. Thereafter, $y_{cr}$ was appended on $y_{m}$, and one corresponding DMP was created, with the method described in the previous subsection. The next step in the work flow was to evaluate the resulting DMP, as shown in \cref{fig:workflow}. 
\subsection{Software implementation}
\label{sec:implementation}
The research interface ExtCtrl \cite{blomdell2005extending,blomdell2010flexible}, was used to send references to the low-level robot joint controller in the ABB IRC5 system \cite{irc5}, at 250 Hz. Most of the programming was done in C++, where DMPs were stored as objects. Among the data members of this class were the parameters $\tau$, $g$ and $w_{1 \dots N_b}$, as well as some description of the context of the DMP and when it was created. It contained member functions for displaying the parameters, and for modifying $g$ and $\tau$. The communication between the C++ program and ExtCtrl was handled by the LabComm protocol \cite{labcomm}. The C++ linear algebra library Armadillo \cite{sanderson2010armadillo} was used in a major part of the implementation. Further, the code generator CVXGEN \cite{mattingley2012cvxgen} was used to generate C code for solving the optimization problem in \cref{eq:opt1,eq:opt2,eq:opt3}. By default, the solver code was optimized with respect to computation time. This resulted in a real-time application, in which the computation times were negligible in teaching scenarios. The optimization problem was typically solved well below one millisecond on an ordinary PC.

\section{Experiments}
\label{sec:experiments}
\begin{figure}
\includegraphics[width=\columnwidth]{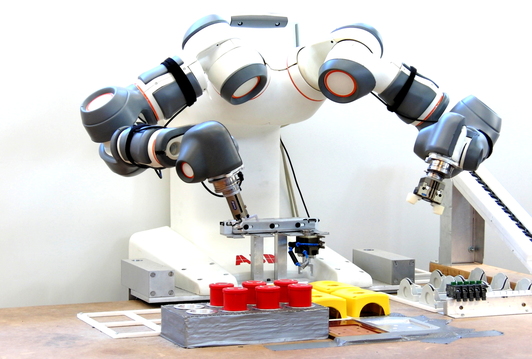}
\caption{The ABB YuMi \cite{yumi} prototype robot used in the experiments. }\label{fig:yumi5}
\end{figure}

The robot used in the experimental setup was a prototype of the dual-arm ABB YuMi \cite{yumi} (previously under the name FRIDA) robot, with 7 joints per arm, see \cref{fig:yumi5}. The experiments were performed in real-time using the implementation described in \cref{sec:implementation}. The computations took place in joint space, and the robot's forward kinematics were used for visualization in Cartesian space in the figures presented. The scenarios in \cref{sec:motivexamples} were used to evaluate the proposed method. For each trial, the following steps were taken:
\begin{itemize}
\item An initial trajectory was taught, deliberately failing to meet the requirements, as explained in \cref{sec:motivexamples};
\item Based on this, a DMP was created;
\item The DMP was used to generate a trajectory similar to the initial one. This formed the deficient trajectory;
\item A corrective trajectory was recorded;
\item Based on the correction, a resulting DMP was formed automatically;
\item The resulting DMP was executed for experimental evaluation.
\end{itemize}

First, Scenario A was set up for evaluation, see \cref{sec:inadequate} and \cref{fig:pm}. The scenario started with execution of a deficient trajectory. For each attempt, a new deficient trajectory was created and modified. A total of 50 attempts were made. 

Similarly, Scenario B (see \cref{sec:novelobstacle} and \cref{fig:as}) was set up, and again, a total of 50 attempts were made.

A video is available as a publication attachment, to facilitate understanding of the experimental setup and results. A version with higher resolution is available on \cite{mod_youtube}.

\section{Results}
\label{sec:results}
For each attempt of Scenario A, the robot was able to place the button properly in the yellow case after the modification. Results from two of these attempts are shown in Figs.~\ref{fig:place1} and \ref{fig:place2}. In the first case, the deficient trajectory went past the goal, whereas in the second case, it did not reach far enough.

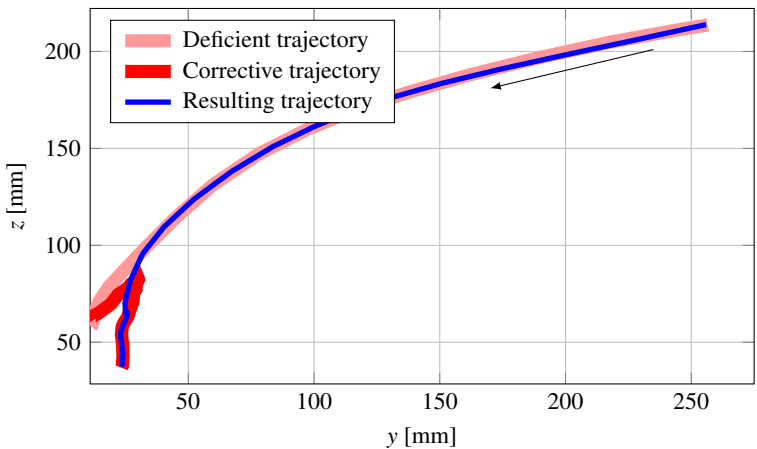
\begin{figure}[b]
	\centering	
	\setlength{\figurewidth}{0.75\linewidth}
	\setlength{\figureheight}{5cm}
	\input{chapters/paper1/figs/place_a1.tex}
	\caption{Trajectories from the experimental evaluation of Scenario A. The deficient trajectory went past the goal in the negative $y$-direction,  preventing the robot from lowering the button into the hole. After correction, the robot was able to reach the target as the modified DMP generated the resulting trajectory.} \label{fig:place1}
\end{figure}
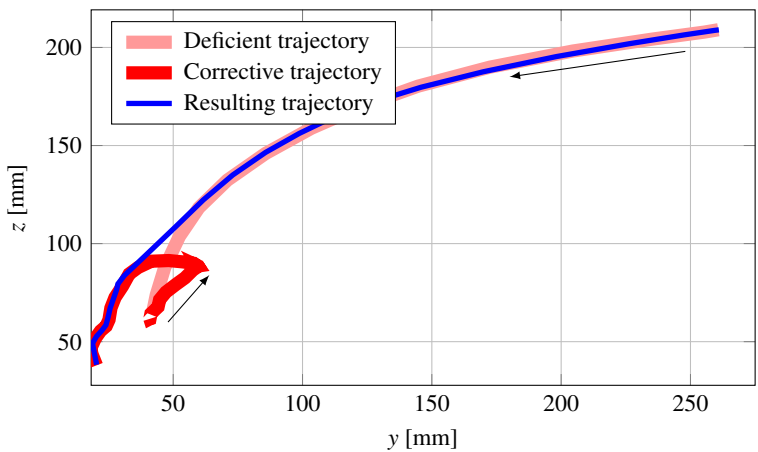
\begin{figure}
	\centering	
	\setlength{\figurewidth}{0.75\linewidth}
	\setlength{\figureheight}{5cm}
	\input{chapters/paper1/figs/place_a2.tex}
	\caption{Similar to Fig. \ref{fig:place1}, except that in this case, the deficient trajectory did not reach far enough in the negative $y$-direction.} \label{fig:place2}
\end{figure}

\begin{figure}
	\centering	
	\setlength{\figurewidth}{0.75\linewidth}
	\setlength{\figureheight}{5cm}
	\input{chapters/paper1/figs/avoid_a1.tex}
	\caption{Trajectories from experimental evaluation of Scenario B. The deficient trajectory was lowered too early, causing a potential collision. After the correction, the robot was able to reach the target while avoiding the obstacles. The movement was also extended to perform the entire peg-in-hole task, rather than stopping above the hole.} \label{fig:avoid1}
\end{figure}
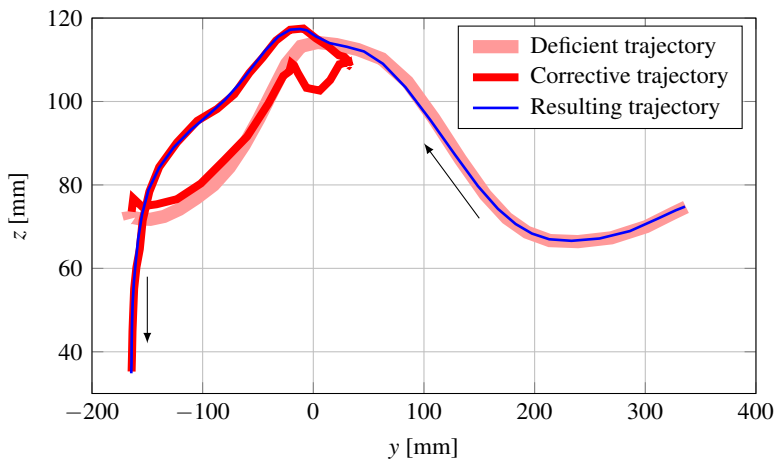

Each of the attempts of Scenario B was also successful. After modification, the DMPs generated trajectories that moved the grasped stop button above the height of potential obstacles, in this case other stop buttons, and subsequently inserted it into the case. The result from one attempt is shown in  \cref{fig:avoid1}.

\section{Discussion}
\label{sec:discussion}

The subsequent step in this work is to integrate the presented framework with trajectory-based reinforcement learning \cite{pastor2013dynamic,stulp2012model}, in order to optimize the motion locally with respect to criteria such as execution time. The program should also be augmented to take the purpose of, and relation between, different DMPs into consideration. This extension will emphasize the necessity of keeping track of different states within the work flow. To this purpose, a state machine implemented in, \emph{e.g.}, JGrafchart \cite{theorin2014sequential}, or the framework of behavior trees, applied on robot control in \cite{marzinotto2014towards}, would be suitable. Extending the user interface with support for natural language, would possibly make this framework more user friendly.

Performing the computations in joint space instead of Cartesian space allowed the operator to determine the entire configuration of the 7~DOF robot arm, rather than the pose of the tool only. However, one could think of situations where the operator is not concerned by the configuration, and the pose of the tool would be more intuitive to consider. It would therefore be valuable if it could be determined whether the operator aimed to adjust the configuration or just the pose of the tool. For example, a large configuration change yielding a small movement of the tool, should promote the hypothesis that the operator aimed to adjust the configuration.

It should be stated that the scenarios evaluated here are not covering the whole range of plausible scenarios related to this method, and it remains as future work to investigate the generalizability, and user experience, more thoroughly. The last part of the resulting movement is guaranteed to follow the retained part of the corrective demonstration accurately, given enough DMP basis functions. Hence, the only source of error on that part is a faulty demonstration. For instance, the movement might require higher accuracy than what is possible to demonstrate using lead-through programming. Another limitation with this method is that it is difficult for the operator to very accurately determine which part of the faulty trajectory to retain, since this is done autonomously. However, for the experiments performed here, the estimation of the operator was sufficient to demonstrate the desired behavior. The benefit with this approach is that it saves time as the operator does not have to specify all details explicitly.

\section{Conclusion}
In this paper, an approach for modification of DMPs, using lead-through programming, was presented. It allowed a robot operator to modify the last part of a faulty generated trajectory, instead of demonstrating a new one from the beginning. Based on the corrective demonstration, modified DMPs were formed automatically. A real-time application, that did not require any additional engineering work by the user, was developed, and verified experimentally. A video showing the functionality is available as a publication attachment, and a version with higher resolution is available on \cite{mod_youtube}.
\label{sec:conclusions}

\section*{Acknowledgments}
The authors would like to thank Fredrik Bagge Carlson, Bj{\"o}rn~Olofsson and Karl Johan {\AA}str{\"o}m at the Department of Automatic Control, Lund University, as well as Maj~Stenmark, Mathias~Haage and Jacek~Malec at Computer Science, Lund University, for valuable discussions throughout this work. The authors are members of the LCCC Linnaeus Center and the ELLIIT Excellence Center at Lund University. The research leading to these results has received funding from the European Commission's Framework Programme Horizon 2020 – under grant agreement No 644938 – SARAFun.


%% file: chapters/paper1/figs/cascade_scheme.tex
\usetikzlibrary{shapes,positioning}
\usetikzlibrary{narrow}

\tikzset{block/.style={draw, rectangle, line width=2pt,
     minimum height=3em, minimum width=3em, outer sep=0pt}}
\tikzset{every picture/.style={auto, line width=1pt,
          >=narrow,font=\small}}
\begin{center}
\begin{tikzpicture}[auto, node distance=1.0cm,>=latex',scale=0.9, every node/.style={scale=0.9}]
\node[text width=2cm,align=center, block,fill=red!20](initdemo){Initial \\ demonstration};
\node[block, right=3.5mm of initdemo](fitdmp){Create DMP};
\node[text width=1.8cm,align=center, block, fill=blue!20, right=10mm of fitdmp](eval){Evaluation};
\coordinate[left=0mm of eval](preeval);
\coordinate[right=6mm of eval](posteval);
\coordinate[right=2mm of posteval](postposteval);
\node[text width=2cm,align=center, block, fill=red!20, below=10mm of posteval](corrdemo){Corrective \\ demonstration};
\node[text width=1cm,align=center, block, below=10mm of preeval](moddmp){Modify \\ DMP};
\node[text width=2cm,align=center, block, fill=blue!20, above=9mm of posteval](reinf){Further \\ improvement};

\node (succ) [above=3.2mm of postposteval]{Successful};
\node (unsucc) [below=4.5mm of postposteval]{Unsuccessful};
\draw[->](initdemo)--(fitdmp);
\draw[->](corrdemo)--(moddmp);
\draw[->](fitdmp)--(eval);
\draw[->](eval)--(reinf);
\draw[->](eval)--(corrdemo);
\draw[->](moddmp)--(eval);

\coordinate[left=2mm of moddmp](tmp1);
\coordinate[below=6mm of tmp1](lowerleft);
\coordinate[above=28mm of lowerleft](upperleft);
\draw[dashed](lowerleft)--(upperleft);
\coordinate[right=42.5mm of upperleft](upperright);
\draw[dashed](upperleft)--(upperright);
\coordinate[below=28mm of upperright](lowerright);
\draw[dashed](upperright)--(lowerright);
\draw[dashed](lowerright)--(lowerleft);
\end{tikzpicture}
\end{center}

%% file: chapters/paper1/figs/adjust_notes.tex
\begin{tikzpicture}
\begin{axis}[%
width=\figurewidth,
height=\figureheight,
scale only axis,
xmin=100,
xmax=140,
xlabel={$x$ [mm]},
xmajorgrids,
ymin=0,
ymax=50,
ylabel={$z$ [mm]},
ymajorgrids,
legend style={draw=black,fill=white,legend cell align=left}
]
\addplot [color=white!60!red,solid,line width=5.0pt]
  table[row sep=crcr]{%
0	140.463091378\\
0.00308358299997735	140.457144087\\
0.0203699589999928	140.393491524\\
0.140442180999969	140.304630024\\
0.701843286999974	140.594079901\\
1.65746168099997	141.265330598\\
2.745038991	141.818598007\\
3.64899754999999	141.993352987\\
4.49078690299996	141.855011206\\
5.48085534299997	141.494089601\\
6.63339217699996	141.066894663\\
7.84344388899996	140.573978829\\
8.98671143199999	140.119895539\\
10.1215486	139.564753132\\
11.267604605	138.83976658\\
12.377478531	137.644423284\\
13.695679894	136.366455469\\
15.147944397	135.158795031\\
16.658526914	134.213695994\\
18.204950258	133.079319887\\
19.766242667	131.908011543\\
21.165397277	130.755744852\\
22.412003624	129.687565062\\
23.593952599	128.81713784\\
24.723244586	128.114076475\\
25.791991447	127.536933495\\
26.896457521	127.055796447\\
28.054210981	126.351845515\\
29.255270749	125.409557757\\
30.629639499	124.31677634\\
32.238966815	122.994514726\\
34.116160245	121.435917863\\
36.004673314	119.735449318\\
37.904143458	117.953023831\\
39.748686065	115.856098276\\
41.511183463	113.967855385\\
43.640898633	111.859602507\\
45.798510669	109.958115579\\
47.89087894	108.091379648\\
49.901278971	106.335849801\\
51.842690536	104.41272511\\
53.619793135	102.582581146\\
55.319071289	100.84745537\\
57.055117286	99.289226351\\
58.764178571	97.803690998\\
60.407900916	96.397761477\\
62.168482104	94.773690652\\
64.022475953	92.707247072\\
65.725969646	90.39483361\\
67.421715631	87.76151879\\
69.386868281	85.109482587\\
71.265634725	82.369057937\\
73.037336999	79.375572928\\
74.869740899	76.152954747\\
76.751255319	73.391965398\\
78.49096103	71.206121781\\
80.206745746	69.104941463\\
81.70867415	67.278573587\\
82.975999917	65.397408884\\
84.144989987	63.239218078\\
85.339230101	61.156261553\\
86.438117602	59.2818322\\
87.417763935	57.457831428\\
88.336582815	55.602149711\\
89.305559714	53.865414481\\
90.363195545	52.335772\\
91.518328297	50.705084892\\
92.762988002	49.13314552\\
94.011306179	47.463430326\\
95.124689592	46.027291838\\
95.946796111	44.889030087\\
96.891294798	43.997309523\\
97.929754621	43.189779511\\
98.928516113	42.331683426\\
99.856037211	41.339081309\\
100.787802673	39.820476724\\
101.659157692	37.976457853\\
102.393521154	36.327219163\\
103.040569741	34.885437686\\
103.634615612	33.513403313\\
104.284363059	32.009457176\\
105.126321255	30.40686201\\
106.015876307	28.864924809\\
106.695638901	27.428275176\\
107.187585637	26.311787104\\
107.850251501	25.783046599\\
108.851260032	25.458293444\\
109.619564484	25.073508086\\
110.311316443	24.489754482\\
110.881495264	23.6236068\\
111.344825923	22.627049989\\
111.811739109	21.746840389\\
112.180791495	20.384286332\\
112.440223779	18.738008878\\
112.544495364	17.006782822\\
112.675978208	15.468238066\\
112.898936644	14.115265447\\
113.193910132	13.031951655\\
113.523546309	12.014580772\\
113.752554988	11.500898599\\
113.873681429	11.173458306\\
114.000789858	10.847273075\\
114.231613013	10.608987533\\
114.564403006	10.105795715\\
115.17909576	8.69745061500001\\
115.656806042	6.49471937499999\\
116.428858987	6.70908925000001\\
116.73453575	6.554459622\\
116.854745781	6.358458644\\
116.894488202	6.33634570500001\\
116.905629789	6.34236913399999\\
};
\addlegendentry{Deficient trajectory};

\addplot [color=red,solid,line width=5.0pt]
  table[row sep=crcr]{%
117.184569534	6.70581914100001\\
117.186539041	6.70670031500001\\
117.190862834	6.71251994400001\\
117.254504573	6.783694156\\
117.329376858	6.87094492400001\\
117.350802766	6.89471633700001\\
117.388538625	6.93600898100001\\
117.508353619	7.07195814299999\\
117.63955769	7.22424777500001\\
117.749100011	7.34907459199999\\
117.891576447	7.51460970400001\\
118.09558241	7.749950867\\
118.326454673	8.109880175\\
118.542299275	9.02013981600001\\
118.734087203	10.078172848\\
118.870170797	10.901255786\\
118.982822641	11.423839612\\
119.027002512	11.586174246\\
119.041771586	11.639647081\\
119.033814289	11.666255525\\
119.030098316	11.664508087\\
119.024742839	11.679326428\\
119.033917565	11.722601903\\
119.019757005	11.758211689\\
118.923925906	11.727290964\\
118.817707075	11.657115261\\
118.74469525	11.613380039\\
118.71738917	11.65805862\\
118.706237561	11.767228303\\
118.713885088	11.997053914\\
118.791830884	12.23768006\\
118.878727966	12.421693481\\
118.851799626	12.593599458\\
118.808122114	13.084272047\\
118.791376066	14.109558858\\
118.782572304	15.184313235\\
118.676977535	16.05282968\\
118.57320573	16.195012051\\
118.375085407	16.014512153\\
118.183118773	15.895817481\\
118.064160467	15.963747039\\
118.000491676	16.395463188\\
117.964809422	17.010160586\\
117.928771798	17.345132121\\
117.876168033	17.461399945\\
117.81401168	17.540037625\\
117.783125657	17.675166955\\
117.763276705	18.094239467\\
117.700091308	19.066263077\\
117.698589536	19.834755164\\
117.658113202	20.538716046\\
117.596151111	21.116180403\\
117.506068536	21.238350283\\
117.265594472	21.471872317\\
117.139251933	22.312386094\\
117.129605661	23.476059961\\
117.061465285	24.719556387\\
117.127689291	25.642733797\\
117.189679661	26.030068524\\
117.273567083	26.154182276\\
117.395564685	26.316505963\\
117.4626943	26.436048973\\
117.470108799	26.492675382\\
117.474266808	26.554619382\\
117.475102729	26.57857974\\
117.481396679	26.595872598\\
117.485346083	26.607079745\\
117.48845576	26.610855433\\
117.503570799	26.621789684\\
117.52804667	26.637354317\\
117.560351775	26.661575023\\
117.639977552	26.721154456\\
117.759534573	26.806627837\\
117.912648456	26.923650125\\
118.131271356	27.113172811\\
118.486342398	27.336752398\\
118.766829438	27.479731519\\
118.950615331	27.392958336\\
119.076573333	27.262195217\\
119.255665376	26.866287165\\
119.474758165	26.163188143\\
119.67876187	25.876659928\\
119.854092169	25.795255525\\
120.039716923	25.761539616\\
120.395281402	25.73492019\\
120.736156616	25.585302033\\
120.937478034	25.352014469\\
121.140877832	25.08267479\\
121.398095543	24.531345722\\
121.759755181	23.89448289\\
122.280203258	23.277685061\\
122.873839209	22.756737628\\
123.452293817	22.354037049\\
124.034807949	22.067471211\\
124.366169658	21.99322345\\
124.622739233	21.732265378\\
124.92464031	21.19422189\\
125.199312599	21.04919097\\
125.428512783	21.186171855\\
125.554733799	21.04241437\\
125.711556647	20.221554981\\
125.892092202	19.304664708\\
126.12761019	18.315338423\\
126.313078715	17.597090207\\
126.474500633	17.272142886\\
126.684518943	16.894069158\\
126.961012846	16.561435299\\
127.080089497	16.105559329\\
127.103296197	15.528138464\\
127.124556412	15.201367179\\
127.176827048	15.151623275\\
127.248720136	15.103513662\\
127.363549548	15.077526579\\
127.555096549	14.981118139\\
127.953045636	14.842804711\\
128.309796794	14.713757429\\
128.559625166	14.449846979\\
128.692937325	14.091784547\\
128.858030629	13.507768025\\
129.074329394	13.095839801\\
129.251740947	12.955957158\\
129.334074163	12.8559682\\
129.426141082	12.691746368\\
129.548085	12.132796296\\
129.68241251	11.589893579\\
129.856434665	11.460557644\\
130.094557026	11.082701668\\
130.356087757	10.388417439\\
130.480069214	10.269245413\\
130.535525101	10.303407794\\
130.602034832	10.284234625\\
130.705614124	10.13912126\\
130.799539906	9.57184424600001\\
130.857640955	8.94036862199999\\
130.896037498	8.86462779999999\\
130.940801659	8.90056696299999\\
131.12760065	9.09238921900001\\
131.237330279	9.17781382500002\\
131.253739215	9.099987505\\
131.287052087	9.063596719\\
131.331307994	8.99944108700001\\
131.385130277	8.57381211699999\\
131.497384446	7.69759403700002\\
131.544306499	7.00061264300001\\
131.631175365	6.424256273\\
131.758577946	6.202933586\\
132.019287385	6.407178335\\
132.298651087	6.52634217600001\\
132.544912172	6.53916063599999\\
132.751855897	6.53335232399999\\
132.828184206	6.316565704\\
132.959412204	6.05255532000001\\
133.183521235	5.94937842600001\\
133.449883345	5.88890010599999\\
133.696777041	5.781465819\\
133.943837981	5.51419529200001\\
134.087565254	5.17151089399999\\
134.120510756	4.92702400799999\\
134.145156896	4.87967854800002\\
134.155480938	4.834942404\\
134.153586084	4.73406413800001\\
134.134256308	4.33591867999999\\
134.102760049	3.70857953699999\\
134.082332068	2.96736787399999\\
134.083850736	2.482875439\\
134.100530564	2.19807138499999\\
134.119829326	1.88030407100001\\
134.128865424	1.43215140699999\\
134.148486923	1.16901198900001\\
134.175901836	1.06403677\\
134.194145427	1.029691235\\
134.206260002	0.994478119999997\\
134.219185404	0.925387341000004\\
134.219217093	0.618446831999989\\
134.26485851	0.0850757810000005\\
134.366105526	0\\
};
\addlegendentry{Corrective trajectory};
\end{axis}

\draw [|-|, line width=.5mm] (2.7,2.1) -- (3.75,2.6);
\node at (3.1,2.6) {$d_m$};

\end{tikzpicture}%

%% file: chapters/paper1/figs/adjust_zoom.tex
\begin{tikzpicture}
\begin{axis}[%
width=\figurewidth,
height=\figureheight,
scale only axis,
xmin=100,
xmax=140,
xlabel={$x$ [mm]},
xmajorgrids,
ymin=0,
ymax=55,
ylabel={$z$ [mm]},
ymajorgrids,
legend style={draw=black,fill=white,legend cell align=left}
]
\addplot [color=white!60!red,solid,line width=5.0pt]
  table[row sep=crcr]{%
0	140.463091378\\
0.00308358299997735	140.457144087\\
0.0203699589999928	140.393491524\\
0.140442180999969	140.304630024\\
0.701843286999974	140.594079901\\
1.65746168099997	141.265330598\\
2.745038991	141.818598007\\
3.64899754999999	141.993352987\\
4.49078690299996	141.855011206\\
5.48085534299997	141.494089601\\
6.63339217699996	141.066894663\\
7.84344388899996	140.573978829\\
8.98671143199999	140.119895539\\
10.1215486	139.564753132\\
11.267604605	138.83976658\\
12.377478531	137.644423284\\
13.695679894	136.366455469\\
15.147944397	135.158795031\\
16.658526914	134.213695994\\
18.204950258	133.079319887\\
19.766242667	131.908011543\\
21.165397277	130.755744852\\
22.412003624	129.687565062\\
23.593952599	128.81713784\\
24.723244586	128.114076475\\
25.791991447	127.536933495\\
26.896457521	127.055796447\\
28.054210981	126.351845515\\
29.255270749	125.409557757\\
30.629639499	124.31677634\\
32.238966815	122.994514726\\
34.116160245	121.435917863\\
36.004673314	119.735449318\\
37.904143458	117.953023831\\
39.748686065	115.856098276\\
41.511183463	113.967855385\\
43.640898633	111.859602507\\
45.798510669	109.958115579\\
47.89087894	108.091379648\\
49.901278971	106.335849801\\
51.842690536	104.41272511\\
53.619793135	102.582581146\\
55.319071289	100.84745537\\
57.055117286	99.289226351\\
58.764178571	97.803690998\\
60.407900916	96.397761477\\
62.168482104	94.773690652\\
64.022475953	92.707247072\\
65.725969646	90.39483361\\
67.421715631	87.76151879\\
69.386868281	85.109482587\\
71.265634725	82.369057937\\
73.037336999	79.375572928\\
74.869740899	76.152954747\\
76.751255319	73.391965398\\
78.49096103	71.206121781\\
80.206745746	69.104941463\\
81.70867415	67.278573587\\
82.975999917	65.397408884\\
84.144989987	63.239218078\\
85.339230101	61.156261553\\
86.438117602	59.2818322\\
87.417763935	57.457831428\\
88.336582815	55.602149711\\
89.305559714	53.865414481\\
90.363195545	52.335772\\
91.518328297	50.705084892\\
92.762988002	49.13314552\\
94.011306179	47.463430326\\
95.124689592	46.027291838\\
95.946796111	44.889030087\\
96.891294798	43.997309523\\
97.929754621	43.189779511\\
98.928516113	42.331683426\\
99.856037211	41.339081309\\
100.787802673	39.820476724\\
101.659157692	37.976457853\\
102.393521154	36.327219163\\
103.040569741	34.885437686\\
103.634615612	33.513403313\\
104.284363059	32.009457176\\
105.126321255	30.40686201\\
106.015876307	28.864924809\\
106.695638901	27.428275176\\
107.187585637	26.311787104\\
107.850251501	25.783046599\\
108.851260032	25.458293444\\
109.619564484	25.073508086\\
110.311316443	24.489754482\\
110.881495264	23.6236068\\
111.344825923	22.627049989\\
111.811739109	21.746840389\\
112.180791495	20.384286332\\
112.440223779	18.738008878\\
112.544495364	17.006782822\\
112.675978208	15.468238066\\
112.898936644	14.115265447\\
113.193910132	13.031951655\\
113.523546309	12.014580772\\
113.752554988	11.500898599\\
113.873681429	11.173458306\\
114.000789858	10.847273075\\
114.231613013	10.608987533\\
114.564403006	10.105795715\\
115.17909576	8.69745061500001\\
115.656806042	6.49471937499999\\
116.428858987	6.70908925000001\\
116.73453575	6.554459622\\
116.854745781	6.358458644\\
116.894488202	6.33634570500001\\
116.905629789	6.34236913399999\\
};
\addlegendentry{Deficient trajectory};

\addplot [color=red,solid,line width=5.0pt]
  table[row sep=crcr]{%
117.184569534	6.70581914100001\\
117.186539041	6.70670031500001\\
117.190862834	6.71251994400001\\
117.254504573	6.783694156\\
117.329376858	6.87094492400001\\
117.350802766	6.89471633700001\\
117.388538625	6.93600898100001\\
117.508353619	7.07195814299999\\
117.63955769	7.22424777500001\\
117.749100011	7.34907459199999\\
117.891576447	7.51460970400001\\
118.09558241	7.749950867\\
118.326454673	8.109880175\\
118.542299275	9.02013981600001\\
118.734087203	10.078172848\\
118.870170797	10.901255786\\
118.982822641	11.423839612\\
119.027002512	11.586174246\\
119.041771586	11.639647081\\
119.033814289	11.666255525\\
119.030098316	11.664508087\\
119.024742839	11.679326428\\
119.033917565	11.722601903\\
119.019757005	11.758211689\\
118.923925906	11.727290964\\
118.817707075	11.657115261\\
118.74469525	11.613380039\\
118.71738917	11.65805862\\
118.706237561	11.767228303\\
118.713885088	11.997053914\\
118.791830884	12.23768006\\
118.878727966	12.421693481\\
118.851799626	12.593599458\\
118.808122114	13.084272047\\
118.791376066	14.109558858\\
118.782572304	15.184313235\\
118.676977535	16.05282968\\
118.57320573	16.195012051\\
118.375085407	16.014512153\\
118.183118773	15.895817481\\
118.064160467	15.963747039\\
118.000491676	16.395463188\\
117.964809422	17.010160586\\
117.928771798	17.345132121\\
117.876168033	17.461399945\\
117.81401168	17.540037625\\
117.783125657	17.675166955\\
117.763276705	18.094239467\\
117.700091308	19.066263077\\
117.698589536	19.834755164\\
117.658113202	20.538716046\\
117.596151111	21.116180403\\
117.506068536	21.238350283\\
117.265594472	21.471872317\\
117.139251933	22.312386094\\
117.129605661	23.476059961\\
117.061465285	24.719556387\\
117.127689291	25.642733797\\
117.189679661	26.030068524\\
117.273567083	26.154182276\\
117.395564685	26.316505963\\
117.4626943	26.436048973\\
117.470108799	26.492675382\\
117.474266808	26.554619382\\
117.475102729	26.57857974\\
117.481396679	26.595872598\\
117.485346083	26.607079745\\
117.48845576	26.610855433\\
117.503570799	26.621789684\\
117.52804667	26.637354317\\
117.560351775	26.661575023\\
117.639977552	26.721154456\\
117.759534573	26.806627837\\
117.912648456	26.923650125\\
118.131271356	27.113172811\\
118.486342398	27.336752398\\
118.766829438	27.479731519\\
118.950615331	27.392958336\\
119.076573333	27.262195217\\
119.255665376	26.866287165\\
119.474758165	26.163188143\\
119.67876187	25.876659928\\
119.854092169	25.795255525\\
120.039716923	25.761539616\\
120.395281402	25.73492019\\
120.736156616	25.585302033\\
120.937478034	25.352014469\\
121.140877832	25.08267479\\
121.398095543	24.531345722\\
121.759755181	23.89448289\\
122.280203258	23.277685061\\
122.873839209	22.756737628\\
123.452293817	22.354037049\\
124.034807949	22.067471211\\
124.366169658	21.99322345\\
124.622739233	21.732265378\\
124.92464031	21.19422189\\
125.199312599	21.04919097\\
125.428512783	21.186171855\\
125.554733799	21.04241437\\
125.711556647	20.221554981\\
125.892092202	19.304664708\\
126.12761019	18.315338423\\
126.313078715	17.597090207\\
126.474500633	17.272142886\\
126.684518943	16.894069158\\
126.961012846	16.561435299\\
127.080089497	16.105559329\\
127.103296197	15.528138464\\
127.124556412	15.201367179\\
127.176827048	15.151623275\\
127.248720136	15.103513662\\
127.363549548	15.077526579\\
127.555096549	14.981118139\\
127.953045636	14.842804711\\
128.309796794	14.713757429\\
128.559625166	14.449846979\\
128.692937325	14.091784547\\
128.858030629	13.507768025\\
129.074329394	13.095839801\\
129.251740947	12.955957158\\
129.334074163	12.8559682\\
129.426141082	12.691746368\\
129.548085	12.132796296\\
129.68241251	11.589893579\\
129.856434665	11.460557644\\
130.094557026	11.082701668\\
130.356087757	10.388417439\\
130.480069214	10.269245413\\
130.535525101	10.303407794\\
130.602034832	10.284234625\\
130.705614124	10.13912126\\
130.799539906	9.57184424600001\\
130.857640955	8.94036862199999\\
130.896037498	8.86462779999999\\
130.940801659	8.90056696299999\\
131.12760065	9.09238921900001\\
131.237330279	9.17781382500002\\
131.253739215	9.099987505\\
131.287052087	9.063596719\\
131.331307994	8.99944108700001\\
131.385130277	8.57381211699999\\
131.497384446	7.69759403700002\\
131.544306499	7.00061264300001\\
131.631175365	6.424256273\\
131.758577946	6.202933586\\
132.019287385	6.407178335\\
132.298651087	6.52634217600001\\
132.544912172	6.53916063599999\\
132.751855897	6.53335232399999\\
132.828184206	6.316565704\\
132.959412204	6.05255532000001\\
133.183521235	5.94937842600001\\
133.449883345	5.88890010599999\\
133.696777041	5.781465819\\
133.943837981	5.51419529200001\\
134.087565254	5.17151089399999\\
134.120510756	4.92702400799999\\
134.145156896	4.87967854800002\\
134.155480938	4.834942404\\
134.153586084	4.73406413800001\\
134.134256308	4.33591867999999\\
134.102760049	3.70857953699999\\
134.082332068	2.96736787399999\\
134.083850736	2.482875439\\
134.100530564	2.19807138499999\\
134.119829326	1.88030407100001\\
134.128865424	1.43215140699999\\
134.148486923	1.16901198900001\\
134.175901836	1.06403677\\
134.194145427	1.029691235\\
134.206260002	0.994478119999997\\
134.219185404	0.925387341000004\\
134.219217093	0.618446831999989\\
134.26485851	0.0850757810000005\\
134.366105526	0\\
};
\addlegendentry{Corrective trajectory};

\addplot [color=black,solid,line width=2.0pt]
  table[row sep=crcr]{%
0	140.463091378\\
0.00308358299997735	140.457144087\\
0.0203699589999928	140.393491524\\
0.140442180999969	140.304630024\\
0.701843286999974	140.594079901\\
1.65746168099997	141.265330598\\
2.745038991	141.818598007\\
3.64899754999999	141.993352987\\
4.49078690299996	141.855011206\\
5.48085534299997	141.494089601\\
6.63339217699996	141.066894663\\
7.84344388899996	140.573978829\\
8.98671143199999	140.119895539\\
10.1215486	139.564753132\\
11.267604605	138.83976658\\
12.377478531	137.644423284\\
13.695679894	136.366455469\\
15.147944397	135.158795031\\
16.658526914	134.213695994\\
18.204950258	133.079319887\\
19.766242667	131.908011543\\
21.165397277	130.755744852\\
22.412003624	129.687565062\\
23.593952599	128.81713784\\
24.723244586	128.114076475\\
25.791991447	127.536933495\\
26.896457521	127.055796447\\
28.054210981	126.351845515\\
29.255270749	125.409557757\\
30.629639499	124.31677634\\
32.238966815	122.994514726\\
34.116160245	121.435917863\\
36.004673314	119.735449318\\
37.904143458	117.953023831\\
39.748686065	115.856098276\\
41.511183463	113.967855385\\
43.640898633	111.859602507\\
45.798510669	109.958115579\\
47.89087894	108.091379648\\
49.901278971	106.335849801\\
51.842690536	104.41272511\\
53.619793135	102.582581146\\
55.319071289	100.84745537\\
57.055117286	99.289226351\\
58.764178571	97.803690998\\
60.407900916	96.397761477\\
62.168482104	94.773690652\\
64.022475953	92.707247072\\
65.725969646	90.39483361\\
67.421715631	87.76151879\\
69.386868281	85.109482587\\
71.265634725	82.369057937\\
73.037336999	79.375572928\\
74.869740899	76.152954747\\
76.751255319	73.391965398\\
78.49096103	71.206121781\\
80.206745746	69.104941463\\
81.70867415	67.278573587\\
82.975999917	65.397408884\\
84.144989987	63.239218078\\
85.339230101	61.156261553\\
86.438117602	59.2818322\\
87.417763935	57.457831428\\
88.336582815	55.602149711\\
89.305559714	53.865414481\\
90.363195545	52.335772\\
91.518328297	50.705084892\\
92.762988002	49.13314552\\
94.011306179	47.463430326\\
95.124689592	46.027291838\\
95.946796111	44.889030087\\
96.891294798	43.997309523\\
97.929754621	43.189779511\\
98.928516113	42.331683426\\
99.856037211	41.339081309\\
100.620633154009	39.3801307974991\\
101.370116849094	37.8022755854935\\
102.074507285708	36.1628327517566\\
102.774663071479	34.5205856100479\\
103.520065003802	32.9223836780777\\
104.356748998797	31.4120906795625\\
105.316929830884	30.0346241498069\\
106.414570671058	28.8304713122816\\
107.648080247925	27.8279206427407\\
109.001697613279	27.0388532064176\\
110.441755221246	26.4577359253435\\
111.913740915068	26.0651473760213\\
113.34682900972	25.8353122323324\\
114.663497139028	25.7409071356765\\
115.785087010621	25.7545031388137\\
116.636768567055	25.8480683596705\\
117.151569480973	25.991858588918\\
117.27356708297	26.1541822758714\\
};
\addlegendentry{Modified trajectory};

\addplot [color=blue,line width=3.0pt,mark size=6.0pt,only marks,mark=x,mark options={solid}]
  table[row sep=crcr]{%
117.273567083	26.154182276\\
};
\addlegendentry{Separation point set by user};

\addplot [color=green,line width=3.0pt,mark size=6.0pt,only marks,mark=x,mark options={solid}]
  table[row sep=crcr]{%
112.180791495	20.384286332\\
};
\addlegendentry{Separation point set automatically};

\addplot [color=black,solid,line width=2.0pt,forget plot]
  table[row sep=crcr]{%
117.273567083	26.154182276\\
117.395564685	26.316505963\\
117.4626943	26.436048973\\
117.470108799	26.492675382\\
117.474266808	26.554619382\\
117.475102729	26.57857974\\
117.481396679	26.595872598\\
117.485346083	26.607079745\\
117.48845576	26.610855433\\
117.503570799	26.621789684\\
117.52804667	26.637354317\\
117.560351775	26.661575023\\
117.639977552	26.721154456\\
117.759534573	26.806627837\\
117.912648456	26.923650125\\
118.131271356	27.113172811\\
118.486342398	27.336752398\\
118.766829438	27.479731519\\
118.950615331	27.392958336\\
119.076573333	27.262195217\\
119.255665376	26.866287165\\
119.474758165	26.163188143\\
119.67876187	25.876659928\\
119.854092169	25.795255525\\
120.039716923	25.761539616\\
120.395281402	25.73492019\\
120.736156616	25.585302033\\
120.937478034	25.352014469\\
121.140877832	25.08267479\\
121.398095543	24.531345722\\
121.759755181	23.89448289\\
122.280203258	23.277685061\\
122.873839209	22.756737628\\
123.452293817	22.354037049\\
124.034807949	22.067471211\\
124.366169658	21.99322345\\
124.622739233	21.732265378\\
124.92464031	21.19422189\\
125.199312599	21.04919097\\
125.428512783	21.186171855\\
125.554733799	21.04241437\\
125.711556647	20.221554981\\
125.892092202	19.304664708\\
126.12761019	18.315338423\\
126.313078715	17.597090207\\
126.474500633	17.272142886\\
126.684518943	16.894069158\\
126.961012846	16.561435299\\
127.080089497	16.105559329\\
127.103296197	15.528138464\\
127.124556412	15.201367179\\
127.176827048	15.151623275\\
127.248720136	15.103513662\\
127.363549548	15.077526579\\
127.555096549	14.981118139\\
127.953045636	14.842804711\\
128.309796794	14.713757429\\
128.559625166	14.449846979\\
128.692937325	14.091784547\\
128.858030629	13.507768025\\
129.074329394	13.095839801\\
129.251740947	12.955957158\\
129.334074163	12.8559682\\
129.426141082	12.691746368\\
129.548085	12.132796296\\
129.68241251	11.589893579\\
129.856434665	11.460557644\\
130.094557026	11.082701668\\
130.356087757	10.388417439\\
130.480069214	10.269245413\\
130.535525101	10.303407794\\
130.602034832	10.284234625\\
130.705614124	10.13912126\\
130.799539906	9.57184424600001\\
130.857640955	8.94036862199999\\
130.896037498	8.86462779999999\\
130.940801659	8.90056696299999\\
131.12760065	9.09238921900001\\
131.237330279	9.17781382500002\\
131.253739215	9.099987505\\
131.287052087	9.063596719\\
131.331307994	8.99944108700001\\
131.385130277	8.57381211699999\\
131.497384446	7.69759403700002\\
131.544306499	7.00061264300001\\
131.631175365	6.424256273\\
131.758577946	6.202933586\\
132.019287385	6.407178335\\
132.298651087	6.52634217600001\\
132.544912172	6.53916063599999\\
132.751855897	6.53335232399999\\
132.828184206	6.316565704\\
132.959412204	6.05255532000001\\
133.183521235	5.94937842600001\\
133.449883345	5.88890010599999\\
133.696777041	5.781465819\\
133.943837981	5.51419529200001\\
134.087565254	5.17151089399999\\
134.120510756	4.92702400799999\\
134.145156896	4.87967854800002\\
134.155480938	4.834942404\\
134.153586084	4.73406413800001\\
134.134256308	4.33591867999999\\
134.102760049	3.70857953699999\\
134.082332068	2.96736787399999\\
134.083850736	2.482875439\\
134.100530564	2.19807138499999\\
134.119829326	1.88030407100001\\
134.128865424	1.43215140699999\\
134.148486923	1.16901198900001\\
134.175901836	1.06403677\\
134.194145427	1.029691235\\
134.206260002	0.994478119999997\\
134.219185404	0.925387341000004\\
134.219217093	0.618446831999989\\
134.26485851	0.0850757810000005\\
134.366105526	0\\
};
\end{axis}

\draw [->] (.35,4) -- (1,3.3);
\draw [->] (3.9,1.1) -- (3.5,2.1);
\draw [->] (5.2,1.85) -- (6.2,0.95);

\end{tikzpicture}%

%% file: chapters/paper1/figs/place_a1.tex
%
%
\begin{tikzpicture}

\begin{axis}[%
width=\figurewidth,
height=\figureheight,
scale only axis,
xmin=10.9775375467,
xmax=275,
xmajorgrids,
ymajorgrids,
xlabel={$y$ [mm]},
ymin=28.4880469855083,
ymax=222.168397523992,
ylabel={$z$ [mm]},
legend pos=north west,
legend style={draw=black,fill=white,legend cell align=left}
]
\addplot [color=white!60!red,solid,line width=5.0pt]
  table[row sep=crcr]{%
256.543626164	213.6734180074\\
243.327411128	210.8414816094\\
217.61217637	204.7677834682\\
188.585181383	196.557414432\\
161.411168318	187.7637455539\\
137.547225384	178.9417460599\\
116.795559654	170.1644688401\\
97.1102754626	159.6792977718\\
77.806137912	147.007751438\\
59.4670445872	130.881550626\\
44.6395492476	113.823093874\\
33.5855511479	99.498715294\\
25.2583739714	88.184954899\\
18.5070571046	78.794129778\\
14.4003228378	70.885169143\\
12.7864500787	64.824791369\\
12.1203985858	60.602786921\\
11.6870184766	58.625051615\\
11.3775710178	58.158323288\\
};
\addlegendentry{Deficient trajectory};

\addplot [color=red,solid,line width=5.0pt]
  table[row sep=crcr]{%
11.2162126985	60.542704133\\
10.9775375467	61.004776456\\
11.6413370186	63.344204566\\
14.08000133	64.780517759\\
19.2580929959	69.72450528\\
22.1714408975	74.660897932\\
25.8863707053	77.892670433\\
27.1558359822	80.036206967\\
29.4469109131	81.869594755\\
30.0646075826	82.752191316\\
30.0872806027	82.842625264\\
30.1063987672	82.78247364\\
30.1531947889	82.585955904\\
29.8883135453	81.355234426\\
29.0756176715	79.350369492\\
29.1965378178	77.333093204\\
28.1606346533	75.012654168\\
28.0640585415	71.659288929\\
27.4419733446	68.851053851\\
26.7708789603	66.618206299\\
25.9340576408	63.124079665\\
24.302701389	59.829593642\\
23.7103348714	57.852392804\\
23.4207051218	53.830939179\\
23.9014041346	48.863948706\\
24.0446392219	42.493501433\\
23.7977728238	37.867857274\\
23.4674894647	36.817308077\\
};
\addlegendentry{Corrective trajectory};

\addplot [color=blue,solid,line width=2.0pt]
  table[row sep=crcr]{%
255.90866926	213.8391364325\\
245.490502729	210.9316240674\\
225.957957911	205.4027569448\\
200.198815945	198.3167203094\\
173.782740354	190.7704545354\\
150.860530792	183.5784736263\\
131.957419355	176.6249019454\\
115.437648226	169.4613129957\\
99.804847042	161.0738669212\\
83.6761778675	150.9234517681\\
67.3857008714	138.244593438\\
52.2981249177	123.82766454\\
40.23961295	109.339937408\\
31.8130663385	95.552775392\\
26.9043072166	81.799131695\\
24.8286442938	70.687003206\\
25.0291033037	65.565806386\\
25.5947720171	65.328032179\\
25.7157303918	63.600630381\\
24.9565108143	61.10169478\\
24.0065905651	58.643777007\\
23.2672096267	56.180846696\\
23.0462084745	53.276865538\\
23.4902102015	49.110300772\\
23.9517280964	43.97395915\\
23.6228083195	39.410597761\\
23.8515416632	37.34942553\\
};
\addlegendentry{Resulting trajectory};

\draw [->] (235,201) -- (170,181);

\end{axis}
\end{tikzpicture}%

%% file: chapters/paper1/figs/place_a2.tex
%
%
\begin{tikzpicture}

\begin{axis}[%
width=\figurewidth,
height=\figureheight,
scale only axis,
xmin=18.2737362278,
xmax=275,
xmajorgrids,
ymajorgrids,
xlabel={$y$ [mm]},
ymin=27.7541896190986,
ymax=219.178118418001,
ylabel={$z$ [mm]},
legend pos=north west,
legend style={draw=black,fill=white,legend cell align=left}
]
\addplot [color=white!60!red,solid,line width=5.0pt]
  table[row sep=crcr]{%
260.775515939	209.0359950458\\
255.311155056	207.8215122446\\
235.182992208	204.2495227445\\
204.481755128	198.1747951035\\
172.476973689	189.9708361185\\
144.728980149	180.2941092231\\
122.549469464	169.8449620442\\
103.424983016	158.8131587019\\
85.7091749471	145.947368014\\
70.9185419217	132.331290603\\
59.8990163847	118.24360018\\
52.3923349609	103.666716218\\
46.7448074255	87.919585178\\
43.459486237	73.776406001\\
41.9069169765	64.035905701\\
\\
};
\addlegendentry{Deficient trajectory};

\addplot [color=red,solid,line width=5.0pt]
  table[row sep=crcr]{%
40.9870088108	61.659336576\\
40.987532389	61.657114616\\
40.990954108	61.659463946\\
40.9983687297	61.667895477\\
41.0117794844	61.673190687\\
41.4088087302	64.318594581\\
44.1324861488	67.706976583\\
44.7550713699	71.398899094\\
47.1887814392	75.272118954\\
51.0340687182	79.155465072\\
54.5352410525	82.504724865\\
57.6089959409	86.498569948\\
59.6890097323	88.417646428\\
59.6023779592	88.603062455\\
59.6157143467	88.594868485\\
59.6189433401	88.586454449\\
59.628261647	88.568439509\\
59.5992410428	88.562649926\\
59.4464757457	88.581728554\\
57.9538964321	89.947150703\\
53.6466199013	90.674228024\\
48.3288670356	91.364532919\\
41.5458970382	91.016124909\\
36.4261736648	87.826396749\\
32.9848159597	84.516038448\\
30.3967501537	78.318012908\\
27.5221143891	72.578096484\\
25.9168463456	67.189832658\\
25.1703577238	61.173488326\\
24.0575596972	57.909625488\\
21.9210099949	55.478314344\\
19.8754822623	51.695623307\\
18.2737362278	47.056463165\\
18.4318495016	44.18328022\\
20.3229438426	37.894213873\\
};
\addlegendentry{Corrective trajectory};

\addplot [color=blue,solid,line width=2.0pt]
  table[row sep=crcr]{%
260.97717017	208.9734724567\\
259.327483559	208.7054163008\\
247.546100225	206.4009586726\\
226.299573333	201.8941980703\\
198.181730669	195.3578930912\\
169.88743422	187.6095651025\\
145.835064451	179.7175449551\\
127.092174452	172.1275581228\\
112.041811383	164.593930684\\
98.766947855	156.1912285397\\
85.7202315329	146.506485135\\
73.0309044427	134.992657338\\
61.6223203737	122.264226846\\
50.6541631934	108.239868679\\
39.1220734789	93.747721993\\
31.096028466	83.702247331\\
28.6216199928	79.439209228\\
28.1566416651	76.671429413\\
26.9228537361	72.054980024\\
25.628199056	67.146282991\\
24.7998829861	62.377539389\\
23.9932208766	58.454277515\\
22.3590902525	55.622940648\\
20.393516298	52.780894881\\
18.7795494648	49.16840099\\
20.3967650102	38.244666645\\
};
\addlegendentry{Resulting trajectory};

\draw [->] (248,198) -- (180,185);
\draw [->] (48,60) -- (64,84);

\end{axis}
\end{tikzpicture}%

%% file: chapters/paper1/figs/avoid_a1.tex
%
%
\begin{tikzpicture}

\begin{axis}[%
width=\figurewidth,
height=\figureheight,
scale only axis,
xmajorgrids,
ymajorgrids,
xmin=-200,
xmax=400,
xlabel={$y$ [mm]},
ymin=30,
ymax=120,
ylabel={$z$ [mm]},
legend style={draw=black,fill=white,legend cell align=left}
]
\addplot [color=white!60!red,solid,line width=5.0pt]
  table[row sep=crcr]{%
336.748093465	74.704676647\\
325.02006008	73.214649502\\
300.883131839	70.007670502\\
269.365063497	67.315893467\\
239.190506578	66.402443469\\
213.649593198	66.657092795\\
193.037274378	68.654630447\\
174.916869911	72.028777842\\
155.452977601	77.635484666\\
133.026860126	86.07461272\\
108.842722669	95.968982884\\
85.0726694917	104.709080803\\
61.9705050259	110.225831188\\
40.2283973731	112.458222942\\
20.8663347026	113.643254317\\
3.76763634336	114.15962358\\
-11.2498315161	113.210159298\\
-24.0953236589	109.060221282\\
-36.4019396114	102.651918485\\
-49.2431309094	95.857556679\\
-62.3923784357	89.579939493\\
-75.8497775393	84.206076732\\
-90.1951205734	79.95954583\\
-105.692670493	76.83034727\\
-120.960774104	74.16122095\\
-134.919712568	72.443522231\\
-147.223100596	71.754638809\\
-157.175655046	71.887677128\\
-163.479803629	72.234084136\\
-165.794650693	72.149324654\\
-165.491595165	71.854641044\\
};
\addlegendentry{Deficient trajectory};

\addplot [color=red,solid,line width=3.0pt]
  table[row sep=crcr]{%
-164.168093636	73.568733941\\
-162.669611459	76.776960542\\
-155.011719657	74.860147259\\
-141.620813875	75.354847243\\
-123.185635928	76.58000853\\
-101.626223589	80.341011441\\
-80.4418546397	85.958161378\\
-59.7490505502	91.612313923\\
-40.6216483059	99.506210842\\
-27.10515921	106.135412502\\
-21.7266127839	107.156615996\\
-19.6422280183	109.174221981\\
-17.3747814785	108.668485411\\
-14.1184042326	106.906764051\\
-6.16986636364	103.258536285\\
5.7819303281	102.617140247\\
15.2498574667	105.122009014\\
23.5602855309	108.918061611\\
30.7819634977	109.393764881\\
32.652644872	108.693617272\\
32.6863156414	108.641202863\\
32.5061025569	108.848854169\\
32.4441064332	108.830516883\\
32.0692019972	108.716917112\\
31.6837388397	108.779628012\\
31.8136639272	108.545927742\\
30.9790621871	109.905098777\\
27.998016175	110.699276623\\
22.9295287149	111.0843779\\
12.7653206589	113.357879568\\
2.40652703378	115.123984381\\
-8.9468248238	117.491474269\\
-21.0247630033	117.196568816\\
-33.6655539225	114.60033323\\
-45.7416673392	110.514417203\\
-57.7049742849	106.867567675\\
-71.4072085756	101.720160084\\
-86.9588550822	98.197646621\\
-104.841617849	95.36872941\\
-123.115681476	90.198326114\\
-139.250918348	84.286014396\\
-148.173994237	78.379148149\\
-154.205332991	71.475499447\\
-156.796854443	64.614679781\\
-159.845563056	60.254374963\\
-161.904945196	55.082173298\\
-162.728902445	49.617382005\\
-163.396723108	44.924524455\\
-163.875058628	38.737457053\\
-164.194377785	35.272364299\\
};
\addlegendentry{Corrective trajectory};

\addplot [color=blue,solid,line width=1pt]
  table[row sep=crcr]{%
335.93946874	74.803364752\\
327.687443693	73.937861775\\
311.012266412	71.889063508\\
286.169844064	68.94197483\\
258.651804325	67.15617382\\
233.363223364	66.60050278\\
213.202698718	66.969741661\\
197.486432438	68.351027367\\
182.82598478	70.630229146\\
166.881073495	74.253433321\\
148.800912578	79.711553641\\
127.881019996	87.291693649\\
105.626302709	95.610754438\\
83.2988548553	103.413611384\\
62.8941430572	109.04752301\\
45.7644074673	111.995577295\\
30.3673378544	113.109199703\\
14.8113267633	113.988134801\\
2.20346406158	115.621310862\\
-5.89413906923	117.071592308\\
-12.0260115364	117.379678248\\
-20.8233416442	117.222427186\\
-31.7397020322	115.5434231\\
-43.2170228671	112.284517056\\
-54.8747243669	108.337527297\\
-67.0511838133	104.17892041\\
-80.0842819309	100.321421909\\
-94.7887738928	96.800789473\\
-110.571010068	93.224190557\\
-126.117531941	88.997300882\\
-139.390793942	84.10018511\\
-148.909421941	78.830433029\\
-154.703961272	72.988624983\\
-158.240294708	67.083345997\\
-160.465888231	61.591180148\\
-162.145074735	56.624833138\\
-163.283862761	51.716487225\\
-163.803264723	46.751251508\\
-164.131488703	41.858002009\\
-164.271837808	37.380804272\\
-164.510999242	34.844254482\\
};
\addlegendentry{Resulting trajectory};

\draw [->] (150,72) -- (100,90);
\draw [->] (-150,58) -- (-150,42);

\end{axis}
\end{tikzpicture}%

%% file: chapters/paper2/paper2.tex
       \paper[Two-Degree-of-Freedom Control for... DMPs]{Two-Degree-of-Freedom Control for Trajectory Tracking and Perturbation Recovery during Execution of Dynamical Movement Primitives}
    \authors{Martin Karlsson \and Fredrik Bagge Carlson \and Anders Robertsson \and Rolf Johansson}
\begin{abstract}
Modeling of robot motion as dynamical movement primitives (DMPs) has become an important framework within robot learning and control. The ability of DMPs to adapt online with respect to the surroundings, \textit{e.g.}, to moving targets, has been used and developed by several researchers. In this work, a method for handling perturbations during execution of DMPs on robots was developed. Two-degree-of-freedom control was introduced in the DMP context, for reference trajectory tracking and perturbation recovery. Benefits compared to the state of the art were demonstrated. The functionality of the method was verified in simulations and in real-world experiments.
\end{abstract}
    \vfill
    Accepted for the IFAC 2017 World Congress, July 9--14, Toulouse, France. Reprinted with permission.
    \newpage

\section{Introduction}
Industrial robots have mostly operated in structured, predictable, environments through sequential execution of predefined motion trajectories. This implies high cost for engineering work, consisting of robot programming and careful work-space preparation. It also limits the range of tasks that are suitable for robots. Improving their ability to operate in unstructured environments with unforeseen events is therefore an important field of research.

This has motivated the development of dynamical movement primitives (DMPs), that are used to model and execute trajectories with an emphasis on online modification.  Early forms were presented in \cite{ijspeert2002humanoid,ijspeert2003learning,schaal2000nonlinear}, and a review can be found in \cite{ijspeert2013dynamical}. The framework has been widely used by robot researchers. For instance, the ability to generalize demonstrated trajectories toward new, although static, goal positions has been used in \cite{niekum2015learning}. Online modulation with respect to a moving goal has been applied in \cite{prada2014handover} for object handover. A method to modify DMP parameters by demonstration has been presented in \cite{karlsson2017autonomous}. Learning and adaptation based on force/torque measurements has been explored in, \textit{e.g.}, \cite{abu2015adaptation,pastor2013dynamic}. Previous work on DMP perturbation recovery in particular is elaborated on in \cref{sec:sota}.

In the standard form, without temporal coupling, a DMP would continue its time evolution regardless of any significant perturbation, as discussed in \cite{ijspeert2013dynamical}. Therefore, its behavior after the perturbation would likely be undesirable and not intuitive.

\begin{figure}
\centering
\includegraphics[width=\columnwidth ]{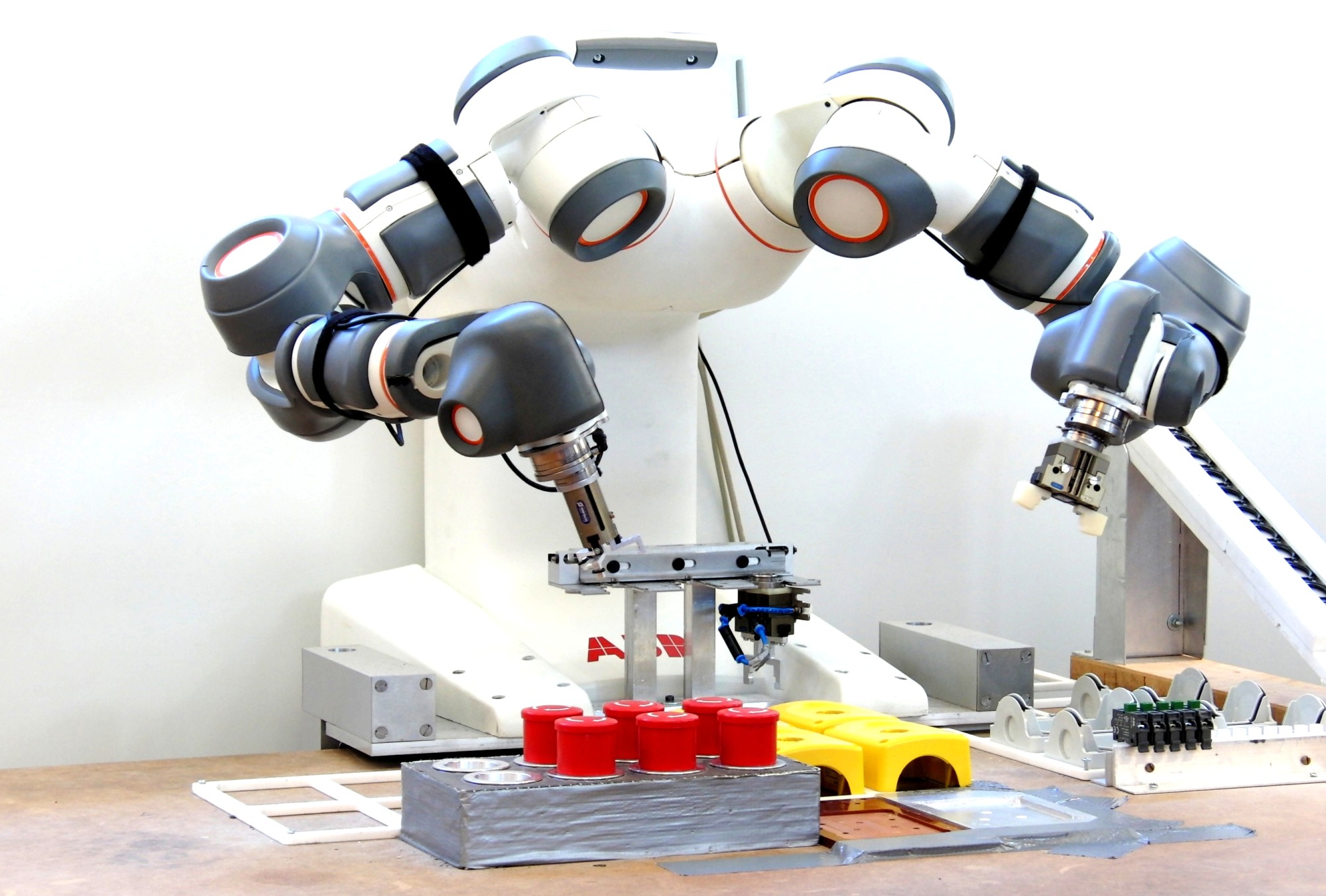}
\caption{The ABB YuMi robot prototype used in the experiments, \cite{yumi}. }\label{fig:yumi5}
\end{figure}

The research described in this paper addressed perturbation recovery for DMPs, and a method was developed where a two-degree-of-freedom controller was integrated with the DMP framework, see, \textit{e.g.}, \cite{aastrom2013computer} for an introduction to the two-degree-of-freedom control structure. The feedforward part of the controller promoted tracking of the DMP trajectory in the absence of significant perturbations, thus mitigating unnecessarily slow trajectory evolution due to temporal coupling acting on small tracking errors. The feedback part suppressed significant errors. The functionality of this method was verified in simulations, as well as in experiments in a real-time robot application. The robot used for experimental evaluation is shown in Fig. \ref{fig:yumi5}.

A code example is available on \cite{dmp_matlab}, to allow exploration of the system proposed. The system was also integrated in the Julia DMP package on \cite{dmp_julia}, originally based on \cite{ijspeert2013dynamical}.

\newpage

\section{Preliminaries}
\label{sec:preliminaries}

\subsection{Dynamical movement primitives}
\label{sec:dmp}
A review of the DMP concept for robotics has been presented in~\cite{ijspeert2013dynamical}, and here follows a condensed description of the fundamentals. A trajectory, $y$, is modeled by the system
\begin{equation}
\tau^2 \ddot{y} = \alpha_z(\beta_z(g-y)-\tau \dot{y}) + f(x)
\label{eq:ddoty}
\end{equation}
Here, $\tau$ is a time constant, $\alpha_z$, $\beta_z$ and $\alpha_x$ are positive constants, and $x$ is a scalar phase parameter that evolves as
\begin{equation}
\tau \dot{x} = -\alpha_x x
\label{eq:x}
\end{equation}
Equation (\ref{eq:ddoty}) is commonly written in the following equivalent form.
\begin{align}
\tau \dot{z} &= \alpha_z(\beta_z(g-y)-z) + f(x) \label{eq:zdot} \\
\tau \dot{y} &= z \label{eq:getting_ydot}
\end{align}
In \cref{eq:ddoty,eq:zdot}, $f(x)$ is given by
\begin{equation}
f(x) = \frac{\sum_{i=1}^{N_b} w_i\Psi_i(x)}{\sum_{i=1}^{N_b} \Psi_i(x)} x \cdot (g-y_0)
\label{eq:f}
\end{equation}
where the basis functions, $\Psi_i(x)$, are determined as
\begin{equation}
\Psi_i(x) = \exp \left(-\frac{1}{2\sigma_i^2}(x-c_i)^2 \right)
\label{eq:psi}
\end{equation}
Here, $N_b$ is the number of basis functions, $w_i$ is the weight for basis function $i$, $y_0$ is the starting point of the trajectory $y$, and $g$ denotes the goal state; $\sigma_i$ and $c_i$ are the width and center of each basis function, respectively. Based on the dynamical system in \cref{eq:getting_ydot,eq:zdot}, a robot trajectory could be generated. Vice versa, given a demonstrated trajectory, $y_{\text{demo}}$, a corresponding DMP could be formed. The goal point $g$ would then be given by the end position of $y_{\text{demo}}$, whereas $\tau$ could be set to get a desired time scale. Further, the weights could be determined by, \textit{e.g.}, locally weighted linear regression, see \cite{atkeson1997locally, schaal1998constructive}, with the solution
\begin{align}
w_i &= \frac{\boldsymbol{s}^T \boldsymbol{\Gamma}_i \boldsymbol{f}_{\text{target}}}{\boldsymbol{s}^T \boldsymbol{\Gamma}_i \boldsymbol{s}}
\intertext{where}
\boldsymbol{s} &= 
\begin{pmatrix}
x^1 (g - y_{\text{demo}}^1) \\
x^2 (g - y_{\text{demo}}^1) \\
\vdots \\
x^{N} (g - y_{\text{demo}}^1)
\end{pmatrix}
\\[2mm]
\boldsymbol{\Gamma_i} &= \text{diag}(\Psi_i^1, \Psi_i^2 \cdots \Psi_i^N) 
\\
f_{\text{target}} &= \tau^2 \ddot{y}_{\text{demo}} - a_z(b_z(g-y_{\text{demo}}) - \tau \dot{y}_{\text{demo}})
\end{align}
Here, $N$ is the number of samples in the demonstrated trajectory.

\subsection{Related work on DMP perturbation recovery}
\label{sec:sota}
We here consider the case where a disturbance is introduced, such that the actual trajectory, denoted $y_a$, evolves differently from $y$, where $y$ evolves according to \cref{eq:ddoty} to \cref{eq:psi}. Without any coupling terms, the time evolution of \cref{eq:x,eq:f} would be unaffected by a perturbation. This behavior is undesired, since it is then likely that the actual trajectory $y_a$ deviates significantly from the intended trajectory even after the cause of the perturbation has vanished. This is more thoroughly described in~\cite{ijspeert2002humanoid,ijspeert2013dynamical}. To mitigate this problem, the solution described in the following paragraph has been suggested in \cite{ijspeert2013dynamical}.

The following coupling terms were introduced.
\begin{align}
\dot{e} &= \alpha_e(y_a - y_c - e) 
\label{eq:edot} \\
C_t &= k_t e \\
\tau_a &= 1 + k_c e^2
\label{eq:taua}
\end{align}
Here, $\alpha_e$, $k_t$ and $k_c$ are constant parameters. The parameter $\tau_a$ was used to determine the evolution rate of the entire dynamical system. Further, the term $C_t$ was added to \cref{eq:zdot} so that the coupled version of $y$, denoted $y_c$, fulfilled the following.
\begin{align}
\tau_a \dot{z} &= \alpha_z(\beta_z(g-y_c)-z) + f(x) + C_t
\label{eq:zc} \\
\tau_a \dot{y_c} &= z
\label{eq:zc_to_ycdot}
\end{align}
A PD controller, given by
\begin{equation}
\ddot{y}_r = K_p(y_c-y_a) + K_v(\dot{y_c} - \dot{y}_a)
\label{eq:sota_ddoty}
\end{equation}
was used to drive $y_a$ to $y$. Here, $\ddot{y}_r$ denotes the reference acceleration, while $K_p$ and $K_v$ are control gains.

This approach from previous research has taken several important parts of disturbance recovery into account, and it should be emphasized that it forms the foundation of this presented work. In this section, however, some aspects are considered where there is room for improvement.

Denote by $y_u$ an unperturbed trajectory generated by an uncoupled DMP, as described in Sec. \ref{sec:dmp}. It is desirable that, in the absence of significant perturbations, $y_a$ should follow $y_c$ closely. If this would not be achieved, in addition to the deviation itself, $y_a$ and $y_c$ would be slowed down, compared to $y_u$, due to the temporal coupling in \cref{eq:taua}. This phenomenon is visualized in \cref{fig:slowgain}. In \cite{ijspeert2013dynamical}, very high controller gains for \cref{eq:sota_ddoty} were suggested, which would have mitigated the issue under ideal conditions and unlimited magnitude of the control signals. Specifically, $K_p=1000$ and $K_v=125$ were chosen. However, even for moderate perturbations, this would imply control signals too large to be realized practically. For instance, a position error in Cartesian space of \SI{1}{\dm} would yield $\ddot{y}_r = \SI{100}{\m / s^2}$. In \cref{fig:sota_stop,fig:sotamove}, two example scenarios are displayed; one where the actual movement was stopped, and one where it was moved away from the nominal path. The method described in \cite{ijspeert2013dynamical} was used for recovery, with prohibitively large values of $\ddot{y}_r$ as a consequence. Moreover, this control system is sensitive to noise and has a dangerously low delay margin of \SI{12}{ms}.

\begin{figure}
	\centering	
	\setlength{\figurewidth}{0.75\linewidth}
	\setlength{\figureheight}{2cm}
	\footnotesize
	\input{chapters/paper2/figs/sota_stop.tex}
	\caption{Simulated trajectories, where $y_a$ was subjected to a stopping perturbation from \SI{2}{s} to \SI{3}{s}, using the approach in \cite{ijspeert2013dynamical}. When $y_a$ was stopped, the evolution of $y_c$ slowed down, and when $y_a$ was released, it was driven to $y_c$ and then behaved like a delayed version of $y_u$. This behavior was desired. However, a prohibitively large accelereration $\ddot{y}_r$ was generated.} \label{fig:sota_stop}
\end{figure}
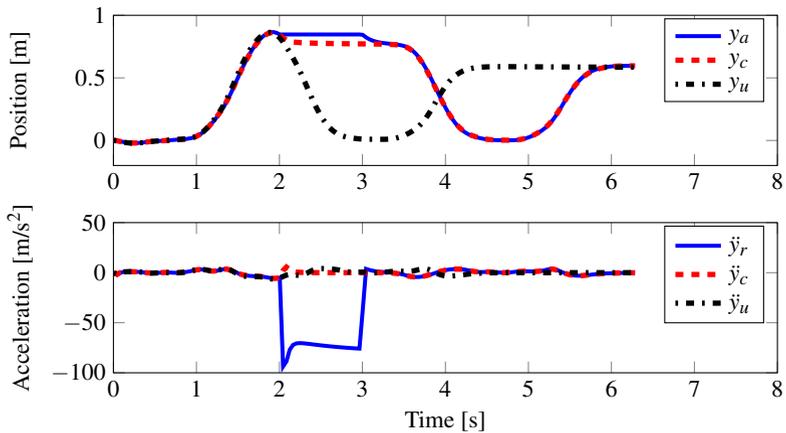

\begin{figure}
	\centering	
	\setlength{\figurewidth}{0.75\linewidth}
	\setlength{\figureheight}{2cm}
	\footnotesize
	\input{chapters/paper2/figs/sotamove.tex}
	\caption{Similar to Fig. \ref{fig:sota_stop}, except that $y_a$ was moved away from the nominal path between \SI{2}{s} to \SI{3}{s}. Again, a prohibitively large accelereration $\ddot{y}_r$ was generated.} \label{fig:sotamove}
\end{figure}
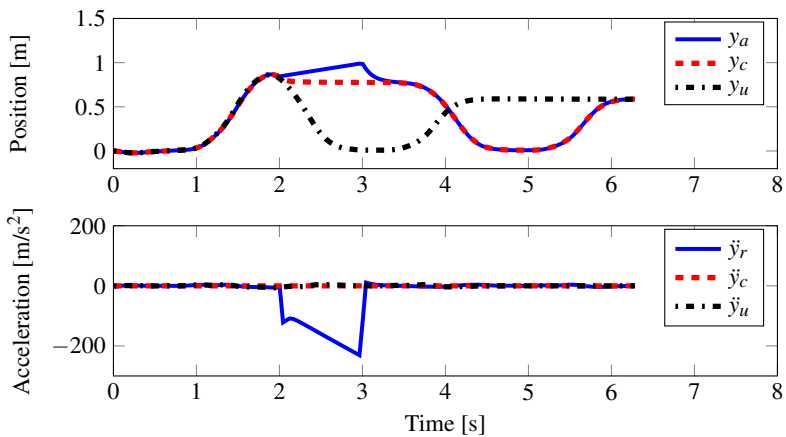

Feedforward control has been used in the DMP context previously, but then only for low-level joint control, with motor torque commands as control signals, see \cite{pastor2009learning,park2008movement}. This control structure was also applied in the internal controller used in the implementation in this present paper, see \cref{sec:implementation}. This inner control design should not be confused with the feedforward control described in \cref{sec:method}, which operated outside the internal robot controller, and was used to determine the reference acceleration for the robot.

\section{Problem Formulation}
\label{sec:problemformulation}
In this paper, we address the question of whether perturbations of DMPs could be recovered from, while fulfilling the following requirements. Only moderate control signals must be used. The benefits of the DMP framework described in \cite{ijspeert2013dynamical}, \textit{i.e.}, scalability in time and space as well as guaranteed convergence to the goal $g$, must be preserved. Further, in  the absence of significant perturbations, the behavior of $y_a$ should resemble that of the original DMP framework described in \cref{sec:dmp}.

\section{Method}
\label{sec:method}
Our proposed method extends that in \cite{ijspeert2013dynamical} as follows. The PD controller in \cref{eq:sota_ddoty} was augmented with feedforward control, as shown in \cref{eq:our_ddoty}. Further, the PD controller gains were moderate, to get a practically realizable control signal. Additionally, the time constant $\tau$ was introduced as a factor in the expression for the adaptive time parameter $\tau_a$, see \cref{eq:taua,eq:our_taua}. Our method is detailed below.

\begin{figure}
	\centering
    \input{chapters/paper2/figs/coupling_scheme.tex}
    \caption{Schematic overview of the control structure described in Sec. \ref{sec:method}. The block denoted 'Robot' includes the internal controller of the robot.}
\label{fig:coupling_scheme}
\end{figure}
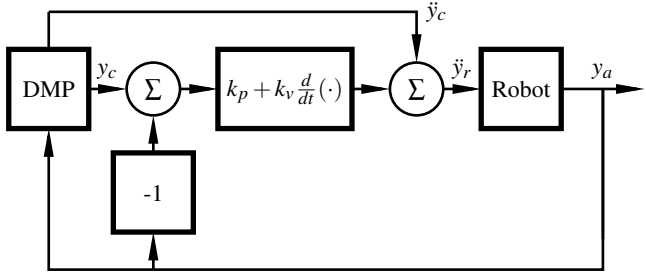

In order for $y_a$ to follow $y_c$, we applied the following control law.
\begin{equation}
\ddot{y}_r = k_p(y_c-y_a) + k_v(\dot{y}_c - \dot{y}_a) + \ddot{y}_c
\label{eq:our_ddoty}
\end{equation}
Here, $\ddot{y}_c$ was obtained by feedforwarding the acceleration of $y_c$. This allowed the controller to act also for zero position- and velocity errors. In turn, the trajectory tracking worked also for moderate controller gains; $k_p = 25$ and $k_v = 10$ are used throughout this paper. With these gains, the closed control loop had a double pole in \linebreak {-5}~rad/s. Since the real parts were negative, the system was asymptotically stable, and since the imaginary parts were 0, it was critically damped. The delay margin was \SI{130}{ms}, which was an improvement compared to \SI{12}{ms} for the previous method, described in \cref{sec:sota}. A~schematic overview of the control system is shown in \cref{fig:coupling_scheme}.

Further, \cref{eq:taua} was modified in order to include the nominal time constant $\tau$, as follows.
\begin{equation}
\tau_a = \tau (1 + k_c e^2)
\label{eq:our_taua}
\end{equation}
The coupling term $C_t$ was omitted in this present method. This choice is elaborated on in \cref{sec:discussion}.

Since $\tau_a$ was not constant over time, determining $\ddot{y}_c$ was more involved than determining $\ddot{y}$ by differentiating \cref{eq:getting_ydot}. One option would be to approximate $\ddot{y}_c$ by discrete-time differentiation of $\dot{y}_c$. However, instead we determined the instantaneous acceleration analytically as follows.

\begin{equation}
\ddot{y}_c = \frac{d}{dt}(\dot{y}_c) = \frac{d}{dt}\left(\frac{z}{\tau_a}\right) = \frac{\dot{z}\tau_a - z\dot{\tau}_a}{\tau_a^2}
= \frac{\dot{z}\tau_a - 2 \tau k_c z e \dot{e}}{\tau_a^2}
\label{eq:inst_acc}
\end{equation}
where $\dot{z}$ and $\dot{e}$ are given by \cref{eq:zc,eq:edot}, respectively. It is noteworthy that the computation of $\ddot{y}_c$ did not require any first- or second-order time-derivative of any measured signal, which would have required prior filtering to mitigate amplification of high-frequency noise. Similarly, $\dot{y}_c$ was determined by \cref{eq:zc,eq:zc_to_ycdot}. In contrast, the computation of $\dot{y}_a$ was complemented with a low-pass filter, to mitigate amplification of measurement noise.

\section{Simulations}
\label{sec:sim}
Two different perturbations were considered in the following simulations; one where $y_a$ was stopped, and one where it was moved. The perturbations took place from time \SI{2}{\s} to \SI{3}{\s}. The systems were sampled at 250 Hz. The same DMP, yielding the same $y_u$, was used in each trial. The adaptive time parameter $\tau_a$ was determined according to \cref{eq:our_taua} in all simulations, to get comparable time scales. First, the controller detailed in \cite{ijspeert2013dynamical} was applied. Except for the perturbations themselves, the conditions were assumed to be ideal, \textit{i.e.}, no delay and no noise were present. The results are shown in \cref{fig:sota_stop,fig:sotamove}. Despite ideal conditions, prohibitively large accelerations were generated by the controller in both cases.

\begin{figure}
	\centering	
	\setlength{\figurewidth}{0.75\linewidth}
	\setlength{\figureheight}{2cm}
	\footnotesize
	\input{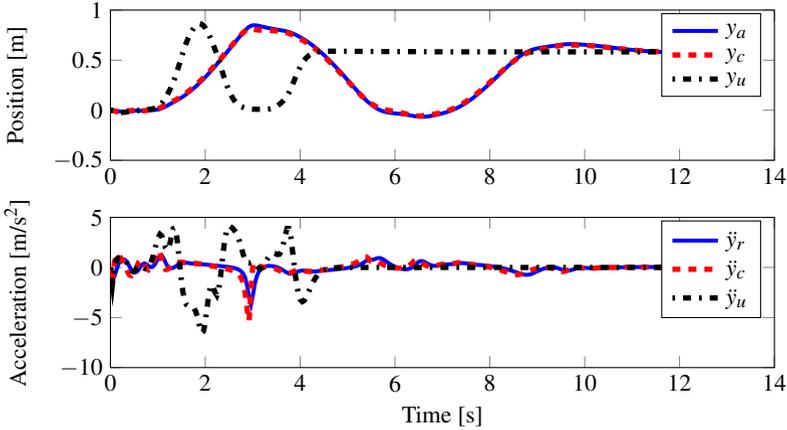}
	\caption{Simulation with the control structure in \cite{ijspeert2013dynamical}, except that the gains were lower ($K_p = 25$ and $K_v = 10$). The reference acceleration was of reasonable magnitude, but the coupled and real systems were slowed down due to small tracking errors combined with temporal coupling.} \label{fig:slowgain}
\end{figure}

\Cref{fig:slowgain} shows the result from a simulation where the controller detailed in \cite{ijspeert2013dynamical} was used, except that the gains were lowered to moderate values. The conditions were ideal, and no perturbation was present. This resulted in reasonable control signals. However, small control errors in combination with the temporal coupling slowed down the evolution of the coupled system as well as the actual movement.

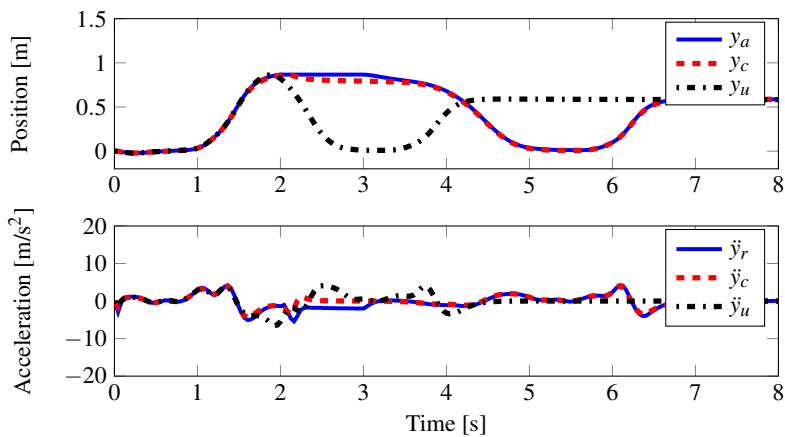
\begin{figure}
	\centering
	\setlength{\figurewidth}{0.75\linewidth}
	\setlength{\figureheight}{2cm}
	\footnotesize
	\input{chapters/paper2/figs/ourstop.tex}
	\caption{Similar to \cref{fig:sota_stop}, but with modeled noise and delay, and using the controller presented in this paper. The behavior was satisfactory both regarding position and acceleration.} \label{fig:ourstop}
\end{figure}

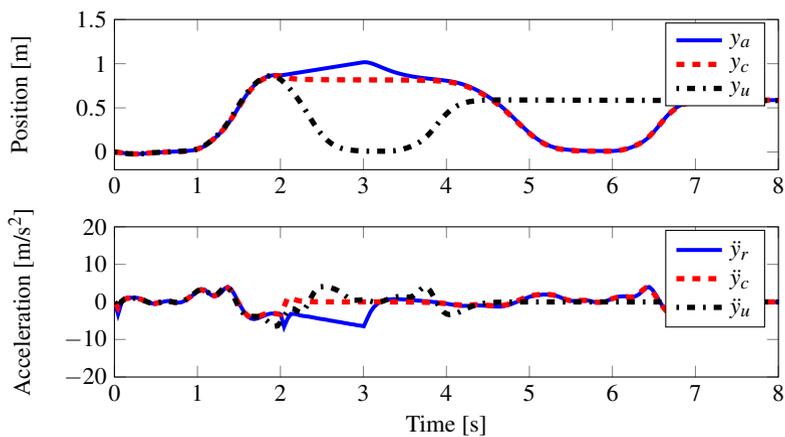
\begin{figure}
	\centering	
	\setlength{\figurewidth}{0.75\linewidth}
	\setlength{\figureheight}{2cm}
	\footnotesize
	\input{chapters/paper2/figs/ourmove.tex}
	\caption{Similar to \cref{fig:ourstop}, except that $y_a$ was moved away from the nominal path between \SI{2}{s} to \SI{3}{s}. Again, the behavior was satisfactory both regarding position and acceleration.} \label{fig:ourmove}
\end{figure}

\begin{figure}
	\centering	
	\setlength{\figurewidth}{0.75\linewidth}
	\setlength{\figureheight}{2cm}
	\footnotesize
	\input{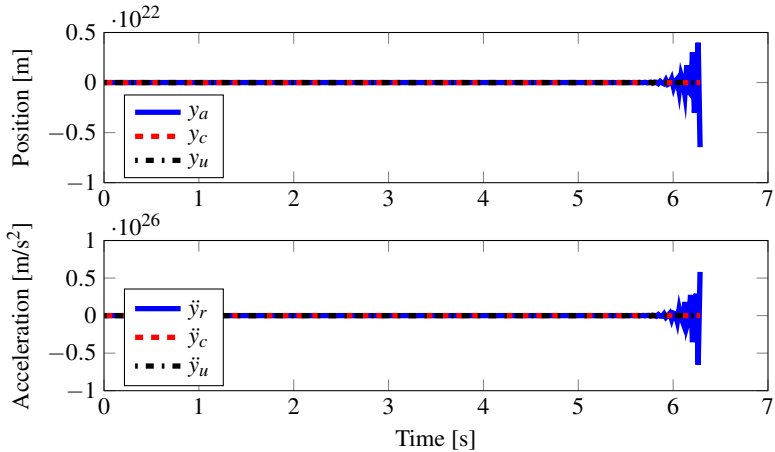}
	\caption{Using the control system in \cite{ijspeert2013dynamical}, subject to the simulated noise and time delay, resulted in unstable behavior.} \label{fig:sota_unstable}
\end{figure}

Thereafter, the controller proposed in this paper, described in \cref{sec:method}, was used. In order to verify robustness under realistic conditions, noise and time delay were introduced. Position measurement noise, and velocity process noise, were modeled as zero mean Gaussian white noise, with standard deviations of \SI{1}{\mm} and \SI{1}{\mm/s}, respectively. Further, an additional configuration dependent forward kinematics error was modeled as a slowly varying position measurement error with standard deviation 1 mm. The time delay between the process and the controller was $L =$ \SI{12}{\ms}. This delay was suitable to simulate since it corresponds both to the delay margin of the method suggested in \cite{ijspeert2013dynamical}, and to the actual delay in the implementation presented in this paper, see \cref{sec:implementation}. (It is, however, a coincidence that these two have the same value. Nevertheless, this shows that a 12 ms delay margin is not necessarily enough.) The results are shown in \cref{fig:ourstop,fig:ourmove}. For comparison, the method in \cite{ijspeert2013dynamical}, with the large gains, was also evaluated under these conditions, although without any perturbation except for the noise. Because of the time delay, this system was unstable, as shown in \cref{fig:sota_unstable}.

\section{Implementation of Real-Time Application}
\label{sec:implementation}
The implementation presented here was performed on a prototype of the dual-arm ABB YuMi robot (previously under the name FRIDA), \cite{yumi}, with 7 joints per arm, see \cref{fig:yumi5}. The method described in \cref{sec:method} was implemented in C++, and the linear algebra library Armadillo, see \cite{sanderson2010armadillo}, was used in a large proportion of the program. The research interface ExtCtrl, \cite{blomdell2005extending,blomdell2010flexible}, was used to send references to the low-level robot joint controller in the ABB IRC5 system, \cite{irc5}. The LabComm protocol, \cite{labcomm}, was used to manage the communication between the C++ program and ExtCtrl. Similar to the simulations, the control system ran at \SI{250}{Hz}, and the delay between process and controller was 3 sample periods, corresponding to \SI{12}{ms}.

\section{Experimental Setup}
\label{sec:experimental_setup}
\begin{figure}[b]
\centering
\includegraphics[width=0.85\columnwidth]{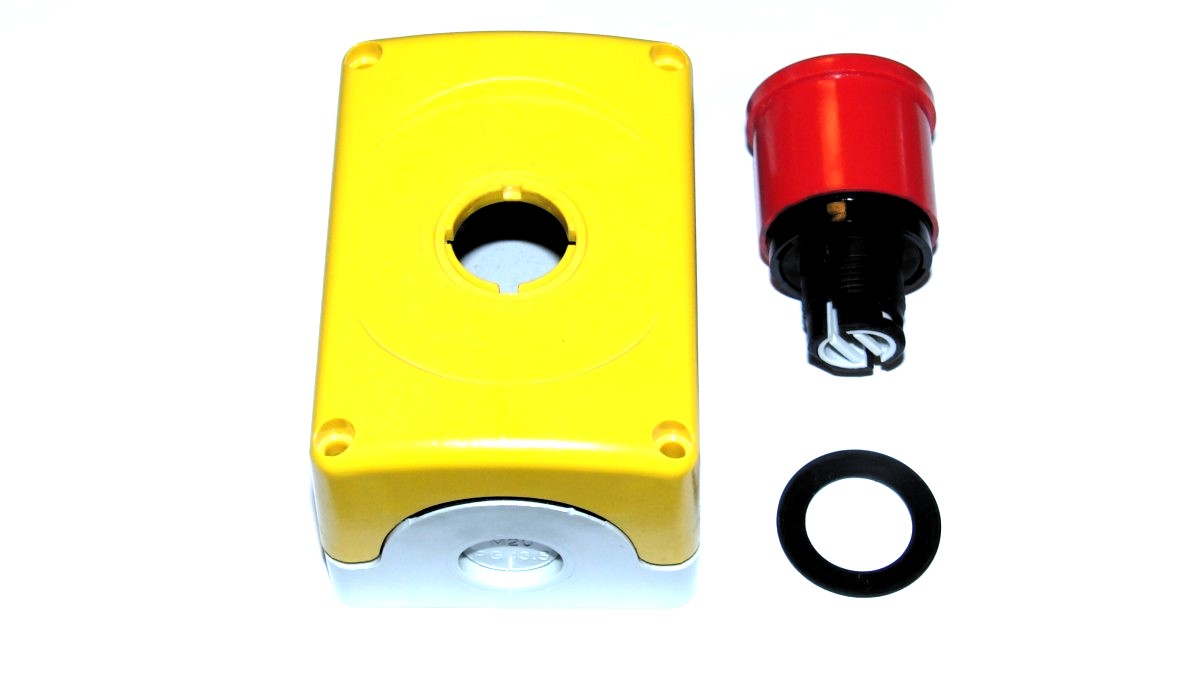}
\caption{Yellow case (left), stop button (upper right) and gasket (lower right) used in the experiments.  }\label{fig:parts}
\end{figure}

The real-time implementation described in \cref{sec:implementation} was used for evaluation. The computations took place in joint space, and the robot's forward kinematics were used for visualization in Cartesian space in the figures presented. The functionality of the method was evaluated in two assembly scenarios. The assembly parts used are shown in \cref{fig:parts}. 

For both scenarios, a new DMP for placing a stop button into the hole of a corresponding case had been taught to the robot by lead-through programming, based on \cite{stolt2015sensorless}, prior to each trial. This implied some variation among the demonstrated trajectories, even though they were qualitatively similar. Subsequently, the DMP was executed on the robot. During the execution, a human perturbed the movement of the robot by physical contact. A wrist-mounted ATI Mini force/torque sensor was used to measure the contact force, and a proportional acceleration, in the same direction as the force, was added to $\ddot{y}_r$ as a load disturbance.

In the first scenario, the human introduced two perturbations during the DMP execution. The first perturbation was formed by moving the end-effector away from its path, and then releasing it. The second perturbation consisted of a longer, unstructured, movement later along the trajectory.

In the second scenario, a human co-worker realized that the stop buttons in the current batch were missing rubber gaskets, and acted to modify the robot trajectory, allowing the co-worker to attach the gasket on the stop button manually. During execution of the DMP, the end-effector was stopped and lifted to a comfortable height by the co-worker. Thereafter, the gasket was attached, and finally the end-effector was released. For the sake of completeness, the modified trajectory was used to form yet another DMP, which allowed the co-worker to attach the gaskets without perturbing the trajectory of the robot, for the remaining buttons in the batch. To verify this functionality, one such modified DMP was executed at the end of each trial.

\begin{figure}
\begin{minipage}{.48\columnwidth}
\label{fig:pert1}
\centering
\includegraphics[height=3cm]{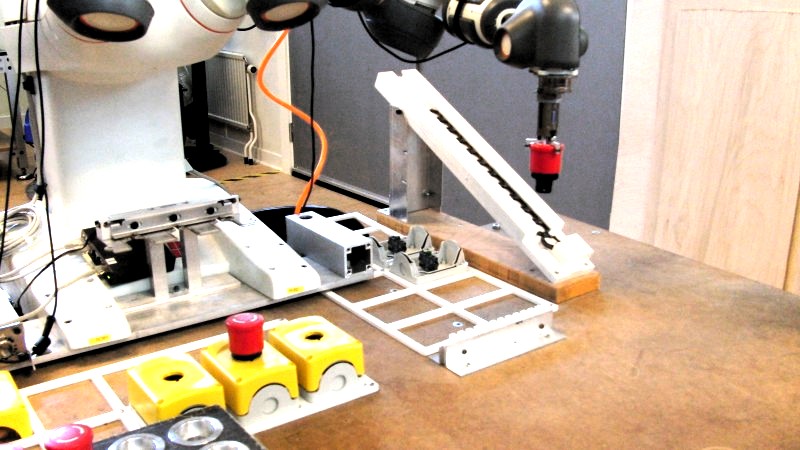}
\subcaption{}
\vspace{2mm}
\end{minipage}
\hfill
\begin{minipage}{.48\columnwidth}
\label{fig:pert2}
\centering
\includegraphics[height=3cm]{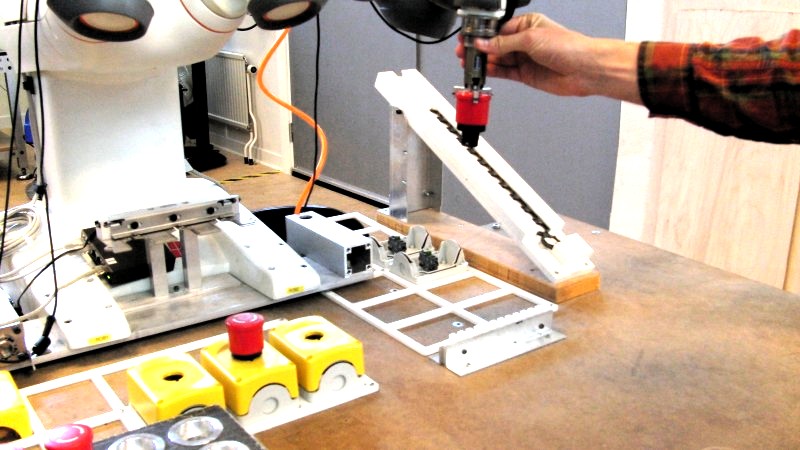}
\subcaption{}
\vspace{2mm}
\end{minipage}
\begin{minipage}{.48\columnwidth}
\label{fig:pert3}
\centering
\includegraphics[height=3cm]{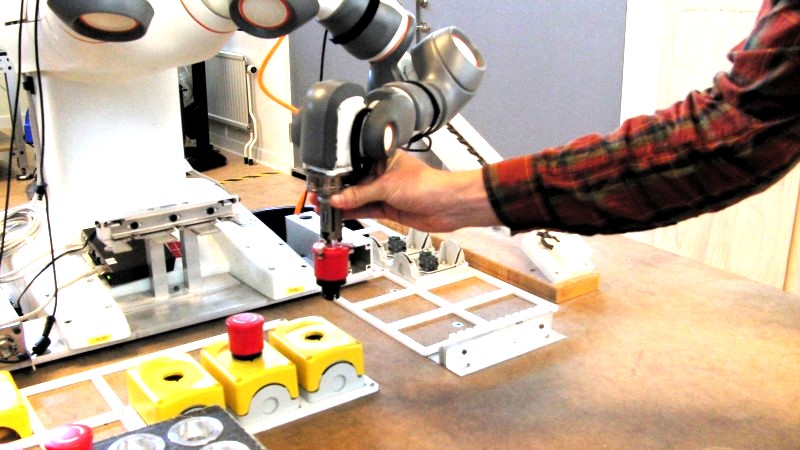}
\subcaption{}
\end{minipage}
\hfill
\begin{minipage}{.48\columnwidth}
\label{fig:pert4}
\centering
\includegraphics[height=3cm]{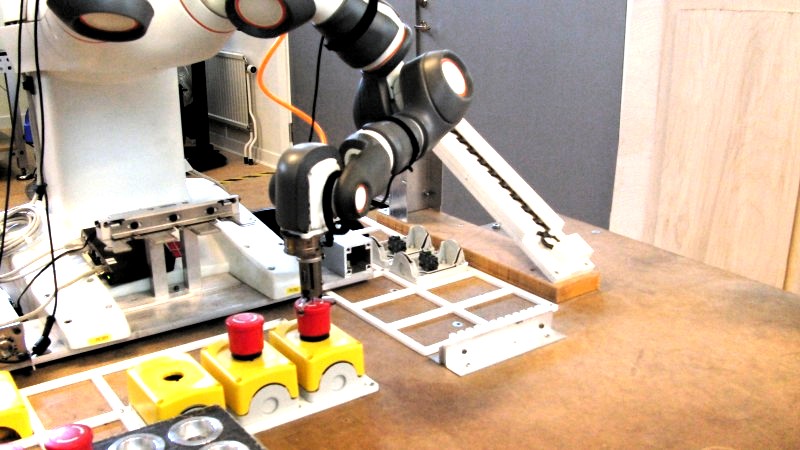}
\subcaption{}
\end{minipage}
\caption{First scenario. In (a), the robot started to execute a DMP for placing the stop button in the rightmost yellow case. A human perturbed the motion twice. The first perturbation (b) was formed by moving the end-effector away from its path, and then releasing it. The second perturbation (c) lasted for a longer time, and consisted of unstructured movement. The robot recovered from both perturbations, and managed to place the stop button in the case (d). Data from one trial are shown in \cref{fig:scen1}.}
\label{fig:perturbation}
\end{figure}

\begin{figure}
\begin{minipage}{.48\columnwidth}
\label{fig:mod1}
\centering
\includegraphics[height=3cm]{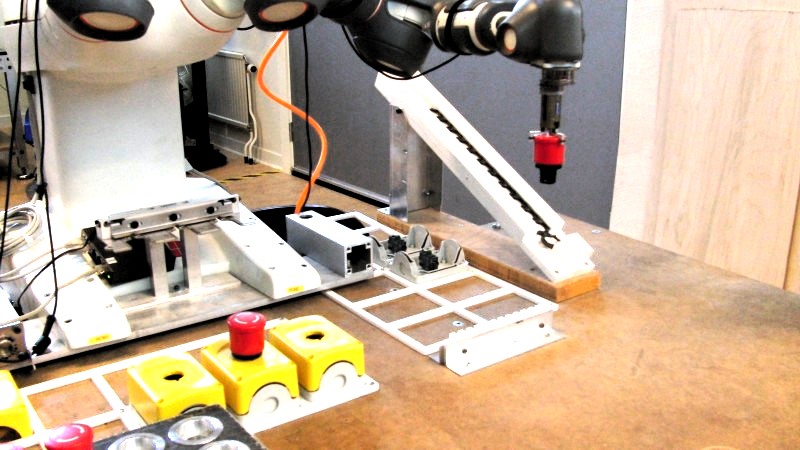}
\subcaption{}
\vspace{2mm}
\end{minipage}
\hfill
\begin{minipage}{.48\columnwidth}
\label{fig:mod2}
\centering
\includegraphics[height=3cm]{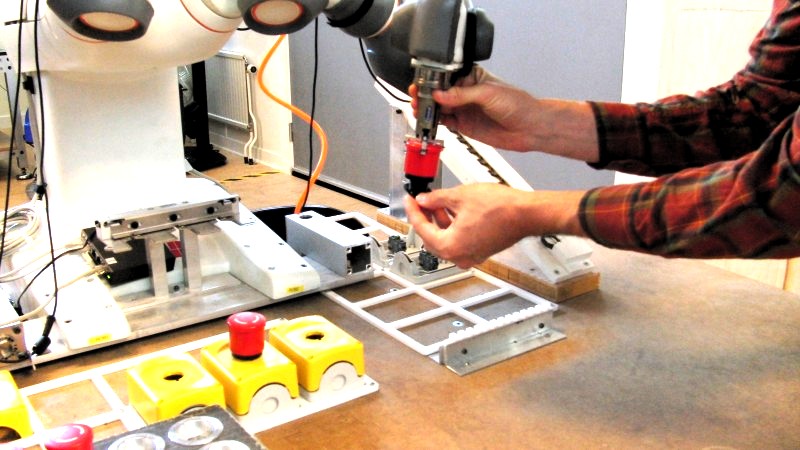}
\subcaption{}
\vspace{2mm}
\end{minipage}
\begin{minipage}{.48\columnwidth}
\label{fig:mod3}
\centering
\includegraphics[height=3cm]{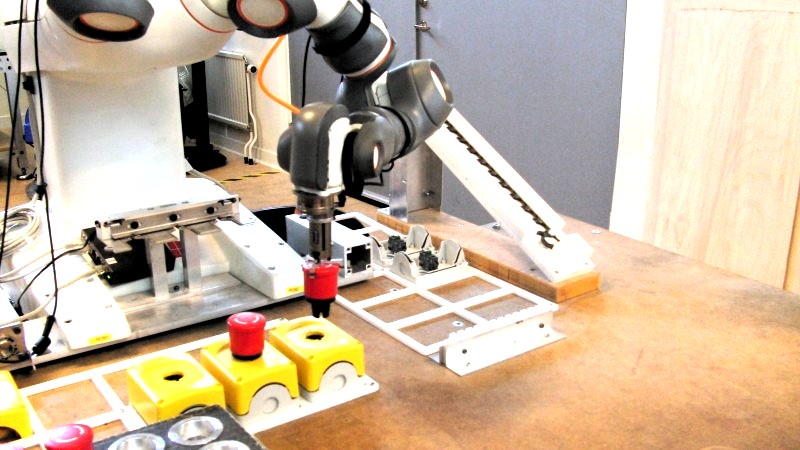}
\subcaption{}
\vspace{2mm}
\end{minipage}
\hfill
\begin{minipage}{.48\columnwidth}
\label{fig:mod4}
\centering
\includegraphics[height=3cm]{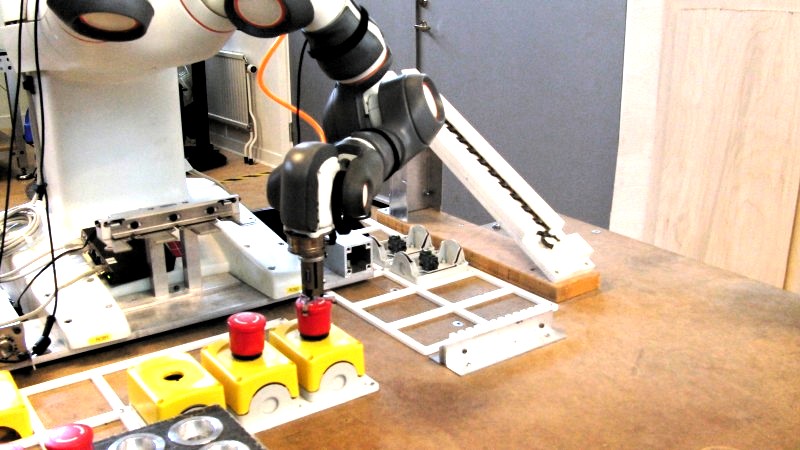}
\subcaption{}
\vspace{2mm}
\end{minipage}
\begin{minipage}{.48\columnwidth}
\label{fig:mod5}
\centering
\includegraphics[height=3cm]{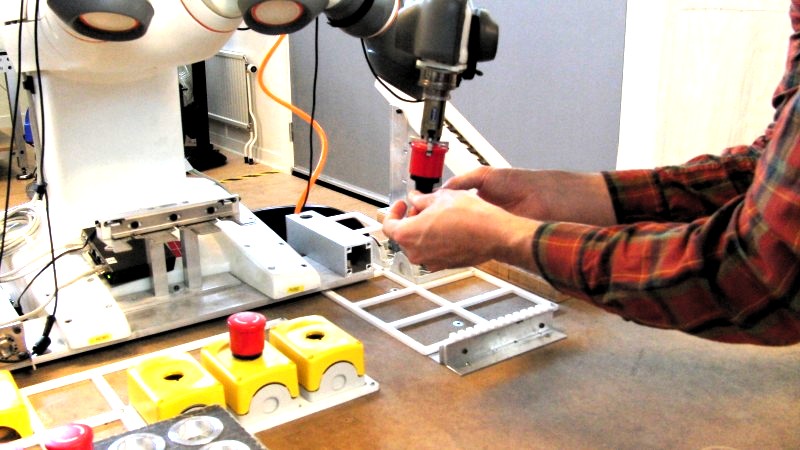}
\subcaption{}
\vspace{2mm}
\end{minipage}
\hfill
\begin{minipage}{.48\columnwidth}
\label{fig:mod6}
\centering
\includegraphics[height=3cm]{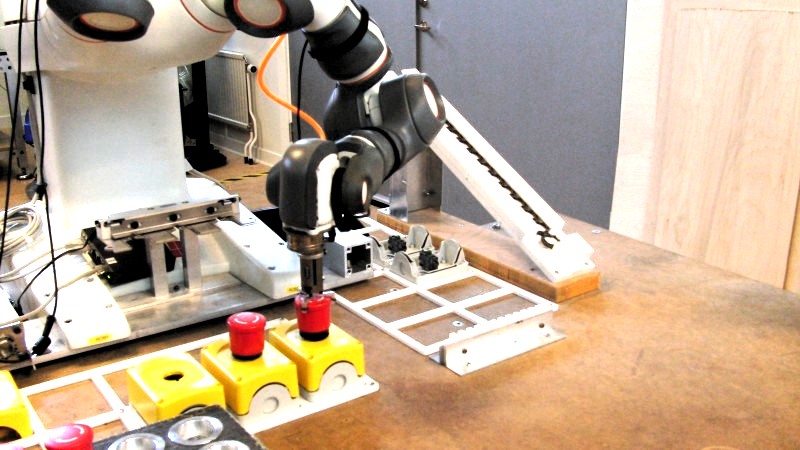}
\subcaption{}
\vspace{2mm}
\end{minipage}
\caption{Second scenario. The robot started its motion toward the rightmost yellow case in (a). The end-effector was stopped and lifted, and the gasket was mounted in (b). The robot was then released, and continued its motion to the case, (c) and (d). The actual trajectory was saved and used to form a modified DMP, and the robot was reset to a configuration similar to that in (a). When executing the modified DMP, the human co-worker could attach the gasket without perturbing the motion of the robot (e). The robot finished the modified DMP in (f). Data from one trial are shown in \cref{fig:scen2}.}
\label{fig:modification}
\end{figure}

The first and second scenarios are visualized in \cref{fig:perturbation,fig:modification}, respectively. To verify repeatability, 50 similar trials were performed for each scenario.

\section{Experimental Results}
\label{sec:experimental_results}
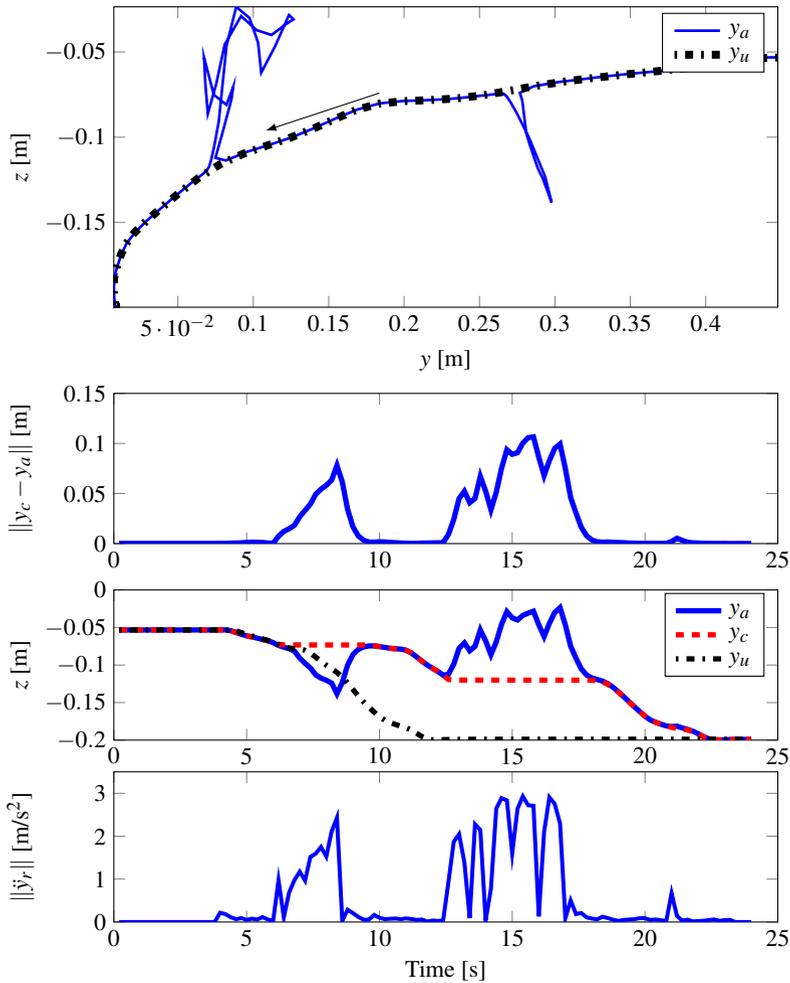
\begin{figure}
	\centering	
	\setlength{\figurewidth}{0.75\linewidth}
	\setlength{\figureheight}{2cm}
	\footnotesize
	\input{chapters/paper2/figs/scen1.tex}
	\caption{Experimental data from a trial of the first scenario. The first (from above) plot shows the path of the end-effector in the Cartesian base frame of the robot, projected on the $yz$-plane. The arrow indicates the movement direction, which started in the upper right and finished in the lower left of the plot. The two perturbations are clearly visible. The second plot shows the distance between $y_a$ and $y_c$ over time. In the third plot, it can be seen that the evolution of $y_c$ slowed down during each perturbation. Subsequently, $y_a$ recovered, and when it was close to $y_c$, the movement continued as a delayed version of $y_u$. The reference acceleration was of reasonable magnitude, as shown in the fourth plot.} \label{fig:scen1}
\end{figure}

Data from a trial of the first scenario are displayed in~\cref{fig:scen1}. The two disturbances were successfully recovered from as intended. The reference acceleration was of reasonable magnitude. The results from all 50 trials were qualitatively mutually similar.

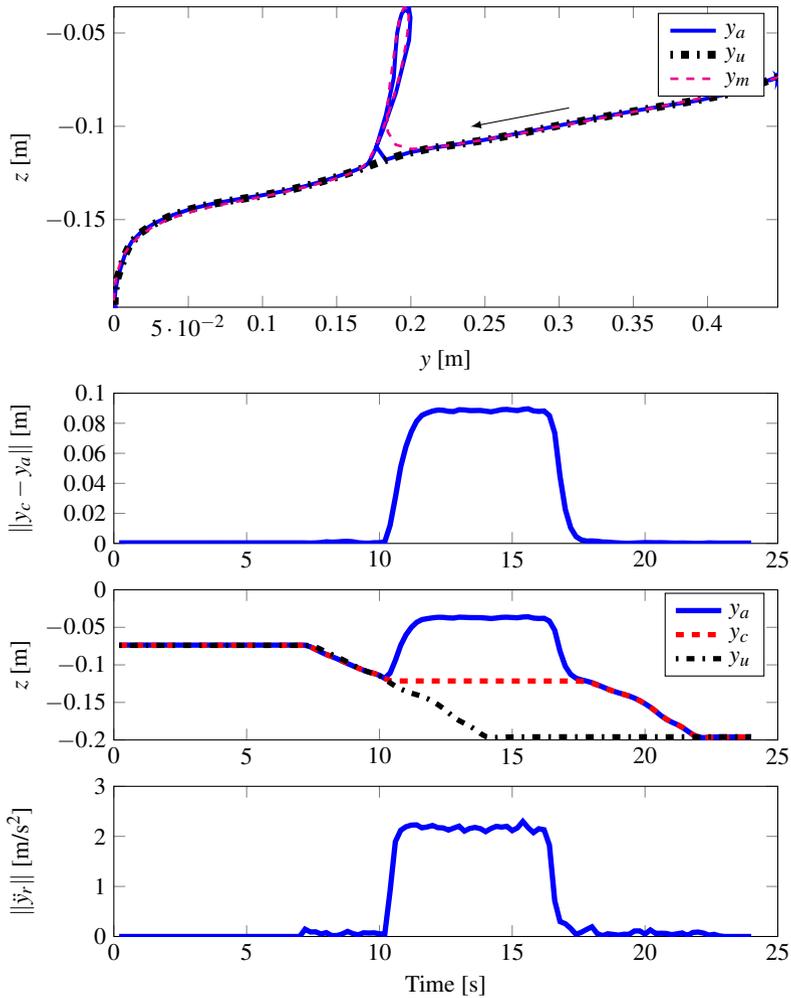
\begin{figure}
	\centering	
	\setlength{\figurewidth}{0.75\linewidth}
	\setlength{\figureheight}{2cm}
	\footnotesize
	\input{chapters/paper2/figs/scen2.tex}
	\caption{Experimental data from a trial of the second scenario. The organization of this figure is similar to that of \cref{fig:scen1}. The perturbation for stopping and lifting the end-effector took place from time \SI{10}{s} to \SI{17.5}{s}, and is clearly visible in each plot. This perturbation was recovered from as intended, and the reference acceleration was of reasonable magnitude. The uppermost plot also displays the measured trajectory obtained by executing the modified DMP, denoted $y_m$. It behaved like a smooth version of the perturbed original trajectory $y_a$.} \label{fig:scen2}
\end{figure}

Data from a trial of the second scenario are displayed in \cref{fig:scen2}. First, the perturbation was successfully recovered from as intended. The reference acceleration was of reasonable magnitude. When the modified DMP was executed, it behaved like a smooth version of the perturbed original trajectory. Again, the results from all 50 trials were qualitatively mutually similar.

To facilitate understanding of the experimental setup and results, a video is publicly available on \cite{pertmovieurl}.
\newpage

\section{Discussion}
\label{sec:discussion}
Compared to previous related research, described in \cite{ijspeert2013dynamical}, the method in this paper contained the following extensions. Feedforward control was added to the PD controller, thus forming a two-degree-of-freedom controller. Further, the PD controller gains were reduced to moderate magnitudes. The expression for $\tau_a$ was also modified, to include the nominal time constant $\tau$ as a factor. These changes resulted in the following benefits, compared to the previous method. The feedforward part allowed the controller to act also for insignificant position- and velocity error, thus improving the trajectory tracking. Because of this, the large controller gains used in \cite{ijspeert2013dynamical}, that were used to mitigate significant tracking errors, could be reduced to moderate magnitudes. In turn, using moderate gains instead of very large ones, resulted in control signals that were practically realizable, instead of prohibitively large. It also improved the delay margin significantly. The aspects above form the main contribution of this paper. In contrast, the modification of the expression for $\tau_a$ was not the main focus of this paper, but it was necessary since it allowed the actual trajectory to converge to the trajectory defined by the DMP, with time constant $\tau$. Without this modification, the time parameter $\tau$ would not have affected the trajectory generated by the DMP. Instead, $\tau_a$ would have converged to 1, regardless of $\tau$, which would not have been desirable.

The work presented here focused on the control structure for trajectory tracking and perturbation recovery, rather than on the perturbations themselves. Even though the perturbations in the experiments considered here emerged from physical contact with a human, the control structure would work similarly for any type of perturbation. There are many other possible perturbations, \textit{e.g.}, a pause of the movement until a certain condition is fulfilled, superpositioned motion control signals to explore the surroundings with a force/torque sensor, a detour to allow line-of-sight between a camera and a part of the work-space, or any other unforeseen deviation from the reference trajectory defined by a DMP. 

It is necessary to implement saturation on the control signals, in order to prevent too large acceleration and velocity for large perturbations. Such boundaries were implemented, but never reached in the experiments in this work.

The coupling term $C_t$ has been introduced in previous research to drive $y_c$ toward $y_a$ when these were different, see \cref{sec:sota}. However, whether this effect is desired, and to what extent, is context dependent. Further, the effect would be mitigated by the temporal coupling, that would slow down the evolution of $y_c$ in \cref{eq:zc,eq:zc_to_ycdot}. Which of these effects that would be dominant in different cases would be difficult to predict intuitively. For these two reasons, the coupling term was not included in the method proposed here, though it would be straight forward to implement. It was, however, included in the simulations where the previous method, described in Sec.~\ref{sec:sota}, was evaluated. During the perturbations in Figs.~\ref{fig:sota_stop} and \ref{fig:sotamove}, the effect of the temporal coupling was dominant, as $y_c$ did not approach $y_a$ significantly.

Apart from the perturbations induced by the human, the motion of the robot was affected by process- and measurement noise. After applying the forward kinematics to determine the position of the end-effector, the accuracy was typically $\pm$ 1 mm. Furthermore, some movement might require higher precision than what would be possible to demonstrate using lead-through programming. Then, \textit{e.g.}, teleoperation could be used for demonstration instead.

In the current implementation, the actual trajectory returned to the reference trajectory, approximately where it started to deviate. This might not always be desired. For instance, it might sometimes be more practical to connect further along the reference trajectory, \textit{e.g.}, after avoiding an obstacle. A lower value of $k_c$ would result in such behavior, however, it must then be known what value of $k_c$ that should be used. Further, one could think of scenarios where it would not be desirable to connect to the reference trajectory, \textit{e.g.}, if a human would modify the last part of the trajectory to a new end point. Hence, future work includes development of a method to determine the desired behavior after a perturbation. \newpage The method presented in this paper would be useful for executing the desired behavior, once it could be determined. Nevertheless, one can think of various scenarios where the recovery presented here would be desirable, such as those in \cref{sec:experimental_setup}.

\section{Conclusion}
\label{sec:conclusions}
In this work, it was shown how perturbations of DMPs could be recovered from, while preserving the characteristics of the original DMP framework in the absence of significant perturbations. Feedforward control was used to track the reference trajectory generated by a DMP. Feedback control with moderate gains was used to suppress deviations. This design is the first, to the best of our knowledge, that takes the following aspects into account. In the absence of significant disturbances, the position error must be small enough, so that the dynamical system would not slow down unnecessarily due to the temporal coupling. Very large controller gains would result in small errors under ideal conditions, but are not practically realizable. On the other hand, if the gains are moderate and only feedback control is used, too large errors occur. 

Feedforward allowed the controller to act even without significant error, which in turn allowed for moderate controller gains.
The suggested method was verified in simulations, and a real-time application was implemented and evaluated, with satisfactory results. A video of the experiments is available on \cite{pertmovieurl}.

\section*{Acknowledgments}
The authors would like to thank Bj{\"o}rn~Olofsson and Fredrik~Magnusson at the Department of Automatic Control, Lund University, as well as Maj~Stenmark, Mathias~Haage and Jacek~Malec at Computer Science, Lund University, for valuable discussions throughout this work. Anthony Remazeilles at Tecnalia, Donostia, and Diogo Almeida at KTH, Stockholm, are gratefully acknowledged for pointing out some of the previous research. The authors are members of the LCCC Linnaeus Center and the ELLIIT Excellence Center at Lund University. The research leading to these results has received funding from the European Commission's Framework Programme Horizon 2020 -- under grant agreement No 644938 -- SARAFun.


%% file: chapters/paper2/figs/sota_stop.tex
%
%
\begin{tikzpicture}

\begin{axis}[%
width=\figurewidth,
height=\figureheight,
ylabel absolute,
scale only axis,
xmin=0,
xmax=8,
ymin=-0.2,
ymax=1,
ylabel={Position [m]},
name=plot1,
legend style={draw=black,fill=white,legend cell align=left}
]
\addplot [color=blue,solid,line width=1.5pt]
  table[row sep=crcr]{%
0.004	0\\
0.044	-0.00232905437719802\\
0.084	-0.00785410448889758\\
0.124	-0.0133909059026196\\
0.164	-0.0176250408214456\\
0.204	-0.0202246762339355\\
0.244	-0.0211987003216157\\
0.284	-0.0206522996359101\\
0.324	-0.0187828986794106\\
0.364	-0.0159474897556613\\
0.404	-0.0126752656603488\\
0.444	-0.0095530420160225\\
0.484	-0.00700441502188302\\
0.524	-0.00510804987790153\\
0.564	-0.00360808838214159\\
0.604	-0.00211041148844772\\
0.644	-0.000322733075829668\\
0.684	0.00179729768267549\\
0.724000000000001	0.00405455359400184\\
0.764000000000001	0.00618876773371098\\
0.804000000000001	0.0081126716961986\\
0.844000000000001	0.0100647050705266\\
0.884000000000001	0.0126121942815933\\
0.924000000000001	0.0165816989340353\\
0.964000000000001	0.0230013797498268\\
1.004	0.0330157117280035\\
1.044	0.0476420052875482\\
1.084	0.0673651228496672\\
1.124	0.0918631649737578\\
1.164	0.120196675920396\\
1.204	0.151403406530137\\
1.244	0.185119700867397\\
1.284	0.221893927221889\\
1.324	0.262986611106964\\
1.364	0.309602268857467\\
1.404	0.361925733670103\\
1.444	0.418709791076907\\
1.484	0.477726649960897\\
1.524	0.536570969680405\\
1.564	0.593209216790925\\
1.604	0.6461623761844\\
1.644	0.694477570211482\\
1.684	0.737597854059099\\
1.724	0.77518532353088\\
1.764	0.80696175068272\\
1.804	0.832589268894963\\
1.844	0.85155329991694\\
1.884	0.863066780542137\\
1.924	0.866156037479039\\
1.964	0.86008798989273\\
2.004	0.846858972787071\\
2.044	0.846858972787071\\
2.084	0.846858972787071\\
2.124	0.846858972787071\\
2.164	0.846858972787071\\
2.204	0.846858972787071\\
2.244	0.846858972787071\\
2.284	0.846858972787071\\
2.324	0.846858972787071\\
2.364	0.846858972787071\\
2.404	0.846858972787071\\
2.444	0.846858972787071\\
2.484	0.846858972787071\\
2.524	0.846858972787071\\
2.564	0.846858972787071\\
2.604	0.846858972787071\\
2.644	0.846858972787071\\
2.684	0.846858972787071\\
2.724	0.846858972787071\\
2.764	0.846858972787071\\
2.804	0.846858972787071\\
2.844	0.846858972787071\\
2.884	0.846858972787071\\
2.924	0.846858972787071\\
2.964	0.846858972787071\\
3.004	0.843837109914562\\
3.044	0.823470553332151\\
3.084	0.80830481199823\\
3.124	0.797379784679224\\
3.164	0.789455861329061\\
3.204	0.783638152273621\\
3.244	0.779272246265719\\
3.284	0.775863758588392\\
3.324	0.773009819486324\\
3.364	0.770329329281511\\
3.404	0.767378910996818\\
3.444	0.763548469770838\\
3.484	0.757955208725162\\
3.524	0.749400873894229\\
3.564	0.736484844222188\\
3.604	0.717892384579584\\
3.644	0.692712512271355\\
3.684	0.660594178353641\\
3.724	0.621727545939995\\
3.764	0.576781509086887\\
3.804	0.526858799283136\\
3.844	0.473423223625389\\
3.884	0.41816718493674\\
3.924	0.362848666674767\\
3.964	0.309137431553693\\
4.004	0.258487227772576\\
4.044	0.212047215925073\\
4.08399999999999	0.170636646496767\\
4.12399999999999	0.134778714682445\\
4.16399999999998	0.104727967462729\\
4.20399999999998	0.0804337397785013\\
4.24399999999998	0.0614755875108854\\
4.28399999999997	0.0470598621667091\\
4.32399999999997	0.0361338891234092\\
4.36399999999996	0.0276133996802454\\
4.40399999999996	0.0206488157048774\\
4.44399999999995	0.0147960054918742\\
4.48399999999995	0.0099907456950543\\
4.52399999999994	0.00635517940661059\\
4.56399999999994	0.00395993018412327\\
4.60399999999994	0.00266254800426591\\
4.64399999999993	0.00209550886752506\\
4.68399999999993	0.00181861836143016\\
4.72399999999992	0.00155630039367909\\
4.76399999999992	0.00135794910697621\\
4.80399999999991	0.00157583540701078\\
4.84399999999991	0.00271388298408358\\
4.88399999999991	0.00527438626727818\\
4.9239999999999	0.00966921928217162\\
4.9639999999999	0.0161760638464908\\
5.00399999999989	0.0249036896998481\\
5.04399999999989	0.0357866614017891\\
5.08399999999988	0.0486715190620128\\
5.12399999999988	0.0635132262799517\\
5.16399999999987	0.0806206922554898\\
5.20399999999987	0.100825689636996\\
5.24399999999987	0.125388203497447\\
5.28399999999986	0.155484827105655\\
5.32399999999986	0.191449241320912\\
5.36399999999985	0.23233777231273\\
5.40399999999985	0.276209163143707\\
5.44399999999984	0.320812505085358\\
5.48399999999984	0.364128283559859\\
5.52399999999983	0.404578929582966\\
5.56399999999983	0.441048920153899\\
5.60399999999983	0.472865062923084\\
5.64399999999982	0.499779826185326\\
5.68399999999982	0.521932124124246\\
5.72399999999981	0.539758583596798\\
5.76399999999981	0.55386556863703\\
5.8039999999998	0.56490191859528\\
5.8439999999998	0.573468106292331\\
5.8839999999998	0.580072109609585\\
5.92399999999979	0.58512148957309\\
5.96399999999979	0.588935055849637\\
6.00399999999978	0.59176105021282\\
6.04399999999978	0.593794609156304\\
6.08399999999977	0.595191703948524\\
6.12399999999977	0.596079209880539\\
6.16399999999976	0.596561781554101\\
6.20399999999976	0.596726449609238\\
6.24399999999976	0.596645743900006\\
6.28399999999975	0.59637993110052\\
};
\addlegendentry{$y_a$};

\addplot [color=red,dashed,line width=2.0pt]
  table[row sep=crcr]{%
0.004	0\\
0.044	-0.00312605637372244\\
0.084	-0.00860630397796379\\
0.124	-0.0137753944847239\\
0.164	-0.0176383151150567\\
0.204	-0.019955234142136\\
0.244	-0.0207359584975827\\
0.284	-0.0200734445554688\\
0.324	-0.0181665710093824\\
0.364	-0.0153821938507908\\
0.404	-0.0122474427787491\\
0.444	-0.00931535368618757\\
0.484	-0.00694611522641168\\
0.524	-0.00515775279179926\\
0.564	-0.00367297378148887\\
0.604	-0.0021239152037738\\
0.644	-0.000271805288423189\\
0.684	0.0018763286715186\\
0.724000000000001	0.00410824551579141\\
0.764000000000001	0.00619077535154867\\
0.804000000000001	0.00809205228891615\\
0.844000000000001	0.0101014924244369\\
0.884000000000001	0.0128138240741553\\
0.924000000000001	0.0170641805425108\\
0.964000000000001	0.0238778036098416\\
1.004	0.0343705168812195\\
1.044	0.0494775232744344\\
1.084	0.0695565338395129\\
1.124	0.0941868697449519\\
1.164	0.122436962301211\\
1.204	0.153464610971588\\
1.244	0.187075617795854\\
1.284	0.223949429045886\\
1.324	0.265350116963133\\
1.364	0.312317190654953\\
1.404	0.364786462837473\\
1.444	0.421348912931743\\
1.484	0.47979019068251\\
1.524	0.537835413997251\\
1.564	0.593599987754415\\
1.604	0.645719566354765\\
1.644	0.693308629027047\\
1.684	0.735831195855455\\
1.724	0.772934865366832\\
1.764	0.804308466473712\\
1.804	0.829571193814298\\
1.844	0.848156754330393\\
1.884	0.859235801198635\\
1.924	0.861850254476609\\
1.964	0.855379044942116\\
2.004	0.840086673172293\\
2.044	0.818820910248653\\
2.084	0.801479878383619\\
2.124	0.79205684855251\\
2.164	0.787043063398955\\
2.204	0.784022009884817\\
2.244	0.781980783125702\\
2.284	0.78047998731019\\
2.324	0.779307238130663\\
2.364	0.778349132089973\\
2.404	0.777540018650384\\
2.444	0.776839362927341\\
2.484	0.776220810914303\\
2.524	0.775666494922086\\
2.564	0.775163875339093\\
2.604	0.774703893939289\\
2.644	0.774279844656335\\
2.684	0.773886656214498\\
2.724	0.773520421226555\\
2.764	0.773178078221546\\
2.804	0.772857191620639\\
2.844	0.772555796222291\\
2.884	0.772272285234163\\
2.924	0.772005328349327\\
2.964	0.771753810954971\\
3.004	0.771516773391199\\
3.044	0.771287314088925\\
3.084	0.77104534645163\\
3.124	0.770772411441065\\
3.164	0.770449804626567\\
3.204	0.770054255066475\\
3.244	0.769552457783211\\
3.284	0.768891238496567\\
3.324	0.767978409074191\\
3.364	0.766646758867917\\
3.404	0.764592758473537\\
3.444	0.761289455979302\\
3.484	0.755900451796858\\
3.524	0.747267051241848\\
3.564	0.734055498118496\\
3.604	0.715057534518888\\
3.644	0.689481249732645\\
3.684	0.657068863350115\\
3.724	0.61806767647359\\
3.764	0.573179904988013\\
3.804	0.523523999140961\\
3.844	0.470556954460587\\
3.884	0.415939549186727\\
3.924	0.361379306638325\\
3.964	0.308485493949665\\
4.004	0.258649350645462\\
4.044	0.212965018042857\\
4.08399999999999	0.172214174980744\\
4.12399999999999	0.136900057047655\\
4.16399999999998	0.107261388705154\\
4.20399999999998	0.0832247063990344\\
4.24399999999998	0.064345682905233\\
4.28399999999997	0.0498234307887712\\
4.32399999999997	0.0386313689852489\\
4.36399999999996	0.0297496779136796\\
4.40399999999996	0.0224154424474559\\
4.44399999999995	0.0162585790918759\\
4.48399999999995	0.0112454786842962\\
4.52399999999994	0.00747981073128393\\
4.56399999999994	0.00498330284034119\\
4.60399999999994	0.00356381147280266\\
4.64399999999993	0.00283227032285265\\
4.68399999999993	0.00237065342005856\\
4.72399999999992	0.00195676692954754\\
4.76399999999992	0.00168884141612409\\
4.80399999999991	0.00193502939407388\\
4.84399999999991	0.00318201152201886\\
4.88399999999991	0.00589956034432419\\
4.9239999999999	0.010466498464579\\
4.9639999999999	0.0171302519261579\\
5.00399999999989	0.0259729072702429\\
5.04399999999989	0.0369160440754446\\
5.08399999999988	0.0498227441140133\\
5.12399999999988	0.0647007025129204\\
5.16399999999987	0.0819357429763882\\
5.20399999999987	0.102424485789472\\
5.24399999999987	0.127418990183048\\
5.28399999999986	0.157967368095752\\
5.32399999999986	0.194195063401233\\
5.36399999999985	0.235002090198917\\
5.40399999999985	0.278436937942376\\
5.44399999999984	0.322354413375857\\
5.48399999999984	0.364869694032878\\
5.52399999999983	0.404515216676291\\
5.56399999999983	0.440252718087217\\
5.60399999999983	0.471462589383904\\
5.64399999999982	0.497932446706093\\
5.68399999999982	0.519815684309707\\
5.72399999999981	0.5375398011891\\
5.76399999999981	0.551681728961\\
5.8039999999998	0.56285023967004\\
5.8439999999998	0.57160587355701\\
5.8839999999998	0.578423991967265\\
5.92399999999979	0.583689353981063\\
5.96399999999979	0.587707003358237\\
6.00399999999978	0.590718258209151\\
6.04399999999978	0.592915839573311\\
6.08399999999977	0.594455922098744\\
6.12399999999977	0.595466894254786\\
6.16399999999976	0.596055429745908\\
6.20399999999976	0.59631064656608\\
6.24399999999976	0.596307026321855\\
6.28399999999975	0.596106580880537\\
};
\addlegendentry{$y_c$};

\addplot [color=black,dash pattern=on 1pt off 3pt on 3pt off 3pt,line width=2.0pt]
  table[row sep=crcr]{%
0.004	0\\
0.044	-0.00313272655695536\\
0.084	-0.00867844462227884\\
0.124	-0.0140079047712132\\
0.164	-0.018091443462256\\
0.204	-0.0206214541016379\\
0.244	-0.0215460958014989\\
0.284	-0.0209199058929612\\
0.324	-0.0189288412516289\\
0.364	-0.015949225187372\\
0.404	-0.0125364827854573\\
0.444	-0.00928587187324214\\
0.484	-0.00660490116139399\\
0.524	-0.00455275981396711\\
0.564	-0.00287791149070371\\
0.604	-0.00121856217740639\\
0.644	0.000675507091553434\\
0.684	0.00281821873137434\\
0.724000000000001	0.00501782179761879\\
0.764000000000001	0.00705430598784005\\
0.804000000000001	0.0089009669828089\\
0.844000000000001	0.0108496495199843\\
0.884000000000001	0.0135027071825912\\
0.924000000000001	0.0177145101636838\\
0.964000000000001	0.0245442194882406\\
1.004	0.0351584542788934\\
1.044	0.0505651174397519\\
1.084	0.0712175715250734\\
1.124	0.0967873452975105\\
1.164	0.126377917375725\\
1.204	0.159107331646396\\
1.244	0.19473414267802\\
1.284	0.233975796668523\\
1.324	0.278229120554815\\
1.364	0.328655313658648\\
1.404	0.385146443575777\\
1.444	0.445966207036838\\
1.484	0.508295240820383\\
1.524	0.569173884000304\\
1.564	0.626243935671721\\
1.604	0.678041809117235\\
1.644	0.723895161403655\\
1.684	0.763610171346251\\
1.724	0.797153637776484\\
1.764	0.824432001191227\\
1.804	0.845132457709113\\
1.844	0.858576312498352\\
1.884	0.863666267137082\\
1.924	0.859141864970853\\
1.964	0.84423425664056\\
2.004	0.819358542274757\\
2.044	0.78622945679985\\
2.084	0.747218369146149\\
2.124	0.704371483894412\\
2.164	0.658650613968891\\
2.204	0.609793434432555\\
2.244	0.556952305088282\\
2.284	0.49981192190462\\
2.324	0.439397521503897\\
2.364	0.377985881712436\\
2.404	0.31832900396078\\
2.444	0.262823213018213\\
2.484	0.213042663253684\\
2.524	0.169700804580705\\
2.564	0.132896193454363\\
2.604	0.102417827002018\\
2.644	0.0779235479195955\\
2.684	0.0589534648286063\\
2.724	0.0448635203897635\\
2.764	0.0347808226458393\\
2.804	0.0276544923332249\\
2.844	0.0224377518846748\\
2.884	0.0183416579056262\\
2.924	0.0149935215607505\\
2.964	0.0123777371182891\\
3.004	0.0106128542915693\\
3.044	0.00971674076270573\\
3.084	0.00948222516471695\\
3.124	0.00952432539596022\\
3.164	0.00948221684459078\\
3.204	0.00924552470914172\\
3.244	0.00903862266841597\\
3.284	0.00932813881865353\\
3.324	0.0106622106472434\\
3.364	0.0135484125036182\\
3.404	0.0183889873633274\\
3.444	0.0254366848064866\\
3.484	0.0347598850355408\\
3.524	0.0462675774782377\\
3.564	0.0598479899864983\\
3.604	0.075604999756669\\
3.644	0.0941007610994917\\
3.684	0.116443112162265\\
3.724	0.144012012790866\\
3.764	0.177774432816555\\
3.804	0.217554868477214\\
3.844	0.261826820226545\\
3.884	0.308163713612062\\
3.924	0.353960674636309\\
3.964	0.397003318091964\\
4.004	0.435737176789083\\
4.044	0.469300324575242\\
4.08399999999999	0.497429417593016\\
4.12399999999999	0.520316345183389\\
4.16399999999998	0.538455166955703\\
4.20399999999998	0.552501266272817\\
4.24399999999998	0.563157691454206\\
4.28399999999997	0.571096227647175\\
4.32399999999997	0.576912126204277\\
4.36399999999996	0.581105440285701\\
4.40399999999996	0.584080070596744\\
4.44399999999995	0.586152698548648\\
4.48399999999995	0.587565979059748\\
4.52399999999994	0.588502507589093\\
4.56399999999994	0.589097701949801\\
4.60399999999994	0.589450803534764\\
4.64399999999993	0.589633821927855\\
4.68399999999993	0.58969856458187\\
4.72399999999992	0.589682026713169\\
4.76399999999992	0.589610447505546\\
4.80399999999991	0.589502318627651\\
4.84399999999991	0.58937058997529\\
4.88399999999991	0.589224271641686\\
4.9239999999999	0.589069588266586\\
4.9639999999999	0.588910805289955\\
5.00399999999989	0.588750816931741\\
5.04399999999989	0.588591562448516\\
5.08399999999988	0.588434319428065\\
5.12399999999988	0.58827990952738\\
5.16399999999987	0.588128842171085\\
5.20399999999987	0.587981414485425\\
5.24399999999987	0.587837780485823\\
5.28399999999986	0.587697998746986\\
5.32399999999986	0.587562065070179\\
5.36399999999985	0.587429934727532\\
5.40399999999985	0.587301537490166\\
5.44399999999984	0.587176787676088\\
5.48399999999984	0.587055590769329\\
5.52399999999983	0.586937847680388\\
5.56399999999983	0.586823457379866\\
5.60399999999983	0.58671231839948\\
5.64399999999982	0.586604329527232\\
5.68399999999982	0.586499389904842\\
5.72399999999981	0.58639739865048\\
5.76399999999981	0.586298254067992\\
5.8039999999998	0.586201852457742\\
5.8439999999998	0.586108086509405\\
5.8839999999998	0.586016843230676\\
5.92399999999979	0.585928001346959\\
5.96399999999979	0.585841428096486\\
6.00399999999978	0.585756975345647\\
6.04399999999978	0.585674474965833\\
6.08399999999977	0.585593733454047\\
6.12399999999977	0.585514525856706\\
6.16399999999976	0.585436589185521\\
6.20399999999976	0.585359615715103\\
6.24399999999976	0.585283246843323\\
6.28399999999975	0.585207068589478\\
};
\addlegendentry{$y_u$};

\end{axis}

\begin{axis}[%
width=\figurewidth,
height=\figureheight,
scale only axis,
xmin=0,
xmax=8,
xlabel={Time [s]},
ymin=-100,
ymax=50,
ylabel absolute,
ylabel={$\text{Acceleration [m/s}^\text{2}\text{]}$},
at=(plot1.below south west),
anchor=above north west,
legend style={draw=black,fill=white,legend cell align=left}
]
\addplot [color=blue,solid,line width=1.5pt]
  table[row sep=crcr]{%
0.004	0\\
0.044	-2.24441851435179\\
0.084	-0.035862424953484\\
0.124	0.820763306047547\\
0.164	1.03174337615905\\
0.204	1.02396906238998\\
0.244	0.962759618635294\\
0.284	0.851337737369418\\
0.324	0.641321032406193\\
0.364	0.31442227988053\\
0.404	-0.0664745796055358\\
0.444	-0.359266430430853\\
0.484	-0.432733907090107\\
0.524	-0.278984373595448\\
0.564	-0.0206504307216746\\
0.604	0.18279489056686\\
0.644	0.226687725737236\\
0.684	0.107371203844738\\
0.724000000000001	-0.071524643642649\\
0.764000000000001	-0.151954026945848\\
0.804000000000001	-0.0249582519871137\\
0.844000000000001	0.315157027158943\\
0.884000000000001	0.820306450007415\\
0.924000000000001	1.454648652668\\
0.964000000000001	2.1783121474279\\
1.004	2.85522753847599\\
1.044	3.22158763102927\\
1.084	3.06381824393409\\
1.124	2.46865038448647\\
1.164	1.81422034797335\\
1.204	1.51669890638853\\
1.244	1.80834196890188\\
1.284	2.60703265977347\\
1.324	3.44844419581763\\
1.364	3.67750268465761\\
1.404	2.958806795089\\
1.444	1.55837788563384\\
1.484	0.0154777778075545\\
1.524	-1.29650037891238\\
1.564	-2.25467929862852\\
1.604	-2.87365429490605\\
1.644	-3.23213892827126\\
1.684	-3.44336478771642\\
1.724	-3.61147254527547\\
1.764	-3.80988988770196\\
1.804	-4.11059862495003\\
1.844	-4.58629054277129\\
1.884	-5.21238592845543\\
1.924	-5.73816411366597\\
1.964	-5.75718286608145\\
2.004	-5.06700478207366\\
2.044	-93.7009642420475\\
2.084	-88.6511736821627\\
2.124	-76.9695763256207\\
2.164	-72.1456761217629\\
2.204	-70.6381885162057\\
2.244	-70.343756312808\\
2.284	-70.5021027860357\\
2.324	-70.8335708026792\\
2.364	-71.227501131509\\
2.404	-71.6373208787544\\
2.444	-72.0427154368597\\
2.484	-72.4348253504645\\
2.524	-72.8100188686004\\
2.564	-73.1671132114984\\
2.604	-73.5060769931603\\
2.644	-73.8274043319945\\
2.684	-74.1318072054473\\
2.724	-74.4200637920419\\
2.764	-74.6929451502103\\
2.804	-74.9511817954934\\
2.844	-75.1954506391092\\
2.884	-75.4263721672587\\
2.924	-75.6445125604426\\
2.964	-75.8503879724855\\
3.004	-36.8152704964046\\
3.044	3.76855717955409\\
3.084	2.71788686635777\\
3.124	1.92404866839991\\
3.164	1.3519179604767\\
3.204	0.934495358512967\\
3.244	0.620965820793129\\
3.284	0.368796567796545\\
3.324	0.135573801618412\\
3.364	-0.12953472123723\\
3.404	-0.491278744042585\\
3.444	-1.02017239704219\\
3.484	-1.75501628203586\\
3.524	-2.63958023882015\\
3.564	-3.49488054574666\\
3.604	-4.10545460558698\\
3.644	-4.35900384199538\\
3.684	-4.26823664380957\\
3.724	-3.87688893011137\\
3.764	-3.21018399295597\\
3.804	-2.30790647019272\\
3.844	-1.25124738254762\\
3.884	-0.143928468567283\\
3.924	0.915881214622494\\
3.964	1.84601104812665\\
4.004	2.58649991534846\\
4.044	3.11655122583482\\
4.08399999999999	3.45856460815359\\
4.12399999999999	3.63656769958318\\
4.16399999999998	3.63006101811504\\
4.20399999999998	3.3926342887535\\
4.24399999999998	2.91281659431096\\
4.28399999999997	2.25396588429408\\
4.32399999999997	1.55577589760435\\
4.36399999999996	0.99310931176181\\
4.40399999999996	0.689112150584485\\
4.44399999999995	0.639774315949982\\
4.48399999999995	0.724627166378528\\
4.52399999999994	0.787242513739766\\
4.56399999999994	0.715310514678244\\
4.60399999999994	0.489398602873166\\
4.64399999999993	0.19889933071066\\
4.68399999999993	0.000235734178399283\\
4.72399999999992	0.0100513680093403\\
4.76399999999992	0.223341716904683\\
4.80399999999991	0.542326482037198\\
4.84399999999991	0.864304353834435\\
4.88399999999991	1.13084767992464\\
4.9239999999999	1.31571645131865\\
4.9639999999999	1.39612001809624\\
5.00399999999989	1.36126231883514\\
5.04399999999989	1.25625959126718\\
5.08399999999988	1.20099440161637\\
5.12399999999988	1.35483250611421\\
5.16399999999987	1.84244479711614\\
5.20399999999987	2.6378163747964\\
5.24399999999987	3.44297123451968\\
5.28399999999986	3.75241956608714\\
5.32399999999986	3.22869022885411\\
5.36399999999985	2.01987650711341\\
5.40399999999985	0.582842808824258\\
5.44399999999984	-0.713827682038398\\
5.48399999999984	-1.72938942064424\\
5.52399999999983	-2.45359955174448\\
5.56399999999983	-2.90093486144317\\
5.60399999999983	-3.07999416535371\\
5.64399999999982	-3.01102331045791\\
5.68399999999982	-2.74659134143948\\
5.72399999999981	-2.36746637054747\\
5.76399999999981	-1.95655757014739\\
5.8039999999998	-1.5741180523506\\
5.8439999999998	-1.24990604452632\\
5.8839999999998	-0.989738025773675\\
5.92399999999979	-0.786402913312155\\
5.96399999999979	-0.628293372692969\\
6.00399999999978	-0.504188436723419\\
6.04399999999978	-0.405112368948865\\
6.08399999999977	-0.324580851823097\\
6.12399999999977	-0.258172339550689\\
6.16399999999976	-0.202939284419436\\
6.20399999999976	-0.156887883894814\\
6.24399999999976	-0.118597020218513\\
6.28399999999975	-0.0869725078307657\\
};
\addlegendentry{$\ddot{y}_{r}$};

\addplot [color=red,dashed,line width=2.0pt]
  table[row sep=crcr]{%
0.004	0\\
0.044	-1.56603597321581\\
0.084	0.175880517165133\\
0.124	0.822570263274668\\
0.164	0.973648558492374\\
0.204	0.96668162274975\\
0.244	0.914948826216802\\
0.284	0.803286471370074\\
0.324	0.586057598082003\\
0.364	0.257813568412393\\
0.404	-0.104399930276573\\
0.444	-0.357010944529498\\
0.484	-0.388977449433295\\
0.524	-0.218156530894563\\
0.564	0.0252902423954638\\
0.604	0.194442851817833\\
0.644	0.204767901638717\\
0.684	0.0713027972330856\\
0.724000000000001	-0.0925563756631019\\
0.764000000000001	-0.137295974564125\\
0.804000000000001	0.0238708605171719\\
0.844000000000001	0.382317822087832\\
0.884000000000001	0.893023044497374\\
0.924000000000001	1.52697688344222\\
0.964000000000001	2.2369975331142\\
1.004	2.86662259835423\\
1.044	3.1508704083377\\
1.084	2.92195897178946\\
1.124	2.32330710796244\\
1.164	1.74159899065626\\
1.204	1.55381486277487\\
1.244	1.93727867208821\\
1.284	2.75155517238049\\
1.324	3.49584702621897\\
1.364	3.55957286979308\\
1.404	2.72417650519108\\
1.444	1.32506717571368\\
1.484	-0.134702577069212\\
1.524	-1.34981129253164\\
1.564	-2.23300949699878\\
1.604	-2.80686880643938\\
1.644	-3.15003914785905\\
1.684	-3.36962018783321\\
1.724	-3.55807311544662\\
1.764	-3.782574810103\\
1.804	-4.11587977823202\\
1.844	-4.6222883577051\\
1.884	-5.24673129130318\\
1.924	-5.70665253440826\\
1.964	-5.61724497361801\\
2.004	-4.85746833596581\\
2.044	2.15577864937039\\
2.084	5.55064192008227\\
2.124	2.85350216347064\\
2.164	1.26996732625037\\
2.204	0.625298212801679\\
2.244	0.345410712403937\\
2.284	0.209700600624073\\
2.324	0.137109130843607\\
2.364	0.0950615824677195\\
2.404	0.0691045301836812\\
2.444	0.0522380462173169\\
2.484	0.0408115705734643\\
2.524	0.0327992745469911\\
2.564	0.0270165349190838\\
2.604	0.022739163353244\\
2.644	0.0195073133164403\\
2.684	0.017019300133282\\
2.724	0.0150715404964442\\
2.764	0.0135232521280676\\
2.804	0.0122749909387398\\
2.844	0.0112552016283407\\
2.884	0.0104115588447911\\
2.924	0.00970525529518282\\
2.964	0.00910715026940581\\
3.004	0.00765325450032342\\
3.044	-0.00685524508254026\\
3.084	-0.0182396962457052\\
3.124	-0.0295849313449313\\
3.164	-0.0434659633578435\\
3.204	-0.0629978785024704\\
3.244	-0.0937059259078442\\
3.284	-0.146407227293609\\
3.324	-0.241813944183351\\
3.364	-0.416250536059285\\
3.404	-0.723902193902093\\
3.444	-1.22371635082293\\
3.484	-1.93640026784871\\
3.524	-2.78263754979688\\
3.564	-3.57206271319839\\
3.604	-4.10537019850798\\
3.644	-4.29767654124622\\
3.684	-4.17007648422535\\
3.724	-3.757183022634\\
3.764	-3.07910862683142\\
3.804	-2.17931882409661\\
3.844	-1.14130167125376\\
3.884	-0.0650834234006692\\
3.924	0.956924086400579\\
3.964	1.84741096191288\\
4.004	2.55211926733395\\
4.044	3.05728015890055\\
4.08399999999999	3.38677933625586\\
4.12399999999999	3.55585869162842\\
4.16399999999998	3.53530183333678\\
4.20399999999998	3.28158504128803\\
4.24399999999998	2.79471193195957\\
4.28399999999997	2.1493506457132\\
4.32399999999997	1.48993913044679\\
4.36399999999996	0.982828776590311\\
4.40399999999996	0.728812457311038\\
4.44399999999995	0.702216635992514\\
4.48399999999995	0.777279168735063\\
4.52399999999994	0.808873319468786\\
4.56399999999994	0.70346808886617\\
4.60399999999994	0.460633070406722\\
4.64399999999993	0.18174579030279\\
4.68399999999993	0.0174184313876291\\
4.72399999999992	0.0608575744133442\\
4.76399999999992	0.286505219649124\\
4.80399999999991	0.595297647941834\\
4.84399999999991	0.896489598129019\\
4.88399999999991	1.14205676281217\\
4.9239999999999	1.30796534069839\\
4.9639999999999	1.37082406732219\\
5.00399999999989	1.3255870373478\\
5.04399999999989	1.22824364055817\\
5.08399999999988	1.20427821175628\\
5.12399999999988	1.40709089647248\\
5.16399999999987	1.94072372857763\\
5.20399999999987	2.74087754984484\\
5.24399999999987	3.47210618048449\\
5.28399999999986	3.64555312858397\\
5.32399999999986	3.01080413857455\\
5.36399999999985	1.78766311388923\\
5.40399999999985	0.417148132170475\\
5.44399999999984	-0.791986805371681\\
5.48399999999984	-1.73633680567536\\
5.52399999999983	-2.41103721251046\\
5.56399999999983	-2.8239775963555\\
5.60399999999983	-2.97976813148024\\
5.64399999999982	-2.90010758599124\\
5.68399999999982	-2.64004300259538\\
5.72399999999981	-2.27878979777199\\
5.76399999999981	-1.89318062424829\\
5.8039999999998	-1.53636181358344\\
5.8439999999998	-1.23324650409894\\
5.8839999999998	-0.987903298191761\\
5.92399999999979	-0.793562976095726\\
5.96399999999979	-0.640014486127764\\
6.00399999999978	-0.517540760893876\\
6.04399999999978	-0.418388753112203\\
6.08399999999977	-0.336923457338163\\
6.12399999999977	-0.269246063401089\\
6.16399999999976	-0.212694366227935\\
6.20399999999976	-0.165407984386165\\
6.24399999999976	-0.126012673491519\\
6.28399999999975	-0.0934180728984439\\
};
\addlegendentry{$\ddot{y}_c$};

\addplot [color=black,dash pattern=on 1pt off 3pt on 3pt off 3pt,line width=2.0pt]
  table[row sep=crcr]{%
0.004	0\\
0.044	-1.59954299014054\\
0.084	0.112244330600025\\
0.124	0.779415907514183\\
0.164	0.974340781299854\\
0.204	1.00789613448221\\
0.244	0.981889560654181\\
0.284	0.879638712474928\\
0.324	0.657613876951821\\
0.364	0.312544034052864\\
0.404	-0.0758329244056085\\
0.444	-0.358629876102407\\
0.484	-0.417524254956118\\
0.524	-0.264206835738674\\
0.564	-0.0257165361475424\\
0.604	0.150091393858293\\
0.644	0.173756653586864\\
0.684	0.053778451992183\\
0.724000000000001	-0.10124004023411\\
0.764000000000001	-0.142621185086257\\
0.804000000000001	0.0195788496590049\\
0.844000000000001	0.38188929410743\\
0.884000000000001	0.903616576576362\\
0.924000000000001	1.55773651233641\\
0.964000000000001	2.29808452337993\\
1.004	2.97130090659434\\
1.044	3.31527592263806\\
1.084	3.14885627773208\\
1.124	2.57879902609129\\
1.164	1.97236638978646\\
1.204	1.74674893040044\\
1.244	2.1456580863959\\
1.284	3.0444733206634\\
1.324	3.87834680857241\\
1.364	3.93610841509374\\
1.404	2.9168773676545\\
1.444	1.14358753926628\\
1.484	-0.764403753417956\\
1.524	-2.30597443251475\\
1.564	-3.27206161988472\\
1.604	-3.71687799024603\\
1.644	-3.84059935886702\\
1.684	-3.8525528877986\\
1.724	-3.89478474842458\\
1.764	-4.06423935751359\\
1.804	-4.45687962283568\\
1.844	-5.12915890554929\\
1.884	-5.95601930841547\\
1.924	-6.53180027601709\\
1.964	-6.36993936717497\\
2.004	-5.33484858491606\\
2.044	-3.81215973869931\\
2.084	-2.44277722246786\\
2.124	-1.74955873361838\\
2.164	-1.87761372606672\\
2.204	-2.4633627045808\\
2.244	-2.77458234492599\\
2.284	-2.22377642961008\\
2.324	-0.81810751367034\\
2.364	0.944986001325874\\
2.404	2.50945045890828\\
2.444	3.55271066632738\\
2.484	4.03406092180096\\
2.524	4.10749549652899\\
2.564	3.97765797891656\\
2.604	3.76949623980906\\
2.644	3.49504351763457\\
2.684	3.10903057082453\\
2.724	2.57597537684592\\
2.764	1.91752823255133\\
2.804	1.24161514436171\\
2.844	0.715836816805452\\
2.884	0.458476204716666\\
2.924	0.443889617692125\\
2.964	0.530088973054349\\
3.004	0.560236435709329\\
3.044	0.443916317665101\\
3.084	0.199235153293816\\
3.124	-0.048229934474968\\
3.164	-0.142991794742527\\
3.204	-0.0175968810182347\\
3.244	0.272536434599432\\
3.284	0.620900420695983\\
3.324	0.946160560318816\\
3.364	1.20753590658553\\
3.404	1.37837125443067\\
3.444	1.43259368586402\\
3.484	1.37645822694343\\
3.524	1.2895048949391\\
3.564	1.3197337210566\\
3.604	1.62999041531844\\
3.644	2.30220663159971\\
3.684	3.19951464678653\\
3.724	3.89643692109493\\
3.764	3.88738929490973\\
3.804	2.98700635306512\\
3.844	1.46618110466839\\
3.884	-0.20140742222897\\
3.924	-1.63525048821201\\
3.964	-2.65261452398771\\
4.004	-3.22782464849213\\
4.044	-3.41853565217931\\
4.08399999999999	-3.31468084666273\\
4.12399999999999	-3.01313164538305\\
4.16399999999998	-2.6039818075494\\
4.20399999999998	-2.16071827720884\\
4.24399999999998	-1.73486070086359\\
4.28399999999997	-1.35627483383283\\
4.32399999999997	-1.03756002418565\\
4.36399999999996	-0.779758927640912\\
4.40399999999996	-0.577422640633427\\
4.44399999999995	-0.422268031144218\\
4.48399999999995	-0.305449104685459\\
4.52399999999994	-0.218773868783872\\
4.56399999999994	-0.155230113215969\\
4.60399999999994	-0.10910820459114\\
4.64399999999993	-0.0759164167060615\\
4.68399999999993	-0.0522083282671145\\
4.72399999999992	-0.0353889175410091\\
4.76399999999992	-0.0235326320042473\\
4.80399999999991	-0.015227247463255\\
4.84399999999991	-0.00944681845785937\\
4.88399999999991	-0.00545189279219722\\
4.9239999999999	-0.0027130791075056\\
4.9639999999999	-0.000853580154019766\\
5.00399999999989	0.000393402360879092\\
5.04399999999989	0.00121588659909386\\
5.08399999999988	0.00174578449573312\\
5.12399999999988	0.00207524660316089\\
5.16399999999987	0.00226837041189437\\
5.20399999999987	0.0023695419900939\\
5.24399999999987	0.00240935067914656\\
5.28399999999986	0.00240876482939938\\
5.32399999999986	0.00238206788462978\\
5.36399999999985	0.0023389145947109\\
5.40399999999985	0.00228576502053481\\
5.44399999999984	0.00222687987980816\\
5.48399999999984	0.00216500735019061\\
5.52399999999983	0.00210185313781284\\
5.56399999999983	0.0020383982802707\\
5.60399999999983	0.00197510972168548\\
5.64399999999982	0.00191207494202302\\
5.68399999999982	0.00184908223414411\\
5.72399999999981	0.0017856614549477\\
5.76399999999981	0.00172109543329775\\
5.8039999999998	0.00165440915847103\\
5.8439999999998	0.0015843420668739\\
5.8839999999998	0.0015093080257962\\
5.92399999999979	0.00142734796885444\\
5.96399999999979	0.00133608169853978\\
6.00399999999978	0.00123266840717009\\
6.04399999999978	0.00111379036147059\\
6.08399999999977	0.00097568137134586\\
6.12399999999977	0.000814231380579757\\
6.16399999999976	0.000625210464398114\\
6.20399999999976	0.000404668050803393\\
6.24399999999976	0.000149572112847882\\
6.28399999999975	-0.000141249881787784\\
};
\addlegendentry{$\ddot{y}_u$};

\end{axis}
\end{tikzpicture}%

%% file: chapters/paper2/figs/sotamove.tex
%
%
\begin{tikzpicture}

\begin{axis}[%
width=\figurewidth,
height=\figureheight,
scale only axis,
xmin=0,
xmax=8,
ymin=-0.2,
ymax=1.5,
ylabel absolute,
ylabel={Position [m]},
name=plot1,
legend style={draw=black,fill=white,legend cell align=left}
]
\addplot [color=blue,solid,line width=1.5pt]
  table[row sep=crcr]{%
0.004	0\\
0.044	-0.00233246259619767\\
0.084	-0.00790801182495154\\
0.124	-0.0135906424655786\\
0.164	-0.0180450186677938\\
0.204	-0.0208727674705647\\
0.244	-0.0220139372628951\\
0.284	-0.0215271014968462\\
0.324	-0.0195914587247132\\
0.364	-0.0165712184021968\\
0.404	-0.0130228316297129\\
0.444	-0.00957459807484681\\
0.484	-0.00669866155845976\\
0.524	-0.00451945538799168\\
0.564	-0.00281291650955862\\
0.604	-0.00119512070312189\\
0.644	0.00063708329475463\\
0.684	0.00274940659803003\\
0.724000000000001	0.00497066182297357\\
0.764000000000001	0.00705590811730964\\
0.804000000000001	0.0089225779593786\\
0.844000000000001	0.0108087775727521\\
0.884000000000001	0.0132856250485682\\
0.924000000000001	0.0171949392877268\\
0.964000000000001	0.0235943365166535\\
1.004	0.0336703139206177\\
1.044	0.0484889434958954\\
1.084	0.0685787687178643\\
1.124	0.0936394067117278\\
1.164	0.122717030441281\\
1.204	0.15479865949261\\
1.244	0.189446803039214\\
1.284	0.227134675530866\\
1.324	0.269072593921805\\
1.364	0.316461216265379\\
1.404	0.369525340198863\\
1.444	0.42706639961917\\
1.484	0.48686987875984\\
1.524	0.546488116531058\\
1.564	0.603797495395387\\
1.604	0.657202102245772\\
1.644	0.70563097730756\\
1.684	0.748435546227803\\
1.724	0.785235476410712\\
1.764	0.815768838397808\\
1.804	0.83976735096802\\
1.844	0.856823643058336\\
1.884	0.866275212071096\\
1.924	0.867266198340583\\
1.964	0.859144679141348\\
2.004	0.844724547948453\\
2.044	0.850724547948454\\
2.084	0.856724547948454\\
2.124	0.862724547948455\\
2.164	0.868724547948455\\
2.204	0.874724547948456\\
2.244	0.880724547948456\\
2.284	0.886724547948456\\
2.324	0.892724547948457\\
2.364	0.898724547948457\\
2.404	0.904724547948458\\
2.444	0.910724547948458\\
2.484	0.916724547948459\\
2.524	0.922724547948459\\
2.564	0.92872454794846\\
2.604	0.93472454794846\\
2.644	0.94072454794846\\
2.684	0.946724547948461\\
2.724	0.952724547948461\\
2.764	0.958724547948462\\
2.804	0.964724547948462\\
2.844	0.970724547948463\\
2.884	0.976724547948463\\
2.924	0.982724547948464\\
2.964	0.988724547948464\\
3.004	0.98531750775692\\
3.044	0.927080813349931\\
3.084	0.883762125307048\\
3.124	0.85278325945234\\
3.164	0.830599404148172\\
3.204	0.814671161096688\\
3.244	0.803176015063052\\
3.284	0.794798850957195\\
3.324	0.788580527668587\\
3.364	0.783807051322169\\
3.404	0.779926630804968\\
3.444	0.776485281184632\\
3.484	0.773074219998743\\
3.524	0.769284684192716\\
3.564	0.764668770726979\\
3.604	0.758709400548514\\
3.644	0.750808851732485\\
3.684	0.740310960627246\\
3.724	0.726568562691093\\
3.764	0.709043608364384\\
3.804	0.687387441925205\\
3.844	0.661433488801027\\
3.884	0.631089009134483\\
3.924	0.596215001475896\\
3.964	0.556639771171471\\
4.004	0.512374282911972\\
4.044	0.463908473155385\\
4.08399999999999	0.41235548840749\\
4.12399999999999	0.359332801757604\\
4.16399999999998	0.306680737449715\\
4.20399999999998	0.256177763038518\\
4.24399999999998	0.209331361849355\\
4.28399999999997	0.16726050369647\\
4.32399999999997	0.130677959296093\\
4.36399999999996	0.0999492660768437\\
4.40399999999996	0.0751455063580261\\
4.44399999999995	0.0560304369037656\\
4.48399999999995	0.0420229105078923\\
4.52399999999994	0.0322157073922335\\
4.56399999999994	0.0254933616998726\\
4.60399999999994	0.0207432945820728\\
4.64399999999993	0.0170997614248944\\
4.68399999999993	0.0141034068928615\\
4.72399999999992	0.0116807142099724\\
4.76399999999992	0.00996133905498778\\
4.80399999999991	0.00904432071621664\\
4.84399999999991	0.00883075909875793\\
4.88399999999991	0.00900233163259742\\
4.9239999999999	0.00916800385816969\\
4.9639999999999	0.009100840243422\\
5.00399999999989	0.00889917440875828\\
5.04399999999989	0.00896474865762006\\
5.08399999999988	0.00985221653376101\\
5.12399999999988	0.0121161694241904\\
5.16399999999987	0.0162195640227268\\
5.20399999999987	0.0224836129736625\\
5.24399999999987	0.031046614166225\\
5.28399999999986	0.0418571083000185\\
5.32399999999986	0.0547651542992693\\
5.36399999999985	0.069727853376418\\
5.40399999999985	0.0870636509314935\\
5.44399999999984	0.107623713581963\\
5.48399999999984	0.13268629521339\\
5.52399999999983	0.16342351313775\\
5.56399999999983	0.200130621718432\\
5.60399999999983	0.241805388072754\\
5.64399999999982	0.28644744811951\\
5.68399999999982	0.331747542484491\\
5.72399999999981	0.375617639258864\\
5.76399999999981	0.416397055192101\\
5.8039999999998	0.452881943785997\\
5.8439999999998	0.484325350526078\\
5.8839999999998	0.510442754547636\\
5.92399999999979	0.531387572080561\\
5.96399999999979	0.547661953355238\\
6.00399999999978	0.559973533263781\\
6.04399999999978	0.569086584911203\\
6.08399999999977	0.575712686616016\\
6.12399999999977	0.580455341217851\\
6.16399999999976	0.583797145682192\\
6.20399999999976	0.586110389588467\\
6.24399999999976	0.587676170728017\\
6.28399999999975	0.588703928096262\\
};
\addlegendentry{$y_a$};

\addplot [color=red,dashed,line width=2.0pt]
  table[row sep=crcr]{%
0.004	0\\
0.044	-0.00313264405209774\\
0.084	-0.00867682633102401\\
0.124	-0.0140031847120632\\
0.164	-0.0180854002897918\\
0.204	-0.0206173162560986\\
0.244	-0.0215457962147642\\
0.284	-0.0209236004639331\\
0.324	-0.0189359757770729\\
0.364	-0.0159595017196166\\
0.404	-0.0125494435169922\\
0.444	-0.00929989633302403\\
0.484	-0.00661782935974226\\
0.524	-0.00456385323207541\\
0.564	-0.00288844079274261\\
0.604	-0.00123042051626604\\
0.644	0.000661442932907893\\
0.684	0.00280270853076064\\
0.724000000000001	0.00500260390000517\\
0.764000000000001	0.00704055434450856\\
0.804000000000001	0.00888786474266576\\
0.844000000000001	0.0108339494621362\\
0.884000000000001	0.0134790657893445\\
0.924000000000001	0.0176756856717105\\
0.964000000000001	0.0244800336297711\\
1.004	0.0350504167492637\\
1.044	0.0503696250214311\\
1.084	0.0708355519433688\\
1.124	0.0960435887580705\\
1.164	0.125042634757202\\
1.204	0.1569373811213\\
1.244	0.191461255184107\\
1.284	0.229223854392233\\
1.324	0.271447316316598\\
1.364	0.319175171836931\\
1.404	0.372385125687316\\
1.444	0.429710347856407\\
1.484	0.488939357995618\\
1.524	0.547747027714983\\
1.564	0.604155877759473\\
1.604	0.656686688897622\\
1.644	0.704341971218924\\
1.684	0.746503034187057\\
1.724	0.782784269837734\\
1.764	0.812897614033207\\
1.804	0.836534723636821\\
1.844	0.853235411064301\\
1.884	0.86229037035354\\
1.924	0.862851358300491\\
1.964	0.854367515708611\\
2.004	0.83711531953096\\
2.044	0.814548643126199\\
2.084	0.798303250494728\\
2.124	0.790453698801662\\
2.164	0.786481229080551\\
2.204	0.784150712363132\\
2.244	0.782614553717948\\
2.284	0.781515450945922\\
2.324	0.780681868659778\\
2.364	0.780022163917303\\
2.404	0.779483092582232\\
2.444	0.779031601834224\\
2.484	0.778646052086546\\
2.524	0.778311645947612\\
2.564	0.778017892033901\\
2.604	0.777757121460287\\
2.644	0.77752358055491\\
2.684	0.777312854484029\\
2.724	0.777121488884727\\
2.764	0.776946734273996\\
2.804	0.776786368991559\\
2.844	0.776638573772689\\
2.884	0.776501841100373\\
2.924	0.776374908501789\\
2.964	0.776256708656259\\
3.004	0.776146322889739\\
3.044	0.776039520399873\\
3.084	0.775924665500245\\
3.124	0.775790687803273\\
3.164	0.775625631977263\\
3.204	0.775414237050095\\
3.244	0.775135605858039\\
3.284	0.774760266597877\\
3.324	0.774246117905694\\
3.364	0.77353271110124\\
3.404	0.772533224652991\\
3.444	0.771123431693348\\
3.484	0.76912712726874\\
3.524	0.766298220933281\\
3.564	0.762301629977999\\
3.604	0.756698979740967\\
3.644	0.748950873547749\\
3.684	0.738451767384502\\
3.724	0.724606937660734\\
3.764	0.706933568956783\\
3.804	0.685129844139781\\
3.844	0.659051197025478\\
3.884	0.628597637983877\\
3.924	0.593613861490167\\
3.964	0.553938897916712\\
4.004	0.509641948105484\\
4.044	0.461296568385962\\
4.08399999999999	0.410075197295957\\
4.12399999999999	0.357597872100143\\
4.16399999999998	0.305655475567515\\
4.20399999999998	0.255948849153063\\
4.24399999999998	0.209902197287478\\
4.28399999999997	0.168563468297544\\
4.32399999999997	0.132599744904972\\
4.36399999999996	0.102355243107497\\
4.40399999999996	0.0778881578104291\\
4.44399999999995	0.0589459213477886\\
4.48399999999995	0.0449327549087704\\
4.52399999999994	0.03494316292386\\
4.56399999999994	0.0278927601989245\\
4.60399999999994	0.0227331922016758\\
4.64399999999993	0.0186821622532635\\
4.68399999999993	0.0153518315586021\\
4.72399999999992	0.0126992782057672\\
4.76399999999992	0.0108370842703684\\
4.80399999999991	0.00981606061619452\\
4.84399999999991	0.009486524039729\\
4.88399999999991	0.00950643994133791\\
4.9239999999999	0.00950498202775134\\
4.9639999999999	0.00930732540320872\\
5.00399999999989	0.0090597732429919\\
5.04399999999989	0.00917938482048681\\
5.08399999999988	0.0102033138152683\\
5.12399999999988	0.0126532537217619\\
5.16399999999987	0.0169579726620259\\
5.20399999999987	0.023406372349706\\
5.24399999999987	0.0321078413283012\\
5.28399999999986	0.042996726720346\\
5.32399999999986	0.055940104398898\\
5.36399999999985	0.0709497895734562\\
5.40399999999985	0.0884236676783915\\
5.44399999999984	0.10927884453975\\
5.48399999999984	0.134783177886434\\
5.52399999999983	0.165975135801012\\
5.56399999999983	0.202938602130689\\
5.60399999999983	0.244516550041282\\
5.64399999999982	0.288701242295518\\
5.68399999999982	0.333289714195216\\
5.72399999999981	0.37632744069471\\
5.76399999999981	0.416263479760896\\
5.8039999999998	0.451973827872119\\
5.8439999999998	0.48277083684382\\
5.8839999999998	0.50841260859742\\
5.92399999999979	0.529073618893768\\
5.96399999999979	0.545250013504141\\
6.00399999999978	0.557618784585831\\
6.04399999999978	0.566899713902964\\
6.08399999999977	0.573758778082609\\
6.12399999999977	0.578761510080412\\
6.16399999999976	0.582363407753573\\
6.20399999999976	0.584919907159826\\
6.24399999999976	0.586703210542915\\
6.28399999999975	0.587919334841555\\
};
\addlegendentry{$y_c$};

\addplot [color=black,dash pattern=on 1pt off 3pt on 3pt off 3pt,line width=2.0pt]
  table[row sep=crcr]{%
0.004	0\\
0.044	-0.00313272655695536\\
0.084	-0.00867844462227884\\
0.124	-0.0140079047712132\\
0.164	-0.018091443462256\\
0.204	-0.0206214541016379\\
0.244	-0.0215460958014989\\
0.284	-0.0209199058929612\\
0.324	-0.0189288412516289\\
0.364	-0.015949225187372\\
0.404	-0.0125364827854573\\
0.444	-0.00928587187324214\\
0.484	-0.00660490116139399\\
0.524	-0.00455275981396711\\
0.564	-0.00287791149070371\\
0.604	-0.00121856217740639\\
0.644	0.000675507091553434\\
0.684	0.00281821873137434\\
0.724000000000001	0.00501782179761879\\
0.764000000000001	0.00705430598784005\\
0.804000000000001	0.0089009669828089\\
0.844000000000001	0.0108496495199843\\
0.884000000000001	0.0135027071825912\\
0.924000000000001	0.0177145101636838\\
0.964000000000001	0.0245442194882406\\
1.004	0.0351584542788934\\
1.044	0.0505651174397519\\
1.084	0.0712175715250734\\
1.124	0.0967873452975105\\
1.164	0.126377917375725\\
1.204	0.159107331646396\\
1.244	0.19473414267802\\
1.284	0.233975796668523\\
1.324	0.278229120554815\\
1.364	0.328655313658648\\
1.404	0.385146443575777\\
1.444	0.445966207036838\\
1.484	0.508295240820383\\
1.524	0.569173884000304\\
1.564	0.626243935671721\\
1.604	0.678041809117235\\
1.644	0.723895161403655\\
1.684	0.763610171346251\\
1.724	0.797153637776484\\
1.764	0.824432001191227\\
1.804	0.845132457709113\\
1.844	0.858576312498352\\
1.884	0.863666267137082\\
1.924	0.859141864970853\\
1.964	0.84423425664056\\
2.004	0.819358542274757\\
2.044	0.78622945679985\\
2.084	0.747218369146149\\
2.124	0.704371483894412\\
2.164	0.658650613968891\\
2.204	0.609793434432555\\
2.244	0.556952305088282\\
2.284	0.49981192190462\\
2.324	0.439397521503897\\
2.364	0.377985881712436\\
2.404	0.31832900396078\\
2.444	0.262823213018213\\
2.484	0.213042663253684\\
2.524	0.169700804580705\\
2.564	0.132896193454363\\
2.604	0.102417827002018\\
2.644	0.0779235479195955\\
2.684	0.0589534648286063\\
2.724	0.0448635203897635\\
2.764	0.0347808226458393\\
2.804	0.0276544923332249\\
2.844	0.0224377518846748\\
2.884	0.0183416579056262\\
2.924	0.0149935215607505\\
2.964	0.0123777371182891\\
3.004	0.0106128542915693\\
3.044	0.00971674076270573\\
3.084	0.00948222516471695\\
3.124	0.00952432539596022\\
3.164	0.00948221684459078\\
3.204	0.00924552470914172\\
3.244	0.00903862266841597\\
3.284	0.00932813881865353\\
3.324	0.0106622106472434\\
3.364	0.0135484125036182\\
3.404	0.0183889873633274\\
3.444	0.0254366848064866\\
3.484	0.0347598850355408\\
3.524	0.0462675774782377\\
3.564	0.0598479899864983\\
3.604	0.075604999756669\\
3.644	0.0941007610994917\\
3.684	0.116443112162265\\
3.724	0.144012012790866\\
3.764	0.177774432816555\\
3.804	0.217554868477214\\
3.844	0.261826820226545\\
3.884	0.308163713612062\\
3.924	0.353960674636309\\
3.964	0.397003318091964\\
4.004	0.435737176789083\\
4.044	0.469300324575242\\
4.08399999999999	0.497429417593016\\
4.12399999999999	0.520316345183389\\
4.16399999999998	0.538455166955703\\
4.20399999999998	0.552501266272817\\
4.24399999999998	0.563157691454206\\
4.28399999999997	0.571096227647175\\
4.32399999999997	0.576912126204277\\
4.36399999999996	0.581105440285701\\
4.40399999999996	0.584080070596744\\
4.44399999999995	0.586152698548648\\
4.48399999999995	0.587565979059748\\
4.52399999999994	0.588502507589093\\
4.56399999999994	0.589097701949801\\
4.60399999999994	0.589450803534764\\
4.64399999999993	0.589633821927855\\
4.68399999999993	0.58969856458187\\
4.72399999999992	0.589682026713169\\
4.76399999999992	0.589610447505546\\
4.80399999999991	0.589502318627651\\
4.84399999999991	0.58937058997529\\
4.88399999999991	0.589224271641686\\
4.9239999999999	0.589069588266586\\
4.9639999999999	0.588910805289955\\
5.00399999999989	0.588750816931741\\
5.04399999999989	0.588591562448516\\
5.08399999999988	0.588434319428065\\
5.12399999999988	0.58827990952738\\
5.16399999999987	0.588128842171085\\
5.20399999999987	0.587981414485425\\
5.24399999999987	0.587837780485823\\
5.28399999999986	0.587697998746986\\
5.32399999999986	0.587562065070179\\
5.36399999999985	0.587429934727532\\
5.40399999999985	0.587301537490166\\
5.44399999999984	0.587176787676088\\
5.48399999999984	0.587055590769329\\
5.52399999999983	0.586937847680388\\
5.56399999999983	0.586823457379866\\
5.60399999999983	0.58671231839948\\
5.64399999999982	0.586604329527232\\
5.68399999999982	0.586499389904842\\
5.72399999999981	0.58639739865048\\
5.76399999999981	0.586298254067992\\
5.8039999999998	0.586201852457742\\
5.8439999999998	0.586108086509405\\
5.8839999999998	0.586016843230676\\
5.92399999999979	0.585928001346959\\
5.96399999999979	0.585841428096486\\
6.00399999999978	0.585756975345647\\
6.04399999999978	0.585674474965833\\
6.08399999999977	0.585593733454047\\
6.12399999999977	0.585514525856706\\
6.16399999999976	0.585436589185521\\
6.20399999999976	0.585359615715103\\
6.24399999999976	0.585283246843323\\
6.28399999999975	0.585207068589478\\
};
\addlegendentry{$y_u$};

\end{axis}

\begin{axis}[%
width=\figurewidth,
height=\figureheight,
scale only axis,
xmin=0,
xmax=8,
xlabel={Time [s]},
ymin=-300,
ymax=200,
ylabel absolute,
ylabel={$\text{Acceleration [m/s}^\text{2}\text{]}$},
at=(plot1.below south west),
anchor=above north west,
legend style={draw=black,fill=white,legend cell align=left}
]
\addplot [color=blue,solid,line width=1.5pt]
  table[row sep=crcr]{%
0.004	0\\
0.044	-2.26856168194922\\
0.084	-0.0986343456604433\\
0.124	0.768511329523994\\
0.164	1.02207656327262\\
0.204	1.05946000086215\\
0.244	1.02940780322211\\
0.284	0.93094515436386\\
0.324	0.717574057507032\\
0.364	0.374443371080146\\
0.404	-0.0320155340091197\\
0.444	-0.355452799626829\\
0.484	-0.45863472923532\\
0.524	-0.326376164171574\\
0.564	-0.075865390787311\\
0.604	0.133671302687038\\
0.644	0.192370327577408\\
0.684	0.0888308848960486\\
0.724000000000001	-0.0796514173893245\\
0.764000000000001	-0.156853909201929\\
0.804000000000001	-0.0305110752222932\\
0.844000000000001	0.311048830213738\\
0.884000000000001	0.82473051769753\\
0.924000000000001	1.47667070689226\\
0.964000000000001	2.22618688299622\\
1.004	2.93398289300119\\
1.044	3.32934065870746\\
1.084	3.18772861030425\\
1.124	2.58610502864062\\
1.164	1.90131024016087\\
1.204	1.55743157148101\\
1.244	1.8000837495048\\
1.284	2.56378470363107\\
1.324	3.39945967715488\\
1.364	3.65329415300672\\
1.404	2.96841431724668\\
1.444	1.58064271073041\\
1.484	0.0143347426744267\\
1.524	-1.35258356433368\\
1.564	-2.38413591505543\\
1.604	-3.07908888862105\\
1.644	-3.49852639578629\\
1.684	-3.74036309167336\\
1.724	-3.9016940280641\\
1.764	-4.05876685198116\\
1.804	-4.29323989044909\\
1.844	-4.69036578644442\\
1.884	-5.24004468243053\\
1.924	-5.71039897395354\\
1.964	-5.71254914568919\\
2.004	-5.04022779935609\\
2.044	-121.800381294875\\
2.084	-114.745374311499\\
2.124	-108.38350281054\\
2.164	-110.048666419563\\
2.204	-114.686956685342\\
2.244	-120.333404822801\\
2.284	-126.34669083612\\
2.324	-132.49903781213\\
2.364	-138.701678183125\\
2.404	-144.917653884161\\
2.444	-151.131083897258\\
2.484	-157.335249767859\\
2.524	-163.527600688229\\
2.564	-169.707526186626\\
2.604	-175.875315314426\\
2.644	-182.03165402276\\
2.684	-188.17737793614\\
2.724	-194.313350648843\\
2.764	-200.440405301719\\
2.804	-206.559318553603\\
2.844	-212.670801191955\\
2.884	-218.775497178138\\
2.924	-224.873986797607\\
2.964	-230.966791619481\\
3.004	-115.048324754392\\
3.044	10.9060406873995\\
3.084	7.90523314826936\\
3.124	5.63371403552367\\
3.164	4.00808032313604\\
3.204	2.84180241331713\\
3.244	2.00084401865313\\
3.284	1.38840019856993\\
3.324	0.933732433083192\\
3.364	0.584040620616674\\
3.404	0.298490459143656\\
3.444	0.043857675101628\\
3.484	-0.2083852095604\\
3.524	-0.483236897927196\\
3.564	-0.800680109704563\\
3.604	-1.17023703549815\\
3.644	-1.58138092904918\\
3.684	-1.99450588810116\\
3.724	-2.34539794866138\\
3.764	-2.57651084384179\\
3.804	-2.68401381616677\\
3.844	-2.73501321916252\\
3.884	-2.81633040561918\\
3.924	-2.93664996835133\\
3.964	-2.96716014210779\\
4.004	-2.70727267365317\\
4.044	-2.04407153848574\\
4.08399999999999	-1.04216365704984\\
4.12399999999999	0.118173509400771\\
4.16399999999998	1.25163679051127\\
4.20399999999998	2.22168462740976\\
4.24399999999998	2.94825418748831\\
4.28399999999997	3.41449254555137\\
4.32399999999997	3.65906286692845\\
4.36399999999996	3.72222764119386\\
4.40399999999996	3.59770107304189\\
4.44399999999995	3.25619305283716\\
4.48399999999995	2.70178274397082\\
4.52399999999994	2.00167211834076\\
4.56399999999994	1.28559934156\\
4.60399999999994	0.713277228434646\\
4.64399999999993	0.399436816733882\\
4.68399999999993	0.34292889513283\\
4.72399999999992	0.430910212272769\\
4.76399999999992	0.510360079441751\\
4.80399999999991	0.465526035422907\\
4.84399999999991	0.270574156523631\\
4.88399999999991	0.0108608343126964\\
4.9239999999999	-0.157461555768395\\
4.9639999999999	-0.117081945132658\\
5.00399999999989	0.127043720389133\\
5.04399999999989	0.477580706649601\\
5.08399999999988	0.832297674854822\\
5.12399999999988	1.13133073694394\\
5.16399999999987	1.3444707121305\\
5.20399999999987	1.44444970577988\\
5.24399999999987	1.4189639229258\\
5.28399999999986	1.31609504196899\\
5.32399999999986	1.2614763937748\\
5.36399999999985	1.42040042749963\\
5.40399999999985	1.92013904482242\\
5.44399999999984	2.7288261487108\\
5.48399999999984	3.53402070324513\\
5.52399999999983	3.82066639708253\\
5.56399999999983	3.25964403790768\\
5.60399999999983	2.0135417844219\\
5.64399999999982	0.540988275751636\\
5.68399999999982	-0.797266326603263\\
5.72399999999981	-1.86442941919818\\
5.76399999999981	-2.64503296390127\\
5.8039999999998	-3.1412527639056\\
5.8439999999998	-3.34715354071977\\
5.8839999999998	-3.27256683121713\\
5.92399999999979	-2.96927420201662\\
5.96399999999979	-2.52731404253108\\
6.00399999999978	-2.04354213967289\\
6.04399999999978	-1.59038740621604\\
6.08399999999977	-1.20504705943693\\
6.12399999999977	-0.896586698251369\\
6.16399999999976	-0.658537029059379\\
6.20399999999976	-0.478756318427805\\
6.24399999999976	-0.344771215950536\\
6.28399999999975	-0.245854366125953\\
};
\addlegendentry{$\ddot{y}_{r}$};

\addplot [color=red,dashed,line width=2.0pt]
  table[row sep=crcr]{%
0.004	0\\
0.044	-0.0820116994644396\\
0.084	-0.0820116994644396\\
0.124	-0.0820116994644396\\
0.164	-0.0820116994644396\\
0.204	-0.0820116994644396\\
0.244	-0.0820116994644396\\
0.284	-0.0820116994644396\\
0.324	-0.0820116994644396\\
0.364	-0.0820116994644396\\
0.404	-0.0820116994644396\\
0.444	-0.0820116994644396\\
0.484	-0.0820116994644396\\
0.524	-0.0820116994644396\\
0.564	-0.0820116994644396\\
0.604	-0.0820116994644396\\
0.644	-0.0820116994644396\\
0.684	-0.0820116994644396\\
0.724000000000001	-0.0820116994644396\\
0.764000000000001	-0.0820116994644396\\
0.804000000000001	-0.0820116994644396\\
0.844000000000001	-0.0820116994644396\\
0.884000000000001	-0.0820116994644396\\
0.924000000000001	-0.0820116994644396\\
0.964000000000001	-0.0820116994644396\\
1.004	-0.0820116994644396\\
1.044	-0.0820116994644396\\
1.084	-0.0820116994644396\\
1.124	-0.0820116994644396\\
1.164	-0.0820116994644396\\
1.204	-0.0820116994644396\\
1.244	-0.0820116994644396\\
1.284	-0.0820116994644396\\
1.324	-0.0820116994644396\\
1.364	-0.0820116994644396\\
1.404	-0.0820116994644396\\
1.444	-0.0820116994644396\\
1.484	-0.0820116994644396\\
1.524	-0.0820116994644396\\
1.564	-0.0820116994644396\\
1.604	-0.0820116994644396\\
1.644	-0.0820116994644396\\
1.684	-0.0820116994644396\\
1.724	-0.0820116994644396\\
1.764	-0.0820116994644396\\
1.804	-0.0820116994644396\\
1.844	-0.0820116994644396\\
1.884	-0.0820116994644396\\
1.924	-0.0820116994644396\\
1.964	-0.0820116994644396\\
2.004	-0.0820116994644396\\
2.044	-0.0820116994644396\\
2.084	-0.0820116994644396\\
2.124	-0.0820116994644396\\
2.164	-0.0820116994644396\\
2.204	-0.0820116994644396\\
2.244	-0.0820116994644396\\
2.284	-0.0820116994644396\\
2.324	-0.0820116994644396\\
2.364	-0.0820116994644396\\
2.404	-0.0820116994644396\\
2.444	-0.0820116994644396\\
2.484	-0.0820116994644396\\
2.524	-0.0820116994644396\\
2.564	-0.0820116994644396\\
2.604	-0.0820116994644396\\
2.644	-0.0820116994644396\\
2.684	-0.0820116994644396\\
2.724	-0.0820116994644396\\
2.764	-0.0820116994644396\\
2.804	-0.0820116994644396\\
2.844	-0.0820116994644396\\
2.884	-0.0820116994644396\\
2.924	-0.0820116994644396\\
2.964	-0.0820116994644396\\
3.004	-0.0820116994644396\\
3.044	-0.0820116994644396\\
3.084	-0.0820116994644396\\
3.124	-0.0820116994644396\\
3.164	-0.0820116994644396\\
3.204	-0.0820116994644396\\
3.244	-0.0820116994644396\\
3.284	-0.0820116994644396\\
3.324	-0.0820116994644396\\
3.364	-0.0820116994644396\\
3.404	-0.0820116994644396\\
3.444	-0.0820116994644396\\
3.484	-0.0820116994644396\\
3.524	-0.0820116994644396\\
3.564	-0.0820116994644396\\
3.604	-0.0820116994644396\\
3.644	-0.0820116994644396\\
3.684	-0.0820116994644396\\
3.724	-0.0820116994644396\\
3.764	-0.0820116994644396\\
3.804	-0.0820116994644396\\
3.844	-0.0820116994644396\\
3.884	-0.0820116994644396\\
3.924	-0.0820116994644396\\
3.964	-0.0820116994644396\\
4.004	-0.0820116994644396\\
4.044	-0.0820116994644396\\
4.08399999999999	-0.0820116994644396\\
4.12399999999999	-0.0820116994644396\\
4.16399999999998	-0.0820116994644396\\
4.20399999999998	-0.0820116994644396\\
4.24399999999998	-0.0820116994644396\\
4.28399999999997	-0.0820116994644396\\
4.32399999999997	-0.0820116994644396\\
4.36399999999996	-0.0820116994644396\\
4.40399999999996	-0.0820116994644396\\
4.44399999999995	-0.0820116994644396\\
4.48399999999995	-0.0820116994644396\\
4.52399999999994	-0.0820116994644396\\
4.56399999999994	-0.0820116994644396\\
4.60399999999994	-0.0820116994644396\\
4.64399999999993	-0.0820116994644396\\
4.68399999999993	-0.0820116994644396\\
4.72399999999992	-0.0820116994644396\\
4.76399999999992	-0.0820116994644396\\
4.80399999999991	-0.0820116994644396\\
4.84399999999991	-0.0820116994644396\\
4.88399999999991	-0.0820116994644396\\
4.9239999999999	-0.0820116994644396\\
4.9639999999999	-0.0820116994644396\\
5.00399999999989	-0.0820116994644396\\
5.04399999999989	-0.0820116994644396\\
5.08399999999988	-0.0820116994644396\\
5.12399999999988	-0.0820116994644396\\
5.16399999999987	-0.0820116994644396\\
5.20399999999987	-0.0820116994644396\\
5.24399999999987	-0.0820116994644396\\
5.28399999999986	-0.0820116994644396\\
5.32399999999986	-0.0820116994644396\\
5.36399999999985	-0.0820116994644396\\
5.40399999999985	-0.0820116994644396\\
5.44399999999984	-0.0820116994644396\\
5.48399999999984	-0.0820116994644396\\
5.52399999999983	-0.0820116994644396\\
5.56399999999983	-0.0820116994644396\\
5.60399999999983	-0.0820116994644396\\
5.64399999999982	-0.0820116994644396\\
5.68399999999982	-0.0820116994644396\\
5.72399999999981	-0.0820116994644396\\
5.76399999999981	-0.0820116994644396\\
5.8039999999998	-0.0820116994644396\\
5.8439999999998	-0.0820116994644396\\
5.8839999999998	-0.0820116994644396\\
5.92399999999979	-0.0820116994644396\\
5.96399999999979	-0.0820116994644396\\
6.00399999999978	-0.0820116994644396\\
6.04399999999978	-0.0820116994644396\\
6.08399999999977	-0.0820116994644396\\
6.12399999999977	-0.0820116994644396\\
6.16399999999976	-0.0820116994644396\\
6.20399999999976	-0.0820116994644396\\
6.24399999999976	-0.0820116994644396\\
6.28399999999975	-0.0820116994644396\\
};
\addlegendentry{$\ddot{y}_c$};

\addplot [color=black,dash pattern=on 1pt off 3pt on 3pt off 3pt,line width=2.0pt]
  table[row sep=crcr]{%
0.004	0\\
0.044	-1.59954299014054\\
0.084	0.112244330600025\\
0.124	0.779415907514183\\
0.164	0.974340781299854\\
0.204	1.00789613448221\\
0.244	0.981889560654181\\
0.284	0.879638712474928\\
0.324	0.657613876951821\\
0.364	0.312544034052864\\
0.404	-0.0758329244056085\\
0.444	-0.358629876102407\\
0.484	-0.417524254956118\\
0.524	-0.264206835738674\\
0.564	-0.0257165361475424\\
0.604	0.150091393858293\\
0.644	0.173756653586864\\
0.684	0.053778451992183\\
0.724000000000001	-0.10124004023411\\
0.764000000000001	-0.142621185086257\\
0.804000000000001	0.0195788496590049\\
0.844000000000001	0.38188929410743\\
0.884000000000001	0.903616576576362\\
0.924000000000001	1.55773651233641\\
0.964000000000001	2.29808452337993\\
1.004	2.97130090659434\\
1.044	3.31527592263806\\
1.084	3.14885627773208\\
1.124	2.57879902609129\\
1.164	1.97236638978646\\
1.204	1.74674893040044\\
1.244	2.1456580863959\\
1.284	3.0444733206634\\
1.324	3.87834680857241\\
1.364	3.93610841509374\\
1.404	2.9168773676545\\
1.444	1.14358753926628\\
1.484	-0.764403753417956\\
1.524	-2.30597443251475\\
1.564	-3.27206161988472\\
1.604	-3.71687799024603\\
1.644	-3.84059935886702\\
1.684	-3.8525528877986\\
1.724	-3.89478474842458\\
1.764	-4.06423935751359\\
1.804	-4.45687962283568\\
1.844	-5.12915890554929\\
1.884	-5.95601930841547\\
1.924	-6.53180027601709\\
1.964	-6.36993936717497\\
2.004	-5.33484858491606\\
2.044	-3.81215973869931\\
2.084	-2.44277722246786\\
2.124	-1.74955873361838\\
2.164	-1.87761372606672\\
2.204	-2.4633627045808\\
2.244	-2.77458234492599\\
2.284	-2.22377642961008\\
2.324	-0.81810751367034\\
2.364	0.944986001325874\\
2.404	2.50945045890828\\
2.444	3.55271066632738\\
2.484	4.03406092180096\\
2.524	4.10749549652899\\
2.564	3.97765797891656\\
2.604	3.76949623980906\\
2.644	3.49504351763457\\
2.684	3.10903057082453\\
2.724	2.57597537684592\\
2.764	1.91752823255133\\
2.804	1.24161514436171\\
2.844	0.715836816805452\\
2.884	0.458476204716666\\
2.924	0.443889617692125\\
2.964	0.530088973054349\\
3.004	0.560236435709329\\
3.044	0.443916317665101\\
3.084	0.199235153293816\\
3.124	-0.048229934474968\\
3.164	-0.142991794742527\\
3.204	-0.0175968810182347\\
3.244	0.272536434599432\\
3.284	0.620900420695983\\
3.324	0.946160560318816\\
3.364	1.20753590658553\\
3.404	1.37837125443067\\
3.444	1.43259368586402\\
3.484	1.37645822694343\\
3.524	1.2895048949391\\
3.564	1.3197337210566\\
3.604	1.62999041531844\\
3.644	2.30220663159971\\
3.684	3.19951464678653\\
3.724	3.89643692109493\\
3.764	3.88738929490973\\
3.804	2.98700635306512\\
3.844	1.46618110466839\\
3.884	-0.20140742222897\\
3.924	-1.63525048821201\\
3.964	-2.65261452398771\\
4.004	-3.22782464849213\\
4.044	-3.41853565217931\\
4.08399999999999	-3.31468084666273\\
4.12399999999999	-3.01313164538305\\
4.16399999999998	-2.6039818075494\\
4.20399999999998	-2.16071827720884\\
4.24399999999998	-1.73486070086359\\
4.28399999999997	-1.35627483383283\\
4.32399999999997	-1.03756002418565\\
4.36399999999996	-0.779758927640912\\
4.40399999999996	-0.577422640633427\\
4.44399999999995	-0.422268031144218\\
4.48399999999995	-0.305449104685459\\
4.52399999999994	-0.218773868783872\\
4.56399999999994	-0.155230113215969\\
4.60399999999994	-0.10910820459114\\
4.64399999999993	-0.0759164167060615\\
4.68399999999993	-0.0522083282671145\\
4.72399999999992	-0.0353889175410091\\
4.76399999999992	-0.0235326320042473\\
4.80399999999991	-0.015227247463255\\
4.84399999999991	-0.00944681845785937\\
4.88399999999991	-0.00545189279219722\\
4.9239999999999	-0.0027130791075056\\
4.9639999999999	-0.000853580154019766\\
5.00399999999989	0.000393402360879092\\
5.04399999999989	0.00121588659909386\\
5.08399999999988	0.00174578449573312\\
5.12399999999988	0.00207524660316089\\
5.16399999999987	0.00226837041189437\\
5.20399999999987	0.0023695419900939\\
5.24399999999987	0.00240935067914656\\
5.28399999999986	0.00240876482939938\\
5.32399999999986	0.00238206788462978\\
5.36399999999985	0.0023389145947109\\
5.40399999999985	0.00228576502053481\\
5.44399999999984	0.00222687987980816\\
5.48399999999984	0.00216500735019061\\
5.52399999999983	0.00210185313781284\\
5.56399999999983	0.0020383982802707\\
5.60399999999983	0.00197510972168548\\
5.64399999999982	0.00191207494202302\\
5.68399999999982	0.00184908223414411\\
5.72399999999981	0.0017856614549477\\
5.76399999999981	0.00172109543329775\\
5.8039999999998	0.00165440915847103\\
5.8439999999998	0.0015843420668739\\
5.8839999999998	0.0015093080257962\\
5.92399999999979	0.00142734796885444\\
5.96399999999979	0.00133608169853978\\
6.00399999999978	0.00123266840717009\\
6.04399999999978	0.00111379036147059\\
6.08399999999977	0.00097568137134586\\
6.12399999999977	0.000814231380579757\\
6.16399999999976	0.000625210464398114\\
6.20399999999976	0.000404668050803393\\
6.24399999999976	0.000149572112847882\\
6.28399999999975	-0.000141249881787784\\
};
\addlegendentry{$\ddot{y}_u$};

\end{axis}
\end{tikzpicture}%

%% file: chapters/paper2/figs/coupling_scheme.tex
\tikzset{block/.style={draw, rectangle, line width=2pt,
     minimum height=3em, minimum width=3em, outer sep=0pt}}
\tikzset{sumcircle/.style={draw, circle, outer sep=0pt, 
     label=center:{{$\sum$}}, minimum width=2em}}
\tikzset{every picture/.style={auto, line width=1pt,
          >=narrow,font=\small}}

\begin{tikzpicture}
\node[block](Robot){Robot};
\node[sumcircle, left=5mm of Robot](sum3){};
\node[block, left=5mm of sum3](PD){$k_p + k_v \frac{d}{dt}(\cdot)$ };
\node[sumcircle, left=5mm of PD](sum2){};
\node[block, left=5mm of sum2](DMP){DMP};
\coordinate[right=11mm of Robot](c1);
\draw[->](DMP)--node[](yc){$y_c$}(sum2);
\draw[->](sum2)--(PD);
\draw[->](PD)--(sum3);
\draw[->](sum3)--node[](yardd){$\ddot{y}_{r}$}(Robot);
\draw[->](Robot)--node[](ya){$y_a$}(c1);
\node[block, below=5mm of sum2](M1){-1};

\coordinate[below=20mm of ya](c2);
\coordinate[below=5mm of M1](c3);
\coordinate[above=5mm of DMP](c4);

\draw[-](DMP)--(c4);
\draw[->](c4)-|node[](ycdd){$\ddot{y}_c$}(sum3);
\draw[-](ya)--(c2);
\draw[-](ya)|-(c3);
\draw[->](c3)-|(DMP);
\draw[->](c3)--(M1);
\draw[->](M1)--(sum2);
\end{tikzpicture}

%% file: chapters/paper2/figs/ourstop.tex
%
%
\begin{tikzpicture}

\begin{axis}[%
width=\figurewidth,
height=\figureheight,
scale only axis,
xmin=0,
xmax=8,
ymin=-0.2,
ymax=1.5,
ylabel absolute,
ylabel={Position [m]},
name=plot1,
legend style={draw=black,fill=white,legend cell align=left}
]
\addplot [color=blue,solid,line width=1.5pt]
  table[row sep=crcr]{%
0.004	0\\
0.044	-0.00194098344387841\\
0.084	-0.00782983362888778\\
0.124	-0.0139518805290859\\
0.164	-0.0186948524072939\\
0.204	-0.021617816371952\\
0.244	-0.022707775339871\\
0.284	-0.0220500083229208\\
0.324	-0.0198935585098084\\
0.364	-0.016609635343215\\
0.404	-0.0127726974480613\\
0.444	-0.00908907347735457\\
0.484	-0.00609033686179491\\
0.524	-0.00389526265465208\\
0.564	-0.00228487026705418\\
0.604	-0.000799908737614542\\
0.644	0.000932153384380087\\
0.684	0.00298630327863743\\
0.724000000000001	0.00519340188930127\\
0.764000000000001	0.00728362393912803\\
0.804000000000001	0.00913364798699571\\
0.844000000000001	0.010959590347298\\
0.884000000000001	0.0133693614783284\\
0.924000000000001	0.0171939128603915\\
0.964000000000001	0.0235154336013563\\
1.004	0.0335876261123646\\
1.044	0.0485694080392404\\
1.084	0.0690661970711065\\
1.124	0.0948283276672182\\
1.164	0.124878185315395\\
1.204	0.158129050283887\\
1.244	0.194079210325368\\
1.284	0.233242945614622\\
1.324	0.277043758283998\\
1.364	0.326979062947231\\
1.404	0.383450840814114\\
1.444	0.444994913554764\\
1.484	0.508585195332597\\
1.524	0.570575684095293\\
1.564	0.627698074041692\\
1.604	0.677691591620068\\
1.644	0.719643612731094\\
1.684	0.753818239496937\\
1.724	0.781175459347899\\
1.764	0.802976622139004\\
1.804	0.820407879091593\\
1.844	0.834449199674216\\
1.884	0.845798189410951\\
1.924	0.854904538163826\\
1.964	0.861951504712177\\
2.004	0.866519229943154\\
2.044	0.866519229943154\\
2.084	0.866519229943154\\
2.124	0.866519229943154\\
2.164	0.866519229943154\\
2.204	0.866519229943154\\
2.244	0.866519229943154\\
2.284	0.866519229943154\\
2.324	0.866519229943154\\
2.364	0.866519229943154\\
2.404	0.866519229943154\\
2.444	0.866519229943154\\
2.484	0.866519229943154\\
2.524	0.866519229943154\\
2.564	0.866519229943154\\
2.604	0.866519229943154\\
2.644	0.866519229943154\\
2.684	0.866519229943154\\
2.724	0.866519229943154\\
2.764	0.866519229943154\\
2.804	0.866519229943154\\
2.844	0.866519229943154\\
2.884	0.866519229943154\\
2.924	0.866519229943154\\
2.964	0.866519229943154\\
3.004	0.866423695341092\\
3.044	0.86415512751961\\
3.084	0.859728442611004\\
3.124	0.854070129472266\\
3.164	0.847788179140404\\
3.204	0.8413255271673\\
3.244	0.834932203723497\\
3.284	0.828731424133716\\
3.324	0.822790571430647\\
3.364	0.817136848217959\\
3.404	0.81173142002559\\
3.444	0.806515903375569\\
3.484	0.801436980346584\\
3.524	0.796362667015281\\
3.564	0.791184022340115\\
3.604	0.785752052460969\\
3.644	0.779888960752747\\
3.684	0.773416975269687\\
3.724	0.766155683612008\\
3.764	0.757886553309777\\
3.804	0.748404486993639\\
3.844	0.7375251935488\\
3.884	0.725117334025882\\
3.924	0.711070925404022\\
3.964	0.695331939206104\\
4.004	0.677883676679025\\
4.044	0.658705269659662\\
4.08399999999999	0.637763335386476\\
4.12399999999999	0.614975238971181\\
4.16399999999998	0.590242941557273\\
4.20399999999998	0.563469128590592\\
4.24399999999998	0.534588727436013\\
4.28399999999997	0.503604681013327\\
4.32399999999997	0.470638463197518\\
4.36399999999996	0.435970412190136\\
4.40399999999996	0.399999767698436\\
4.44399999999995	0.363257318199201\\
4.48399999999995	0.326363276471725\\
4.52399999999994	0.289999040550052\\
4.56399999999994	0.254826238805445\\
4.60399999999994	0.221414685298219\\
4.64399999999993	0.190258320917725\\
4.68399999999993	0.161721156303999\\
4.72399999999992	0.136041386996525\\
4.76399999999992	0.113317192623986\\
4.80399999999991	0.0936421110636448\\
4.84399999999991	0.0769815188934989\\
4.88399999999991	0.0632406997365275\\
4.9239999999999	0.0522293577259354\\
4.9639999999999	0.0436353089952307\\
5.00399999999989	0.0370450880438161\\
5.04399999999989	0.0319923475107127\\
5.08399999999988	0.0279853857783332\\
5.12399999999988	0.0246028498518221\\
5.16399999999987	0.0216139201034053\\
5.20399999999987	0.0189499169580151\\
5.24399999999987	0.0166568700556283\\
5.28399999999986	0.0148412782075617\\
5.32399999999986	0.0135775161041342\\
5.36399999999985	0.0128621518612652\\
5.40399999999985	0.0125355076428548\\
5.44399999999984	0.0123772554590808\\
5.48399999999984	0.0121404049226822\\
5.52399999999983	0.0117087811893283\\
5.56399999999983	0.0111817248721505\\
5.60399999999983	0.010881289757547\\
5.64399999999982	0.0112596223391889\\
5.68399999999982	0.0127992053710582\\
5.72399999999981	0.015971408124097\\
5.76399999999981	0.0211781188216064\\
5.8039999999998	0.0286673522442162\\
5.8439999999998	0.0385057442841402\\
5.8839999999998	0.0505889168356921\\
5.92399999999979	0.0647820138354843\\
5.96399999999979	0.0812015146151279\\
6.00399999999978	0.100461157517192\\
6.04399999999978	0.123777291284445\\
6.08399999999977	0.152678711657008\\
6.12399999999977	0.188190874922414\\
6.16399999999976	0.230001171995316\\
6.20399999999976	0.276262686239185\\
6.24399999999976	0.324173677988096\\
6.28399999999975	0.3707764835493\\
6.32399999999975	0.413517954679694\\
6.36399999999974	0.450653884436483\\
6.40399999999974	0.481410852740581\\
6.44399999999973	0.505964570933056\\
6.48399999999973	0.525159610934476\\
6.52399999999972	0.540040747602014\\
6.56399999999972	0.551588131803536\\
6.60399999999972	0.560584218596018\\
6.64399999999971	0.567629921139762\\
6.68399999999971	0.573133567716924\\
6.7239999999997	0.577404881128176\\
6.7639999999997	0.58068277544352\\
6.80399999999969	0.583149107288519\\
6.84399999999969	0.584923469173441\\
6.88399999999969	0.586168060804461\\
6.92399999999968	0.587011830843006\\
6.96399999999968	0.587555945139412\\
7.00399999999967	0.587894272001029\\
7.04399999999967	0.588101619510754\\
7.08399999999966	0.588229857518999\\
7.12399999999966	0.588298449808088\\
7.16399999999965	0.588349071726582\\
7.20399999999965	0.588372208429985\\
7.24399999999965	0.588374830617273\\
7.28399999999964	0.588352543587876\\
7.32399999999964	0.588324874836166\\
7.36399999999963	0.588291732434115\\
7.40399999999963	0.58824617219725\\
7.44399999999962	0.588187111918725\\
7.48399999999962	0.588111599241082\\
7.52399999999961	0.58802564760893\\
7.56399999999961	0.587933559766826\\
7.60399999999961	0.587842408941076\\
7.6439999999996	0.587746985617772\\
7.6839999999996	0.587644844045693\\
7.72399999999959	0.587517239388999\\
7.76399999999959	0.587408051417718\\
7.80399999999958	0.587311316770111\\
7.84399999999958	0.587202000820351\\
7.88399999999957	0.587092295997216\\
7.92399999999957	0.586990693963218\\
7.96399999999957	0.586881398997511\\
8.00399999999956	0.58677528581289\\
8.04399999999956	0.586670325333826\\
8.08399999999955	0.58656736900179\\
8.12399999999955	0.586456377408921\\
8.16399999999954	0.586353158751982\\
8.20399999999954	0.586264308733178\\
8.24399999999954	0.586181264169124\\
8.28399999999953	0.586103635905949\\
8.32399999999953	0.586022891920469\\
8.36399999999952	0.585937539595794\\
};
\addlegendentry{$y_a$};

\addplot [color=red,dashed,line width=2.0pt]
  table[row sep=crcr]{%
0.004	0\\
0.044	-0.00313254143200675\\
0.084	-0.00867508689363015\\
0.124	-0.0139997682573689\\
0.164	-0.0180831399840514\\
0.204	-0.0206168217631547\\
0.244	-0.0215456939015792\\
0.284	-0.0209247417265336\\
0.324	-0.0189428715470429\\
0.364	-0.0159763433570373\\
0.404	-0.0125746457062455\\
0.444	-0.00932629373233514\\
0.484	-0.00663934776499803\\
0.524	-0.00457993032161797\\
0.564	-0.0029021447323103\\
0.604	-0.00124501255051504\\
0.644	0.000644348568312007\\
0.684	0.00278367526449344\\
0.724000000000001	0.00498360267422708\\
0.764000000000001	0.0070229743189355\\
0.804000000000001	0.00887057497218791\\
0.844000000000001	0.0108124517573353\\
0.884000000000001	0.0134456600509057\\
0.924000000000001	0.0176204812365966\\
0.964000000000001	0.0243934843217142\\
1.004	0.0349280809647184\\
1.044	0.0502192528433035\\
1.084	0.0706908821051007\\
1.124	0.0959942138822332\\
1.164	0.125267650413091\\
1.204	0.157703933131217\\
1.244	0.19309594987446\\
1.284	0.232120892434127\\
1.324	0.27611780189086\\
1.364	0.326244233530784\\
1.404	0.382445138026023\\
1.444	0.443013780230124\\
1.484	0.505014665540246\\
1.524	0.565104145755034\\
1.564	0.620296163222391\\
1.604	0.668558155848552\\
1.644	0.709116898295447\\
1.684	0.742319900795619\\
1.724	0.769189246074382\\
1.764	0.790959473217279\\
1.804	0.808759900149883\\
1.844	0.823476295867237\\
1.884	0.835721868860209\\
1.924	0.845854108805441\\
1.964	0.853993131332526\\
2.004	0.860026566677056\\
2.044	0.863627115300331\\
2.084	0.863991962781572\\
2.124	0.85979625469864\\
2.164	0.850744133164297\\
2.204	0.839281633339258\\
2.244	0.829101975796349\\
2.284	0.821677251880624\\
2.324	0.816493158155498\\
2.364	0.812754209221596\\
2.404	0.809916280077193\\
2.444	0.807655956934182\\
2.484	0.805781787192292\\
2.524	0.804176647821129\\
2.564	0.802765927464895\\
2.604	0.8015002083646\\
2.644	0.800345599683794\\
2.684	0.79927814396693\\
2.724	0.79828045733294\\
2.764	0.797339638630053\\
2.804	0.796445925369136\\
2.844	0.7955918040299\\
2.884	0.794771405461551\\
2.924	0.793980084261472\\
2.964	0.793214119953721\\
3.004	0.792470499368197\\
3.044	0.791745529073654\\
3.084	0.791029205743003\\
3.124	0.790305890355921\\
3.164	0.78955741459254\\
3.204	0.788763853695407\\
3.244	0.787903224781075\\
3.284	0.786950530993583\\
3.324	0.785876404196223\\
3.364	0.784645537577323\\
3.404	0.783214926969803\\
3.444	0.781531635163818\\
3.484	0.779530292490725\\
3.524	0.777130694670233\\
3.564	0.774234876891824\\
3.604	0.770725094257224\\
3.644	0.766462711681274\\
3.684	0.761288882630591\\
3.724	0.755029610779033\\
3.764	0.747506866024486\\
3.804	0.738553424464252\\
3.844	0.728029336354358\\
3.884	0.715836416676926\\
3.924	0.701925100866446\\
3.964	0.68628457072058\\
4.004	0.668921663602325\\
4.044	0.649835976959277\\
4.08399999999999	0.628999914585129\\
4.12399999999999	0.606350425088704\\
4.16399999999998	0.581792718516507\\
4.20399999999998	0.555225162143121\\
4.24399999999998	0.526581037927647\\
4.28399999999997	0.495872793439085\\
4.32399999999997	0.463222587708559\\
4.36399999999996	0.428882897794346\\
4.40399999999996	0.393248113527238\\
4.44399999999995	0.356837924099934\\
4.48399999999995	0.32026003582272\\
4.52399999999994	0.284164680792208\\
4.56399999999994	0.249200508535358\\
4.60399999999994	0.215957649349394\\
4.64399999999993	0.184915595613198\\
4.68399999999993	0.156433864135672\\
4.72399999999992	0.130764424991616\\
4.76399999999992	0.10806353422617\\
4.80399999999991	0.0883962089600921\\
4.84399999999991	0.0717502678573673\\
4.88399999999991	0.0580238772510905\\
4.9239999999999	0.0470131844545388\\
4.9639999999999	0.03840661270453\\
5.00399999999989	0.0317943660327127\\
5.04399999999989	0.0267086801145449\\
5.08399999999988	0.0226940607629541\\
5.12399999999988	0.0193855622738237\\
5.16399999999987	0.0165622269243535\\
5.20399999999987	0.0141511915022278\\
5.24399999999987	0.0121843846165698\\
5.28399999999986	0.0107310004769245\\
5.32399999999986	0.00982830544851598\\
5.36399999999985	0.00942722853197591\\
5.40399999999985	0.00937176068811989\\
5.44399999999984	0.00943411813212054\\
5.48399999999984	0.00940919744162558\\
5.52399999999983	0.00922714284094251\\
5.56399999999983	0.00901718427978985\\
5.60399999999983	0.00909058775621242\\
5.64399999999982	0.0098700752659486\\
5.68399999999982	0.0118131176204807\\
5.72399999999981	0.0153528070409947\\
5.76399999999981	0.0208466552909839\\
5.8039999999998	0.0285177835282867\\
5.8439999999998	0.0384128403221908\\
5.8839999999998	0.0504424681256374\\
5.92399999999979	0.0645464513729595\\
5.96399999999979	0.0809453668298041\\
6.00399999999978	0.100362539851396\\
6.04399999999978	0.124034491000433\\
6.08399999999977	0.153315053768853\\
6.12399999999977	0.188908374197207\\
6.16399999999976	0.230220336065847\\
6.20399999999976	0.275373620575802\\
6.24399999999976	0.321789032598611\\
6.28399999999975	0.366803529310035\\
6.32399999999975	0.408116122051191\\
6.36399999999974	0.444145698247992\\
6.40399999999974	0.47425033525469\\
6.44399999999973	0.498664001903547\\
6.48399999999973	0.518168684553299\\
6.52399999999972	0.533709160840735\\
6.56399999999972	0.546143042615527\\
6.60399999999972	0.556146031049726\\
6.64399999999971	0.564213605346866\\
6.68399999999971	0.570699354140663\\
6.7239999999997	0.575861351125297\\
6.7639999999997	0.579900573368947\\
6.80399999999969	0.582988637292687\\
6.84399999999969	0.585283786715199\\
6.88399999999969	0.58693538273003\\
6.92399999999968	0.588080579510606\\
6.96399999999968	0.588839714269811\\
7.00399999999967	0.589313148315183\\
7.04399999999967	0.589580619965232\\
7.08399999999966	0.589702919749239\\
7.12399999999966	0.589724879541389\\
7.16399999999965	0.589678724059665\\
7.20399999999965	0.589587182219319\\
7.24399999999965	0.589466074283249\\
7.28399999999964	0.589326315866715\\
7.32399999999964	0.589175395606613\\
7.36399999999963	0.589018421710101\\
7.40399999999963	0.588858860171037\\
7.44399999999962	0.58869904886913\\
7.48399999999962	0.588540551862243\\
7.52399999999961	0.588384401733781\\
7.56399999999961	0.588231263466137\\
7.60399999999961	0.588081548385904\\
7.6439999999996	0.587935492869971\\
7.6839999999996	0.587793215835166\\
7.72399999999959	0.587654758366465\\
7.76399999999959	0.587520110459654\\
7.80399999999958	0.587389223276987\\
7.84399999999958	0.587262024406858\\
7.88399999999957	0.587138429402229\\
7.92399999999957	0.587018345924291\\
7.96399999999957	0.586901676799956\\
8.00399999999956	0.58678832329276\\
8.04399999999956	0.586678185845685\\
8.08399999999955	0.586571164593029\\
8.12399999999955	0.586467159635667\\
8.16399999999954	0.586366070706832\\
8.20399999999954	0.586267795990313\\
8.24399999999954	0.586172230868718\\
8.28399999999953	0.586079266498442\\
8.32399999999953	0.585988787662354\\
8.36399999999952	0.585900670007578\\
};
\addlegendentry{$y_c$};

\addplot [color=black,dash pattern=on 1pt off 3pt on 3pt off 3pt,line width=2.0pt]
  table[row sep=crcr]{%
0.004	0\\
0.044	-0.00313272655695536\\
0.084	-0.00867844462227884\\
0.124	-0.0140079047712132\\
0.164	-0.018091443462256\\
0.204	-0.0206214541016379\\
0.244	-0.0215460958014989\\
0.284	-0.0209199058929612\\
0.324	-0.0189288412516289\\
0.364	-0.015949225187372\\
0.404	-0.0125364827854573\\
0.444	-0.00928587187324214\\
0.484	-0.00660490116139399\\
0.524	-0.00455275981396711\\
0.564	-0.00287791149070371\\
0.604	-0.00121856217740639\\
0.644	0.000675507091553434\\
0.684	0.00281821873137434\\
0.724000000000001	0.00501782179761879\\
0.764000000000001	0.00705430598784005\\
0.804000000000001	0.0089009669828089\\
0.844000000000001	0.0108496495199843\\
0.884000000000001	0.0135027071825912\\
0.924000000000001	0.0177145101636838\\
0.964000000000001	0.0245442194882406\\
1.004	0.0351584542788934\\
1.044	0.0505651174397519\\
1.084	0.0712175715250734\\
1.124	0.0967873452975105\\
1.164	0.126377917375725\\
1.204	0.159107331646396\\
1.244	0.19473414267802\\
1.284	0.233975796668523\\
1.324	0.278229120554815\\
1.364	0.328655313658648\\
1.404	0.385146443575777\\
1.444	0.445966207036838\\
1.484	0.508295240820383\\
1.524	0.569173884000304\\
1.564	0.626243935671721\\
1.604	0.678041809117235\\
1.644	0.723895161403655\\
1.684	0.763610171346251\\
1.724	0.797153637776484\\
1.764	0.824432001191227\\
1.804	0.845132457709113\\
1.844	0.858576312498352\\
1.884	0.863666267137082\\
1.924	0.859141864970853\\
1.964	0.84423425664056\\
2.004	0.819358542274757\\
2.044	0.78622945679985\\
2.084	0.747218369146149\\
2.124	0.704371483894412\\
2.164	0.658650613968891\\
2.204	0.609793434432555\\
2.244	0.556952305088282\\
2.284	0.49981192190462\\
2.324	0.439397521503897\\
2.364	0.377985881712436\\
2.404	0.31832900396078\\
2.444	0.262823213018213\\
2.484	0.213042663253684\\
2.524	0.169700804580705\\
2.564	0.132896193454363\\
2.604	0.102417827002018\\
2.644	0.0779235479195955\\
2.684	0.0589534648286063\\
2.724	0.0448635203897635\\
2.764	0.0347808226458393\\
2.804	0.0276544923332249\\
2.844	0.0224377518846748\\
2.884	0.0183416579056262\\
2.924	0.0149935215607505\\
2.964	0.0123777371182891\\
3.004	0.0106128542915693\\
3.044	0.00971674076270573\\
3.084	0.00948222516471695\\
3.124	0.00952432539596022\\
3.164	0.00948221684459078\\
3.204	0.00924552470914172\\
3.244	0.00903862266841597\\
3.284	0.00932813881865353\\
3.324	0.0106622106472434\\
3.364	0.0135484125036182\\
3.404	0.0183889873633274\\
3.444	0.0254366848064866\\
3.484	0.0347598850355408\\
3.524	0.0462675774782377\\
3.564	0.0598479899864983\\
3.604	0.075604999756669\\
3.644	0.0941007610994917\\
3.684	0.116443112162265\\
3.724	0.144012012790866\\
3.764	0.177774432816555\\
3.804	0.217554868477214\\
3.844	0.261826820226545\\
3.884	0.308163713612062\\
3.924	0.353960674636309\\
3.964	0.397003318091964\\
4.004	0.435737176789083\\
4.044	0.469300324575242\\
4.08399999999999	0.497429417593016\\
4.12399999999999	0.520316345183389\\
4.16399999999998	0.538455166955703\\
4.20399999999998	0.552501266272817\\
4.24399999999998	0.563157691454206\\
4.28399999999997	0.571096227647175\\
4.32399999999997	0.576912126204277\\
4.36399999999996	0.581105440285701\\
4.40399999999996	0.584080070596744\\
4.44399999999995	0.586152698548648\\
4.48399999999995	0.587565979059748\\
4.52399999999994	0.588502507589093\\
4.56399999999994	0.589097701949801\\
4.60399999999994	0.589450803534764\\
4.64399999999993	0.589633821927855\\
4.68399999999993	0.58969856458187\\
4.72399999999992	0.589682026713169\\
4.76399999999992	0.589610447505546\\
4.80399999999991	0.589502318627651\\
4.84399999999991	0.58937058997529\\
4.88399999999991	0.589224271641686\\
4.9239999999999	0.589069588266586\\
4.9639999999999	0.588910805289955\\
5.00399999999989	0.588750816931741\\
5.04399999999989	0.588591562448516\\
5.08399999999988	0.588434319428065\\
5.12399999999988	0.58827990952738\\
5.16399999999987	0.588128842171085\\
5.20399999999987	0.587981414485425\\
5.24399999999987	0.587837780485823\\
5.28399999999986	0.587697998746986\\
5.32399999999986	0.587562065070179\\
5.36399999999985	0.587429934727532\\
5.40399999999985	0.587301537490166\\
5.44399999999984	0.587176787676088\\
5.48399999999984	0.587055590769329\\
5.52399999999983	0.586937847680388\\
5.56399999999983	0.586823457379866\\
5.60399999999983	0.58671231839948\\
5.64399999999982	0.586604329527232\\
5.68399999999982	0.586499389904842\\
5.72399999999981	0.58639739865048\\
5.76399999999981	0.586298254067992\\
5.8039999999998	0.586201852457742\\
5.8439999999998	0.586108086509405\\
5.8839999999998	0.586016843230676\\
5.92399999999979	0.585928001346959\\
5.96399999999979	0.585841428096486\\
6.00399999999978	0.585756975345647\\
6.04399999999978	0.585674474965833\\
6.08399999999977	0.585593733454047\\
6.12399999999977	0.585514525856706\\
6.16399999999976	0.585436589185521\\
6.20399999999976	0.585359615715103\\
6.24399999999976	0.585283246843323\\
6.28399999999975	0.585207068589478\\
6.32399999999975	0.585130610292267\\
6.36399999999974	0.58505334859578\\
6.40399999999974	0.58497471924966\\
6.44399999999973	0.584894139373195\\
6.48399999999973	0.584811042320703\\
6.52399999999972	0.584724925789564\\
6.56399999999972	0.584635411129029\\
6.60399999999972	0.58454230813178\\
6.64399999999971	0.584445675738449\\
6.68399999999971	0.584345866477008\\
6.7239999999997	0.584243542659453\\
6.7639999999997	0.584139656253521\\
6.80399999999969	0.58403539139365\\
6.84399999999969	0.583932076643043\\
6.88399999999969	0.583831080641208\\
6.92399999999968	0.583733707564214\\
6.96399999999968	0.583641107342456\\
7.00399999999967	0.583554210875005\\
7.04399999999967	0.583473694463474\\
7.08399999999966	0.583399972199393\\
7.12399999999966	0.583333211221083\\
7.16399999999965	0.583273362927566\\
7.20399999999965	0.583220203106023\\
7.24399999999965	0.58317337492077\\
7.28399999999964	0.583132430236163\\
7.32399999999964	0.583096866348323\\
7.36399999999963	0.5830661566021\\
7.40399999999963	0.583039774445797\\
7.44399999999962	0.583017211209402\\
7.48399999999962	0.582997988325663\\
7.52399999999961	0.582981664915265\\
7.56399999999961	0.58296784169779\\
7.60399999999961	0.582956162129656\\
7.6439999999996	0.5829463115556\\
7.6839999999996	0.582938015024331\\
7.72399999999959	0.5829310342829\\
7.76399999999959	0.582925164340255\\
7.80399999999958	0.582920229884124\\
7.84399999999958	0.582916081748512\\
7.88399999999957	0.582912593560856\\
7.92399999999957	0.582909658646104\\
7.96399999999957	0.582907187227148\\
8.00399999999956	0.582905103934294\\
8.04399999999956	0.582903345618514\\
8.08399999999955	0.582901859451768\\
8.12399999999955	0.582900601291172\\
8.16399999999954	0.582899534280566\\
8.20399999999954	0.582898627662306\\
8.24399999999954	0.582897855772765\\
8.28399999999953	0.582897197196757\\
8.32399999999953	0.582896634058223\\
8.36399999999952	0.582896151426968\\
};
\addlegendentry{$y_u$};

\end{axis}

\begin{axis}[%
width=\figurewidth,
height=\figureheight,
scale only axis,
xmin=0,
xmax=8,
xlabel={Time [s]},
ymin=-20,
ymax=20,
ylabel absolute,
ylabel={$\text{Acceleration [m/s}^\text{2}\text{]}$},
at=(plot1.below south west),
anchor=above north west,
legend style={draw=black,fill=white,legend cell align=left}
]
\addplot [color=blue,solid,line width=1.5pt]
  table[row sep=crcr]{%
0.004	0\\
0.044	-2.9027536415637\\
0.084	-0.207769668167929\\
0.124	0.881208650046085\\
0.164	1.12449632466387\\
0.204	1.12460808955466\\
0.244	1.12823634798183\\
0.284	1.01307399288362\\
0.324	0.763145174443566\\
0.364	0.414980344915637\\
0.404	-0.0686367227711918\\
0.444	-0.441408881581333\\
0.484	-0.514047348699834\\
0.524	-0.411640454651934\\
0.564	-0.0999324696707817\\
0.604	0.180771575354148\\
0.644	0.181459886897203\\
0.684	0.141294747923578\\
0.724000000000001	-0.0696064020614331\\
0.764000000000001	-0.211830259775723\\
0.804000000000001	-0.0307645363064899\\
0.844000000000001	0.283978387823865\\
0.884000000000001	0.828510227188342\\
0.924000000000001	1.49504879980592\\
0.964000000000001	2.27494170741295\\
1.004	3.05444195716109\\
1.044	3.47634328481466\\
1.084	3.41197944931322\\
1.124	2.82903156436117\\
1.164	2.02554429929457\\
1.204	1.64418494269888\\
1.244	1.8859108439715\\
1.284	2.77510763945156\\
1.324	3.78100710864371\\
1.364	4.19118755336548\\
1.404	3.4205594314191\\
1.444	1.54882984042133\\
1.484	-0.815396440757918\\
1.524	-2.89298593844169\\
1.564	-4.43595866420882\\
1.604	-5.07087978587714\\
1.644	-4.92072308843771\\
1.684	-4.37466596218751\\
1.724	-3.51048045090243\\
1.764	-2.75732102003503\\
1.804	-2.1744265648088\\
1.844	-1.75097447826027\\
1.884	-1.39988268008332\\
1.924	-1.2631623002362\\
1.964	-1.28611281290623\\
2.004	-1.39981273164328\\
2.044	-1.03126928792754\\
2.084	-2.64623252520437\\
2.124	-4.59467839917493\\
2.164	-5.40978610418076\\
2.204	-4.03567409372921\\
2.244	-2.3705951523928\\
2.284	-1.83949746488297\\
2.324	-1.73047360035245\\
2.364	-1.77452522393808\\
2.404	-1.81624351840956\\
2.444	-1.79887425035857\\
2.484	-1.78397797966384\\
2.524	-1.88608707202277\\
2.564	-1.8423961729156\\
2.604	-1.90181887954909\\
2.644	-1.89218467591261\\
2.684	-1.93035369194052\\
2.724	-1.93257025614812\\
2.764	-1.94371497198818\\
2.804	-1.93368211315791\\
2.844	-1.97086737649564\\
2.884	-1.95185307122789\\
2.924	-2.01155841021689\\
2.964	-1.96482260469466\\
3.004	-2.00140431129826\\
3.044	-1.39478262081573\\
3.084	-0.822064734540804\\
3.124	-0.430549296485583\\
3.164	-0.119156979458528\\
3.204	0.0410286854006283\\
3.244	0.141865417789495\\
3.284	0.157687289814043\\
3.324	0.188829005691818\\
3.364	0.136447895861986\\
3.404	0.112589699522618\\
3.444	0.100589277476642\\
3.484	-0.0642695170702011\\
3.524	-0.0603972967763051\\
3.564	-0.115066486696592\\
3.604	-0.229777358221886\\
3.644	-0.389023061976059\\
3.684	-0.532084430638367\\
3.724	-0.600641628185225\\
3.764	-0.75951872653301\\
3.804	-0.868731772433633\\
3.844	-0.932347775527374\\
3.884	-1.05511076011829\\
3.924	-1.06622933425011\\
3.964	-1.0540997704643\\
4.004	-1.11280866806112\\
4.044	-1.07849855889618\\
4.08399999999999	-1.11376890933085\\
4.12399999999999	-1.15934249142997\\
4.16399999999998	-1.26463011078145\\
4.20399999999998	-1.34232387854895\\
4.24399999999998	-1.32578632329502\\
4.28399999999997	-1.25406135667534\\
4.32399999999997	-1.07145364809507\\
4.36399999999996	-0.87958227805485\\
4.40399999999996	-0.509985475732647\\
4.44399999999995	-0.141385466058728\\
4.48399999999995	0.271846381043308\\
4.52399999999994	0.73238833102394\\
4.56399999999994	1.0930075249337\\
4.60399999999994	1.37616558335663\\
4.64399999999993	1.67477823354949\\
4.68399999999993	1.78228898441653\\
4.72399999999992	1.85213942946394\\
4.76399999999992	1.93701090182511\\
4.80399999999991	1.88627873323226\\
4.84399999999991	1.83107462046373\\
4.88399999999991	1.76911470465063\\
4.9239999999999	1.53601556498047\\
4.9639999999999	1.28894561978675\\
5.00399999999989	0.987800810296902\\
5.04399999999989	0.709722045720179\\
5.08399999999988	0.396516944782944\\
5.12399999999988	0.283087819000548\\
5.16399999999987	0.205799408292322\\
5.20399999999987	0.221767100339811\\
5.24399999999987	0.241515869113469\\
5.28399999999986	0.338016487044539\\
5.32399999999986	0.39695747043936\\
5.36399999999985	0.282630726395867\\
5.40399999999985	0.092344415182896\\
5.44399999999984	-0.0367430521025134\\
5.48399999999984	-0.152270686395203\\
5.52399999999983	-0.0992958250400601\\
5.56399999999983	0.117328714987911\\
5.60399999999983	0.447048725614749\\
5.64399999999982	0.650743624195656\\
5.68399999999982	0.979057662711682\\
5.72399999999981	1.21286476440146\\
5.76399999999981	1.42916224394288\\
5.8039999999998	1.41234405490334\\
5.8439999999998	1.42952556882788\\
5.8839999999998	1.31353836559292\\
5.92399999999979	1.35648043746273\\
5.96399999999979	1.65715422296169\\
6.00399999999978	2.44311138020331\\
6.04399999999978	3.4166780866227\\
6.08399999999977	4.19049654637951\\
6.12399999999977	4.11553668143992\\
6.16399999999976	3.00430523205945\\
6.20399999999976	1.21375207741664\\
6.24399999999976	-0.648343514189284\\
6.28399999999975	-2.34022279984987\\
6.32399999999975	-3.48210654158497\\
6.36399999999974	-3.98216764247147\\
6.40399999999974	-3.92562966815789\\
6.44399999999973	-3.35351238975908\\
6.48399999999973	-2.74188717908495\\
6.52399999999972	-2.14822701044239\\
6.56399999999972	-1.62941106693178\\
6.60399999999972	-1.23451426611824\\
6.64399999999971	-0.939392655227228\\
6.68399999999971	-0.773690115457287\\
6.7239999999997	-0.63576801840217\\
6.7639999999997	-0.468872076904962\\
6.80399999999969	-0.41639025654904\\
6.84399999999969	-0.338987161791503\\
6.88399999999969	-0.299167312363662\\
6.92399999999968	-0.209056538897203\\
6.96399999999968	-0.126429852433932\\
7.00399999999967	-0.0895741499611\\
7.04399999999967	-0.0598278215082402\\
7.08399999999966	-0.0142795323516778\\
7.12399999999966	0.0315166337371501\\
7.16399999999965	-0.0273631166644214\\
7.20399999999965	-0.0110851291279945\\
7.24399999999965	-0.0314864955982562\\
7.28399999999964	0.0130106911349725\\
7.32399999999964	0.00322195067234316\\
7.36399999999963	-0.0122614011730241\\
7.40399999999963	-0.000205827676424687\\
7.44399999999962	-0.0177849029535832\\
7.48399999999962	-0.0238403438166052\\
7.52399999999961	-0.0524037110842531\\
7.56399999999961	-0.0108626495418214\\
7.60399999999961	0.00268628293938206\\
7.6439999999996	-0.0146301740094456\\
7.6839999999996	-0.0568532127048712\\
7.72399999999959	0.0162356435261315\\
7.76399999999959	0.0146600487176354\\
7.80399999999958	0.0475105656600779\\
7.84399999999958	0.0245002026488293\\
7.88399999999957	-0.00483756867206194\\
7.92399999999957	0.0202729852564928\\
7.96399999999957	0.0379758755533685\\
8.00399999999956	-0.0195175255186057\\
8.04399999999956	0.0283978207613002\\
8.08399999999955	-0.0109337087862377\\
8.12399999999955	-0.00974675880772044\\
8.16399999999954	0.0307153904250279\\
8.20399999999954	0.0124669742882541\\
8.24399999999954	0.0208114628447497\\
8.28399999999953	-0.0305679348271995\\
8.32399999999953	0.0428519744178768\\
8.36399999999952	0.0179447605483786\\
};
\addlegendentry{$\ddot{y}_{r}$};

\addplot [color=red,dashed,line width=2.0pt]
  table[row sep=crcr]{%
0.004	0\\
0.044	-1.59807938252395\\
0.084	0.113383510223496\\
0.124	0.776500313869451\\
0.164	0.972128866652748\\
0.204	1.00776566720201\\
0.244	0.981506692096663\\
0.284	0.877335203785843\\
0.324	0.655013808902723\\
0.364	0.313366384309601\\
0.404	-0.0705858287901475\\
0.444	-0.353456587077216\\
0.484	-0.41679389058884\\
0.524	-0.267148458538123\\
0.564	-0.0291915294113499\\
0.604	0.148286430051225\\
0.644	0.174447185705549\\
0.684	0.0562351968454356\\
0.724000000000001	-0.0992520311087818\\
0.764000000000001	-0.143403693611697\\
0.804000000000001	0.015086346970987\\
0.844000000000001	0.373962842979537\\
0.884000000000001	0.892949086528421\\
0.924000000000001	1.54528900866533\\
0.964000000000001	2.28448077233513\\
1.004	2.95294933145411\\
1.044	3.28252257093076\\
1.084	3.10225741334615\\
1.124	2.54656399426813\\
1.164	1.97850065842784\\
1.204	1.7700614585188\\
1.244	2.14511765863693\\
1.284	3.01209097995189\\
1.324	3.84700489158919\\
1.364	3.94269164175787\\
1.404	2.95703318599866\\
1.444	1.15281205839461\\
1.484	-0.907356102121407\\
1.524	-2.70552905830204\\
1.564	-3.87608369325218\\
1.604	-4.27573954392183\\
1.644	-4.04027859539291\\
1.684	-3.46137351285897\\
1.724	-2.79690270985585\\
1.764	-2.20887150485158\\
1.804	-1.75887241944408\\
1.844	-1.455573878862\\
1.884	-1.28698453664913\\
1.924	-1.24428872636667\\
1.964	-1.32553012780275\\
2.004	-1.53895419384541\\
2.044	-1.9700983914397\\
2.084	-2.80993724920091\\
2.124	-3.23591628536018\\
2.164	-2.1031802617703\\
2.204	0.0622200517903645\\
2.244	1.05447445721146\\
2.284	0.918685426530208\\
2.324	0.602916721533253\\
2.364	0.379890942049619\\
2.404	0.245246178662545\\
2.444	0.164658623604367\\
2.484	0.114973365329358\\
2.524	0.08315129745068\\
2.564	0.0619972751359355\\
2.604	0.0474547873096962\\
2.644	0.0371592203235286\\
2.684	0.0296824343322234\\
2.724	0.0241320036568057\\
2.764	0.0199324438574729\\
2.804	0.0167019011560511\\
2.844	0.014180401901436\\
2.884	0.0121868215812662\\
2.924	0.0105923402838919\\
2.964	0.00930363606171948\\
3.004	0.00819398616077506\\
3.044	0.00408678900889337\\
3.084	-0.00270163899939539\\
3.124	-0.0106216620433328\\
3.164	-0.0193444507129082\\
3.204	-0.0289769322053002\\
3.244	-0.0399335434568711\\
3.284	-0.0528201506753893\\
3.324	-0.068344348911414\\
3.364	-0.0872897843804866\\
3.404	-0.110667598158895\\
3.444	-0.139674338024764\\
3.484	-0.175375396590795\\
3.524	-0.219265432749178\\
3.564	-0.272434378587481\\
3.604	-0.335471834182809\\
3.644	-0.408509882203961\\
3.684	-0.48962433351221\\
3.724	-0.574094864231227\\
3.764	-0.655168248449409\\
3.804	-0.725067286362149\\
3.844	-0.777030210580636\\
3.884	-0.806972681823741\\
3.924	-0.819783070643485\\
3.964	-0.826406881675136\\
4.004	-0.839137804083457\\
4.044	-0.87046441049501\\
4.08399999999999	-0.922721850836268\\
4.12399999999999	-0.993446983446545\\
4.16399999999998	-1.06649266494585\\
4.20399999999998	-1.11648646643692\\
4.24399999999998	-1.1153124843884\\
4.28399999999997	-1.04149114660166\\
4.32399999999997	-0.883979481721362\\
4.36399999999996	-0.635057646365973\\
4.40399999999996	-0.308179694181016\\
4.44399999999995	0.0738380597095754\\
4.48399999999995	0.478523062256755\\
4.52399999999994	0.876140981127762\\
4.56399999999994	1.23230342513313\\
4.60399999999994	1.51442247379952\\
4.64399999999993	1.7177175529066\\
4.68399999999993	1.85139544985668\\
4.72399999999992	1.92796798806618\\
4.76399999999992	1.95227407569702\\
4.80399999999991	1.93479761780841\\
4.84399999999991	1.86617832413951\\
4.88399999999991	1.73728652547981\\
4.9239999999999	1.54249838970842\\
4.9639999999999	1.28513007096056\\
5.00399999999989	0.987864159348867\\
5.04399999999989	0.693772213864755\\
5.08399999999988	0.454067469898802\\
5.12399999999988	0.305969918881016\\
5.16399999999987	0.255031669669022\\
5.20399999999987	0.274786640994484\\
5.24399999999987	0.322040324197592\\
5.28399999999986	0.351911963062988\\
5.32399999999986	0.328052658818581\\
5.36399999999985	0.234063103303595\\
5.40399999999985	0.0880924895167839\\
5.44399999999984	-0.0515045288174138\\
5.48399999999984	-0.109821750175002\\
5.52399999999983	-0.0406443177550898\\
5.56399999999983	0.148107640634947\\
5.60399999999983	0.410909388779961\\
5.64399999999982	0.697311528425587\\
5.68399999999982	0.969924187676388\\
5.72399999999981	1.19882646050631\\
5.76399999999981	1.34812766808711\\
5.8039999999998	1.38603166133971\\
5.8439999999998	1.32838379370529\\
5.8839999999998	1.27155907540227\\
5.92399999999979	1.37420704158174\\
5.96399999999979	1.78920776940103\\
6.00399999999978	2.55347777253097\\
6.04399999999978	3.45100603847029\\
6.08399999999977	3.99451135985984\\
6.12399999999977	3.71881754947761\\
6.16399999999976	2.58409284516374\\
6.20399999999976	0.964732054924272\\
6.24399999999976	-0.71268898553769\\
6.28399999999975	-2.14582797891336\\
6.32399999999975	-3.1153835277827\\
6.36399999999974	-3.50641495851974\\
6.40399999999974	-3.37646525205722\\
6.44399999999973	-2.93096363261532\\
6.48399999999973	-2.39684715713071\\
6.52399999999972	-1.91348803440093\\
6.56399999999972	-1.53047472059747\\
6.60399999999972	-1.24485856408992\\
6.64399999999971	-1.0350207998931\\
6.68399999999971	-0.875215250137145\\
6.7239999999997	-0.744951569692599\\
6.7639999999997	-0.629913370507563\\
6.80399999999969	-0.522553152852712\\
6.84399999999969	-0.421531198541547\\
6.88399999999969	-0.329718608281867\\
6.92399999999968	-0.250053182068075\\
6.96399999999968	-0.184295562867443\\
7.00399999999967	-0.132498216230352\\
7.04399999999967	-0.0932597344464556\\
7.08399999999966	-0.0644464767868446\\
7.12399999999966	-0.0437850680005339\\
7.16399999999965	-0.0292221634230656\\
7.20399999999965	-0.0190799351448403\\
7.24399999999965	-0.0120727216745475\\
7.28399999999964	-0.00725648897053083\\
7.32399999999964	-0.00396487041031101\\
7.36399999999963	-0.00172875168358663\\
7.40399999999963	-0.000218308625837052\\
7.44399999999962	0.000791171147413489\\
7.48399999999962	0.00145704506276112\\
7.52399999999961	0.00188477893426435\\
7.56399999999961	0.00214903710562687\\
7.60399999999961	0.0023003666155097\\
7.6439999999996	0.0023761069645005\\
7.6839999999996	0.00240117583982096\\
7.72399999999959	0.00239418164903221\\
7.76399999999959	0.00236273735490477\\
7.80399999999958	0.00231659550563493\\
7.84399999999958	0.00226285202497641\\
7.88399999999957	0.00220419576935434\\
7.92399999999957	0.00214267685680585\\
7.96399999999957	0.00208027521228802\\
8.00399999999956	0.00201748727396661\\
8.04399999999956	0.00195473647186749\\
8.08399999999955	0.0018920341246632\\
8.12399999999955	0.0018292590671851\\
8.16399999999954	0.0017656758157067\\
8.20399999999954	0.00170052428982195\\
8.24399999999954	0.00163289546842067\\
8.28399999999953	0.00156145758210562\\
8.32399999999953	0.00148453414059768\\
8.36399999999952	0.00139999099217402\\
};
\addlegendentry{$\ddot{y}_c$};

\addplot [color=black,dash pattern=on 1pt off 3pt on 3pt off 3pt,line width=2.0pt]
  table[row sep=crcr]{%
0.004	0\\
0.044	-1.59954299014054\\
0.084	0.112244330600025\\
0.124	0.779415907514183\\
0.164	0.974340781299854\\
0.204	1.00789613448221\\
0.244	0.981889560654181\\
0.284	0.879638712474928\\
0.324	0.657613876951821\\
0.364	0.312544034052864\\
0.404	-0.0758329244056085\\
0.444	-0.358629876102407\\
0.484	-0.417524254956118\\
0.524	-0.264206835738674\\
0.564	-0.0257165361475424\\
0.604	0.150091393858293\\
0.644	0.173756653586864\\
0.684	0.053778451992183\\
0.724000000000001	-0.10124004023411\\
0.764000000000001	-0.142621185086257\\
0.804000000000001	0.0195788496590049\\
0.844000000000001	0.38188929410743\\
0.884000000000001	0.903616576576362\\
0.924000000000001	1.55773651233641\\
0.964000000000001	2.29808452337993\\
1.004	2.97130090659434\\
1.044	3.31527592263806\\
1.084	3.14885627773208\\
1.124	2.57879902609129\\
1.164	1.97236638978646\\
1.204	1.74674893040044\\
1.244	2.1456580863959\\
1.284	3.0444733206634\\
1.324	3.87834680857241\\
1.364	3.93610841509374\\
1.404	2.9168773676545\\
1.444	1.14358753926628\\
1.484	-0.764403753417956\\
1.524	-2.30597443251475\\
1.564	-3.27206161988472\\
1.604	-3.71687799024603\\
1.644	-3.84059935886702\\
1.684	-3.8525528877986\\
1.724	-3.89478474842458\\
1.764	-4.06423935751359\\
1.804	-4.45687962283568\\
1.844	-5.12915890554929\\
1.884	-5.95601930841547\\
1.924	-6.53180027601709\\
1.964	-6.36993936717497\\
2.004	-5.33484858491606\\
2.044	-3.81215973869931\\
2.084	-2.44277722246786\\
2.124	-1.74955873361838\\
2.164	-1.87761372606672\\
2.204	-2.4633627045808\\
2.244	-2.77458234492599\\
2.284	-2.22377642961008\\
2.324	-0.81810751367034\\
2.364	0.944986001325874\\
2.404	2.50945045890828\\
2.444	3.55271066632738\\
2.484	4.03406092180096\\
2.524	4.10749549652899\\
2.564	3.97765797891656\\
2.604	3.76949623980906\\
2.644	3.49504351763457\\
2.684	3.10903057082453\\
2.724	2.57597537684592\\
2.764	1.91752823255133\\
2.804	1.24161514436171\\
2.844	0.715836816805452\\
2.884	0.458476204716666\\
2.924	0.443889617692125\\
2.964	0.530088973054349\\
3.004	0.560236435709329\\
3.044	0.443916317665101\\
3.084	0.199235153293816\\
3.124	-0.048229934474968\\
3.164	-0.142991794742527\\
3.204	-0.0175968810182347\\
3.244	0.272536434599432\\
3.284	0.620900420695983\\
3.324	0.946160560318816\\
3.364	1.20753590658553\\
3.404	1.37837125443067\\
3.444	1.43259368586402\\
3.484	1.37645822694343\\
3.524	1.2895048949391\\
3.564	1.3197337210566\\
3.604	1.62999041531844\\
3.644	2.30220663159971\\
3.684	3.19951464678653\\
3.724	3.89643692109493\\
3.764	3.88738929490973\\
3.804	2.98700635306512\\
3.844	1.46618110466839\\
3.884	-0.20140742222897\\
3.924	-1.63525048821201\\
3.964	-2.65261452398771\\
4.004	-3.22782464849213\\
4.044	-3.41853565217931\\
4.08399999999999	-3.31468084666273\\
4.12399999999999	-3.01313164538305\\
4.16399999999998	-2.6039818075494\\
4.20399999999998	-2.16071827720884\\
4.24399999999998	-1.73486070086359\\
4.28399999999997	-1.35627483383283\\
4.32399999999997	-1.03756002418565\\
4.36399999999996	-0.779758927640912\\
4.40399999999996	-0.577422640633427\\
4.44399999999995	-0.422268031144218\\
4.48399999999995	-0.305449104685459\\
4.52399999999994	-0.218773868783872\\
4.56399999999994	-0.155230113215969\\
4.60399999999994	-0.10910820459114\\
4.64399999999993	-0.0759164167060615\\
4.68399999999993	-0.0522083282671145\\
4.72399999999992	-0.0353889175410091\\
4.76399999999992	-0.0235326320042473\\
4.80399999999991	-0.015227247463255\\
4.84399999999991	-0.00944681845785937\\
4.88399999999991	-0.00545189279219722\\
4.9239999999999	-0.0027130791075056\\
4.9639999999999	-0.000853580154019766\\
5.00399999999989	0.000393402360879092\\
5.04399999999989	0.00121588659909386\\
5.08399999999988	0.00174578449573312\\
5.12399999999988	0.00207524660316089\\
5.16399999999987	0.00226837041189437\\
5.20399999999987	0.0023695419900939\\
5.24399999999987	0.00240935067914656\\
5.28399999999986	0.00240876482939938\\
5.32399999999986	0.00238206788462978\\
5.36399999999985	0.0023389145947109\\
5.40399999999985	0.00228576502053481\\
5.44399999999984	0.00222687987980816\\
5.48399999999984	0.00216500735019061\\
5.52399999999983	0.00210185313781284\\
5.56399999999983	0.0020383982802707\\
5.60399999999983	0.00197510972168548\\
5.64399999999982	0.00191207494202302\\
5.68399999999982	0.00184908223414411\\
5.72399999999981	0.0017856614549477\\
5.76399999999981	0.00172109543329775\\
5.8039999999998	0.00165440915847103\\
5.8439999999998	0.0015843420668739\\
5.8839999999998	0.0015093080257962\\
5.92399999999979	0.00142734796885444\\
5.96399999999979	0.00133608169853978\\
6.00399999999978	0.00123266840717009\\
6.04399999999978	0.00111379036147059\\
6.08399999999977	0.00097568137134586\\
6.12399999999977	0.000814231380579757\\
6.16399999999976	0.000625210464398114\\
6.20399999999976	0.000404668050803393\\
6.24399999999976	0.000149572112847882\\
6.28399999999975	-0.000141249881787784\\
6.32399999999975	-0.000465835029304582\\
6.36399999999974	-0.000817513468245631\\
6.40399999999974	-0.00118321517247105\\
6.44399999999973	-0.00154202797694196\\
6.48399999999973	-0.00186460715242412\\
6.52399999999972	-0.0021141933158696\\
6.56399999999972	-0.00224992007488204\\
6.60399999999972	-0.00223262652288668\\
6.64399999999971	-0.00203251138606636\\
6.68399999999971	-0.00163693118670242\\
6.7239999999997	-0.00105597835435195\\
6.7639999999997	-0.00032370298084723\\
6.80399999999969	0.000505914625030705\\
6.84399999999969	0.00136736350198578\\
6.88399999999969	0.00219405173660347\\
6.92399999999968	0.00292802861854526\\
6.96399999999968	0.00352681380396087\\
7.00399999999967	0.00396639009652421\\
7.04399999999967	0.00424062744071652\\
7.08399999999966	0.00435818059148931\\
7.12399999999966	0.0043381382345849\\
7.16399999999965	0.00420552984141242\\
7.20399999999965	0.00398743127308889\\
7.24399999999965	0.00371002906526718\\
7.28399999999964	0.00339670610411125\\
7.32399999999964	0.00306702750355998\\
7.36399999999963	0.0027364217136277\\
7.40399999999963	0.00241633827143159\\
7.44399999999962	0.00211468987863567\\
7.48399999999962	0.0018364294065429\\
7.52399999999961	0.00158415720284034\\
7.56399999999961	0.00135869295960074\\
7.60399999999961	0.00115957650639598\\
7.6439999999996	0.000985483151911338\\
7.6839999999996	0.000834552917464963\\
7.72399999999959	0.000704640937363636\\
7.76399999999959	0.00059350013546641\\
7.80399999999958	0.000498908439203837\\
7.84399999999958	0.000418752334455796\\
7.88399999999957	0.000351077256536209\\
7.92399999999957	0.000294113656343146\\
7.96399999999957	0.000246285882532864\\
8.00399999999956	0.000206209453611982\\
8.04399999999956	0.000172680937576856\\
8.08399999999955	0.000144663536039873\\
8.12399999999955	0.000121270575991518\\
8.16399999999954	0.000101748420861654\\
8.20399999999954	8.54597920047183e-05\\
8.24399999999954	7.1868109581084e-05\\
8.28399999999953	6.05231894001987e-05\\
8.32399999999953	5.1048443177436e-05\\
8.36399999999952	4.3129604157568e-05\\
};
\addlegendentry{$\ddot{y}_u$};

\end{axis}
\end{tikzpicture}%

%% file: chapters/paper2/figs/ourmove.tex
%
%
\begin{tikzpicture}

\begin{axis}[%
width=\figurewidth,
height=\figureheight,
scale only axis,
xmin=0,
xmax=8,
ymin=-0.2,
ymax=1.5,
ylabel absolute,
ylabel={Position [m]},
name=plot1,
legend style={draw=black,fill=white,legend cell align=left}
]
\addplot [color=blue,solid,line width=1.5pt]
  table[row sep=crcr]{%
0.004	0\\
0.044	-0.00149007650688207\\
0.084	-0.00718417703465765\\
0.124	-0.0133754529038121\\
0.164	-0.0182715037921661\\
0.204	-0.0213967145350968\\
0.244	-0.022688422593945\\
0.284	-0.0222219815231073\\
0.324	-0.0201923720791924\\
0.364	-0.0169856303393988\\
0.404	-0.0131752224640764\\
0.444	-0.00944252461819106\\
0.484	-0.00631698329266422\\
0.524	-0.00401569258056918\\
0.564	-0.00232808176527732\\
0.604	-0.000825547771199802\\
0.644	0.000881974217450561\\
0.684	0.00292384259425027\\
0.724000000000001	0.00515414619298134\\
0.764000000000001	0.00728045746545773\\
0.804000000000001	0.00915188762160465\\
0.844000000000001	0.0109434868454917\\
0.884000000000001	0.0132282303313121\\
0.924000000000001	0.0168354578510506\\
0.964000000000001	0.0228398542988777\\
1.004	0.0324920171842281\\
1.044	0.0469614407522817\\
1.084	0.0668414430077267\\
1.124	0.0918092195801368\\
1.164	0.120847174924954\\
1.204	0.152859413262914\\
1.244	0.187348595074413\\
1.284	0.224695815155656\\
1.324	0.266143193910852\\
1.364	0.313190324002816\\
1.404	0.366473267987174\\
1.444	0.425065396183653\\
1.484	0.486728961110872\\
1.524	0.548762191954327\\
1.564	0.608558486774525\\
1.604	0.663859943146762\\
1.644	0.713004570775106\\
1.684	0.755078931293662\\
1.724	0.789862576294827\\
1.764	0.817663893242637\\
1.804	0.839059952130222\\
1.844	0.854664622197637\\
1.884	0.864981485353106\\
1.924	0.870274856262413\\
1.964	0.870564041065266\\
2.004	0.867009430375245\\
2.044	0.873009430375245\\
2.084	0.879009430375246\\
2.124	0.885009430375246\\
2.164	0.891009430375246\\
2.204	0.897009430375247\\
2.244	0.903009430375247\\
2.284	0.909009430375248\\
2.324	0.915009430375248\\
2.364	0.921009430375249\\
2.404	0.927009430375249\\
2.444	0.93300943037525\\
2.484	0.93900943037525\\
2.524	0.945009430375251\\
2.564	0.951009430375251\\
2.604	0.957009430375251\\
2.644	0.963009430375252\\
2.684	0.969009430375252\\
2.724	0.975009430375253\\
2.764	0.981009430375253\\
2.804	0.987009430375254\\
2.844	0.993009430375254\\
2.884	0.999009430375255\\
2.924	1.00500943037525\\
2.964	1.01100943037525\\
3.004	1.01669711116837\\
3.044	1.01537043454851\\
3.084	1.00700943783036\\
3.124	0.994481942564756\\
3.164	0.979765891641821\\
3.204	0.964217410398364\\
3.244	0.948720436558853\\
3.284	0.933842839063582\\
3.324	0.919905306374577\\
3.364	0.907087196387934\\
3.404	0.895452285982698\\
3.444	0.884967371698335\\
3.484	0.875582426783011\\
3.524	0.867225087586943\\
3.564	0.859830092224389\\
3.604	0.853268379105453\\
3.644	0.847422549999469\\
3.684	0.842153118392589\\
3.724	0.837375547595964\\
3.764	0.832961780883529\\
3.804	0.828804308479764\\
3.844	0.824776317552518\\
3.884	0.820721963433818\\
3.924	0.816503931338501\\
3.964	0.811938414405252\\
4.004	0.806823913630111\\
4.044	0.800920750443537\\
4.08399999999999	0.793981961163223\\
4.12399999999999	0.785766899007308\\
4.16399999999998	0.776073692776362\\
4.20399999999998	0.764749110347733\\
4.24399999999998	0.751695087702882\\
4.28399999999997	0.736884596434419\\
4.32399999999997	0.720380234964469\\
4.36399999999996	0.702262184109686\\
4.40399999999996	0.682575152038007\\
4.44399999999995	0.661355785266757\\
4.48399999999995	0.638563236065124\\
4.52399999999994	0.614091015700582\\
4.56399999999994	0.587846079003744\\
4.60399999999994	0.559695591048448\\
4.64399999999993	0.529598909289965\\
4.68399999999993	0.497572180217168\\
4.72399999999992	0.463778704873771\\
4.76399999999992	0.428506562910324\\
4.80399999999991	0.392168483790882\\
4.84399999999991	0.35527236385504\\
4.88399999999991	0.318370855157496\\
4.9239999999999	0.282100287653021\\
4.9639999999999	0.247044602336694\\
5.00399999999989	0.213754198441423\\
5.04399999999989	0.182687992554048\\
5.08399999999988	0.154193042047243\\
5.12399999999988	0.128512620052217\\
5.16399999999987	0.105822971288077\\
5.20399999999987	0.0862471344055697\\
5.24399999999987	0.0697993870042092\\
5.28399999999986	0.0564114675057759\\
5.32399999999986	0.0459098015107865\\
5.36399999999985	0.0379633400754392\\
5.40399999999985	0.0320422229175078\\
5.44399999999984	0.02755821961928\\
5.48399999999984	0.0239327389412755\\
5.52399999999983	0.0208177474601405\\
5.56399999999983	0.0180711083010842\\
5.60399999999983	0.0157178886970853\\
5.64399999999982	0.0139076263299026\\
5.68399999999982	0.0127301814058253\\
5.72399999999981	0.0121664844190914\\
5.76399999999981	0.0119979712741979\\
5.8039999999998	0.0119213505915937\\
5.8439999999998	0.0116849426680516\\
5.8839999999998	0.0112371505212149\\
5.92399999999979	0.0107987126272506\\
5.96399999999979	0.01077617429322\\
6.00399999999978	0.0116801841935651\\
6.04399999999978	0.0140515908861926\\
6.08399999999977	0.018331868876258\\
6.12399999999977	0.024851373693442\\
6.16399999999976	0.0337903862518339\\
6.20399999999976	0.0450955936670279\\
6.24399999999976	0.0586069830619882\\
6.28399999999975	0.0742628569704861\\
6.32399999999975	0.0924339857704552\\
6.36399999999974	0.114112315921437\\
6.40399999999974	0.140708652963582\\
6.44399999999973	0.173415612933884\\
6.48399999999973	0.212334017208187\\
6.52399999999972	0.256076222370583\\
6.56399999999972	0.302400098506869\\
6.60399999999972	0.348942049816365\\
6.64399999999971	0.393535181579826\\
6.68399999999971	0.434281418755484\\
6.7239999999997	0.469707485995973\\
6.7639999999997	0.498999296014995\\
6.80399999999969	0.522162041815546\\
6.84399999999969	0.539871620300882\\
6.88399999999969	0.553135688058113\\
6.92399999999968	0.562985846197341\\
6.96399999999968	0.570271511497115\\
7.00399999999967	0.575669268018902\\
7.04399999999967	0.579680029386752\\
7.08399999999966	0.582634830871889\\
7.12399999999966	0.584779930030068\\
7.16399999999965	0.58628121906424\\
7.20399999999965	0.587312385203043\\
7.24399999999965	0.587998961834411\\
7.28399999999964	0.588420104416366\\
7.32399999999964	0.588652179978842\\
7.36399999999963	0.588759484989703\\
7.40399999999963	0.588792429574274\\
7.44399999999962	0.588800381053681\\
7.48399999999962	0.58876980651275\\
7.52399999999961	0.588718274230565\\
7.56399999999961	0.588638978867804\\
7.60399999999961	0.588553518594681\\
7.6439999999996	0.588458246469626\\
7.6839999999996	0.588365291957639\\
7.72399999999959	0.588286067159156\\
7.76399999999959	0.588196238835585\\
7.80399999999958	0.588091318913074\\
7.84399999999958	0.5879934802302\\
7.88399999999957	0.587897975353531\\
7.92399999999957	0.587810078975999\\
7.96399999999957	0.587729733301312\\
8.00399999999956	0.587637010842206\\
8.04399999999956	0.587528833305478\\
8.08399999999955	0.58741912681674\\
8.12399999999955	0.587292198494406\\
8.16399999999954	0.587164703678012\\
8.20399999999954	0.587044618158207\\
8.24399999999954	0.586951919947081\\
8.28399999999953	0.586849295984493\\
8.32399999999953	0.586728532395892\\
8.36399999999952	0.586601426827453\\
};
\addlegendentry{$y_a$};

\addplot [color=red,dashed,line width=2.0pt]
  table[row sep=crcr]{%
0.004	0\\
0.044	-0.00318904077583302\\
0.084	-0.00875007451985893\\
0.124	-0.0140698745536684\\
0.164	-0.0181428819861451\\
0.204	-0.0206713502526308\\
0.244	-0.02159980879444\\
0.284	-0.0209779474168539\\
0.324	-0.0189912708828956\\
0.364	-0.0160172868241699\\
0.404	-0.0126086921485921\\
0.444	-0.00935503316393776\\
0.484	-0.00666316437489098\\
0.524	-0.00459908734663793\\
0.564	-0.00291818201121589\\
0.604	-0.00125984620009107\\
0.644	0.000629766630858564\\
0.684	0.00276967850458165\\
0.724000000000001	0.00497114312455852\\
0.764000000000001	0.0070125244363929\\
0.804000000000001	0.00886135127462666\\
0.844000000000001	0.0108023566537318\\
0.884000000000001	0.0134316522050041\\
0.924000000000001	0.0175989289850452\\
0.964000000000001	0.0243589645713636\\
1.004	0.0348630336803219\\
1.044	0.0500582871763218\\
1.084	0.0702605703791762\\
1.124	0.0949892848054932\\
1.164	0.123327858835251\\
1.204	0.154528997763876\\
1.244	0.188460065756344\\
1.284	0.225772157610198\\
1.324	0.267692472469263\\
1.364	0.31531734037054\\
1.404	0.368720601342495\\
1.444	0.426641978348144\\
1.484	0.486977979790369\\
1.524	0.547382406616174\\
1.564	0.605543473064484\\
1.604	0.659370073772813\\
1.644	0.707267695474517\\
1.684	0.748345559573886\\
1.724	0.782404299438242\\
1.764	0.809757240376433\\
1.804	0.83098078329175\\
1.844	0.846669572665373\\
1.884	0.857256116718848\\
1.924	0.862924019262494\\
1.964	0.863624818210068\\
2.004	0.859180366330232\\
2.044	0.850099014124832\\
2.084	0.840205640478546\\
2.124	0.833014149783458\\
2.164	0.828527621748673\\
2.204	0.825697733663512\\
2.244	0.823804383662751\\
2.284	0.822459672160124\\
2.324	0.821455787415719\\
2.364	0.820675918735729\\
2.404	0.820050634010576\\
2.444	0.819536466081427\\
2.484	0.819104939825729\\
2.524	0.818736658752245\\
2.564	0.818417960064876\\
2.604	0.818138937641712\\
2.644	0.817892226995295\\
2.684	0.817672232307152\\
2.724	0.817474619449016\\
2.764	0.817295974328208\\
2.804	0.817133567016427\\
2.844	0.816985185347444\\
2.884	0.816849015214843\\
2.924	0.816723552933496\\
2.964	0.816607540041267\\
3.004	0.816499913827032\\
3.044	0.816399548966137\\
3.084	0.816304339535948\\
3.124	0.816211525451283\\
3.164	0.816118285962955\\
3.204	0.8160218377234\\
3.244	0.815919339314827\\
3.284	0.815807730235449\\
3.324	0.815683541994695\\
3.364	0.815542681283025\\
3.404	0.815380176564671\\
3.444	0.815189865685036\\
3.484	0.814963991821655\\
3.524	0.814692690168611\\
3.564	0.814363336966761\\
3.604	0.813959732260011\\
3.644	0.813461001963129\\
3.684	0.812840157453099\\
3.724	0.812062205479272\\
3.764	0.811081829200991\\
3.804	0.80984045510487\\
3.844	0.808262820425695\\
3.884	0.806253078801876\\
3.924	0.803690824838335\\
3.964	0.800428519483642\\
4.004	0.796291498429398\\
4.044	0.791083200510263\\
4.08399999999999	0.784597044272474\\
4.12399999999999	0.776635918693139\\
4.16399999999998	0.767037363942866\\
4.20399999999998	0.755698967275652\\
4.24399999999998	0.742592306042796\\
4.28399999999997	0.727754769108286\\
4.32399999999997	0.711261056476797\\
4.36399999999996	0.693191150330637\\
4.40399999999996	0.673600779628924\\
4.44399999999995	0.652495536128118\\
4.48399999999995	0.62982498998426\\
4.52399999999994	0.605491339308976\\
4.56399999999994	0.579373985637676\\
4.60399999999994	0.551374749017379\\
4.64399999999993	0.521457581316497\\
4.68399999999993	0.489681939286349\\
4.72399999999992	0.456211325017073\\
4.76399999999992	0.421317629055323\\
4.80399999999991	0.385378930275\\
4.84399999999991	0.348877380901575\\
4.88399999999991	0.312373896466669\\
4.9239999999999	0.276461503423148\\
4.9639999999999	0.241731531024964\\
5.00399999999989	0.208721495579518\\
5.04399999999989	0.17788492121512\\
5.08399999999988	0.149580789934791\\
5.12399999999988	0.124075537774179\\
5.16399999999987	0.101552979928439\\
5.20399999999987	0.0821257539721296\\
5.24399999999987	0.0658256821066115\\
5.28399999999986	0.0525719652165379\\
5.32399999999986	0.0421440744547864\\
5.36399999999985	0.0341700286898389\\
5.40399999999985	0.0281471902200303\\
5.44399999999984	0.0235191993289192\\
5.48399999999984	0.0198007495778968\\
5.52399999999983	0.0166845967561065\\
5.56399999999983	0.0140679543117604\\
5.60399999999983	0.0119901636294746\\
5.64399999999982	0.0105287759736152\\
5.68399999999982	0.00970258919930142\\
5.72399999999981	0.0094088505886455\\
5.76399999999981	0.00942025122160025\\
5.8039999999998	0.00946303927486657\\
5.8439999999998	0.00935466250679183\\
5.8839999999998	0.00911933795042832\\
5.92399999999979	0.00900769247600341\\
5.96399999999979	0.00942805874682446\\
6.00399999999978	0.0108532913883118\\
6.04399999999978	0.0137492181409285\\
6.08399999999977	0.0185221074935293\\
6.12399999999977	0.0254610394339032\\
6.16399999999976	0.0346784072898476\\
6.20399999999976	0.0461099556094224\\
6.24399999999976	0.059644059031085\\
6.28399999999975	0.0753672066071715\\
6.32399999999975	0.0938178798319309\\
6.36399999999974	0.116069162420401\\
6.40399999999974	0.143423583087218\\
6.44399999999973	0.176680589786822\\
6.48399999999973	0.215457255273521\\
6.52399999999972	0.258247864722048\\
6.56399999999972	0.303075264378519\\
6.60399999999972	0.347951771290843\\
6.64399999999971	0.390956074667243\\
6.68399999999971	0.430303491023391\\
6.7239999999997	0.464599738373229\\
6.7639999999997	0.493134115777224\\
6.80399999999969	0.51598035329663\\
6.84399999999969	0.533811538989404\\
6.88399999999969	0.547561551372447\\
6.92399999999968	0.558139747195863\\
6.96399999999968	0.566292597100078\\
7.00399999999967	0.572582884288275\\
7.04399999999967	0.577420232675614\\
7.08399999999966	0.581105265201747\\
7.12399999999966	0.583868497846887\\
7.16399999999965	0.585895797186066\\
7.20399999999965	0.587342484918754\\
7.24399999999965	0.588339314418087\\
7.28399999999964	0.588995161265515\\
7.32399999999964	0.58939852005224\\
7.36399999999963	0.589619071609626\\
7.40399999999963	0.589710027221724\\
7.44399999999962	0.589710939453238\\
7.48399999999962	0.58965054510464\\
7.52399999999961	0.589549332116663\\
7.56399999999961	0.589421690449669\\
7.60399999999961	0.589277611560089\\
7.6439999999996	0.589123966855166\\
7.6839999999996	0.588965440518048\\
7.72399999999959	0.588805194087751\\
7.76399999999959	0.588645335328549\\
7.80399999999958	0.588487253233475\\
7.84399999999958	0.588331846617911\\
7.88399999999957	0.588179676740253\\
7.92399999999957	0.588031076456024\\
7.96399999999957	0.587886224150011\\
8.00399999999956	0.587745196200793\\
8.04399999999956	0.587608004862204\\
8.08399999999955	0.587474619590747\\
8.12399999999955	0.587344980864169\\
8.16399999999954	0.587219011737995\\
8.20399999999954	0.587096623444077\\
8.24399999999954	0.586977719652984\\
8.28399999999953	0.586862199962954\\
8.32399999999953	0.586749963731796\\
8.36399999999952	0.586640910982968\\
};
\addlegendentry{$y_c$};

\addplot [color=black,dash pattern=on 1pt off 3pt on 3pt off 3pt,line width=2.0pt]
  table[row sep=crcr]{%
0.004	0\\
0.044	-0.00313272655695536\\
0.084	-0.00867844462227884\\
0.124	-0.0140079047712132\\
0.164	-0.018091443462256\\
0.204	-0.0206214541016379\\
0.244	-0.0215460958014989\\
0.284	-0.0209199058929612\\
0.324	-0.0189288412516289\\
0.364	-0.015949225187372\\
0.404	-0.0125364827854573\\
0.444	-0.00928587187324214\\
0.484	-0.00660490116139399\\
0.524	-0.00455275981396711\\
0.564	-0.00287791149070371\\
0.604	-0.00121856217740639\\
0.644	0.000675507091553434\\
0.684	0.00281821873137434\\
0.724000000000001	0.00501782179761879\\
0.764000000000001	0.00705430598784005\\
0.804000000000001	0.0089009669828089\\
0.844000000000001	0.0108496495199843\\
0.884000000000001	0.0135027071825912\\
0.924000000000001	0.0177145101636838\\
0.964000000000001	0.0245442194882406\\
1.004	0.0351584542788934\\
1.044	0.0505651174397519\\
1.084	0.0712175715250734\\
1.124	0.0967873452975105\\
1.164	0.126377917375725\\
1.204	0.159107331646396\\
1.244	0.19473414267802\\
1.284	0.233975796668523\\
1.324	0.278229120554815\\
1.364	0.328655313658648\\
1.404	0.385146443575777\\
1.444	0.445966207036838\\
1.484	0.508295240820383\\
1.524	0.569173884000304\\
1.564	0.626243935671721\\
1.604	0.678041809117235\\
1.644	0.723895161403655\\
1.684	0.763610171346251\\
1.724	0.797153637776484\\
1.764	0.824432001191227\\
1.804	0.845132457709113\\
1.844	0.858576312498352\\
1.884	0.863666267137082\\
1.924	0.859141864970853\\
1.964	0.84423425664056\\
2.004	0.819358542274757\\
2.044	0.78622945679985\\
2.084	0.747218369146149\\
2.124	0.704371483894412\\
2.164	0.658650613968891\\
2.204	0.609793434432555\\
2.244	0.556952305088282\\
2.284	0.49981192190462\\
2.324	0.439397521503897\\
2.364	0.377985881712436\\
2.404	0.31832900396078\\
2.444	0.262823213018213\\
2.484	0.213042663253684\\
2.524	0.169700804580705\\
2.564	0.132896193454363\\
2.604	0.102417827002018\\
2.644	0.0779235479195955\\
2.684	0.0589534648286063\\
2.724	0.0448635203897635\\
2.764	0.0347808226458393\\
2.804	0.0276544923332249\\
2.844	0.0224377518846748\\
2.884	0.0183416579056262\\
2.924	0.0149935215607505\\
2.964	0.0123777371182891\\
3.004	0.0106128542915693\\
3.044	0.00971674076270573\\
3.084	0.00948222516471695\\
3.124	0.00952432539596022\\
3.164	0.00948221684459078\\
3.204	0.00924552470914172\\
3.244	0.00903862266841597\\
3.284	0.00932813881865353\\
3.324	0.0106622106472434\\
3.364	0.0135484125036182\\
3.404	0.0183889873633274\\
3.444	0.0254366848064866\\
3.484	0.0347598850355408\\
3.524	0.0462675774782377\\
3.564	0.0598479899864983\\
3.604	0.075604999756669\\
3.644	0.0941007610994917\\
3.684	0.116443112162265\\
3.724	0.144012012790866\\
3.764	0.177774432816555\\
3.804	0.217554868477214\\
3.844	0.261826820226545\\
3.884	0.308163713612062\\
3.924	0.353960674636309\\
3.964	0.397003318091964\\
4.004	0.435737176789083\\
4.044	0.469300324575242\\
4.08399999999999	0.497429417593016\\
4.12399999999999	0.520316345183389\\
4.16399999999998	0.538455166955703\\
4.20399999999998	0.552501266272817\\
4.24399999999998	0.563157691454206\\
4.28399999999997	0.571096227647175\\
4.32399999999997	0.576912126204277\\
4.36399999999996	0.581105440285701\\
4.40399999999996	0.584080070596744\\
4.44399999999995	0.586152698548648\\
4.48399999999995	0.587565979059748\\
4.52399999999994	0.588502507589093\\
4.56399999999994	0.589097701949801\\
4.60399999999994	0.589450803534764\\
4.64399999999993	0.589633821927855\\
4.68399999999993	0.58969856458187\\
4.72399999999992	0.589682026713169\\
4.76399999999992	0.589610447505546\\
4.80399999999991	0.589502318627651\\
4.84399999999991	0.58937058997529\\
4.88399999999991	0.589224271641686\\
4.9239999999999	0.589069588266586\\
4.9639999999999	0.588910805289955\\
5.00399999999989	0.588750816931741\\
5.04399999999989	0.588591562448516\\
5.08399999999988	0.588434319428065\\
5.12399999999988	0.58827990952738\\
5.16399999999987	0.588128842171085\\
5.20399999999987	0.587981414485425\\
5.24399999999987	0.587837780485823\\
5.28399999999986	0.587697998746986\\
5.32399999999986	0.587562065070179\\
5.36399999999985	0.587429934727532\\
5.40399999999985	0.587301537490166\\
5.44399999999984	0.587176787676088\\
5.48399999999984	0.587055590769329\\
5.52399999999983	0.586937847680388\\
5.56399999999983	0.586823457379866\\
5.60399999999983	0.58671231839948\\
5.64399999999982	0.586604329527232\\
5.68399999999982	0.586499389904842\\
5.72399999999981	0.58639739865048\\
5.76399999999981	0.586298254067992\\
5.8039999999998	0.586201852457742\\
5.8439999999998	0.586108086509405\\
5.8839999999998	0.586016843230676\\
5.92399999999979	0.585928001346959\\
5.96399999999979	0.585841428096486\\
6.00399999999978	0.585756975345647\\
6.04399999999978	0.585674474965833\\
6.08399999999977	0.585593733454047\\
6.12399999999977	0.585514525856706\\
6.16399999999976	0.585436589185521\\
6.20399999999976	0.585359615715103\\
6.24399999999976	0.585283246843323\\
6.28399999999975	0.585207068589478\\
6.32399999999975	0.585130610292267\\
6.36399999999974	0.58505334859578\\
6.40399999999974	0.58497471924966\\
6.44399999999973	0.584894139373195\\
6.48399999999973	0.584811042320703\\
6.52399999999972	0.584724925789564\\
6.56399999999972	0.584635411129029\\
6.60399999999972	0.58454230813178\\
6.64399999999971	0.584445675738449\\
6.68399999999971	0.584345866477008\\
6.7239999999997	0.584243542659453\\
6.7639999999997	0.584139656253521\\
6.80399999999969	0.58403539139365\\
6.84399999999969	0.583932076643043\\
6.88399999999969	0.583831080641208\\
6.92399999999968	0.583733707564214\\
6.96399999999968	0.583641107342456\\
7.00399999999967	0.583554210875005\\
7.04399999999967	0.583473694463474\\
7.08399999999966	0.583399972199393\\
7.12399999999966	0.583333211221083\\
7.16399999999965	0.583273362927566\\
7.20399999999965	0.583220203106023\\
7.24399999999965	0.58317337492077\\
7.28399999999964	0.583132430236163\\
7.32399999999964	0.583096866348323\\
7.36399999999963	0.5830661566021\\
7.40399999999963	0.583039774445797\\
7.44399999999962	0.583017211209402\\
7.48399999999962	0.582997988325663\\
7.52399999999961	0.582981664915265\\
7.56399999999961	0.58296784169779\\
7.60399999999961	0.582956162129656\\
7.6439999999996	0.5829463115556\\
7.6839999999996	0.582938015024331\\
7.72399999999959	0.5829310342829\\
7.76399999999959	0.582925164340255\\
7.80399999999958	0.582920229884124\\
7.84399999999958	0.582916081748512\\
7.88399999999957	0.582912593560856\\
7.92399999999957	0.582909658646104\\
7.96399999999957	0.582907187227148\\
8.00399999999956	0.582905103934294\\
8.04399999999956	0.582903345618514\\
8.08399999999955	0.582901859451768\\
8.12399999999955	0.582900601291172\\
8.16399999999954	0.582899534280566\\
8.20399999999954	0.582898627662306\\
8.24399999999954	0.582897855772765\\
8.28399999999953	0.582897197196757\\
8.32399999999953	0.582896634058223\\
8.36399999999952	0.582896151426968\\
};
\addlegendentry{$y_u$};

\end{axis}

\begin{axis}[%
width=\figurewidth,
height=\figureheight,
scale only axis,
xmin=0,
xmax=8,
xlabel={Time [s]},
ymin=-20,
ymax=20,
ylabel absolute,
ylabel={$\text{Acceleration [m/s}^\text{2}\text{]}$},
at=(plot1.below south west),
anchor=above north west,
legend style={draw=black,fill=white,legend cell align=left}
]
\addplot [color=blue,solid,line width=1.5pt]
  table[row sep=crcr]{%
0.004	0\\
0.044	-3.29219257418834\\
0.084	-0.339241062817682\\
0.124	0.793190095788521\\
0.164	1.16250288694352\\
0.204	1.10827384539289\\
0.244	1.1220276516137\\
0.284	1.03931141459611\\
0.324	0.773032040210077\\
0.364	0.43535176132479\\
0.404	0.019352909179075\\
0.444	-0.352422978282893\\
0.484	-0.522680803356865\\
0.524	-0.402140402935841\\
0.564	-0.136953396789317\\
0.604	0.130710794482936\\
0.644	0.24315021869687\\
0.684	0.180838037367052\\
0.724000000000001	-0.0625047836616467\\
0.764000000000001	-0.209201006270452\\
0.804000000000001	-0.106313855320988\\
0.844000000000001	0.271397200994723\\
0.884000000000001	0.752822108216127\\
0.924000000000001	1.40780031360311\\
0.964000000000001	2.21457924572806\\
1.004	3.05010150051141\\
1.044	3.40449519353419\\
1.084	3.28925790510263\\
1.124	2.61797862430263\\
1.164	1.84439330114262\\
1.204	1.49131584958858\\
1.244	1.69504657283283\\
1.284	2.40994182226492\\
1.324	3.4772992384938\\
1.364	3.9925263385074\\
1.404	3.50467584798831\\
1.444	2.08912805235478\\
1.484	0.383586414144981\\
1.524	-1.25996960305722\\
1.564	-2.72042231655742\\
1.604	-3.83206418211717\\
1.644	-4.43435027429392\\
1.684	-4.5830126422034\\
1.724	-4.39276660499343\\
1.764	-4.00995822988521\\
1.804	-3.66520314774916\\
1.844	-3.30574996416438\\
1.884	-3.10769874399792\\
1.924	-3.10186173833946\\
1.964	-3.19821755223557\\
2.004	-3.35993049991521\\
2.044	-6.61718756094303\\
2.084	-4.56187794802485\\
2.124	-3.19829111800346\\
2.164	-3.18637730924054\\
2.204	-3.36604418435544\\
2.244	-3.63484362593311\\
2.284	-3.76363884132088\\
2.324	-3.86266057747742\\
2.364	-4.01136969056145\\
2.404	-4.18213133854816\\
2.444	-4.36175101076941\\
2.484	-4.5301447753326\\
2.524	-4.62925891894498\\
2.564	-4.79618790791703\\
2.604	-4.98333563706841\\
2.644	-5.10116570689995\\
2.684	-5.29010476674795\\
2.724	-5.40395971533064\\
2.764	-5.60174094717589\\
2.804	-5.73677197097909\\
2.844	-5.91663021037132\\
2.884	-6.00842159615803\\
2.924	-6.22251695139491\\
2.964	-6.31468297396222\\
3.004	-6.50203644153859\\
3.044	-4.55699586111632\\
3.084	-2.6917367625998\\
3.124	-1.41200695268916\\
3.164	-0.580842923484326\\
3.204	0.0235244967900634\\
3.244	0.383230170765539\\
3.284	0.55938717390245\\
3.324	0.690441320360178\\
3.364	0.739181684873644\\
3.404	0.748164122342982\\
3.444	0.717141198881362\\
3.484	0.663338615604376\\
3.524	0.593509780431352\\
3.564	0.527828141489233\\
3.604	0.470857589421543\\
3.644	0.362383747949766\\
3.684	0.32130376510922\\
3.724	0.201314049725922\\
3.764	0.204326873981414\\
3.804	0.0745411242879177\\
3.844	-0.0286419773425833\\
3.884	-0.0622179015734772\\
3.924	-0.20084950908644\\
3.964	-0.372695631998273\\
4.004	-0.480130560001684\\
4.044	-0.631663210168325\\
4.08399999999999	-0.778996175235718\\
4.12399999999999	-0.906052814106616\\
4.16399999999998	-1.03196538369922\\
4.20399999999998	-1.05559549800315\\
4.24399999999998	-1.04331044719939\\
4.28399999999997	-1.0518620222541\\
4.32399999999997	-0.996361967323993\\
4.36399999999996	-0.960132850440561\\
4.40399999999996	-0.946549244478326\\
4.44399999999995	-0.997548130554762\\
4.48399999999995	-1.03326396923461\\
4.52399999999994	-1.08340871698097\\
4.56399999999994	-1.22019280769024\\
4.60399999999994	-1.23175329855321\\
4.64399999999993	-1.25886756701869\\
4.68399999999993	-1.16013785719676\\
4.72399999999992	-0.931934276487687\\
4.76399999999992	-0.74185428484316\\
4.80399999999991	-0.38089696726806\\
4.84399999999991	-0.0438209510352884\\
4.88399999999991	0.34018872180357\\
4.9239999999999	0.708901838790263\\
4.9639999999999	1.04864477047569\\
5.00399999999989	1.38157343497686\\
5.04399999999989	1.59319091268588\\
5.08399999999988	1.80773044797949\\
5.12399999999988	1.83887833489195\\
5.16399999999987	1.9711107887903\\
5.20399999999987	1.95060201462196\\
5.24399999999987	1.93590782479824\\
5.28399999999986	1.82615499820666\\
5.32399999999986	1.61137437383672\\
5.36399999999985	1.31441293480934\\
5.40399999999985	0.926067314699253\\
5.44399999999984	0.53007770594678\\
5.48399999999984	0.334518292667202\\
5.52399999999983	0.247038028642956\\
5.56399999999983	0.212572857303438\\
5.60399999999983	0.327649447027658\\
5.64399999999982	0.366372969325337\\
5.68399999999982	0.453472427497998\\
5.72399999999981	0.253494570493083\\
5.76399999999981	0.0728590456956464\\
5.8039999999998	-0.143513015951338\\
5.8439999999998	-0.179385979308134\\
5.8839999999998	-0.026434590464287\\
5.92399999999979	0.203412561348431\\
5.96399999999979	0.572003684865777\\
6.00399999999978	0.917537546297228\\
6.04399999999978	1.14995635070504\\
6.08399999999977	1.43204084620388\\
6.12399999999977	1.48863076427438\\
6.16399999999976	1.49587950860304\\
6.20399999999976	1.38290606668128\\
6.24399999999976	1.34288821671675\\
6.28399999999975	1.55939244152412\\
6.32399999999975	2.08774655914339\\
6.36399999999974	2.98990507544396\\
6.40399999999974	3.78568723290533\\
6.44399999999973	3.97523290807974\\
6.48399999999973	3.1811530811732\\
6.52399999999972	1.79620572890944\\
6.56399999999972	0.303146616325129\\
6.60399999999972	-1.07940187685426\\
6.64399999999971	-2.30285921245101\\
6.68399999999971	-3.28135699105255\\
6.7239999999997	-3.84058679722921\\
6.7639999999997	-3.87991590997341\\
6.80399999999969	-3.48986366627019\\
6.84399999999969	-2.84836035901151\\
6.88399999999969	-2.21654829024325\\
6.92399999999968	-1.63500597009393\\
6.96399999999968	-1.25387289995012\\
7.00399999999967	-0.875165755829886\\
7.04399999999967	-0.651121682086963\\
7.08399999999966	-0.499557777919571\\
7.12399999999966	-0.379520502955069\\
7.16399999999965	-0.296840398328532\\
7.20399999999965	-0.241846381324202\\
7.24399999999965	-0.135561989661881\\
7.28399999999964	-0.0954458384083245\\
7.32399999999964	-0.0625548273334691\\
7.36399999999963	-0.0354048295026325\\
7.40399999999963	-0.0274645568606415\\
7.44399999999962	0.0235871248022514\\
7.48399999999962	-0.0248237054483876\\
7.52399999999961	-0.0291919173444513\\
7.56399999999961	0.0529301740609702\\
7.60399999999961	0.0149516629550022\\
7.6439999999996	0.00608769061090144\\
7.6839999999996	0.0392496966431356\\
7.72399999999959	-0.0286519065760162\\
7.76399999999959	-0.0171186983819018\\
7.80399999999958	-0.000473872911753834\\
7.84399999999958	0.0110963713744038\\
7.88399999999957	-0.0116345419252449\\
7.92399999999957	0.0497177243434817\\
7.96399999999957	-0.00994800914241519\\
8.00399999999956	-0.028806231192631\\
8.04399999999956	0.0286701241588729\\
8.08399999999955	-0.0148910639456286\\
8.12399999999955	-0.0158707876556207\\
8.16399999999954	-0.0227809825218335\\
8.20399999999954	0.0708363505039247\\
8.24399999999954	-0.0173880213379449\\
8.28399999999953	-0.018100140454299\\
8.32399999999953	0.0141096922464102\\
8.36399999999952	0.0324927277130193\\
};
\addlegendentry{$\ddot{y}_{r}$};

\addplot [color=red,dashed,line width=2.0pt]
  table[row sep=crcr]{%
0.004	0\\
0.044	-1.57437280489045\\
0.084	0.128163200057461\\
0.124	0.779774008397672\\
0.164	0.968961785904813\\
0.204	1.00482466914398\\
0.244	0.981756854310344\\
0.284	0.879868385588202\\
0.324	0.657041645549083\\
0.364	0.31346208517137\\
0.404	-0.0712566649471207\\
0.444	-0.35346260479738\\
0.484	-0.416865854506512\\
0.524	-0.268233125011895\\
0.564	-0.0305807644813577\\
0.604	0.147551804926067\\
0.644	0.174611657304748\\
0.684	0.0568841454679858\\
0.724000000000001	-0.0988468799792545\\
0.764000000000001	-0.143794170661615\\
0.804000000000001	0.0138382474672118\\
0.844000000000001	0.372096923457199\\
0.884000000000001	0.890713812571873\\
0.924000000000001	1.54251246836075\\
0.964000000000001	2.2773365938375\\
1.004	2.92480832697614\\
1.044	3.19980991244301\\
1.084	2.94588522111097\\
1.124	2.35460775818995\\
1.164	1.81598276929412\\
1.204	1.64076813146842\\
1.244	1.98265065251613\\
1.284	2.76597746626507\\
1.324	3.56355529562542\\
1.364	3.74763945244969\\
1.404	3.01691591555881\\
1.444	1.6532852090403\\
1.484	0.0917481662251384\\
1.524	-1.42268915614832\\
1.564	-2.73072139879109\\
1.604	-3.65895542763446\\
1.644	-4.11653404059202\\
1.684	-4.16455073712029\\
1.724	-3.94417498713878\\
1.764	-3.60594619587301\\
1.804	-3.28234937457219\\
1.844	-3.06588104767158\\
1.884	-2.99852362570142\\
1.924	-3.0654394475585\\
1.964	-3.20336139916605\\
2.004	-3.3212371826804\\
2.044	-1.2914229448083\\
2.084	0.948507959510521\\
2.124	1.11540512823047\\
2.164	0.692599146431349\\
2.204	0.396038343566046\\
2.244	0.234280981627198\\
2.284	0.146566735835155\\
2.324	0.0967961362393253\\
2.364	0.0670107373119468\\
2.404	0.0482738614665742\\
2.444	0.0359573547848251\\
2.484	0.0275470924767814\\
2.524	0.0216124206203225\\
2.564	0.0173039295617203\\
2.604	0.0140977933074068\\
2.644	0.0116598927241605\\
2.684	0.00977063520205245\\
2.724	0.00828180053026289\\
2.764	0.00709093231554443\\
2.804	0.00612567806441228\\
2.844	0.00533394971940902\\
2.884	0.00467758769880291\\
2.924	0.00412818905754156\\
2.964	0.00366430430252021\\
3.004	0.00325847718528098\\
3.044	0.00236490476472868\\
3.084	0.00115531711309011\\
3.124	-8.03698152925373e-05\\
3.164	-0.00129833146056816\\
3.204	-0.00253957073728916\\
3.244	-0.00387406644155774\\
3.284	-0.00538351300872639\\
3.324	-0.00716337429837332\\
3.364	-0.00932340067219002\\
3.404	-0.0119996277782916\\
3.444	-0.0153693128480132\\
3.484	-0.0196569573154348\\
3.524	-0.0251516743983422\\
3.564	-0.0322117149923661\\
3.604	-0.041324720732504\\
3.644	-0.0531255617984399\\
3.684	-0.0684857483829097\\
3.724	-0.088437404518003\\
3.764	-0.114371619317606\\
3.804	-0.147888711742423\\
3.844	-0.190889524023044\\
3.884	-0.245498113900281\\
3.924	-0.313239439259852\\
3.964	-0.394704272067845\\
4.004	-0.488217152901593\\
4.044	-0.588899803799628\\
4.08399999999999	-0.687943395770214\\
4.12399999999999	-0.773276463429356\\
4.16399999999998	-0.831637740750605\\
4.20399999999998	-0.85437075545723\\
4.24399999999998	-0.843458141205964\\
4.28399999999997	-0.812787889136028\\
4.32399999999997	-0.778576265620794\\
4.36399999999996	-0.757762809842208\\
4.40399999999996	-0.76586051911841\\
4.44399999999995	-0.806589914931847\\
4.48399999999995	-0.876190293675969\\
4.52399999999994	-0.961933415920906\\
4.56399999999994	-1.03302727467588\\
4.60399999999994	-1.0669596972733\\
4.64399999999993	-1.03501759250046\\
4.68399999999993	-0.932023073235301\\
4.72399999999992	-0.753355623730734\\
4.76399999999992	-0.503009376537406\\
4.80399999999991	-0.188257863192918\\
4.84399999999991	0.175327017090453\\
4.88399999999991	0.553014264522586\\
4.9239999999999	0.924257521550024\\
4.9639999999999	1.25448077588833\\
5.00399999999989	1.52483194174032\\
5.04399999999989	1.7297402929671\\
5.08399999999988	1.87413004393044\\
5.12399999999988	1.96878003218255\\
5.16399999999987	2.02175891723106\\
5.20399999999987	2.02967376593283\\
5.24399999999987	1.97212327522032\\
5.28399999999986	1.83057043916272\\
5.32399999999986	1.59446816713972\\
5.36399999999985	1.27283225627715\\
5.40399999999985	0.910746594170551\\
5.44399999999984	0.588586612253154\\
5.48399999999984	0.379631304024814\\
5.52399999999983	0.306353349340571\\
5.56399999999983	0.330974412739364\\
5.60399999999983	0.386099050540141\\
5.64399999999982	0.407608970928966\\
5.68399999999982	0.351830221135388\\
5.72399999999981	0.211474230107631\\
5.76399999999981	0.0315271794760908\\
5.8039999999998	-0.099158651382164\\
5.8439999999998	-0.0998522629247263\\
5.8839999999998	0.0479118355969244\\
5.92399999999979	0.300905853650153\\
5.96399999999979	0.597245858690055\\
6.00399999999978	0.890187843514116\\
6.04399999999978	1.14830186775736\\
6.08399999999977	1.33795455959362\\
6.12399999999977	1.41947088208528\\
6.16399999999976	1.38332432430963\\
6.20399999999976	1.30116577012175\\
6.24399999999976	1.32473630634852\\
6.28399999999975	1.62399450799305\\
6.32399999999975	2.28068287579071\\
6.36399999999974	3.1442737638699\\
6.40399999999974	3.76135788268566\\
6.44399999999973	3.63520971807385\\
6.48399999999973	2.72092311723977\\
6.52399999999972	1.42616209860704\\
6.56399999999972	0.0961165585959717\\
6.60399999999972	-1.16777460564761\\
6.64399999999971	-2.2987371294059\\
6.68399999999971	-3.14656963222683\\
6.7239999999997	-3.5542094379058\\
6.7639999999997	-3.48389657531181\\
6.80399999999969	-3.06368242008867\\
6.84399999999969	-2.50150841661711\\
6.88399999999969	-1.96216426360184\\
6.92399999999968	-1.52075718157191\\
6.96399999999968	-1.18524063075704\\
7.00399999999967	-0.936426284351783\\
7.04399999999967	-0.749072672013174\\
7.08399999999966	-0.601641228939159\\
7.12399999999966	-0.480522983784885\\
7.16399999999965	-0.378362417642156\\
7.20399999999965	-0.292258799908116\\
7.24399999999965	-0.220767644545211\\
7.28399999999964	-0.162976249764292\\
7.32399999999964	-0.117728117100713\\
7.36399999999963	-0.0833394575694871\\
7.40399999999963	-0.0578744438867819\\
7.44399999999962	-0.0394196984260206\\
7.48399999999962	-0.026280956302471\\
7.52399999999961	-0.0170563508674129\\
7.56399999999961	-0.0106485874640654\\
7.60399999999961	-0.00623767751615298\\
7.6439999999996	-0.0032265330786622\\
7.6839999999996	-0.00119031729907167\\
7.72399999999959	0.000169734065922144\\
7.76399999999959	0.00106750825818804\\
7.80399999999958	0.00164964878050805\\
7.84399999999958	0.00201397036231597\\
7.88399999999957	0.00223084937162768\\
7.92399999999957	0.00234833744041859\\
7.96399999999957	0.00239917980561743\\
8.00399999999956	0.00240801116556787\\
8.04399999999956	0.00238914020722046\\
8.08399999999955	0.00235143942136285\\
8.12399999999955	0.00230261011046474\\
8.16399999999954	0.00224639832989597\\
8.20399999999954	0.00218581425334412\\
8.24399999999954	0.00212266968628262\\
8.28399999999953	0.00205942297786866\\
8.32399999999953	0.00199653474212911\\
8.36399999999952	0.00193381130473087\\
};
\addlegendentry{$\ddot{y}_c$};

\addplot [color=black,dash pattern=on 1pt off 3pt on 3pt off 3pt,line width=2.0pt]
  table[row sep=crcr]{%
0.004	0\\
0.044	-1.59954299014054\\
0.084	0.112244330600025\\
0.124	0.779415907514183\\
0.164	0.974340781299854\\
0.204	1.00789613448221\\
0.244	0.981889560654181\\
0.284	0.879638712474928\\
0.324	0.657613876951821\\
0.364	0.312544034052864\\
0.404	-0.0758329244056085\\
0.444	-0.358629876102407\\
0.484	-0.417524254956118\\
0.524	-0.264206835738674\\
0.564	-0.0257165361475424\\
0.604	0.150091393858293\\
0.644	0.173756653586864\\
0.684	0.053778451992183\\
0.724000000000001	-0.10124004023411\\
0.764000000000001	-0.142621185086257\\
0.804000000000001	0.0195788496590049\\
0.844000000000001	0.38188929410743\\
0.884000000000001	0.903616576576362\\
0.924000000000001	1.55773651233641\\
0.964000000000001	2.29808452337993\\
1.004	2.97130090659434\\
1.044	3.31527592263806\\
1.084	3.14885627773208\\
1.124	2.57879902609129\\
1.164	1.97236638978646\\
1.204	1.74674893040044\\
1.244	2.1456580863959\\
1.284	3.0444733206634\\
1.324	3.87834680857241\\
1.364	3.93610841509374\\
1.404	2.9168773676545\\
1.444	1.14358753926628\\
1.484	-0.764403753417956\\
1.524	-2.30597443251475\\
1.564	-3.27206161988472\\
1.604	-3.71687799024603\\
1.644	-3.84059935886702\\
1.684	-3.8525528877986\\
1.724	-3.89478474842458\\
1.764	-4.06423935751359\\
1.804	-4.45687962283568\\
1.844	-5.12915890554929\\
1.884	-5.95601930841547\\
1.924	-6.53180027601709\\
1.964	-6.36993936717497\\
2.004	-5.33484858491606\\
2.044	-3.81215973869931\\
2.084	-2.44277722246786\\
2.124	-1.74955873361838\\
2.164	-1.87761372606672\\
2.204	-2.4633627045808\\
2.244	-2.77458234492599\\
2.284	-2.22377642961008\\
2.324	-0.81810751367034\\
2.364	0.944986001325874\\
2.404	2.50945045890828\\
2.444	3.55271066632738\\
2.484	4.03406092180096\\
2.524	4.10749549652899\\
2.564	3.97765797891656\\
2.604	3.76949623980906\\
2.644	3.49504351763457\\
2.684	3.10903057082453\\
2.724	2.57597537684592\\
2.764	1.91752823255133\\
2.804	1.24161514436171\\
2.844	0.715836816805452\\
2.884	0.458476204716666\\
2.924	0.443889617692125\\
2.964	0.530088973054349\\
3.004	0.560236435709329\\
3.044	0.443916317665101\\
3.084	0.199235153293816\\
3.124	-0.048229934474968\\
3.164	-0.142991794742527\\
3.204	-0.0175968810182347\\
3.244	0.272536434599432\\
3.284	0.620900420695983\\
3.324	0.946160560318816\\
3.364	1.20753590658553\\
3.404	1.37837125443067\\
3.444	1.43259368586402\\
3.484	1.37645822694343\\
3.524	1.2895048949391\\
3.564	1.3197337210566\\
3.604	1.62999041531844\\
3.644	2.30220663159971\\
3.684	3.19951464678653\\
3.724	3.89643692109493\\
3.764	3.88738929490973\\
3.804	2.98700635306512\\
3.844	1.46618110466839\\
3.884	-0.20140742222897\\
3.924	-1.63525048821201\\
3.964	-2.65261452398771\\
4.004	-3.22782464849213\\
4.044	-3.41853565217931\\
4.08399999999999	-3.31468084666273\\
4.12399999999999	-3.01313164538305\\
4.16399999999998	-2.6039818075494\\
4.20399999999998	-2.16071827720884\\
4.24399999999998	-1.73486070086359\\
4.28399999999997	-1.35627483383283\\
4.32399999999997	-1.03756002418565\\
4.36399999999996	-0.779758927640912\\
4.40399999999996	-0.577422640633427\\
4.44399999999995	-0.422268031144218\\
4.48399999999995	-0.305449104685459\\
4.52399999999994	-0.218773868783872\\
4.56399999999994	-0.155230113215969\\
4.60399999999994	-0.10910820459114\\
4.64399999999993	-0.0759164167060615\\
4.68399999999993	-0.0522083282671145\\
4.72399999999992	-0.0353889175410091\\
4.76399999999992	-0.0235326320042473\\
4.80399999999991	-0.015227247463255\\
4.84399999999991	-0.00944681845785937\\
4.88399999999991	-0.00545189279219722\\
4.9239999999999	-0.0027130791075056\\
4.9639999999999	-0.000853580154019766\\
5.00399999999989	0.000393402360879092\\
5.04399999999989	0.00121588659909386\\
5.08399999999988	0.00174578449573312\\
5.12399999999988	0.00207524660316089\\
5.16399999999987	0.00226837041189437\\
5.20399999999987	0.0023695419900939\\
5.24399999999987	0.00240935067914656\\
5.28399999999986	0.00240876482939938\\
5.32399999999986	0.00238206788462978\\
5.36399999999985	0.0023389145947109\\
5.40399999999985	0.00228576502053481\\
5.44399999999984	0.00222687987980816\\
5.48399999999984	0.00216500735019061\\
5.52399999999983	0.00210185313781284\\
5.56399999999983	0.0020383982802707\\
5.60399999999983	0.00197510972168548\\
5.64399999999982	0.00191207494202302\\
5.68399999999982	0.00184908223414411\\
5.72399999999981	0.0017856614549477\\
5.76399999999981	0.00172109543329775\\
5.8039999999998	0.00165440915847103\\
5.8439999999998	0.0015843420668739\\
5.8839999999998	0.0015093080257962\\
5.92399999999979	0.00142734796885444\\
5.96399999999979	0.00133608169853978\\
6.00399999999978	0.00123266840717009\\
6.04399999999978	0.00111379036147059\\
6.08399999999977	0.00097568137134586\\
6.12399999999977	0.000814231380579757\\
6.16399999999976	0.000625210464398114\\
6.20399999999976	0.000404668050803393\\
6.24399999999976	0.000149572112847882\\
6.28399999999975	-0.000141249881787784\\
6.32399999999975	-0.000465835029304582\\
6.36399999999974	-0.000817513468245631\\
6.40399999999974	-0.00118321517247105\\
6.44399999999973	-0.00154202797694196\\
6.48399999999973	-0.00186460715242412\\
6.52399999999972	-0.0021141933158696\\
6.56399999999972	-0.00224992007488204\\
6.60399999999972	-0.00223262652288668\\
6.64399999999971	-0.00203251138606636\\
6.68399999999971	-0.00163693118670242\\
6.7239999999997	-0.00105597835435195\\
6.7639999999997	-0.00032370298084723\\
6.80399999999969	0.000505914625030705\\
6.84399999999969	0.00136736350198578\\
6.88399999999969	0.00219405173660347\\
6.92399999999968	0.00292802861854526\\
6.96399999999968	0.00352681380396087\\
7.00399999999967	0.00396639009652421\\
7.04399999999967	0.00424062744071652\\
7.08399999999966	0.00435818059148931\\
7.12399999999966	0.0043381382345849\\
7.16399999999965	0.00420552984141242\\
7.20399999999965	0.00398743127308889\\
7.24399999999965	0.00371002906526718\\
7.28399999999964	0.00339670610411125\\
7.32399999999964	0.00306702750355998\\
7.36399999999963	0.0027364217136277\\
7.40399999999963	0.00241633827143159\\
7.44399999999962	0.00211468987863567\\
7.48399999999962	0.0018364294065429\\
7.52399999999961	0.00158415720284034\\
7.56399999999961	0.00135869295960074\\
7.60399999999961	0.00115957650639598\\
7.6439999999996	0.000985483151911338\\
7.6839999999996	0.000834552917464963\\
7.72399999999959	0.000704640937363636\\
7.76399999999959	0.00059350013546641\\
7.80399999999958	0.000498908439203837\\
7.84399999999958	0.000418752334455796\\
7.88399999999957	0.000351077256536209\\
7.92399999999957	0.000294113656343146\\
7.96399999999957	0.000246285882532864\\
8.00399999999956	0.000206209453611982\\
8.04399999999956	0.000172680937576856\\
8.08399999999955	0.000144663536039873\\
8.12399999999955	0.000121270575991518\\
8.16399999999954	0.000101748420861654\\
8.20399999999954	8.54597920047183e-05\\
8.24399999999954	7.1868109581084e-05\\
8.28399999999953	6.05231894001987e-05\\
8.32399999999953	5.1048443177436e-05\\
8.36399999999952	4.3129604157568e-05\\
};
\addlegendentry{$\ddot{y}_u$};

\end{axis}
\end{tikzpicture}%

%% file: chapters/paper2/figs/scen1.tex
%
%
\begin{tikzpicture}

\begin{axis}[%
width=\figurewidth,
height=\figureheight,
scale only axis,
scaled y ticks = false,
xmin=0,
xmax=25,
ymin=0,
ymax=0.15,
ylabel absolute,
ylabel={$\Vert y_c-y_a \Vert$ [m]},
name=plot2,
yticklabel style={/pgf/number format/fixed},
]
\addplot [color=blue,solid,line width=2.0pt,forget plot]
  table[row sep=crcr]{%
0.2	0.000201924263681773\\
0.4	0.000205289200636057\\
0.6	0.000205451638611946\\
0.8	0.00020910510275252\\
1	0.000205169621013415\\
1.2	0.000206829450569128\\
1.4	0.000208881127613434\\
1.6	0.000208878332118546\\
1.8	0.000210387460448896\\
2	0.000210754345901642\\
2.2	0.000213651981285948\\
2.4	0.000213046331765038\\
2.6	0.000216074843783327\\
2.8	0.000215122269761611\\
3	0.000215781573729068\\
3.2	0.000218689187718693\\
3.4	0.00021766548681128\\
3.6	0.000219541986589998\\
3.8	0.000220334021099837\\
4	0.000301207539604625\\
4.2	0.000198993017051539\\
4.4	0.000485459065700724\\
4.6	0.000601025092793945\\
4.8	0.000748293574184471\\
5	0.00118590845928814\\
5.2	0.00142913810529916\\
5.4	0.00130690189606812\\
5.6	0.00115687482304836\\
5.8	0.000555589438448961\\
6	0.000349617399174821\\
6.2	0.00735627273822242\\
6.4	0.0121996022182495\\
6.6	0.0148768001744771\\
6.8	0.0183256048518655\\
7	0.0269490671686216\\
7.2	0.0331811295966416\\
7.4	0.0394358742222389\\
7.6	0.0492778919256006\\
7.8	0.0543861375497324\\
8	0.0582751615268536\\
8.2	0.063419863479227\\
8.4	0.0780723073117661\\
8.6	0.0616514632221268\\
8.8	0.0342841066080814\\
9	0.0168340430205868\\
9.2	0.00781117328879743\\
9.4	0.00348078429513553\\
9.6	0.00156563258868508\\
9.8	0.0014381639594945\\
10	0.00179525768418808\\
10.2	0.0016340293674686\\
10.4	0.00125096215133951\\
10.6	0.000882528706310177\\
10.8	0.000584364132659376\\
11	0.000356589317664908\\
11.2	0.000182182856436602\\
11.4	0.000292466150264233\\
11.6	0.000515922711736408\\
11.8	0.000554935381537148\\
12	0.000671319307451199\\
12.2	0.000815594299036985\\
12.4	0.000831512153756882\\
12.6	0.00920699546992081\\
12.8	0.0241246964325781\\
13	0.045050422707716\\
13.2	0.0522643601524335\\
13.4	0.0409085951360471\\
13.6	0.0451110381076322\\
13.8	0.0671424069138221\\
14	0.0523754968030311\\
14.2	0.0338388322724313\\
14.4	0.0513567061104034\\
14.6	0.0752839559332199\\
14.8	0.0937671979544935\\
15	0.0891580910104786\\
15.2	0.0907243746062221\\
15.4	0.100372344641088\\
15.6	0.10577983713723\\
15.8	0.106584354570863\\
16	0.0862517586016189\\
16.2	0.0679570811018168\\
16.4	0.0829824855033686\\
16.6	0.0950670379956018\\
16.8	0.0994582344933773\\
17	0.0748677441040177\\
17.2	0.0460295301185005\\
17.4	0.0305699015363723\\
17.6	0.0162779431231964\\
17.8	0.0077788928183167\\
18	0.00360140253673097\\
18.2	0.00178903075370137\\
18.4	0.00143102788969387\\
18.6	0.00161027224673808\\
18.8	0.00135006102715242\\
19	0.00106802376003246\\
19.2	0.00086859101577528\\
19.4	0.000571328361076809\\
19.6	0.000447361286150497\\
19.8	0.000246533741838229\\
20	0.000249026971033873\\
20.2	0.000264729575890371\\
20.4	0.000271208676032973\\
20.6	0.000248942520774268\\
20.8	0.000234204976181672\\
21	0.00207340736958118\\
21.2	0.00533764337743387\\
21.4	0.0032425045291769\\
21.6	0.00144601050044856\\
21.8	0.000758527364164086\\
22	0.000551511348393007\\
22.2	0.000335813139295693\\
22.4	0.000193050495980139\\
22.6	0.000207649319113286\\
22.8	0.000174618398084297\\
23	0.000221113612101784\\
23.1999999999999	4.61037174295445e-05\\
23.3999999999999	0.000108055583092637\\
23.5999999999999	0.000104728908529431\\
23.7999999999999	0.000104866890871494\\
23.9999999999999	0.000105760863863979\\
};
\end{axis}

\begin{axis}[%
width=\figurewidth,
height=2\figureheight,
scale only axis,
xmin=0.00753680614383,
xmax=0.447789444596,
xlabel={$y$ [m]},
ymin=-0.199706399445,
ymax=-0.0235422904677,
ylabel absolute,
ylabel={$z$ [m]},
at=(plot2.above north west),
anchor=below south west,
legend style={draw=black,fill=white,legend cell align=left},
yticklabel style={/pgf/number format/fixed},
]
\addplot [color=blue,solid,line width=1pt]
  table[row sep=crcr]{%
0.447728444727	-0.0533229633798\\
0.447725873136	-0.0533243669732\\
0.447726248968	-0.0533256408499\\
0.44772268359	-0.0533251305364\\
0.447724770587	-0.0533224245921\\
0.447723112198	-0.0533230582869\\
0.4477216918	-0.0533237304655\\
0.447721507749	-0.0533236068717\\
0.447719786988	-0.0533223307397\\
0.447718709269	-0.0533221571517\\
0.447716773007	-0.0533236735565\\
0.447717579287	-0.0533246276995\\
0.447715253604	-0.0533263513485\\
0.447714784017	-0.0533234841779\\
0.44771386657	-0.0533228713568\\
0.447711566683	-0.0533256671978\\
0.447711653208	-0.0533218187563\\
0.447710048518	-0.0533223008898\\
0.44770968555	-0.0533256280189\\
0.447130936478	-0.0531383681401\\
0.440276475966	-0.0532407576105\\
0.427447522568	-0.0542224020736\\
0.410363348985	-0.0564497032399\\
0.39071594655	-0.0590803936664\\
0.370484801224	-0.0611340024051\\
0.351580643487	-0.0626274910118\\
0.334432814256	-0.0642211848062\\
0.3181992553	-0.0659735424409\\
0.302202587352	-0.0676731506139\\
0.285954339769	-0.070114964197\\
0.27674925753	-0.0739762132699\\
0.277562369052	-0.0762498851087\\
0.27828033693	-0.0782495055049\\
0.278791489923	-0.0814889465765\\
0.28005304754	-0.0907925480718\\
0.281047566321	-0.0977661311755\\
0.283126638896	-0.103859595794\\
0.286633365714	-0.113058213729\\
0.288331570299	-0.117403835615\\
0.289708375776	-0.120255050889\\
0.291856048164	-0.124001135527\\
0.297794775497	-0.138236760538\\
0.293552770617	-0.123280325871\\
0.282809844614	-0.100244515377\\
0.27490766579	-0.0861625225908\\
0.27041776901	-0.0791940748216\\
0.267905740245	-0.0759556118491\\
0.265977846781	-0.0745042545034\\
0.262100229799	-0.074605175237\\
0.254059671829	-0.0753501523331\\
0.242542066347	-0.0764978447659\\
0.22909995489	-0.0776396841743\\
0.214340878873	-0.0782991255993\\
0.198938488335	-0.0787738519091\\
0.183549152895	-0.0802050082146\\
0.168759656311	-0.0838134811716\\
0.154108601168	-0.0892232915585\\
0.139850044837	-0.0948606841402\\
0.12581992945	-0.0999853710156\\
0.111108025802	-0.104353559831\\
0.0960322515827	-0.108509424245\\
0.0818101082676	-0.113639296491\\
0.0748970180938	-0.112103771217\\
0.0791622899086	-0.0987960576695\\
0.0855579824468	-0.0777149225386\\
0.086653667226	-0.0705090583782\\
0.0822275167102	-0.0810009629113\\
0.073840621578	-0.0751071874693\\
0.0673687357664	-0.0533133835361\\
0.0681552795559	-0.0678460894056\\
0.069859348517	-0.0862479140856\\
0.07532887191	-0.0691137493489\\
0.0810538141006	-0.0456467689364\\
0.0919043385736	-0.0289648524846\\
0.102281407209	-0.0372429808019\\
0.112101877341	-0.040049298813\\
0.121256865397	-0.0339102672234\\
0.126636640812	-0.0308827333808\\
0.12365021588	-0.028321290903\\
0.11458968771	-0.0464078626437\\
0.105279389333	-0.0621117634349\\
0.103147091434	-0.044421800408\\
0.097445779041	-0.0300254745908\\
0.0886678873225	-0.0235422904677\\
0.0822153796711	-0.0466458922011\\
0.0789478321527	-0.075015216404\\
0.0770805489516	-0.0909630362893\\
0.0741256374132	-0.105131388121\\
0.0721067320306	-0.113260803283\\
0.0707248052895	-0.1172097342\\
0.0693272191049	-0.11928010113\\
0.0669070528138	-0.121342050507\\
0.0618854388639	-0.124744389185\\
0.0547942207278	-0.129732901645\\
0.0466771345926	-0.135691305704\\
0.0380363369644	-0.142268273861\\
0.029351578972	-0.149002347243\\
0.0218576167667	-0.15564192215\\
0.0161491666658	-0.161998645435\\
0.0127904957156	-0.168182097278\\
0.0108619421676	-0.173110704521\\
0.00963005611818	-0.176373725975\\
0.00881527736081	-0.178835565179\\
0.00824508327576	-0.181111323668\\
0.00820235934795	-0.18189619387\\
0.00889040454564	-0.181421650544\\
0.00835251880375	-0.183704901467\\
0.00776192395008	-0.185681439595\\
0.00759586022104	-0.188092184461\\
0.00766768495376	-0.191469884508\\
0.00789673640531	-0.195297855255\\
0.00825322233494	-0.198157295918\\
0.00854917085605	-0.199415229461\\
0.00873538817106	-0.199706399445\\
0.00903356915835	-0.199558905198\\
0.00895505258395	-0.199230713771\\
0.00887145831394	-0.199076638124\\
0.00887703513247	-0.199077526755\\
0.00887984758485	-0.199075979394\\
0.0088805458403	-0.19907451549\\
};
\addlegendentry{$y_a$};

\addplot [color=black,dash pattern=on 1pt off 3pt on 3pt off 3pt,line width=3.0pt]
  table[row sep=crcr]{%
0.447789444596	-0.0532388497689\\
0.447789444596	-0.0532388497689\\
0.447789444596	-0.0532388497689\\
0.447789444596	-0.0532388497689\\
0.447789444596	-0.0532388497689\\
0.447789444596	-0.0532388497689\\
0.447789444596	-0.0532388497689\\
0.447789444596	-0.0532388497689\\
0.447789444596	-0.0532388497689\\
0.447789444596	-0.0532388497689\\
0.447789444596	-0.0532388497689\\
0.447789444596	-0.0532388497689\\
0.447789444596	-0.0532388497689\\
0.447789444596	-0.0532388497689\\
0.447789444596	-0.0532388497689\\
0.447789444596	-0.0532388497689\\
0.447789444596	-0.0532388497689\\
0.447789444596	-0.0532388497689\\
0.447789444596	-0.0532388497689\\
0.447789444596	-0.0532388497689\\
0.446882293098	-0.0532770817831\\
0.440237713281	-0.0534396613276\\
0.427608422945	-0.0546205204587\\
0.410446147773	-0.0567485192239\\
0.390067108444	-0.0591252003342\\
0.368263406358	-0.061235389829\\
0.346993394978	-0.0631498829873\\
0.32669087205	-0.0650979277094\\
0.307588215921	-0.0671794599954\\
0.288766780624	-0.0697172413699\\
0.271071729159	-0.0725631234566\\
0.25320586621	-0.0752901572742\\
0.234994493924	-0.0772321719472\\
0.217044440893	-0.0782345208557\\
0.199754928306	-0.0788354176606\\
0.183476772171	-0.080322673718\\
0.168290741354	-0.0839574938062\\
0.153715516057	-0.0894523920388\\
0.139343026189	-0.095151991112\\
0.124637118051	-0.100250915748\\
0.109042184274	-0.104923188617\\
0.093018525743	-0.109841234873\\
0.077759749493	-0.116003433963\\
0.0638844286105	-0.123782432857\\
0.0512044730876	-0.132426441252\\
0.0395407036342	-0.14105635452\\
0.0292334900834	-0.14900315893\\
0.0210394232757	-0.156090167589\\
0.01555510292	-0.1625766481\\
0.0122820118682	-0.168814214357\\
0.0104942561151	-0.173462355745\\
0.00929746674277	-0.176602287273\\
0.00850022610738	-0.179135834958\\
0.00796608575767	-0.181394631984\\
0.00762074918832	-0.183482925959\\
0.00753680614383	-0.186537420671\\
0.00757076114173	-0.190900072841\\
0.00784145271376	-0.195298099939\\
0.00834495354344	-0.198349512952\\
0.00871864411952	-0.199550640197\\
0.00885465354514	-0.199583537874\\
0.00889834173182	-0.199384917734\\
0.00891432346256	-0.199228245681\\
0.00892053570776	-0.199119380605\\
0.0089229922396	-0.199042641509\\
0.00892350792301	-0.198988277744\\
0.00892311517122	-0.198949638908\\
0.0089225975123	-0.198921483898\\
0.00892225681644	-0.198900542386\\
0.00892199988682	-0.198885063389\\
0.00892186471486	-0.19887359031\\
0.00892177617049	-0.198865115208\\
0.00892168019601	-0.198858845419\\
0.00892166501201	-0.198854242292\\
0.00892159789751	-0.198850841193\\
0.00892159513621	-0.198848349121\\
0.00892157747851	-0.198846513899\\
0.00892158152277	-0.198845165731\\
0.00892159109261	-0.198844190838\\
0.00892159356383	-0.198843476438\\
0.00892156645181	-0.198842946112\\
0.00892157382415	-0.1988425576\\
0.00892156698587	-0.198842272798\\
0.00892156637271	-0.198842066712\\
0.0089215749889	-0.198841931671\\
0.00892157783254	-0.198841809949\\
0.0089215632415	-0.198841716939\\
0.00892158087992	-0.198841676279\\
0.00892155842003	-0.198841621177\\
0.00892154012116	-0.198841593138\\
0.00892159365685	-0.198841568302\\
0.0089215837234	-0.198841554558\\
0.0089215839272	-0.19884154261\\
0.0089215839272	-0.19884154261\\
0.00892157399375	-0.198841528865\\
0.00892157119693	-0.1988415132\\
0.00892157119693	-0.1988415132\\
0.00892157119693	-0.1988415132\\
0.00892157119693	-0.1988415132\\
0.00892157119693	-0.1988415132\\
0.00892157119693	-0.1988415132\\
0.00892157119693	-0.1988415132\\
0.00892157119693	-0.1988415132\\
0.00892157119693	-0.1988415132\\
0.00892157119693	-0.1988415132\\
0.00892157119693	-0.1988415132\\
0.00892157119693	-0.1988415132\\
0.00892157119693	-0.1988415132\\
0.00892157119693	-0.1988415132\\
0.00892157119693	-0.1988415132\\
0.00892157119693	-0.1988415132\\
0.00892157119693	-0.1988415132\\
0.00892157119693	-0.1988415132\\
0.00892157119693	-0.1988415132\\
0.00892157119693	-0.1988415132\\
0.00892157119693	-0.1988415132\\
0.00892157119693	-0.1988415132\\
0.00892157119693	-0.1988415132\\
0.00892157119693	-0.1988415132\\
0.00892157119693	-0.1988415132\\
};
\addlegendentry{$y_u$};

\end{axis}

\begin{axis}[%
width=\figurewidth,
height=\figureheight,
scale only axis,
xmin=0,
xmax=25,
ymin=-0.2,
ymax=0,
ylabel absolute,
ylabel={$z$ [m]},
name=plot3,
at=(plot2.below south west),
anchor=above north west,
legend style={draw=black,fill=white,legend cell align=left},
yticklabel style={/pgf/number format/fixed},
]
\addplot [color=blue,solid,line width=2.0pt]
  table[row sep=crcr]{%
0.2	-0.0533229633798\\
0.4	-0.0533243669732\\
0.6	-0.0533256408499\\
0.8	-0.0533251305364\\
1	-0.0533224245921\\
1.2	-0.0533230582869\\
1.4	-0.0533237304655\\
1.6	-0.0533236068717\\
1.8	-0.0533223307397\\
2	-0.0533221571517\\
2.2	-0.0533236735565\\
2.4	-0.0533246276995\\
2.6	-0.0533263513485\\
2.8	-0.0533234841779\\
3	-0.0533228713568\\
3.2	-0.0533256671978\\
3.4	-0.0533218187563\\
3.6	-0.0533223008898\\
3.8	-0.0533256280189\\
4	-0.0531383681401\\
4.2	-0.0532407576105\\
4.4	-0.0542224020736\\
4.6	-0.0564497032399\\
4.8	-0.0590803936664\\
5	-0.0611340024051\\
5.2	-0.0626274910118\\
5.4	-0.0642211848062\\
5.6	-0.0659735424409\\
5.8	-0.0676731506139\\
6	-0.070114964197\\
6.2	-0.0739762132699\\
6.4	-0.0762498851087\\
6.6	-0.0782495055049\\
6.8	-0.0814889465765\\
7	-0.0907925480718\\
7.2	-0.0977661311755\\
7.4	-0.103859595794\\
7.6	-0.113058213729\\
7.8	-0.117403835615\\
8	-0.120255050889\\
8.2	-0.124001135527\\
8.4	-0.138236760538\\
8.6	-0.123280325871\\
8.8	-0.100244515377\\
9	-0.0861625225908\\
9.2	-0.0791940748216\\
9.4	-0.0759556118491\\
9.6	-0.0745042545034\\
9.8	-0.074605175237\\
10	-0.0753501523331\\
10.2	-0.0764978447659\\
10.4	-0.0776396841743\\
10.6	-0.0782991255993\\
10.8	-0.0787738519091\\
11	-0.0802050082146\\
11.2	-0.0838134811716\\
11.4	-0.0892232915585\\
11.6	-0.0948606841402\\
11.8	-0.0999853710156\\
12	-0.104353559831\\
12.2	-0.108509424245\\
12.4	-0.113639296491\\
12.6	-0.112103771217\\
12.8	-0.0987960576695\\
13	-0.0777149225386\\
13.2	-0.0705090583782\\
13.4	-0.0810009629113\\
13.6	-0.0751071874693\\
13.8	-0.0533133835361\\
14	-0.0678460894056\\
14.2	-0.0862479140856\\
14.4	-0.0691137493489\\
14.6	-0.0456467689364\\
14.8	-0.0289648524846\\
15	-0.0372429808019\\
15.2	-0.040049298813\\
15.4	-0.0339102672234\\
15.6	-0.0308827333808\\
15.8	-0.028321290903\\
16	-0.0464078626437\\
16.2	-0.0621117634349\\
16.4	-0.044421800408\\
16.6	-0.0300254745908\\
16.8	-0.0235422904677\\
17	-0.0466458922011\\
17.2	-0.075015216404\\
17.4	-0.0909630362893\\
17.6	-0.105131388121\\
17.8	-0.113260803283\\
18	-0.1172097342\\
18.2	-0.11928010113\\
18.4	-0.121342050507\\
18.6	-0.124744389185\\
18.8	-0.129732901645\\
19	-0.135691305704\\
19.2	-0.142268273861\\
19.4	-0.149002347243\\
19.6	-0.15564192215\\
19.8	-0.161998645435\\
20	-0.168182097278\\
20.2	-0.173110704521\\
20.4	-0.176373725975\\
20.6	-0.178835565179\\
20.8	-0.181111323668\\
21	-0.18189619387\\
21.2	-0.181421650544\\
21.4	-0.183704901467\\
21.6	-0.185681439595\\
21.8	-0.188092184461\\
22	-0.191469884508\\
22.2	-0.195297855255\\
22.4	-0.198157295918\\
22.6	-0.199415229461\\
22.8	-0.199706399445\\
23	-0.199558905198\\
23.1999999999999	-0.199230713771\\
23.3999999999999	-0.199076638124\\
23.5999999999999	-0.199077526755\\
23.7999999999999	-0.199075979394\\
23.9999999999999	-0.19907451549\\
};
\addlegendentry{$y_a$};

\addplot [color=red,dashed,line width=2.0pt]
  table[row sep=crcr]{%
0.2	-0.0532359469319\\
0.4	-0.0532359469319\\
0.6	-0.0532359469319\\
0.8	-0.0532359469319\\
1	-0.0532359469319\\
1.2	-0.0532359469319\\
1.4	-0.0532359469319\\
1.6	-0.0532359469319\\
1.8	-0.0532359469319\\
2	-0.0532359469319\\
2.2	-0.0532359469319\\
2.4	-0.0532359469319\\
2.6	-0.0532359469319\\
2.8	-0.0532359469319\\
3	-0.0532359469319\\
3.2	-0.0532359469319\\
3.4	-0.0532359469319\\
3.6	-0.0532359469319\\
3.8	-0.0532359469319\\
4	-0.0532771060022\\
4.2	-0.053438427829\\
4.4	-0.0546093591266\\
4.6	-0.0566953298835\\
4.8	-0.0589769576852\\
5	-0.0609290463148\\
5.2	-0.0626022904793\\
5.4	-0.0642110184437\\
5.6	-0.065855584301\\
5.8	-0.0677766298532\\
6	-0.0700924255731\\
6.2	-0.0726275274421\\
6.4	-0.0730535820904\\
6.6	-0.0731384929114\\
6.8	-0.0731890328527\\
7	-0.0732205684791\\
7.2	-0.0732383721021\\
7.4	-0.0732500111696\\
7.6	-0.0732579990868\\
7.8	-0.0732634976259\\
8	-0.0732677552319\\
8.2	-0.0732712987339\\
8.4	-0.0732741568616\\
8.6	-0.0732763667194\\
8.8	-0.0732791715753\\
9	-0.0732851900638\\
9.2	-0.0733022858824\\
9.4	-0.0733596021233\\
9.6	-0.0735580513499\\
9.8	-0.0742082335726\\
10	-0.0754060711078\\
10.2	-0.0767012686203\\
10.4	-0.0777247874097\\
10.6	-0.0783513208026\\
10.8	-0.0788962684581\\
11	-0.0803413970429\\
11.2	-0.0838295952027\\
11.4	-0.0891927888773\\
11.6	-0.0947833187784\\
11.8	-0.0997290164022\\
12	-0.104139015589\\
12.2	-0.108584130422\\
12.4	-0.113789338651\\
12.6	-0.119211850919\\
12.8	-0.11992485606\\
13	-0.120002366789\\
13.2	-0.120022737476\\
13.4	-0.120034730272\\
13.6	-0.120048322337\\
13.8	-0.120060145877\\
14	-0.120067628926\\
14.2	-0.120076710748\\
14.4	-0.120090554164\\
14.6	-0.120100921737\\
14.8	-0.12010645493\\
15	-0.120109888005\\
15.2	-0.120113019508\\
15.4	-0.120116006364\\
15.6	-0.120118624315\\
15.8	-0.120120902507\\
16	-0.120123106923\\
16.2	-0.120126004151\\
16.4	-0.120129896064\\
16.6	-0.120133384008\\
16.8	-0.120136195922\\
17	-0.120138747845\\
17.2	-0.120142455649\\
17.4	-0.120149752911\\
17.6	-0.120165856757\\
17.8	-0.120210013562\\
18	-0.120353797064\\
18.2	-0.120856013036\\
18.4	-0.122440541011\\
18.6	-0.125824367107\\
18.8	-0.130580152587\\
19	-0.136323810538\\
19.2	-0.142717068386\\
19.4	-0.149270606275\\
19.6	-0.155685174974\\
19.8	-0.16201772691\\
20	-0.168097807971\\
20.2	-0.172918985412\\
20.4	-0.176180675266\\
20.6	-0.178726865421\\
20.8	-0.181005416876\\
21	-0.183027950522\\
21.2	-0.184605897091\\
21.4	-0.185375823811\\
21.6	-0.186504552079\\
21.8	-0.18866152982\\
22	-0.191977806048\\
22.2	-0.195631877321\\
22.4	-0.198328958575\\
22.6	-0.199531740668\\
22.8	-0.199581545031\\
23	-0.199386609232\\
23.1999999999999	-0.19923067883\\
23.3999999999999	-0.19917207588\\
23.5999999999999	-0.19917207588\\
23.7999999999999	-0.19917207588\\
23.9999999999999	-0.19917207588\\
};
\addlegendentry{$y_c$};

\addplot [color=black,dash pattern=on 1pt off 3pt on 3pt off 3pt,line width=2.0pt]
  table[row sep=crcr]{%
0.2	-0.0532388497689\\
0.4	-0.0532388497689\\
0.6	-0.0532388497689\\
0.8	-0.0532388497689\\
1	-0.0532388497689\\
1.2	-0.0532388497689\\
1.4	-0.0532388497689\\
1.6	-0.0532388497689\\
1.8	-0.0532388497689\\
2	-0.0532388497689\\
2.2	-0.0532388497689\\
2.4	-0.0532388497689\\
2.6	-0.0532388497689\\
2.8	-0.0532388497689\\
3	-0.0532388497689\\
3.2	-0.0532388497689\\
3.4	-0.0532388497689\\
3.6	-0.0532388497689\\
3.8	-0.0532388497689\\
4	-0.0532388497689\\
4.2	-0.0532770817831\\
4.4	-0.0534396613276\\
4.6	-0.0546205204587\\
4.8	-0.0567485192239\\
5	-0.0591252003342\\
5.2	-0.061235389829\\
5.4	-0.0631498829873\\
5.6	-0.0650979277094\\
5.8	-0.0671794599954\\
6	-0.0697172413699\\
6.2	-0.0725631234566\\
6.4	-0.0752901572742\\
6.6	-0.0772321719472\\
6.8	-0.0782345208557\\
7	-0.0788354176606\\
7.2	-0.080322673718\\
7.4	-0.0839574938062\\
7.6	-0.0894523920388\\
7.8	-0.095151991112\\
8	-0.100250915748\\
8.2	-0.104923188617\\
8.4	-0.109841234873\\
8.6	-0.116003433963\\
8.8	-0.123782432857\\
9	-0.132426441252\\
9.2	-0.14105635452\\
9.4	-0.14900315893\\
9.6	-0.156090167589\\
9.8	-0.1625766481\\
10	-0.168814214357\\
10.2	-0.173462355745\\
10.4	-0.176602287273\\
10.6	-0.179135834958\\
10.8	-0.181394631984\\
11	-0.183482925959\\
11.2	-0.186537420671\\
11.4	-0.190900072841\\
11.6	-0.195298099939\\
11.8	-0.198349512952\\
12	-0.199550640197\\
12.2	-0.199583537874\\
12.4	-0.199384917734\\
12.6	-0.199228245681\\
12.8	-0.199119380605\\
13	-0.199042641509\\
13.2	-0.198988277744\\
13.4	-0.198949638908\\
13.6	-0.198921483898\\
13.8	-0.198900542386\\
14	-0.198885063389\\
14.2	-0.19887359031\\
14.4	-0.198865115208\\
14.6	-0.198858845419\\
14.8	-0.198854242292\\
15	-0.198850841193\\
15.2	-0.198848349121\\
15.4	-0.198846513899\\
15.6	-0.198845165731\\
15.8	-0.198844190838\\
16	-0.198843476438\\
16.2	-0.198842946112\\
16.4	-0.1988425576\\
16.6	-0.198842272798\\
16.8	-0.198842066712\\
17	-0.198841931671\\
17.2	-0.198841809949\\
17.4	-0.198841716939\\
17.6	-0.198841676279\\
17.8	-0.198841621177\\
18	-0.198841593138\\
18.2	-0.198841568302\\
18.4	-0.198841554558\\
18.6	-0.19884154261\\
18.8	-0.19884154261\\
19	-0.198841528865\\
19.2	-0.1988415132\\
19.4	-0.1988415132\\
19.6	-0.1988415132\\
19.8	-0.1988415132\\
20	-0.1988415132\\
20.2	-0.1988415132\\
20.4	-0.1988415132\\
20.6	-0.1988415132\\
20.8	-0.1988415132\\
21	-0.1988415132\\
21.2	-0.1988415132\\
21.4	-0.1988415132\\
21.6	-0.1988415132\\
21.8	-0.1988415132\\
22	-0.1988415132\\
22.2	-0.1988415132\\
22.4	-0.1988415132\\
22.6	-0.1988415132\\
22.8	-0.1988415132\\
23	-0.1988415132\\
23.1999999999999	-0.1988415132\\
23.3999999999999	-0.1988415132\\
23.5999999999999	-0.1988415132\\
23.7999999999999	-0.1988415132\\
23.9999999999999	-0.1988415132\\
};
\addlegendentry{$y_u$};

\end{axis}

\begin{axis}[%
width=\figurewidth,
height=\figureheight,
scale only axis,
xmin=0,
xmax=25,
xlabel={Time [s]},
ymin=0,
ymax=3.5,
ylabel absolute,
ylabel={$\Vert \ddot{y}_{r} \Vert$ [m/s$^2$]},
at=(plot3.below south west),
anchor=above north west,
yticklabel style={/pgf/number format/fixed},
]
\addplot [color=blue,solid,line width=1.5pt,forget plot]
  table[row sep=crcr]{%
0.2	0\\
0.4	0\\
0.6	0\\
0.8	0\\
1	0\\
1.2	0\\
1.4	0\\
1.6	0\\
1.8	0\\
2	0\\
2.2	0\\
2.4	0\\
2.6	0\\
2.8	0\\
3	0\\
3.2	0\\
3.4	0\\
3.6	0\\
3.8	0\\
4	0.215630457138618\\
4.2	0.183532545669181\\
4.4	0.112964800830739\\
4.6	0.0660874034092386\\
4.8	0.0919120214487712\\
5	0.0499824214355521\\
5.2	0.0775820807431204\\
5.4	0.0586053891674744\\
5.6	0.114809936855621\\
5.8	0.0572829754718231\\
6	0.0321677352328744\\
6.2	0.970190099243693\\
6.4	0.111424037711193\\
6.6	0.688840211795642\\
6.8	0.978629291260656\\
7	1.17341598006761\\
7.2	0.965928158798193\\
7.4	1.51920175126052\\
7.6	1.59304642226496\\
7.8	1.74789926296163\\
8	1.52705121220997\\
8.2	2.11282149420179\\
8.4	2.43081886314547\\
8.6	0.0306568562420203\\
8.8	0.286774832237394\\
9	0.208570406631437\\
9.2	0.120620733176198\\
9.4	0.0547242708966397\\
9.6	0.0536541949484007\\
9.8	0.164339128007777\\
10	0.0995011849814585\\
10.2	0.0661976134321894\\
10.4	0.0745436977661301\\
10.6	0.0816331918679677\\
10.8	0.0663679543718014\\
11	0.0837870634545484\\
11.2	0.0702399989355227\\
11.4	0.0496420768395497\\
11.6	0.0588207829407025\\
11.8	0.0105565991938198\\
12	0.0349791124605229\\
12.2	0.0255784042047986\\
12.4	0.0347622657034332\\
12.6	0.931534904659203\\
12.8	1.86963097852128\\
13	2.04719898634495\\
13.2	1.38959315338278\\
13.4	0.0908882201635765\\
13.6	2.27778378750004\\
13.8	2.14710681027638\\
14	0.0375732343893648\\
14.2	0.7749437455585\\
14.4	2.64394368315133\\
14.6	2.8898097939077\\
14.8	2.8367801149357\\
15	1.60257949626978\\
15.2	2.64078117376496\\
15.4	2.92966246477479\\
15.6	2.72193877616668\\
15.8	2.70847765826456\\
16	0.1311204231244\\
16.2	2.09847165550063\\
16.4	2.90654439540028\\
16.6	2.7524009731186\\
16.8	2.29855809720561\\
17	0.0897614368487041\\
17.2	0.50354711533474\\
17.4	0.185003330371181\\
17.6	0.209093925419585\\
17.8	0.109067466294002\\
18	0.0801808466916367\\
18.2	0.0623784668783835\\
18.4	0.122906550834614\\
18.6	0.0914015900796619\\
18.8	0.0541413920630687\\
19	0.0260059901762671\\
19.2	0.0487353859373358\\
19.4	0.0582733207821273\\
19.6	0.0650760560034079\\
19.8	0.0907599705110287\\
20	0.0874093674223185\\
20.2	0.0477007880552645\\
20.4	0.0587098616725932\\
20.6	0.040419547799168\\
20.8	0.0250384651802559\\
21	0.670068890854827\\
21.2	0.14577746493467\\
21.4	0.0320627438496575\\
21.6	0.0555234680208893\\
21.8	0.0160196124303074\\
22	0.0213910439888626\\
22.2	0.0287175704513051\\
22.4	0.0439197690139041\\
22.6	0.0597778714014099\\
22.8	0.0266409283856713\\
23	0.0449610583800005\\
23.1999999999999	0.049918446629071\\
23.3999999999999	0\\
23.5999999999999	0\\
23.7999999999999	0\\
23.9999999999999	0\\
};
\end{axis}

\draw [->] (3.5,6.0) -- (2,5.5);

\end{tikzpicture}%

%% file: chapters/paper2/figs/scen2.tex
%
%
\begin{tikzpicture}

\begin{axis}[%
width=\figurewidth,
height=\figureheight,
scale only axis,
xmin=0,
xmax=25,
ymin=0,
ymax=0.1,
ylabel absolute,
ylabel={$\Vert y_c-y_a \Vert$ [m]},
yticklabel style={/pgf/number format/fixed},
name=plot2
]
\addplot [color=blue,solid,line width=2.0pt,forget plot]
  table[row sep=crcr]{%
0.2	0.000364089159783116\\
0.4	0.000363544249070649\\
0.6	0.000365462430629543\\
0.8	0.000365408122418706\\
1	0.000363270339869484\\
1.2	0.000366054372336287\\
1.4	0.000365067438729697\\
1.6	0.000366151864229917\\
1.8	0.000366430598029409\\
2	0.000365403523638064\\
2.2	0.000365664131798211\\
2.4	0.000365850428817542\\
2.6	0.000364677871740482\\
2.8	0.000366297124715977\\
3	0.000366907612329772\\
3.2	0.000364097721823597\\
3.4	0.000363819981871198\\
3.6	0.000364621398369268\\
3.8	0.000367124557036632\\
4	0.000365476914501064\\
4.2	0.000361093139517437\\
4.4	0.000364870172800485\\
4.6	0.00036676779174014\\
4.8	0.000364086003918821\\
5	0.000364758064209577\\
5.2	0.000364983109205548\\
5.4	0.000364568768981469\\
5.6	0.00036524166756858\\
5.8	0.000365389885471511\\
6	0.000366476174799447\\
6.2	0.000364014804604384\\
6.4	0.000364795489020359\\
6.6	0.000365734927175758\\
6.8	0.000363572424623379\\
7	0.00036625213223787\\
7.2	0.000342491004315742\\
7.4	0.000248660177163144\\
7.6	0.000370973452838497\\
7.8	0.000568004163787507\\
8	0.00107819683029797\\
8.2	0.000846050923004864\\
8.4	0.000969147064533933\\
8.6	0.00126460216954975\\
8.8	0.00138192258109411\\
9	0.00109294981929434\\
9.2	0.000471505110699585\\
9.4	0.000234979913981644\\
9.6	0.00016953350657917\\
9.8	0.000454463659847094\\
10	0.00045840455839205\\
10.2	0.00105484179604041\\
10.4	0.0119534516477861\\
10.6	0.0305674044440157\\
10.8	0.0509125177397114\\
11	0.0645767176487799\\
11.2	0.0742895309875981\\
11.4	0.0813160785008112\\
11.6	0.0852845970151826\\
11.8	0.0868766042843275\\
12	0.0881016868290004\\
12.2	0.0888019259985254\\
12.4	0.0886462456673304\\
12.6	0.0879409792698235\\
12.8	0.0878082115714995\\
13	0.0890209323651448\\
13.2	0.0888742549122217\\
13.4	0.0884513277827341\\
13.6	0.088477233717888\\
13.8	0.0885465922554187\\
14	0.0880848536452647\\
14.2	0.0876210461794438\\
14.4	0.0882894267385283\\
14.6	0.0890856994996922\\
14.8	0.0892369719470561\\
15	0.0887964002623204\\
15.2	0.0884594555405201\\
15.4	0.0892571140837982\\
15.6	0.0896375588500428\\
15.8	0.0882663962428944\\
16	0.0878997356168885\\
16.2	0.0881747032440973\\
16.4	0.0851010284858044\\
16.6	0.0735896411059794\\
16.8	0.0445209325008874\\
17	0.0225461759564046\\
17.2	0.0104361257063117\\
17.4	0.00459967551159265\\
17.6	0.00232754737892941\\
17.8	0.0015726703653266\\
18	0.00165476662251359\\
18.2	0.00143793335848187\\
18.4	0.0009551683838119\\
18.6	0.000863013528142028\\
18.8	0.000542308756423696\\
19	0.00033601290185447\\
19.2	0.000138417845707246\\
19.4	0.000108914864219911\\
19.6	0.000264897961376642\\
19.8	0.000402143845486736\\
20	0.000542428620370628\\
20.2	0.000513083955531034\\
20.4	0.000436476894265814\\
20.6	0.000490447483439045\\
20.8	0.000405580090383798\\
21	0.000258186563901506\\
21.2	0.000203387257483675\\
21.4	0.000288865775718599\\
21.6	0.000205686411268026\\
21.8	0.00010089839213889\\
22	6.65923462132708e-05\\
22.2	0.000127682426894539\\
22.4	0.000357351757156814\\
22.6	0.000277558743609088\\
22.8	0.000209772713458636\\
23	0.000212282639574521\\
23.1999999999999	0.00021528119169006\\
23.3999999999999	0.000217629569496219\\
23.5999999999999	0.000221568553637652\\
23.7999999999999	0.000223312644135338\\
23.9999999999999	0.000223942915939466\\
};
\end{axis}

\begin{axis}[%
width=\figurewidth,
height=2\figureheight,
scale only axis,
xmin=-0.000234721669811,
xmax=0.447396908321,
xlabel={$y$ [m]},
ymin=-0.196939583697,
ymax=-0.0359867843259,
ylabel absolute,
ylabel={$z$ [m]},
at=(plot2.above north west),
anchor=below south west,
yticklabel style={/pgf/number format/fixed},
legend style={draw=black,fill=white,legend cell align=left}
]
\addplot [color=blue,solid,line width=1.5pt]
  table[row sep=crcr]{%
0.447394966896	-0.0736824386683\\
0.447395517801	-0.0736828530775\\
0.447393986296	-0.0736817847327\\
0.447394857405	-0.0736812588064\\
0.447394267938	-0.0736840829701\\
0.447395025956	-0.0736805998471\\
0.447395749447	-0.0736814676659\\
0.447393973193	-0.073681322668\\
0.447396264556	-0.0736795129387\\
0.447395517221	-0.0736810495959\\
0.447394623784	-0.0736813336102\\
0.44739596147	-0.0736803039701\\
0.447392994287	-0.0736825762481\\
0.447395353344	-0.0736801140853\\
0.447395697696	-0.0736794881811\\
0.447396012279	-0.0736821111345\\
0.447395010722	-0.073682658068\\
0.447394610076	-0.0736821381249\\
0.447395954228	-0.0736790574426\\
0.447394647489	-0.0736812625696\\
0.447395329744	-0.0736853282261\\
0.447396692167	-0.0736810279113\\
0.447394996177	-0.0736800175341\\
0.447396205579	-0.0736819974896\\
0.447396908321	-0.0736806479377\\
0.44739547311	-0.0736816901529\\
0.447395222184	-0.0736818331438\\
0.44739498183	-0.0736813763942\\
0.447394352075	-0.073681680453\\
0.447393233955	-0.0736814878636\\
0.44739664215	-0.0736818278231\\
0.447394989069	-0.0736817616149\\
0.447395551337	-0.073680966747\\
0.447394379384	-0.0736833441553\\
0.447394621297	-0.07368069127\\
0.44632955825	-0.0738247817938\\
0.439076579353	-0.0759292784189\\
0.427120725283	-0.0793520361358\\
0.411347365603	-0.083070544136\\
0.392730619147	-0.0862944187053\\
0.372183764536	-0.0888913503855\\
0.350844010617	-0.0917201045889\\
0.329117162487	-0.0949722800874\\
0.308421308141	-0.0982266493457\\
0.288987831751	-0.1013044778\\
0.269982852036	-0.104184992662\\
0.250815873669	-0.107100898669\\
0.232854188919	-0.109484077282\\
0.215941638456	-0.1113912758\\
0.199696835454	-0.113886844027\\
0.183005567965	-0.117946691028\\
0.176781251705	-0.110945714848\\
0.183111944017	-0.0939748345003\\
0.188468965814	-0.0740340737897\\
0.189691190893	-0.0599797600953\\
0.190757266603	-0.0501617298674\\
0.19248528922	-0.0434006131927\\
0.193840582039	-0.0397426328904\\
0.195006216187	-0.0384391663975\\
0.196016059926	-0.037378315716\\
0.196203824357	-0.0366323314909\\
0.195970904752	-0.0367227261786\\
0.195918413165	-0.037470088813\\
0.196306018839	-0.0376384824938\\
0.196748168778	-0.0365095301421\\
0.196941981086	-0.0367050730764\\
0.196696082931	-0.0370472883414\\
0.197134307939	-0.0371129056567\\
0.197431989817	-0.0371404062703\\
0.1972474272	-0.0376262035124\\
0.197984600607	-0.0383043105069\\
0.198179876316	-0.0376290319921\\
0.198020151259	-0.036745562567\\
0.197835206417	-0.0365113045427\\
0.197727000297	-0.0369644656688\\
0.197650745232	-0.0372682907058\\
0.197133233664	-0.0362801557272\\
0.197469318039	-0.0359867843259\\
0.198329463173	-0.0377037347018\\
0.198724598825	-0.0382125338695\\
0.198873326356	-0.037968684897\\
0.19979157731	-0.0416611979152\\
0.198803847199	-0.0539373289678\\
0.189982693874	-0.0819354088721\\
0.181236175839	-0.101963100491\\
0.175849810804	-0.112642502849\\
0.172810214004	-0.117667122369\\
0.17115510508	-0.119947124075\\
0.168416153025	-0.121649342555\\
0.162201679915	-0.123646095415\\
0.152517820287	-0.126353820418\\
0.140961363399	-0.129349521933\\
0.127856455319	-0.132238195981\\
0.113423041189	-0.134877147674\\
0.097981199806	-0.137206265717\\
0.082045929918	-0.139252115286\\
0.066633199211	-0.141445675643\\
0.0520762488756	-0.144198337817\\
0.0389471177735	-0.147704090678\\
0.0276276325817	-0.15172095079\\
0.0184826182297	-0.156404102848\\
0.0118926701651	-0.161793260863\\
0.00778305959407	-0.167568256413\\
0.00570652568709	-0.172521765114\\
0.00426702105975	-0.176159196286\\
0.00313574863015	-0.179322468747\\
0.00190239481711	-0.183144147147\\
0.000863341201903	-0.188027201265\\
0.000271050511569	-0.193026584533\\
0.000329613690186	-0.196153103185\\
0.000556168280847	-0.196939583697\\
0.00108221316537	-0.19665393315\\
0.00107890072705	-0.196074850092\\
0.00103883919273	-0.195910565785\\
0.00102553734879	-0.195891664089\\
0.00102416263423	-0.195886411744\\
0.00102546057162	-0.195884856612\\
0.00102716633017	-0.195882138117\\
0.00102688289676	-0.195879929321\\
0.00102752668346	-0.195879604758\\
};
\addlegendentry{$y_a$};

\addplot [color=black,dash pattern=on 1pt off 3pt on 3pt off 3pt,line width=3.0pt]
  table[row sep=crcr]{%
0.447326709237	-0.0740638202575\\
0.447326709237	-0.0740638202575\\
0.447326709237	-0.0740638202575\\
0.447326709237	-0.0740638202575\\
0.447326709237	-0.0740638202575\\
0.447326709237	-0.0740638202575\\
0.447326709237	-0.0740638202575\\
0.447326709237	-0.0740638202575\\
0.447326709237	-0.0740638202575\\
0.447326709237	-0.0740638202575\\
0.447326709237	-0.0740638202575\\
0.447326709237	-0.0740638202575\\
0.447326709237	-0.0740638202575\\
0.447326709237	-0.0740638202575\\
0.447326709237	-0.0740638202575\\
0.447326709237	-0.0740638202575\\
0.447326709237	-0.0740638202575\\
0.447326709237	-0.0740638202575\\
0.447326709237	-0.0740638202575\\
0.447326709237	-0.0740638202575\\
0.447326709237	-0.0740638202575\\
0.447326709237	-0.0740638202575\\
0.447326709237	-0.0740638202575\\
0.447326709237	-0.0740638202575\\
0.447326709237	-0.0740638202575\\
0.447326709237	-0.0740638202575\\
0.447326709237	-0.0740638202575\\
0.447326709237	-0.0740638202575\\
0.447326709237	-0.0740638202575\\
0.447326709237	-0.0740638202575\\
0.447326709237	-0.0740638202575\\
0.447326709237	-0.0740638202575\\
0.447326709237	-0.0740638202575\\
0.447326709237	-0.0740638202575\\
0.447326709237	-0.0740638202575\\
0.447326709237	-0.0740638202575\\
0.446075855666	-0.0740488153114\\
0.439230365951	-0.0760171772099\\
0.427337940694	-0.0794014749521\\
0.411478622107	-0.0831278021661\\
0.392594788284	-0.0861287084614\\
0.370104391251	-0.0891579114316\\
0.34587320287	-0.0926196609802\\
0.320477984783	-0.0963498745273\\
0.295555338358	-0.100216535778\\
0.272274140585	-0.104030578949\\
0.251051577998	-0.107225425823\\
0.231877074015	-0.109627724098\\
0.214263034382	-0.111648788353\\
0.197182522716	-0.114529493813\\
0.179889629948	-0.118906443975\\
0.162728879271	-0.123839553152\\
0.145225463259	-0.128385412733\\
0.128641249704	-0.132091928028\\
0.112193383102	-0.135179369711\\
0.095601432885	-0.137662557955\\
0.0792801569769	-0.139741664493\\
0.0637037223196	-0.141945714305\\
0.0493226704943	-0.144862516651\\
0.0365664403055	-0.148570770257\\
0.0255381479563	-0.152816481036\\
0.0165626692871	-0.15778268205\\
0.0103763005889	-0.163636448747\\
0.00698803177047	-0.169563860178\\
0.00515595662521	-0.174199034321\\
0.00362730882435	-0.177800959332\\
0.00222882106618	-0.181292767015\\
0.00106509788173	-0.185974125046\\
0.000374484802931	-0.191338201498\\
0.000292840448021	-0.195405432388\\
0.000533409039373	-0.196788199155\\
0.000729487618161	-0.196624045169\\
0.000832514396064	-0.196289215324\\
0.000889788779594	-0.196087629589\\
0.000927552871902	-0.195963898045\\
0.000953500298094	-0.195884827668\\
0.000971436278066	-0.195834310149\\
0.000983683753202	-0.19580213519\\
0.000991810388447	-0.195781860955\\
0.000997257616201	-0.195769175973\\
0.00100084326564	-0.195761234815\\
0.00100324573397	-0.195756298859\\
0.00100485099556	-0.195753249563\\
0.00100590121752	-0.195751373211\\
0.00100665295996	-0.195750247845\\
0.00100711367146	-0.195749532995\\
0.00100744610434	-0.195749140524\\
0.00100764626098	-0.195748890208\\
0.00100779892746	-0.195748736397\\
0.00100790270999	-0.195748679718\\
0.00100796193424	-0.195748630271\\
0.00100800561283	-0.195748588287\\
0.00100802115849	-0.195748580825\\
0.00100803981314	-0.195748577059\\
0.00100806861716	-0.195748586396\\
0.00100806550818	-0.195748582698\\
0.0010080623992	-0.195748579001\\
0.00100809431221	-0.195748592035\\
0.00100809431221	-0.195748592035\\
0.00100809431221	-0.195748592035\\
0.00100809431221	-0.195748592035\\
0.00100809120323	-0.195748588337\\
0.00100809120323	-0.195748588337\\
0.00100809120323	-0.195748588337\\
0.00100809120323	-0.195748588337\\
0.00100809120323	-0.195748588337\\
0.00100809120323	-0.195748588337\\
0.00100809120323	-0.195748588337\\
0.00100809120323	-0.195748588337\\
0.00100809120323	-0.195748588337\\
0.00100809120323	-0.195748588337\\
0.00100809120323	-0.195748588337\\
0.00100809120323	-0.195748588337\\
0.00100809120323	-0.195748588337\\
0.00100809120323	-0.195748588337\\
0.00100809120323	-0.195748588337\\
0.00100809120323	-0.195748588337\\
0.00100809120323	-0.195748588337\\
0.00100809120323	-0.195748588337\\
0.00100809120323	-0.195748588337\\
};
\addlegendentry{$y_u$};

\addplot [color=magenta,dashed,line width=1.0pt]
  table[row sep=crcr]{%
0.447333884954	-0.0740616480804\\
0.447333884954	-0.0740616480804\\
0.447333884954	-0.0740616480804\\
0.447333884954	-0.0740616480804\\
0.447333884954	-0.0740616480804\\
0.447106900457	-0.0738727153396\\
0.443530126898	-0.0747663118556\\
0.435828077326	-0.0766703230459\\
0.424037331648	-0.0793644375339\\
0.408623836573	-0.0825090182989\\
0.390409574082	-0.0857910011532\\
0.370329552192	-0.0890720366002\\
0.349299133396	-0.0923658448874\\
0.328125928118	-0.0956416540366\\
0.307388495854	-0.0988412764918\\
0.287302532763	-0.101879014057\\
0.268295779177	-0.104633688565\\
0.249239178201	-0.107239507252\\
0.23141383864	-0.1095830322\\
0.214824178216	-0.111530919389\\
0.200478470708	-0.112144550218\\
0.189827625863	-0.109909769548\\
0.184470903379	-0.103015998015\\
0.183926223719	-0.0915386093578\\
0.185823555766	-0.0776909607156\\
0.188158022006	-0.0643099004696\\
0.190020574665	-0.0535853865518\\
0.191526404932	-0.0460285745037\\
0.192899478543	-0.041236669199\\
0.194097681904	-0.0384897511454\\
0.194998650672	-0.0370972690476\\
0.195530482835	-0.0365095978136\\
0.195812239653	-0.0363976280145\\
0.19598159666	-0.0365487497804\\
0.196178720788	-0.036670180752\\
0.19641796079	-0.0366751012072\\
0.196627418592	-0.0366848224176\\
0.196798685967	-0.0367811881422\\
0.196971809209	-0.0369718741688\\
0.197166069403	-0.0372330589411\\
0.197384126435	-0.0375001672405\\
0.197618521858	-0.0376527759689\\
0.197806199746	-0.0376069959331\\
0.197870366144	-0.0373649654879\\
0.197817692765	-0.0371185987822\\
0.197690813512	-0.0369992612825\\
0.197551442871	-0.0369396247721\\
0.197482511448	-0.0369131742001\\
0.197589419825	-0.0370277496361\\
0.197876361069	-0.0374237378131\\
0.19825905884	-0.0383764418229\\
0.198554181339	-0.0406192979266\\
0.19834239272	-0.0456486252069\\
0.196814047218	-0.0555355375896\\
0.193247564441	-0.0700953430983\\
0.187604744287	-0.086727993811\\
0.181254723279	-0.101492338833\\
0.175848953184	-0.111832297859\\
0.171627479112	-0.118077817107\\
0.16768272763	-0.121675392658\\
0.162615871723	-0.124144136944\\
0.155516999838	-0.126410681518\\
0.146080813407	-0.128852922269\\
0.1345890688	-0.131450244822\\
0.121519010749	-0.134052603861\\
0.107316539406	-0.136530969725\\
0.09247443688	-0.13885858257\\
0.0775086459517	-0.141134100736\\
0.0629790172	-0.143564418915\\
0.0493828254829	-0.146366152083\\
0.0371574989563	-0.149687231497\\
0.0266430549	-0.153589765437\\
0.0182802176674	-0.157954891877\\
0.0117303282986	-0.162875525328\\
0.00734383435612	-0.167761966553\\
0.00455441641778	-0.172350548014\\
0.00282668458663	-0.176511201601\\
0.00165733834221	-0.180449887129\\
0.000779008696033	-0.184403552472\\
0.000138951588662	-0.188420194238\\
-0.000207480376789	-0.192063155648\\
-0.000234721669811	-0.194799632672\\
-3.49850999367e-05	-0.196344232257\\
0.000248166680725	-0.196845958491\\
0.000481637246816	-0.196776465233\\
0.000629108105383	-0.196530871435\\
0.000714843811131	-0.196329900327\\
0.000770337171546	-0.19620738248\\
0.000814605563134	-0.196133639439\\
0.000852304687627	-0.196084475103\\
0.000884258005728	-0.196048170523\\
0.00091131411287	-0.196020386782\\
0.000934092496247	-0.195998990323\\
0.000950060446509	-0.195985173246\\
0.000950060446509	-0.195985173246\\
0.000950060446509	-0.195985173246\\
0.000950060446509	-0.195985173246\\
0.000950060446509	-0.195985173246\\
0.000950060446509	-0.195985173246\\
0.000950060446509	-0.195985173246\\
0.000950060446509	-0.195985173246\\
};
\addlegendentry{$y_m$};

\end{axis}

\begin{axis}[%
width=\figurewidth,
height=\figureheight,
scale only axis,
xmin=0,
xmax=25,
ymin=-0.2,
ymax=0,
ylabel absolute,
ylabel={$z$ [m]},
name=plot3,
at=(plot2.below south west),
anchor=above north west,
yticklabel style={/pgf/number format/fixed},
legend style={draw=black,fill=white,legend cell align=left}
]
\addplot [color=blue,solid,line width=2.0pt]
  table[row sep=crcr]{%
0.2	-0.0736824386683\\
0.4	-0.0736828530775\\
0.6	-0.0736817847327\\
0.8	-0.0736812588064\\
1	-0.0736840829701\\
1.2	-0.0736805998471\\
1.4	-0.0736814676659\\
1.6	-0.073681322668\\
1.8	-0.0736795129387\\
2	-0.0736810495959\\
2.2	-0.0736813336102\\
2.4	-0.0736803039701\\
2.6	-0.0736825762481\\
2.8	-0.0736801140853\\
3	-0.0736794881811\\
3.2	-0.0736821111345\\
3.4	-0.073682658068\\
3.6	-0.0736821381249\\
3.8	-0.0736790574426\\
4	-0.0736812625696\\
4.2	-0.0736853282261\\
4.4	-0.0736810279113\\
4.6	-0.0736800175341\\
4.8	-0.0736819974896\\
5	-0.0736806479377\\
5.2	-0.0736816901529\\
5.4	-0.0736818331438\\
5.6	-0.0736813763942\\
5.8	-0.073681680453\\
6	-0.0736814878636\\
6.2	-0.0736818278231\\
6.4	-0.0736817616149\\
6.6	-0.073680966747\\
6.8	-0.0736833441553\\
7	-0.07368069127\\
7.2	-0.0738247817938\\
7.4	-0.0759292784189\\
7.6	-0.0793520361358\\
7.8	-0.083070544136\\
8	-0.0862944187053\\
8.2	-0.0888913503855\\
8.4	-0.0917201045889\\
8.6	-0.0949722800874\\
8.8	-0.0982266493457\\
9	-0.1013044778\\
9.2	-0.104184992662\\
9.4	-0.107100898669\\
9.6	-0.109484077282\\
9.8	-0.1113912758\\
10	-0.113886844027\\
10.2	-0.117946691028\\
10.4	-0.110945714848\\
10.6	-0.0939748345003\\
10.8	-0.0740340737897\\
11	-0.0599797600953\\
11.2	-0.0501617298674\\
11.4	-0.0434006131927\\
11.6	-0.0397426328904\\
11.8	-0.0384391663975\\
12	-0.037378315716\\
12.2	-0.0366323314909\\
12.4	-0.0367227261786\\
12.6	-0.037470088813\\
12.8	-0.0376384824938\\
13	-0.0365095301421\\
13.2	-0.0367050730764\\
13.4	-0.0370472883414\\
13.6	-0.0371129056567\\
13.8	-0.0371404062703\\
14	-0.0376262035124\\
14.2	-0.0383043105069\\
14.4	-0.0376290319921\\
14.6	-0.036745562567\\
14.8	-0.0365113045427\\
15	-0.0369644656688\\
15.2	-0.0372682907058\\
15.4	-0.0362801557272\\
15.6	-0.0359867843259\\
15.8	-0.0377037347018\\
16	-0.0382125338695\\
16.2	-0.037968684897\\
16.4	-0.0416611979152\\
16.6	-0.0539373289678\\
16.8	-0.0819354088721\\
17	-0.101963100491\\
17.2	-0.112642502849\\
17.4	-0.117667122369\\
17.6	-0.119947124075\\
17.8	-0.121649342555\\
18	-0.123646095415\\
18.2	-0.126353820418\\
18.4	-0.129349521933\\
18.6	-0.132238195981\\
18.8	-0.134877147674\\
19	-0.137206265717\\
19.2	-0.139252115286\\
19.4	-0.141445675643\\
19.6	-0.144198337817\\
19.8	-0.147704090678\\
20	-0.15172095079\\
20.2	-0.156404102848\\
20.4	-0.161793260863\\
20.6	-0.167568256413\\
20.8	-0.172521765114\\
21	-0.176159196286\\
21.2	-0.179322468747\\
21.4	-0.183144147147\\
21.6	-0.188027201265\\
21.8	-0.193026584533\\
22	-0.196153103185\\
22.2	-0.196939583697\\
22.4	-0.19665393315\\
22.6	-0.196074850092\\
22.8	-0.195910565785\\
23	-0.195891664089\\
23.1999999999999	-0.195886411744\\
23.3999999999999	-0.195884856612\\
23.5999999999999	-0.195882138117\\
23.7999999999999	-0.195879929321\\
23.9999999999999	-0.195879604758\\
};
\addlegendentry{$y_a$};

\addplot [color=red,dashed,line width=2.0pt]
  table[row sep=crcr]{%
0.2	-0.0740187616274\\
0.4	-0.0740187616274\\
0.6	-0.0740187616274\\
0.8	-0.0740187616274\\
1	-0.0740187616274\\
1.2	-0.0740187616274\\
1.4	-0.0740187616274\\
1.6	-0.0740187616274\\
1.8	-0.0740187616274\\
2	-0.0740187616274\\
2.2	-0.0740187616274\\
2.4	-0.0740187616274\\
2.6	-0.0740187616274\\
2.8	-0.0740187616274\\
3	-0.0740187616274\\
3.2	-0.0740187616274\\
3.4	-0.0740187616274\\
3.6	-0.0740187616274\\
3.8	-0.0740187616274\\
4	-0.0740187616274\\
4.2	-0.0740187616274\\
4.4	-0.0740187616274\\
4.6	-0.0740187616274\\
4.8	-0.0740187616274\\
5	-0.0740187616274\\
5.2	-0.0740187616274\\
5.4	-0.0740187616274\\
5.6	-0.0740187616274\\
5.8	-0.0740187616274\\
6	-0.0740187616274\\
6.2	-0.0740187616274\\
6.4	-0.0740187616274\\
6.6	-0.0740187616274\\
6.8	-0.0740187616274\\
7	-0.0740187616274\\
7.2	-0.0740484627922\\
7.4	-0.0760077152746\\
7.6	-0.0793683503518\\
7.8	-0.0830484778586\\
8	-0.0859694795576\\
8.2	-0.0887587696463\\
8.4	-0.0917549582641\\
8.6	-0.0948857987176\\
8.8	-0.0979695171356\\
9	-0.101109457646\\
9.2	-0.104323542235\\
9.4	-0.107228590995\\
9.6	-0.109500602566\\
9.8	-0.11137982578\\
10	-0.113916118891\\
10.2	-0.117750616379\\
10.4	-0.121096107119\\
10.6	-0.121369185213\\
10.8	-0.12140577512\\
11	-0.121417296268\\
11.2	-0.121423385256\\
11.4	-0.12142758798\\
11.6	-0.12143090716\\
11.8	-0.121433774385\\
12	-0.12143646275\\
12.2	-0.121439043181\\
12.4	-0.121441552571\\
12.6	-0.12144405028\\
12.8	-0.121446518166\\
13	-0.121449103197\\
13.2	-0.121451571563\\
13.4	-0.121454051428\\
13.6	-0.121456535973\\
13.8	-0.121459032017\\
14	-0.121461504988\\
14.2	-0.121464005813\\
14.4	-0.121466542597\\
14.6	-0.121469049093\\
14.8	-0.121471529949\\
15	-0.121473986725\\
15.2	-0.121476448547\\
15.4	-0.121478929743\\
15.6	-0.12148138407\\
15.8	-0.121483813772\\
16	-0.121486306702\\
16.2	-0.121488824785\\
16.4	-0.121491347905\\
16.6	-0.121494034026\\
16.8	-0.121497560287\\
17	-0.121504547845\\
17.2	-0.121523416411\\
17.4	-0.121585334992\\
17.6	-0.121810275368\\
17.8	-0.122580571763\\
18	-0.124385969228\\
18.2	-0.126901618231\\
18.4	-0.129593860493\\
18.6	-0.132415663548\\
18.8	-0.135050627405\\
19	-0.137374239213\\
19.2	-0.139389836325\\
19.4	-0.141458405927\\
19.6	-0.144155283373\\
19.8	-0.147664539232\\
20	-0.151692206319\\
20.2	-0.156213322091\\
20.4	-0.161497606625\\
20.6	-0.16719614504\\
20.8	-0.172184423308\\
21	-0.175967633524\\
21.2	-0.17915907779\\
21.4	-0.182901964826\\
21.6	-0.187853296166\\
21.8	-0.192927079096\\
22	-0.196104841181\\
22.2	-0.196820093106\\
22.4	-0.196518549154\\
22.6	-0.196227271984\\
22.8	-0.196057308375\\
23	-0.196052116157\\
23.1999999999999	-0.196052116157\\
23.3999999999999	-0.196052116157\\
23.5999999999999	-0.196052116157\\
23.7999999999999	-0.196052116157\\
23.9999999999999	-0.196052116157\\
};
\addlegendentry{$y_c$};

\addplot [color=black,dash pattern=on 1pt off 3pt on 3pt off 3pt,line width=2.0pt]
  table[row sep=crcr]{%
0.2	-0.0740638202575\\
0.4	-0.0740638202575\\
0.6	-0.0740638202575\\
0.8	-0.0740638202575\\
1	-0.0740638202575\\
1.2	-0.0740638202575\\
1.4	-0.0740638202575\\
1.6	-0.0740638202575\\
1.8	-0.0740638202575\\
2	-0.0740638202575\\
2.2	-0.0740638202575\\
2.4	-0.0740638202575\\
2.6	-0.0740638202575\\
2.8	-0.0740638202575\\
3	-0.0740638202575\\
3.2	-0.0740638202575\\
3.4	-0.0740638202575\\
3.6	-0.0740638202575\\
3.8	-0.0740638202575\\
4	-0.0740638202575\\
4.2	-0.0740638202575\\
4.4	-0.0740638202575\\
4.6	-0.0740638202575\\
4.8	-0.0740638202575\\
5	-0.0740638202575\\
5.2	-0.0740638202575\\
5.4	-0.0740638202575\\
5.6	-0.0740638202575\\
5.8	-0.0740638202575\\
6	-0.0740638202575\\
6.2	-0.0740638202575\\
6.4	-0.0740638202575\\
6.6	-0.0740638202575\\
6.8	-0.0740638202575\\
7	-0.0740638202575\\
7.2	-0.0740638202575\\
7.4	-0.0740488153114\\
7.6	-0.0760171772099\\
7.8	-0.0794014749521\\
8	-0.0831278021661\\
8.2	-0.0861287084614\\
8.4	-0.0891579114316\\
8.6	-0.0926196609802\\
8.8	-0.0963498745273\\
9	-0.100216535778\\
9.2	-0.104030578949\\
9.4	-0.107225425823\\
9.6	-0.109627724098\\
9.8	-0.111648788353\\
10	-0.114529493813\\
10.2	-0.118906443975\\
10.4	-0.123839553152\\
10.6	-0.128385412733\\
10.8	-0.132091928028\\
11	-0.135179369711\\
11.2	-0.137662557955\\
11.4	-0.139741664493\\
11.6	-0.141945714305\\
11.8	-0.144862516651\\
12	-0.148570770257\\
12.2	-0.152816481036\\
12.4	-0.15778268205\\
12.6	-0.163636448747\\
12.8	-0.169563860178\\
13	-0.174199034321\\
13.2	-0.177800959332\\
13.4	-0.181292767015\\
13.6	-0.185974125046\\
13.8	-0.191338201498\\
14	-0.195405432388\\
14.2	-0.196788199155\\
14.4	-0.196624045169\\
14.6	-0.196289215324\\
14.8	-0.196087629589\\
15	-0.195963898045\\
15.2	-0.195884827668\\
15.4	-0.195834310149\\
15.6	-0.19580213519\\
15.8	-0.195781860955\\
16	-0.195769175973\\
16.2	-0.195761234815\\
16.4	-0.195756298859\\
16.6	-0.195753249563\\
16.8	-0.195751373211\\
17	-0.195750247845\\
17.2	-0.195749532995\\
17.4	-0.195749140524\\
17.6	-0.195748890208\\
17.8	-0.195748736397\\
18	-0.195748679718\\
18.2	-0.195748630271\\
18.4	-0.195748588287\\
18.6	-0.195748580825\\
18.8	-0.195748577059\\
19	-0.195748586396\\
19.2	-0.195748582698\\
19.4	-0.195748579001\\
19.6	-0.195748592035\\
19.8	-0.195748592035\\
20	-0.195748592035\\
20.2	-0.195748592035\\
20.4	-0.195748588337\\
20.6	-0.195748588337\\
20.8	-0.195748588337\\
21	-0.195748588337\\
21.2	-0.195748588337\\
21.4	-0.195748588337\\
21.6	-0.195748588337\\
21.8	-0.195748588337\\
22	-0.195748588337\\
22.2	-0.195748588337\\
22.4	-0.195748588337\\
22.6	-0.195748588337\\
22.8	-0.195748588337\\
23	-0.195748588337\\
23.1999999999999	-0.195748588337\\
23.3999999999999	-0.195748588337\\
23.5999999999999	-0.195748588337\\
23.7999999999999	-0.195748588337\\
23.9999999999999	-0.195748588337\\
};
\addlegendentry{$y_u$};

\end{axis}

\begin{axis}[%
width=\figurewidth,
height=\figureheight,
scale only axis,
xmin=0,
xmax=25,
xlabel={Time [s]},
ymin=0,
ymax=3,
ylabel absolute,
ylabel={$\Vert \ddot{y}_{r} \Vert$ [m/s$^2$]},
at=(plot3.below south west),
yticklabel style={/pgf/number format/fixed},
anchor=above north west
]
\addplot [color=blue,solid,line width=2.0pt,forget plot]
  table[row sep=crcr]{%
0.2	0\\
0.4	0\\
0.6	0\\
0.8	0\\
1	0\\
1.2	0\\
1.4	0\\
1.6	0\\
1.8	0\\
2	0\\
2.2	0\\
2.4	0\\
2.6	0\\
2.8	0\\
3	0\\
3.2	0\\
3.4	0\\
3.6	0\\
3.8	0\\
4	0\\
4.2	0\\
4.4	0\\
4.6	0\\
4.8	0\\
5	0\\
5.2	0\\
5.4	0\\
5.6	0\\
5.8	0\\
6	0\\
6.2	0\\
6.4	0\\
6.6	0\\
6.8	0\\
7	0\\
7.2	0.146077526283755\\
7.4	0.0901460467402637\\
7.6	0.093393065331675\\
7.8	0.0496770406112752\\
8	0.0778510123860639\\
8.2	0.0189451512464034\\
8.4	0.0185761307065277\\
8.6	0.0508789470058169\\
8.8	0.101578091256884\\
9	0.0578751522829507\\
9.2	0.0320129463983265\\
9.4	0.0738245928538743\\
9.6	0.0688275320091251\\
9.8	0.0727346115740953\\
10	0.053695918352079\\
10.2	0.0453949451997085\\
10.4	0.904544852549868\\
10.6	1.89475839107848\\
10.8	2.12156483484072\\
11	2.19287520387428\\
11.2	2.22519974796172\\
11.4	2.22971763023413\\
11.6	2.17101028079293\\
11.8	2.18829926004947\\
12	2.21829121934383\\
12.2	2.1742795216364\\
12.4	2.16910693125958\\
12.6	2.12585508639064\\
12.8	2.16664477388921\\
13	2.20411342484425\\
13.2	2.1056450489284\\
13.4	2.1578365103339\\
13.6	2.17902712879975\\
13.8	2.15512320752721\\
14	2.09491165161077\\
14.2	2.15720007651306\\
14.4	2.20030939098548\\
14.6	2.18464777449725\\
14.8	2.20146983255182\\
15	2.11451919084196\\
15.2	2.17046941932001\\
15.4	2.29805326415475\\
15.6	2.17367575184058\\
15.8	2.07524990198352\\
16	2.1524904243853\\
16.2	2.13237223242563\\
16.4	1.82405651837464\\
16.6	0.714832655902606\\
16.8	0.297439699841428\\
17	0.254726900766686\\
17.2	0.162349416736509\\
17.4	0.0436137510678795\\
17.6	0.0841006128762306\\
17.8	0.140340255235677\\
18	0.193059788150322\\
18.2	0.0584768866050502\\
18.4	0.0165492970243241\\
18.6	0.0546637973774661\\
18.8	0.0413363965867126\\
19	0.027352360730871\\
19.2	0.0444643716008912\\
19.4	0.0425782968654029\\
19.6	0.110384781687205\\
19.8	0.111217902728746\\
20	0.0488388679086696\\
20.2	0.1307465366424\\
20.4	0.0616687886702258\\
20.6	0.0647758738565544\\
20.8	0.0604523054251046\\
21	0.0366540000925446\\
21.2	0.0530158075411805\\
21.4	0.0702406781908705\\
21.6	0.0196705793196261\\
21.8	0.0807526379845806\\
22	0.0883241875292854\\
22.2	0.068676510952523\\
22.4	0.0271458311810752\\
22.6	0.0427802658522429\\
22.8	0.0125411808338883\\
23	0\\
23.1999999999999	0\\
23.3999999999999	0\\
23.5999999999999	0\\
23.7999999999999	0\\
23.9999999999999	0\\
};
\end{axis}

\draw [->] (6,5.8) -- (4.7,5.55);

\end{tikzpicture}%

%% file: chapters/paper3/paper3.tex
       \paper[Detection of Contact Force Transients...]{Detection of Contact Force Transients during Robotic Assembly without a Force Sensor}
    \authors{Martin Karlsson \and Anders Robertsson \and Rolf Johansson}
\begin{abstract}
In this research, robot joint torques are used to recognize contact force transients induced by snap-fit assembly, thus detecting when the task is completed. The approach does not assume any external sensor, which is a benefit compared to the state of the art. The joint torque data is used as input to a recurrent neural network (RNN), and the output of the RNN indicates whether a snap-fit has occurred or not. A real-time application for snap-fit detection is developed, and verified experimentally on an industrial robot.
\end{abstract}
    \vfill
    Manuscript prepared for submission to review for publication.
    \newpage

\section{Introduction}
In the context of robotic assembly, robots commonly make series of movements, and switch between these when certain criteria are fulfilled. Such criteria usually consist of thresholds on measured signals, \emph{e.g.}, positions and contact forces. 

In this paper, a method to detect snap-fits during robotic assembly is presented and evaluated. In \cite{stolt2015detection}, this was achieved by detecting contact force transients induced by the snap-fit, using a force/torque sensor. This detection reduced the assembly time, compared to using a force threshold. It also removed the necessity to determine any level of the force threshold, which would have required considerable engineering work and explicit programming of the robot. 

Here, we continue the work presented in \cite{stolt2015detection}, with the following extensions. In \cite{stolt2015detection}, a force/torque sensor was used to measure the contact force/torque. Such sensors and systems are usually expensive, with costs comparable to the robot itself. If attached to the wrist of the robot, it would introduce extra weight that the robot would have to lift and move. Further, some robot models do not support any seamless attachment of such sensors. If the force sensor would be attached to an object in the work space, \emph{e.g.}, a table, this would imply restrictions on where the assembly could take place. In this work, we use robot joint torque measurements for the detection, thus avoiding the requirement of a force/torque sensor. This introduces a new difficulty; due to friction in the robot joints, some information is lost when using joint torques compared to a force/torque sensor. In \cite{stolt2015detection}, a support vector machine (SVM) was used for classification, whereas in the present approach, a recurrent neural network (RNN), which is an artificial neural network specialized in processing sequential data, was used. The RNN was implemented and trained using TensorFlow \cite{abadi2016tensorflow,tensorflowweb}, a software library for numerical computation.

Machine learning for analyzing contact forces in robotic assembly was also applied in \cite{rodriguez2010failure}, where force measurements were used as input to an SVM, to distinguish between successful and failed assemblies. A verification system, specialized in snap-fit assembly, was developed in \cite{rojas2012relative}. Similar to \cite{stolt2015detection} and \cite{rodriguez2010failure}, a force/torque sensor was assumed in \cite{rojas2012relative}. Such a requirement has been avoided in some previous research, by using internal robot sensors instead. For instance, a method to estimate contact forces from joint torques was presented in \cite{linderoth2013robotic}. Further, force controlled assembly without a force sensor was achieved in \cite{stolt2012force}, by estimating contact forces from position errors in the internal controller of the robot.

\subsection{Problem formulation}
In this work, we address the question of whether robot joint torques could be used to recognize contact force transients during robotic assembly, despite uncertainties introduced by, \emph{e.g.}, joint friction. Further, we investigate how long parts of the transients that should be included as input for the detection algorithm, in order to distinguish whether a transient is present or not.


\section{Method}
\label{sec:method}
\begin{figure}
	\includegraphics[width=\columnwidth ]{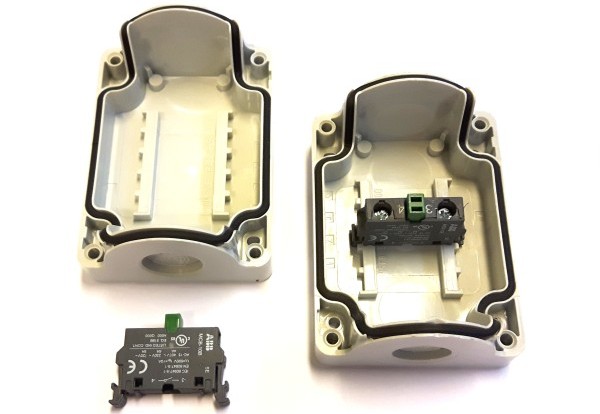}
	\caption{The parts to be assembled. Box (upper left), switch (lower left), and switch attached to box (right).}\label{fig:switch_parts}
\end{figure}

The snap-fit scenario considered here consisted of attaching a switch to a box, see \cref{fig:switch_parts}. The objective of the robot was to move toward the box while holding the switch, thus pushing the switch against the box, until it snapped into place. The robot should detect the snap-fit automatically, stop moving toward the box, and possibly start a new movement.

\subsection{Sequence model}
\label{sec:sequencemodel}
An RNN \cite{deeplearningbook,graves2012neural} was used as a sequence classifier. This choice is discussed in \cref{sec:disc}. It had a sequence of joint torques as input, one single output indicating whether the sequence contained a snap-fit or not, one hidden layer, and recurrent connections between its hidden neurons. Each input torque sequence consisted of $T = n_{\text{pre}}+1+n_{\text{post}}$ time samples, where $n_{\text{pre}}$ and $n_{\text{post}}$ were determined as explained in \cref{sec:modeltraining}. In turn, each time sample consisted of $n_{\text{ch}}=7$ channels; one per robot joint. The dimension of the hidden layer was chosen to be the same as the number of input channels, $n_{\text{ch}}$.

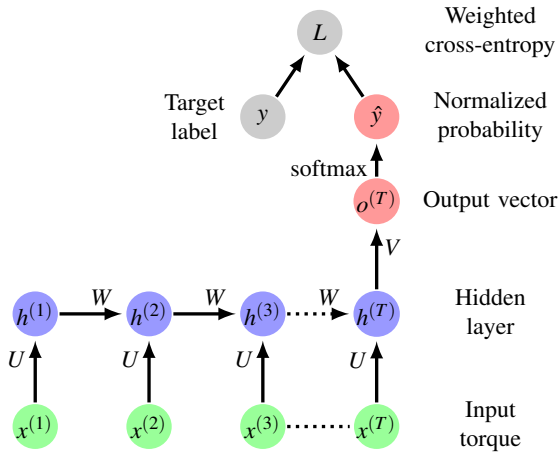
\begin{figure}
	\centering	
	\footnotesize
	\input{chapters/paper3/figs/rnn_sketch.tex}
	\caption{The RNN visualized as an unfolded computational graph, where each node is associated with a certain time step. The biases $b$ and $c$, as well as the activation function tanh($\cdot$), are omitted for a clearer view, but the computations are detailed in \cref{eq:h1,eq:ht,eq:oT,eq:yhat}. The input torque was used to determine the hidden state, which was updated each time step. The last hidden state was used to determine the normalized probability $\hat{y}$ of whether a snap-fit was present in the time sequence or not.} \label{fig:rnn_sketch}
\end{figure}

Denote by $h^{(t)}$ the activation of the hidden units at time step $t$. The activation was defined recursively as 
\begin{align}
\label{eq:h1}
h^{(1)} &= \tanh(b + U x^{(1)}) \\
\label{eq:ht}
h^{(t)} &= \tanh(b + Wh^{(t-1)} + U x^{(t)}) \phantom{space} t \in [2;T]
\end{align}
where $U$ and $W$ are weight matrices, both of size $n_{\text{ch}} \times n_{\text{ch}}$, $b$ is a bias vector with dimension $n_{\text{ch}}$, and $x^{(t)}$ is the input at time $t$. Further, tanh($\cdot$) represents the hyperbolic tangent function. After reading an entire input sequence, the RNN produced one output $o^{(T)}$ given by
\begin{equation}
o^{(T)} = c + Vh^{(T)} \label{eq:oT}
\end{equation}
where $V$ is a weight matrix of size $2 \times n_{\text{ch}}$, and $c$ is a bias vector with dimension 2.
Finally, the softmax operation was applied to generate $\hat{y}$, a vector that represented the normalized probabilities of the output elements.
\begin{equation}
\label{eq:yhat}
\hat{y} = \left[\frac{e^{o_1^{(T)}}}{e^{o_1^{(T)}}+e^{o_2^{(T)}}} \phantom{d} \frac{e^{o_2^{(T)}}}{e^{o_1^{(T)}}+e^{o_2^{(T)}}} \right]^T
\end{equation}
Here, $o_i^{(T)}$ represents the $i$:th element of the output vector. If the first element of $\hat{y}$ was larger than the second, or equivalently, larger than 0.5, the data point was classified as positive, \emph{i.e.}, it was indicated that a snap-fit occurred within the sequence. Vice versa, if the first element was less than than 0.5, the data point was classified as negative, meaning that no snap-fit was present. The RNN architecture is visualized in \cref{fig:rnn_sketch}.

\subsection{Gathering of training data and test data}
\label{sec:get_data}
\begin{figure}[p]
\centering
\begin{minipage}{\textwidth}
\centering
\label{fig:whole_switch}
\includegraphics[width=\textwidth]{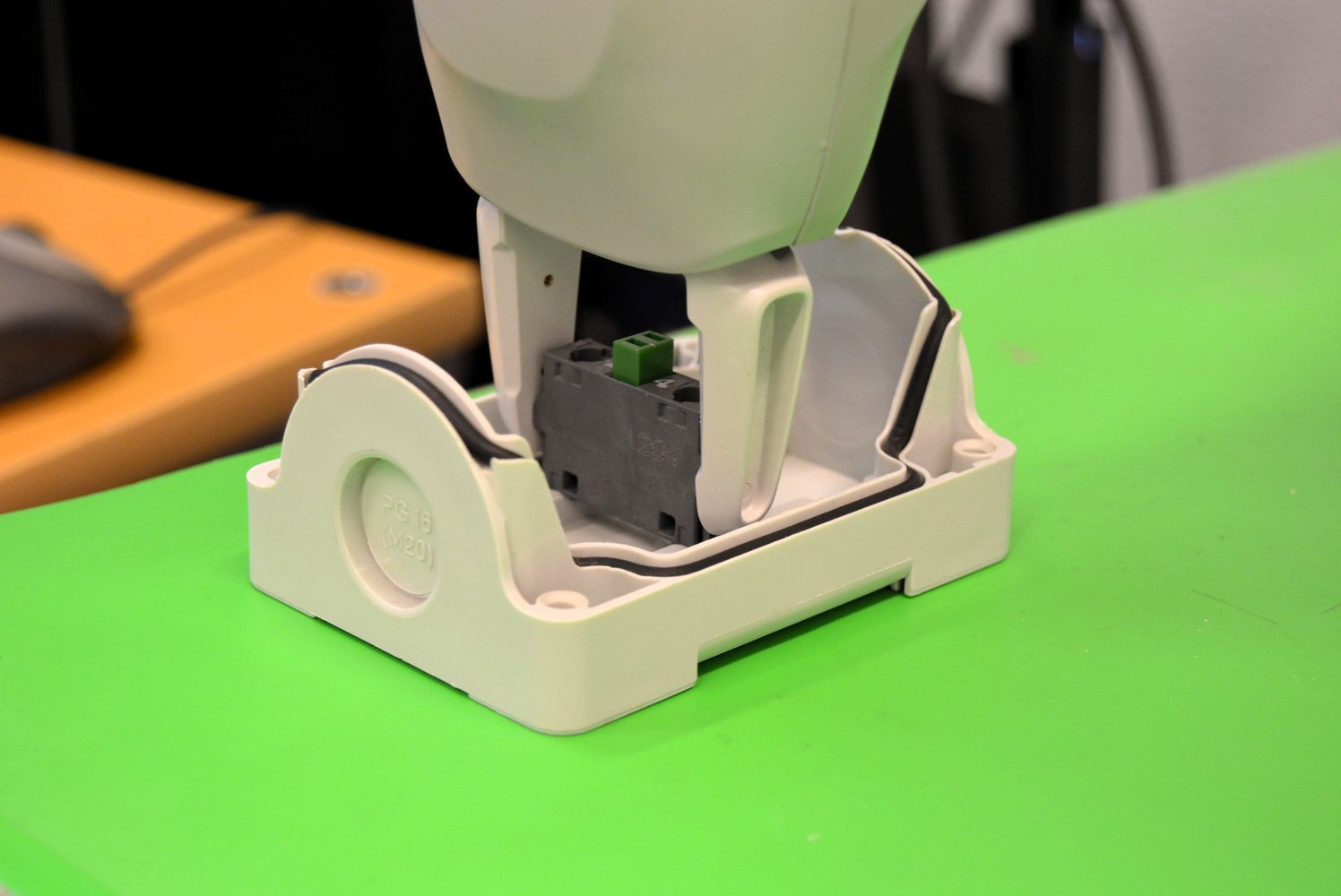}
\subcaption{}
\end{minipage}
\par\vspace{3mm}
\begin{minipage}{.48\textwidth}
\label{fig:switch_start}
\centering
\includegraphics[width=\textwidth]{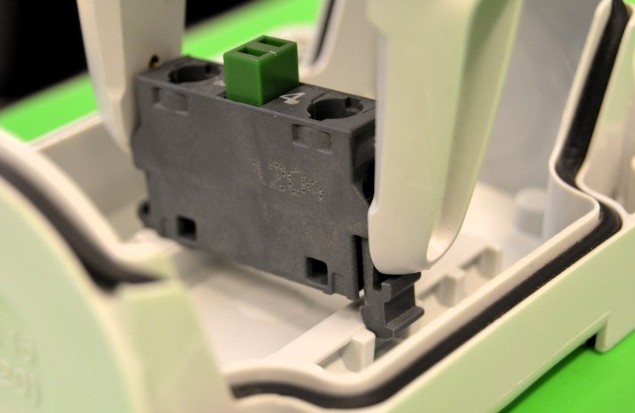}
\subcaption{}
\end{minipage}\hfill
\begin{minipage}{.48\textwidth}
\label{fig:switch_stop}
\centering
\includegraphics[width=\textwidth]{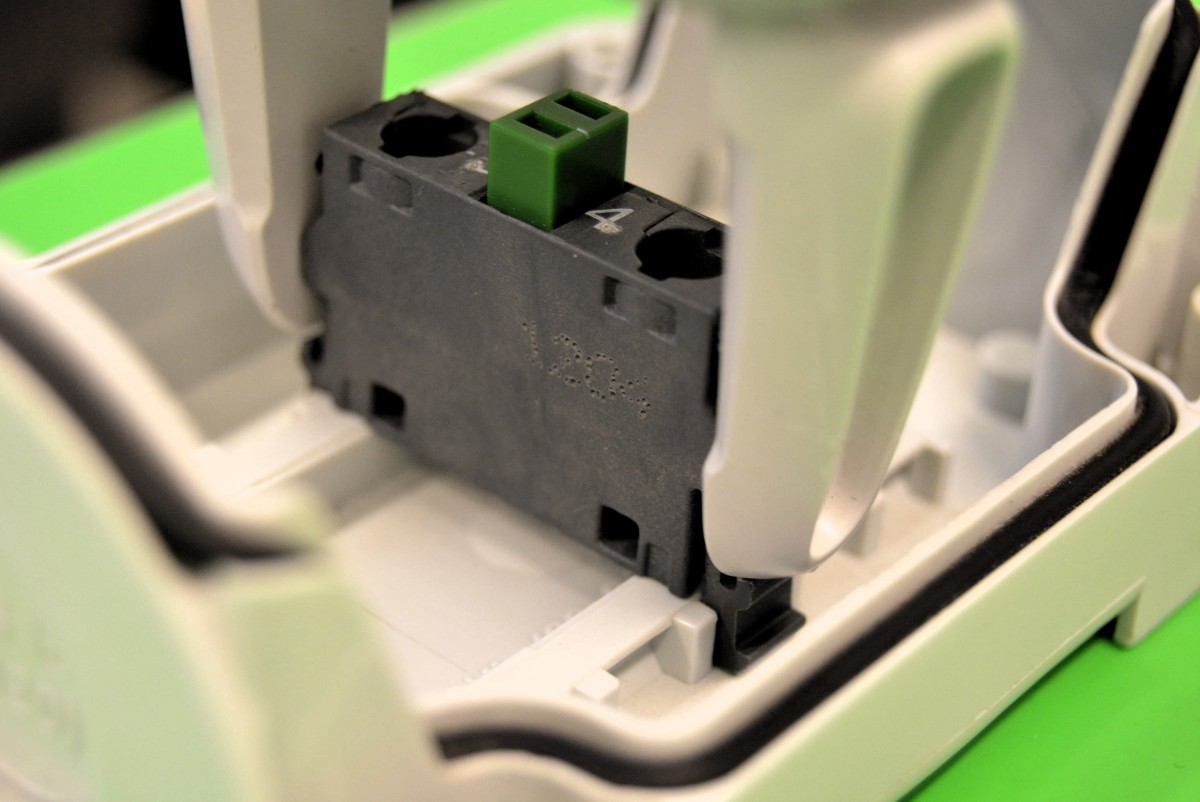}
\subcaption{}
\end{minipage}
\caption{Experimental setup. An overview is shown in (a). The ABB YuMi \cite{yumi} robot was used to grasp the switch, and attaching it to the box by pushing downwards. The downward motion began in (b), where the switch was not yet snapped into place. It ended when the snap-fit assembly was complete (c). These photos were taken during the experimental evaluation (see \cref{sec:exp}), and the same setup was used for gathering training and test data (see \cref{sec:get_data}).}
\label{fig:snap_switch}
\end{figure}

Training data and test data were obtained as follows. The right arm of the robot was used to grasp the switch, just above the box, as shown in \cref{fig:snap_switch}. Thereafter, a reference velocity was sent to the internal controller of the robot, causing the robot gripper to move toward the box at \SI{1.5}{mm/s}, thus pushing the switch against the box. Once the switch was snapped into place, the robot was stopped manually by the robot operator. The robot joint torques were recorded in 250 Hz through the ABB research interface EGMRI. The torque transient, induced by the snap-fit, was labeled manually, and used to form a positive data point. This procedure was repeated $N=50$ times, which yielded 50 positive data points. Data prior to each transient was used to form negative data points.

Given $n_{\text{pre}}$ and $n_{\text{post}}$, a positive data point was formed by extracting a torque sequence, from $n_{\text{pre}}$ samples previous to the peak value of the transient (inclusive), to $n_{\text{post}}$ samples after (inclusive). Negative data points, with the same sequence length $T$ as the positive ones, were extracted from torque measurements that ranged from a couple of seconds before the snap-fit, until the positive data point (exclusive). The negative data points were chosen so that overlap was avoided. For each data point, the target was labeled as a two-dimensional one-hot vector $y$, where $y=[1 \phantom{d} 0]^T$ represented a positive data point, and $y=[0 \phantom{d} 1]^T$ represented a negative one.

Note that with the approach above, it was possible to extract several negative data points, but only one positive data point, for every snap-fit experienced by the robot. 

Half of the positive and negative data points were used in the training set, and the other half was used in the test set.

\subsection{Model training}
\label{sec:modeltraining}
Given the training set, the model parameters $U, V, W, b$, and $c$ were determined by minimizing a loss function $L$.

The training set contained much more negative data points than positive ones. If not taken into account, this type of class imbalance has been reported to obstruct the training procedure of several different classifiers. The phenomenon has been described in more detail in \cite{japkowicz2002class,japkowicz2000class}, and should be taken into account when designing the loss function. Consider first the following loss function $\bar{L}$, which is the ordinary cross-entropy between training data and model predictions, averaged over the training examples.
\begin{equation}
\label{eq:lossnoweight}
\bar{L} = - \frac{1}{D} \sum\limits_{d=1}^D \sum\limits_{a=1}^A y_a^d \log{\hat{y}_a^d}
\end{equation}
Here, $a$ and $d$ are indices for summing over the vector elements and training data points, respectively. This cross-entropy is commonly used as a loss function in machine learning \cite{deeplearningbook,rubinstein2013cross}. Due to the class imbalance in the present training set, it would be possible to yield a relatively low loss $\bar{L}$ by simply classifying all or most of the data points as negative, regardless of the input, even though that strategy would not be desirable. 

In order to take the class imbalance into account, weighted cross-entropy was used as loss function. Denote by $r$ the ratio between negative and positive data points in the training set, and introduce the weight vector $w_r = [r \phantom{d} 1]^T$. The loss function was defined as 
\begin{equation}
\label{eq:weightedloss}
L = - \frac{1}{D} \sum\limits_{d=1}^D \sum\limits_{a=1}^A y_a^d \log{\hat{y}_a^d} \cdot w_r^T y^d
\end{equation}

The RNN in \cref{sec:sequencemodel} was implemented as a computational graph in the Julia programming language \cite{BEKS14}, using TensorFlow \cite{abadi2016tensorflow,tensorflowweb} and the wrapper TensorFlow.jl \cite{tfjl}. The Adam algorithm \cite{kingma2014adam} was used for minimization of the loss function $L$.

The values of $n_{\text{pre}}$ and $n_{\text{post}}$ were determined using both the training set and the test set as follows. All positive data points available were used, and $r=20$ times as many negative data points. Starting with $n_{\text{pre}}=n_{\text{post}}=1$, the model was trained using the training set, and its performance was measured using the test set. Subsequently, both $n_{\text{pre}}$ and $n_{\text{post}}$ were increased by 1, and the training and evaluation procedure was repeated. This continued until perfect classification was achieved, or until the values of $n_{\text{pre}}$ and $n_{\text{post}}$ were large. (30 was chosen as an upper limit, though it was never reached in the experiments presented here.) Thereafter, $n_{\text{pre}}$ was kept constant, and it was investigated how much $n_{\text{post}}$ could be lowered with retained performance. This was done by decreasing $n_{\text{post}}$ one step at a time, while repeating the training and evaluation procedure for each value. Once the performance was decreased, the value just above that was chosen for $n_{\text{post}}$. This way, the lowest possible value of $n_{\text{post}}$ was found, that resulted in retained performance.

Once $n_{\text{pre}}$ and $n_{\text{post}}$ were determined, new model parameters were obtained by training on a larger data set, with $r=100$. The reason for using a lower value for the other iterations, was that it took significantly longer computation time to use such a large data set.

Due to the class imbalance in the test set, ordinary classification accuracy, as defined by the number of correctly classified test data points divided by the total number of test data points, would not be a good model performance measurement. Instead, the F-measurement \cite{hripcsak2005agreement} was used, defined as 
\begin{equation}
F_1 = 2\frac{PR}{P+R}
\end{equation}
where $P$ is the precision, \emph{i.e.}, the number of correctly classified positive data points divided by the number of all data points classified as positive by the model, and $R$ is the recall, \emph{i.e.}, the number of correctly classified positive data points divided by the number of all data points that were truly positive. The value of $F_1$ ranges from 0 to 1, where 1 indicates perfect classification.

\subsection{Implementation of real-time application}
\label{sec:implementation}
\begin{figure}
\includegraphics[width=\columnwidth]{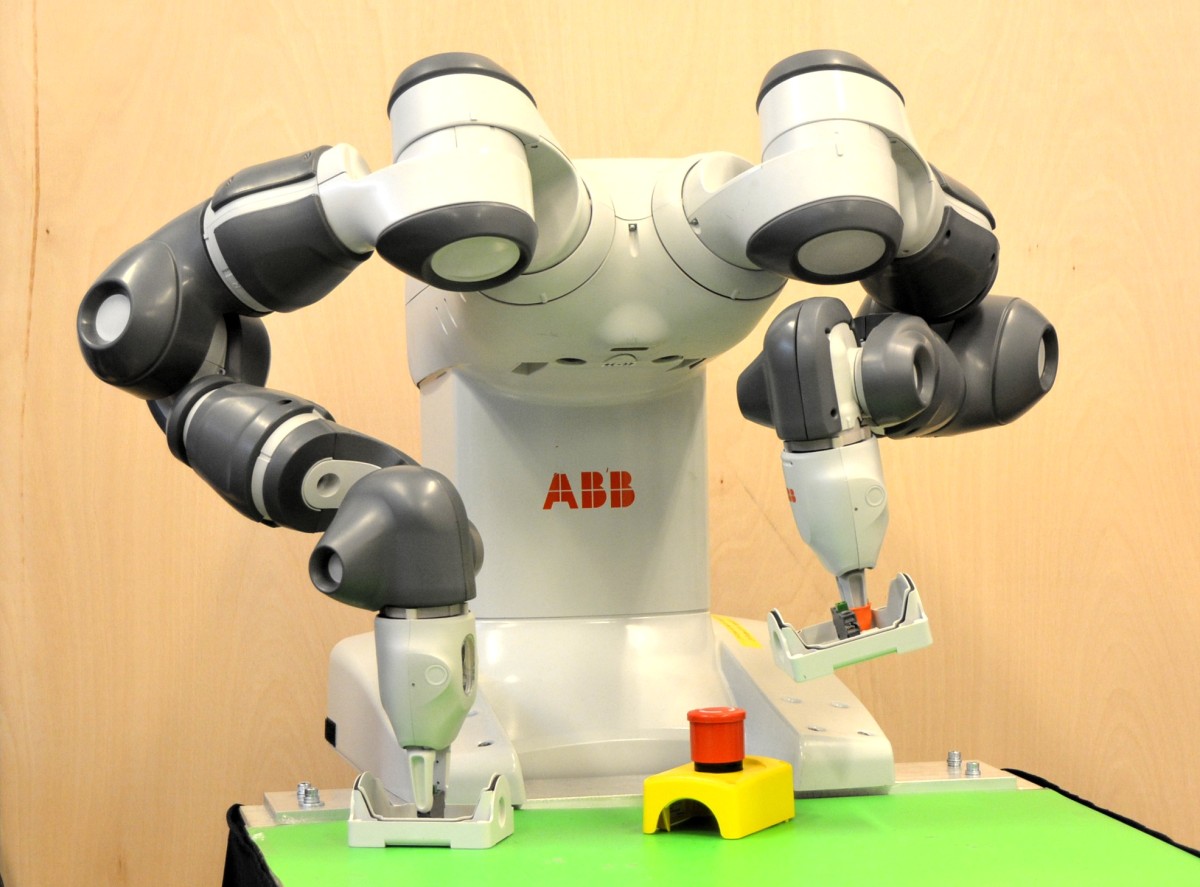}
\caption{The ABB YuMi robot \cite{yumi} used in the experiments.}\label{fig:yumiok}
\end{figure}

The ABB YuMi robot \cite{yumi} was used for experimental evaluation. The robot is shown in \cref{fig:yumiok}.
The RNN model obtained according to \cref{sec:sequencemodel,sec:get_data,sec:modeltraining} was saved on a server, and loaded into a Julia program on a PC, which communicated with the internal controller of the robot through the LabComm protocol \cite{labcomm}. The sample frequency was \SI{250}{Hz}. The robot joint torques were logged and saved, and for each time sample $t$, a joint torque sequence was formed by the samples in $[t-n_{\text{pre}}-n_{\text{post}};t]$, and sent as input to the RNN. Measurements after time $t$ were not available at $t$, which is why the input only contained samples up until $t$. Each sequence was classified in real-time. The computation time for one classification was short; well below the sample period. To move the robot, desired velocity references for the gripper in Cartesian space were first specified in the Julia program. Then, the corresponding joint velocities were computed using the robot Jacobian, and these were sent as references to the internal controller of the robot.

\section{Experiments}
\label{sec:exp}
Since the test set was used to determine the hyper parameters of the RNN, \emph{i.e.}, $n_{\text{pre}}$ and $n_{\text{post}}$, it was necessary to gather new measurements to evaluate the general performance of the RNN. The experimental setup was similar to that in \cref{sec:get_data}, except that the measured torque sequences were saved and classified by the RNN, instead of just saved to the training and test sets. The implementation in \cref{sec:implementation} was used for robot control and snap-fit detection. The robot was programmed to first move its gripper down, thus pushing the switch against the box. Once a snap-fit was detected, it was programmed to stop its downward motion, and instead move the box to the side. The snap-fit assembly was repeated 50 times, to evaluate the robustness of the proposed approach. The experimental setup is visualized in \cref{fig:snap_switch}.

\section{Results}
The performance of the RNN on the test set, for different values of the hyper parameters, is shown in \cref{table:performance}. The abbreviations are as follows: number of true positives (TP), true negatives (TN), false positives (FP), and false negatives (FN).
The hyper parameters were increased until $n_{\text{pre}}~=~n_{\text{post}}~=~5$, for which perfect classification was obtained. Then, $n_{\text{post}}$ was decreased until the performance decreased at $n_{\text{post}}=2$. With this value of $n_{\text{post}}$, larger values of $n_{\text{pre}}$ were tested (see second last row in \cref{table:performance}), which did not yield perfect classification for any values, \emph{i.e.}, $F_1 < 1$. Thus, $(n_{\text{pre}},n_{\text{post}}) = (5,3)$ was chosen for the final model.

\begin{table}
\begin{center}
\caption{RNN performance on the test set, for different values of the hyper parameters. The row with the lowest value of $n_{\text{post}}$ that yielded perfect classification is marked in {\color{blue} \textbf{blue}}. With these values, the model was trained and tested again, but now with more negative data points (see last row, marked~in~{\color{red} \textbf{red}}).} \label{table:performance}
\begin{tabular}{c c | c c c c l l l}$n_{\text{pre}}$ & $n_{\text{post}}$ & TP & TN & FP & FN & $P$ & $R$ & $F_1$  \\
\hline
1 & 1 & 20 & 500 & 0 & 5 & 1 & 0.80 & 0.89\\
2 & 2 & 22 & 500 & 0 & 3 & 1 & 0.88 & 0.94\\
3 & 3 & 23 & 498 & 2 & 2 & 0.92 & 0.92 & 0.92\\
4 & 4 & 23 & 500 & 0 & 2 & 1 & 0.92 & 0.96\\
5 & 5 & 25 & 500 & 0 & 0 & 1 & 1 & 1\\
5 & 4 & 25 & 500 & 0 & 0 & 1 & 1 & 1\\
{\color{blue} \textbf{5}} & {\color{blue} \textbf{3}} & {\color{blue} \textbf{25}} & {\color{blue} \textbf{500}} & {\color{blue} \textbf{0}} & {\color{blue} \textbf{0}} & {\color{blue} \textbf{1}} & {\color{blue} \textbf{1}} & {\color{blue} \textbf{1}}\\
5 & 2 & 22 & 500 & 0 & 3 & 1 & 0.88 & 0.94\\
$[6;30]$ & 2 & - & - & - & - & - & - & <1 \\
{\color{red} \textbf{5}} & {\color{red} \textbf{3}} & {\color{red} \textbf{25}} & {\color{red} \textbf{2500}} & {\color{red} \textbf{0}} & {\color{red} \textbf{0}} & {\color{red} \textbf{1}} & {\color{red} \textbf{1}} & {\color{red} \textbf{1}}\\
\hline
\end{tabular}
\end{center}
\end{table}

\begin{figure}
	\centering	
	\footnotesize
	\input{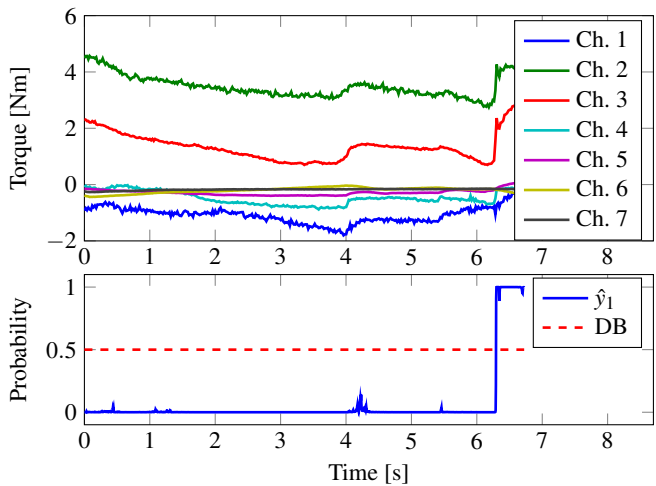}
	\caption{Data from one of the experiments. The robot joint torques (upper plot) were used as input for the RNN. These are represented by one channel (Ch.) per joint. The first element of the RNN output ($\hat{y}_1$ in the lower plot) was close to 0 before the snap-fit occurred, and increased to close to 1 at the time of the snap-fit. A snap-fit was indicated when $\hat{y}_1$ was above the decision boundary (DB, at $0.5$) for the first time. Thus, for detection purposes, the RNN output generated after this event was not relevant.} \label{fig:snap_prob}
\end{figure}

\begin{figure}
	\centering	
	\footnotesize
	\input{chapters/paper3/figs/snap_prob_zoom.tex}
	\caption{Same data as in Fig. \ref{fig:snap_prob}, but zoomed-in on the time of the snap-fit. The legend of the upper plot is absent for better visualization, but can be found in \cref{fig:snap_prob}. The snap-fit was detected at time $t=$ \SI{6.308}{s}. The first torque sequence to be classified as positive was that within the vertical dashed lines in the upper plot.} \label{fig:snap_prob_zoom}
\end{figure}
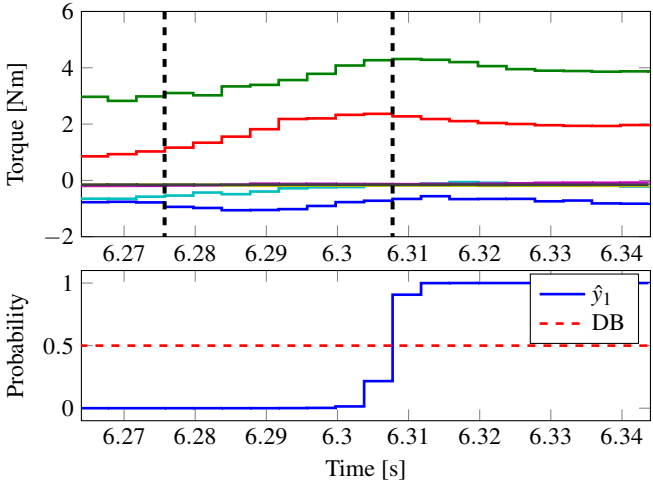

After training, the RNN detected all 50 snap-fits in the experiments in \cref{sec:exp} correctly, without any false positives prior to the snap. The torque data and RNN output from one of the trials are shown in \cref{fig:snap_prob,fig:snap_prob_zoom}. The other trials gave qualitatively similar results.

\section{Discussion}
\label{sec:disc}
There are several alternatives to RNN for classification. Using an RNN is motivated as follows. Two simpler models, matched filter and logistic regression, were tried initially, without achieving satisfactory performance on test data. RNNs are specialized in processing sequential data, and the torque measurements used in this work were sequential. Thanks to the parameter sharing of the RNN, it is possible to estimate a model with significantly fewer training examples, than would be needed without parameter sharing. Compared to models that are not specialized in sequential data, \emph{e.g.}, logit models, SVMs, and ordinary neural networks, the RNN is less sensitive to variations of the exact time step in which some information in the input sequence appears. An RNN can also be generalized to classify data points of sequence lengths not present in the training set, though this was not used in this present work. General properties of RNNs are well described in \cite{deeplearningbook}.

Compared to \cite{stolt2015detection}, the approach proposed here had three new benefits. A force sensor was no longer required, the detection delay was reduced, and the computation time for model training was shortened. In \cite{stolt2015detection}, $n_{\text{post}}>10$ (corresponding to > \SI{40}{ms}) was required for perfect classification, whereas our proposed method required $n_{\text{post}}=3$ (\SI{12}{ms}). The training time of the RNN was in the order of minutes on an ordinary PC, which was an improvement compared to days in \cite{stolt2015detection}.

The concept of weighted loss to compensate for class imbalance in machine learning has been evaluated in \cite{japkowicz2002class}, and successfully applied to a deep neural network in \cite{panchapagesan2016multi}. Equation (\ref{eq:weightedloss}) extends \cref{eq:lossnoweight} by the factor $w_r^T y^d$, which evaluates to $r$ for positive data points, and to 1 for negative ones.  

In \cref{fig:snap_prob}, it should be noted that $\hat{y}_1$ raised significantly above 0 at $t \approx$ \SI{4.2}{s}, even though no snap-fit occurred at that time. Even though the values were still well below the decision boundary, this leaves room for improvement in terms of robustness of the detection.

Given a certain contact force/torque acting on the end-effector of the robot, the corresponding joint torques depend on the configuration, as well as gravity and friction. In order to generalize the detection to other robot states than the one used in the training, it would be a good idea to estimate the contact forces/torques, by first modeling the gravity and friction-induced torques, and subsequently compensating for these.

Whereas the evaluation of the model was done in real-time in this work, the RNN training was performed offline. It therefore remains as future work to create a user interface that allows for an operator to gather training data and test data, label them, and run the training procedure.

It is also important to shorten the reaction time of the robot, \emph{i.e.}, to further reduce the value of $n_{\text{post}}$. This could be done by including more sensors. For instance, the snap-fit assembly generates a sound, easily recognized by a human. Adding a microphone to the current setup, would therefore add information to the detection approach. In turn, this could be used to decrease the required amount of training data, improve robustness of the detection, or detect the snap-fit earlier.

\section{Conclusion}
In this work, we have addressed the question of whether robot joint torque measurements, together with an RNN model, could be used for detection of snap-fits during assembly. First, training and test data were gathered and labeled. Then, these were used to determine the model parameters. Finally, a real-time application for snap-fit detection was implemented and tested. The method presented seems promising, since the resulting model had high performance, both on the test data and during the experiments.

\section*{Acknowledgments}
The authors would like to thank Maj~Stenmark and Jacek~Malec at Computer Science, Lund University, for valuable discussions throughout this work. Fredrik~Bagge Carlson at Dept. Automatic Control, Lund University, and Mathias Haage at Computer Science, Lund University, are gratefully acknowledged both for valuable discussions, and for the development of a bridge to the research interface EGMRI. The authors are members of the LCCC Linnaeus Center and the ELLIIT Excellence Center at Lund University. The research leading to these results has received funding from the European Commission's Framework Programme Horizon 2020 – under grant agreement No 644938 – SARAFun.


%% file: chapters/paper3/figs/rnn_sketch.tex
\def\layersep{1.5cm}
\def\neuronsep{1.5cm}

\tikzset{every picture/.style={black,auto, line width=1.25pt,font=\small}}
\begin{tikzpicture}[shorten >=1pt,draw=black!100, node distance=\layersep]
    \tikzstyle{neuron}=[circle,fill=black!25,minimum size=17pt,inner sep=0pt]
    \tikzstyle{input neuron}=[neuron, fill=green!40];
    \tikzstyle{output neuron}=[neuron, fill=red!40];
    \tikzstyle{hidden neuron}=[neuron, fill=blue!40];
    \tikzstyle{gray neuron}=[neuron, fill=black!20];
    
    \node[input neuron](in1) at (\neuronsep*0,0) {$x^{(1)}$};
    \node[input neuron](in2) at (\neuronsep*1,0) {$x^{(2)}$};
    \node[input neuron](in3) at (\neuronsep*2,0) {$x^{(3)}$};
	\node[input neuron](inT) at (\neuronsep*3,0) {$x^{(T)}$};

    \node[hidden neuron](h1) at (\neuronsep*0,\layersep) {$h^{(1)}$}; 
    \node[hidden neuron](h2) at (\neuronsep*1,\layersep) {$h^{(2)}$}; 
    \node[hidden neuron](h3) at (\neuronsep*2,\layersep) {$h^{(3)}$};    
    \node[hidden neuron](hT) at (\neuronsep*3, \layersep) {$h^{(T)}$};  
   
   	\node[output neuron](output) at (\neuronsep*3, 2*\layersep) {$o^{(T)}$};
   
   	\node[gray neuron](y) at (\neuronsep*2, 2.75*\layersep) {$y$};
   	\node[gray neuron](L) at (\neuronsep*2.5, 3.5*\layersep) {$L$};

	\node[output neuron](yhat) at (\neuronsep*3, 2.75*\layersep) {$\hat{y}$};

	\node[text width=1cm,align=center] at (4*\neuronsep,0) {Input torque};
	\node[text width=1cm,align=center] at (4*\neuronsep,\layersep) {Hidden layer};
	\node[text width=2cm,align=center] at (4*\neuronsep,2*\layersep) {Output vector};
	\node[text width=2.2cm,align=center] at (4*\neuronsep,3.5*\layersep) {Weighted cross-entropy};
	\node[text width=2cm,align=center] at (4*\neuronsep,2.75*\layersep) {Normalized probability};
	\node[text width=1cm,align=center] at (1.4*\neuronsep,2.75*\layersep) {Target label};
	\node[text width=1cm] at (4.5*\neuronsep,2*\layersep) { };
	
	\draw[->](in1)--(h1) node [pos=0.66,left] {$U$};
	\draw[->](in2)--(h2) node [pos=0.66,left] {$U$};
	\draw[->](in3)--(h3) node [pos=0.66,left] {$U$};
	\draw[->](inT)--(hT) node [pos=0.66,left] {$U$};
	
	\draw[->](h1)--(h2) node [pos=0.66,above] {$W$};
	\draw[->](h2)--(h3) node [pos=0.66,above] {$W$};
	\draw[dotted,->](h3)--(hT) node [pos=0.66,above] {$W$};
	\draw[dotted](in3)--(inT);
	\draw[->](hT)--(output) node [pos=0.66,right] {$V$};
	\draw[->](output)--(yhat) node [pos=0.7,below left] {softmax};
	\draw[->](yhat)--(L);
	\draw[->](y)--(L);
\end{tikzpicture}

%% file: chapters/paper3/figs/snap_prob_zoom.tex
%
%
%
\definecolor{mycolor1}{rgb}{0.00000,0.75000,0.75000}%
\definecolor{mycolor2}{rgb}{0.75000,0.00000,0.75000}%
\definecolor{mycolor3}{rgb}{0.75000,0.75000,0.00000}%
\begin{tikzpicture}

\begin{axis}[%
width=7.5cm,
height=3cm,
scale only axis,
ylabel near ticks,
xmin=6.264,
xmax=6.344,
ymin=-2,
ymax=6,
ylabel={Torque [Nm]},
name=plot1
]
\addplot[const plot,color=blue,solid,line width=1.0pt] plot table[row sep=crcr] {%
6.26372619047619	-0.773805141448975\\
6.26772857142857	-0.759521782398224\\
6.27173095238095	-0.781263589859009\\
6.27573333333333	-0.93985778093338\\
6.27973571428571	-0.978891670703888\\
6.2837380952381	-1.05859398841858\\
6.28774047619048	-1.05173766613007\\
6.29174285714286	-1.02090716362\\
6.29574523809524	-0.90735787153244\\
6.29974761904762	-0.771898746490479\\
6.30375	-0.722419142723083\\
6.30775238095238	-0.65771073102951\\
6.31175476190476	-0.561058700084686\\
6.31575714285714	-0.664211869239807\\
6.31975952380952	-0.65760749578476\\
6.3237619047619	-0.655762791633606\\
6.32776428571429	-0.742476522922516\\
6.33176666666667	-0.715067207813263\\
6.33576904761905	-0.81693035364151\\
6.33977142857143	-0.827525079250336\\
6.34377380952381	-0.800156593322754\\
};
\addplot[const plot,color=black!50!green,solid,line width=1.0pt] plot table[row sep=crcr] {%
6.26372619047619	2.96814799308777\\
6.26772857142857	2.82442402839661\\
6.27173095238095	2.98232698440552\\
6.27573333333333	3.10258936882019\\
6.27973571428571	3.02174735069275\\
6.2837380952381	3.33793449401855\\
6.28774047619048	3.39263558387756\\
6.29174285714286	3.56330800056458\\
6.29574523809524	3.78508281707764\\
6.29974761904762	4.07972383499146\\
6.30375	4.26867866516113\\
6.30775238095238	4.31040859222412\\
6.31175476190476	4.28316354751587\\
6.31575714285714	4.20130205154419\\
6.31975952380952	4.05858659744263\\
6.3237619047619	3.95105409622192\\
6.32776428571429	3.89646029472351\\
6.33176666666667	3.8846161365509\\
6.33576904761905	3.86004734039307\\
6.33977142857143	3.87271523475647\\
6.34377380952381	3.82066369056702\\
};
\addplot[const plot,color=red,solid,line width=1.0pt] plot table[row sep=crcr] {%
6.26372619047619	0.854251265525818\\
6.26772857142857	0.931059658527374\\
6.27173095238095	1.02793836593628\\
6.27573333333333	1.16453921794891\\
6.27973571428571	1.33883380889893\\
6.2837380952381	1.55400109291077\\
6.28774047619048	1.81801819801331\\
6.29174285714286	2.18054580688477\\
6.29574523809524	2.20219445228577\\
6.29974761904762	2.32831835746765\\
6.30375	2.36407566070557\\
6.30775238095238	2.27380037307739\\
6.31175476190476	2.17960143089294\\
6.31575714285714	2.10553932189941\\
6.31975952380952	2.03736472129822\\
6.3237619047619	2.00290870666504\\
6.32776428571429	1.95963752269745\\
6.33176666666667	1.93734955787659\\
6.33576904761905	1.9333564043045\\
6.33977142857143	1.9662469625473\\
6.34377380952381	1.99195766448975\\
};
\addplot[const plot,color=mycolor1,solid,line width=1.0pt] plot table[row sep=crcr] {%
6.26372619047619	-0.649566769599915\\
6.26772857142857	-0.651495575904846\\
6.27173095238095	-0.577397882938385\\
6.27573333333333	-0.538933157920837\\
6.27973571428571	-0.431459665298462\\
6.2837380952381	-0.489994734525681\\
6.28774047619048	-0.395639896392822\\
6.29174285714286	-0.28188955783844\\
6.29574523809524	-0.244276389479637\\
6.29974761904762	-0.236580222845078\\
6.30375	-0.170659646391869\\
6.30775238095238	-0.148266837000847\\
6.31175476190476	-0.111814498901367\\
6.31575714285714	-0.0604249089956284\\
6.31975952380952	-0.0824102163314819\\
6.3237619047619	-0.148059397935867\\
6.32776428571429	-0.153522431850433\\
6.33176666666667	-0.170892313122749\\
6.33576904761905	-0.186384260654449\\
6.33977142857143	-0.215603068470955\\
6.34377380952381	-0.189010187983513\\
};
\addplot[const plot,color=mycolor2,solid,line width=1.0pt] plot table[row sep=crcr] {%
6.26372619047619	-0.195000886917114\\
6.26772857142857	-0.193147838115692\\
6.27173095238095	-0.19179305434227\\
6.27573333333333	-0.18011286854744\\
6.27973571428571	-0.172669529914856\\
6.2837380952381	-0.154123365879059\\
6.28774047619048	-0.117280334234238\\
6.29174285714286	-0.125051751732826\\
6.29574523809524	-0.121273264288902\\
6.29974761904762	-0.127290114760399\\
6.30375	-0.129498511552811\\
6.30775238095238	-0.133107334375381\\
6.31175476190476	-0.124629236757755\\
6.31575714285714	-0.126470640301704\\
6.31975952380952	-0.116976514458656\\
6.3237619047619	-0.104716628789902\\
6.32776428571429	-0.0857902988791466\\
6.33176666666667	-0.0828614085912704\\
6.33576904761905	-0.0850665792822838\\
6.33977142857143	-0.077022060751915\\
6.34377380952381	-0.0778784230351448\\
};
\addplot[const plot,color=mycolor3,solid,line width=1.0pt] plot table[row sep=crcr] {%
6.26372619047619	-0.155620619654655\\
6.26772857142857	-0.155827179551125\\
6.27173095238095	-0.151764646172523\\
6.27573333333333	-0.153197452425957\\
6.27973571428571	-0.153444573283195\\
6.2837380952381	-0.181484907865524\\
6.28774047619048	-0.190399080514908\\
6.29174285714286	-0.186281204223633\\
6.29574523809524	-0.180294543504715\\
6.29974761904762	-0.186567217111588\\
6.30375	-0.186667561531067\\
6.30775238095238	-0.181734979152679\\
6.31175476190476	-0.185511901974678\\
6.31575714285714	-0.186366453766823\\
6.31975952380952	-0.191400200128555\\
6.3237619047619	-0.19061504304409\\
6.32776428571429	-0.190224543213844\\
6.33176666666667	-0.191338092088699\\
6.33576904761905	-0.192955061793327\\
6.33977142857143	-0.193429738283157\\
6.34377380952381	-0.190633967518806\\
};
\addplot[const plot,color=darkgray,solid,line width=1.0pt] plot table[row sep=crcr] {%
6.26372619047619	-0.152681455016136\\
6.26772857142857	-0.15039786696434\\
6.27173095238095	-0.156771346926689\\
6.27573333333333	-0.15175524353981\\
6.27973571428571	-0.156984776258469\\
6.2837380952381	-0.156141504645348\\
6.28774047619048	-0.148967906832695\\
6.29174285714286	-0.14944264292717\\
6.29574523809524	-0.14915107190609\\
6.29974761904762	-0.153300985693932\\
6.30375	-0.151143401861191\\
6.30775238095238	-0.15379074215889\\
6.31175476190476	-0.154753863811493\\
6.31575714285714	-0.148626506328583\\
6.31975952380952	-0.159248113632202\\
6.3237619047619	-0.153877601027489\\
6.32776428571429	-0.153743654489517\\
6.33176666666667	-0.15414659678936\\
6.33576904761905	-0.151383846998215\\
6.33977142857143	-0.154748246073723\\
6.34377380952381	-0.155291348695755\\
};

\addplot[const plot,color=black,dashed,line width=1.5pt] plot table[row sep=crcr] {%
6.30775238095238 6 \\
6.30775238095238 -2\\ 
};

\addplot[const plot,color=black,dashed,line width=1.5pt] plot table[row sep=crcr] {%
6.2757 6 \\
6.2757 -2\\ 
};
\end{axis}

\begin{axis}[%
width=7.5cm,
height=2cm,
scale only axis,
ylabel near ticks,
xmin=6.264,
xmax=6.344,
xlabel={Time [s]},
ymin=-0.1,
ymax=1.1,
ylabel={Probability},
at=(plot1.below south west),
anchor=above north west,
legend style={draw=black,fill=white,legend cell align=left}
]
\addplot[const plot,color=blue,solid,line width=1.0pt] plot table[row sep=crcr] {%
6.26372619047619	3.7613731e-06\\
6.26772857142857	9.657439e-06\\
6.27173095238095	9.849772e-06\\
6.27573333333333	1.666575e-05\\
6.27973571428571	2.4238816e-05\\
6.2837380952381	7.5052994e-05\\
6.28774047619048	0.0001335324\\
6.29174285714286	0.0008052557\\
6.29574523809524	0.0023069242\\
6.29974761904762	0.014159188\\
6.30375	0.21624723\\
6.30775238095238	0.90674853\\
6.31175476190476	0.9993686\\
6.31575714285714	0.99999714\\
6.31975952380952	1\\
6.3237619047619	1\\
6.32776428571429	1\\
6.33176666666667	1\\
6.33576904761905	1\\
6.33977142857143	0.9999999\\
6.34377380952381	0.9999865\\
};
\addlegendentry{$\hat{y}_1$};

\addplot[const plot,color=red,dashed,line width=1.0pt] plot table[row sep=crcr] {%
6.26372619047619	0.5\\
6.26772857142857	0.5\\
6.27173095238095	0.5\\
6.27573333333333	0.5\\
6.27973571428571	0.5\\
6.2837380952381	0.5\\
6.28774047619048	0.5\\
6.29174285714286	0.5\\
6.29574523809524	0.5\\
6.29974761904762	0.5\\
6.30375	0.5\\
6.30775238095238	0.5\\
6.31175476190476	0.5\\
6.31575714285714	0.5\\
6.31975952380952	0.5\\
6.3237619047619	0.5\\
6.32776428571429	0.5\\
6.33176666666667	0.5\\
6.33576904761905	0.5\\
6.33977142857143	0.5\\
6.34377380952381	0.5\\
};
\addlegendentry{DB};

\end{axis}
\end{tikzpicture}%